\documentclass{article}

\usepackage[utf8]{inputenc} 
\usepackage[T1]{fontenc}    
\usepackage[colorlinks=true,citecolor=blue,urlcolor=black]{hyperref}        
\usepackage{url}            
\usepackage{booktabs}       
\usepackage{amsfonts}       
\usepackage{nicefrac}       
\usepackage{microtype}      
\usepackage[dvipsnames]{xcolor}         
\usepackage{bm}             
\usepackage{wrapfig}        
\usepackage{varwidth}      

\usepackage[round]{natbib}

\usepackage[left=1in,right=1in,top=1in,bottom=1in]{geometry}
\usepackage{subcaption}
\usepackage{placeins}

\usepackage{algorithm}
\usepackage[noend]{algorithmic}

\usepackage{amsmath}
\usepackage{amssymb}
\usepackage{amsthm}
\usepackage{mathtools}
\usepackage{xspace}

\PassOptionsToPackage{capitalise,noabbrev}{cleveref}
\usepackage{cleveref}
\crefname{equation}{}{}
\Crefname{equation}{}{}

\usepackage[hang,flushmargin]{footmisc}

\newcommand{\ie}{\textit{i.e.,}\xspace}

\newcommand{\E}{\mathbb{E}}

\newcommand{\sgd}{SGD\xspace}

\newcommand{\serveropt}{\textsc{ServerOpt}\xspace}
\newcommand{\fedavg}{\texttt{FedAvg}\xspace}
\newcommand{\fedavgm}{\texttt{FedAvgM}\xspace}
\newcommand{\fedadam}{\texttt{FedAdam}\xspace}
\newcommand{\fedadagrad}{\texttt{FedAdagrad}\xspace}
\newcommand{\fedlars}{\texttt{FedLARS}\xspace}
\newcommand{\fedlamb}{\texttt{FedLamb}\xspace}
\newcommand{\fedsgd}{\texttt{FedSGD}\xspace}

\newcommand{\fedopt}{\texttt{FedOpt}\xspace}

\begin{document}

\title{On Large-Cohort Training for Federated Learning}

\author{%
    \begin{minipage}{0.33\textwidth}
    \centering
    Zachary Charles\\
    Google\\
    \url{zachcharles@google.com}
    \end{minipage}
    \begin{minipage}{0.33\textwidth}
    \centering
    Zachary Garrett\\
    Google\\
    \url{zachgarrett@google.com}
    \end{minipage}
    \begin{minipage}{0.33\textwidth}
    \centering
    Zhouyuan Huo\\
    Google\\
    \url{zhhuo@google.com}
    \end{minipage}\\\\
    \begin{minipage}{0.33\textwidth}
    \centering
    Sergei Shmulyian\\
    Google\\
    \url{sshmulyian@google.com}
    \end{minipage}
    \begin{minipage}{0.33\textwidth}
    \centering
    Virginia Smith\\
    Carnegie Mellon University\\
    \url{smithv@cmu.edu}
    \end{minipage}
}
\date{June 14, 2021}

\maketitle

\begin{abstract}
Federated learning methods typically learn a model by iteratively sampling updates from a population of clients. In this work, we explore how the number of clients sampled at each round (the \emph{cohort size}) impacts the quality of the learned model and the training dynamics of federated learning algorithms. Our work poses three fundamental questions. First, what challenges arise when trying to scale federated learning to larger cohorts? Second, what parallels exist between cohort sizes in federated learning and batch sizes in centralized learning? Last, how can we design federated learning methods that effectively utilize larger cohort sizes? We give partial answers to these questions based on extensive empirical evaluation. Our work highlights a number of challenges stemming from the use of larger cohorts. While some of these (such as generalization issues and diminishing returns) are analogs of large-batch training challenges, others (including training failures and fairness concerns) are unique to federated learning.
\end{abstract}

\section{Introduction}
Federated learning (FL) \citep{mcmahan2017aistats} considers learning a model from multiple clients without directly sharing training data, often under the orchestration of a central server. In this work we focus on \emph{cross-device} FL, in which the aim is to learn across a large population of edge devices \citep[Table 1]{kairouz2019advances}. A distinguishing characteristic of cross-device FL is \emph{partial participation} of the client population: Due to systems constraints such as network size, the  server typically only communicates with a subset of the clients at a time\footnote{In contrast, \emph{cross-silo} settings often have a small set of clients, most of which participate in each round~\citep{{kairouz2019advances}}.}. For example, in the popular \fedavg algorithm~\citep{mcmahan2017aistats}, at each communication round the server broadcasts its current model to a subset of available clients (referred to as a \emph{cohort}), who use the model to initialize local optimization and send their model updates back to the server.

Intuitively, larger cohort sizes have the potential to improve the convergence of FL algorithms. By sampling more clients per round, we can observe a more representative sample of the underlying population---possibly reducing the number of communication rounds needed to achieve a given accuracy. This intuition is reflected in many convergence analyses of FL methods~\citep{khaled2019first,  karimireddy2019scaffold, khaled2020tighter, reddi2021adaptive, yang2021achieving}, which generally show that asymptotic convergence rates improve as the cohort size increases.

Larger cohorts can also provide privacy benefits. For example, when using the distributed differential privacy model~\citep{shi2011privacy, bittau2017prochlo, cheu2019distributed, erlingsson2020encode} in federated learning, noise is typically added to the updates sent from the clients to the server~\citep{mcmahan2017learning}. This helps preserve privacy but can also mar the utility of the learned model.
By dividing the noise among more clients, larger cohorts may mitigate detrimental effects of noise.
Moreover, since  privacy tends to decrease as a function of the number of communication rounds~\citep{abadi2016deep, girgis2020shuffled}, larger cohorts also have the potential to improve privacy in FL by reducing the number of rounds needed for convergence.

Motivated by the potential benefits of large-cohort training, we systematically explore the impact of cohort size in realistic cross-device settings. Our results show that increasing the cohort size may not lead to significant convergence improvements in practice, despite their theoretical benefit~\citep{yang2021achieving}. Moreover, large-cohort training can introduce fundamental optimization and generalization issues. Our results are reminiscent of work on large-batch training in centralized settings, where larger batches can stagnate convergence improvements~\citep{dean2012large, you2017large, golmant2018computational, mccandlish2018empirical, yin2018gradient}, and even lead to generalization issues with deep neural networks~\citep{shallue2019measuring, ma2018power, keskar2017iclr, hoffer2017train, masters2018revisiting, lin2019don, lin2020extrapolation}. While some of the challenges we identify with large-cohort training are parallel to issues that arise in large-batch centralized learning, others are unique to federated learning and have not been previously identified in the literature. 

\textbf{Contributions.} In this work, we provide a novel examination of cohort sizes in federated learning. We give a wide ranging empirical analysis spanning many popular federated algorithms and datasets (Section~\ref{sec:preliminaries}). Despite the many possible benefits of large-cohort training, we find that challenges exist in realizing these benefits (Section~\ref{sec:challenges}). We show that these issues are caused in part by distinctive characteristics of federated training dynamics (Section~\ref{sec:diagnosis}).
Using these insights, we provide partial solutions to the challenges we identify (Section~\ref{sec:better_methods}), focusing on how to adapt techniques from large-batch training, and the limitations of such approaches. Our solutions are designed to serve as simple benchmarks for future work. We conclude by discussing limitations and open problems (\cref{sec:future_work}).
Throughout, we attempt to uncover interesting theoretical questions, but remain firmly grounded in the practical realities of federated learning. 

\subsection{Related Work}

\textbf{Large-batch training.} In non-federated settings, mini-batch stochastic gradient descent (SGD) and its variants are common choices for training machine learning models, particularly deep neural networks. While larger mini-batch sizes ostensibly allow for improved convergence (in terms of the number of steps required to reach a desired accuracy), in practice speedups may quickly saturate when increasing the mini-batch size. This property of diminishing returns has been explored both empirically~\citep{dean2012large,mccandlish2018empirical,golmant2018computational,shallue2019measuring} and theoretically~\citep{ma2018power,yin2018gradient}. Beyond the issue of speedup saturation, numerous works have also observed a \textit{generalization gap} when training deep neural networks with large batches~\citep{keskar2017iclr,hoffer2017train,you2017large,masters2018revisiting,lin2019don,lin2020extrapolation}. Our work differs from these areas by specifically exploring how the cohort size (the number of selected clients) affects \emph{federated} optimization methods. While some of the issues with large-batch training appear in large-cohort training, we also identify a number of new challenges introduced by the federated setting.

\textbf{Optimization for federated learning.} Significant attention has been paid towards developing federated optimization techniques. Such work has focused on various aspects, including communication-efficiency~\citep{konevcny2016federated, mcmahan2017aistats, basu2019qsparse, laguel2020device}, data and systems heterogeneity~\citep{li2020federated, li2019feddane, karimireddy2019scaffold, hsu2019measuring, karimireddy2020mime, li2018federated, wang2020federated, li2019fedmd}, and fairness~\citep{li2019fair, hu2020fedmgda+}. We provide a description of some relevant methods in Section~\ref{sec:preliminaries}, and defer readers to recent surveys such as \citep{kairouz2019advances} and \citep{li2020federated} for additional background. One area pertinent to our work is that of variance reduction for federated learning, which can mitigate negative effects of data heterogeneity~\citep{karimireddy2019scaffold, karimireddy2020mime, zhang2020fedpd}. However, such methods often require clients to maintain state across rounds~\citep{karimireddy2019scaffold, zhang2020fedpd}, which may be infeasible in cross-device settings~\citep{kairouz2019advances}. Moreover, such methods may not perform well in settings with limited client participation~\citep{reddi2021adaptive}.
Many convergence analyses of federated optimization methods show that larger cohort sizes can lead to improved convergence rates, even without explicit variance reduction~\citep{khaled2019first, khaled2020tighter, yang2021achieving}. These analyses typically focus on asymptotic convergence, and require assumptions on learning rates and heterogeneity that may not hold in practice~\citep{kairouz2019advances, charles2021convergence}. In this work, we attempt to see whether increasing the cohort size leads to improved convergence in practical, communication-limited settings.

\textbf{Client sampling.} 
A number of works have explored how to select cohorts of a fixed size in cross-device FL~\citep{nishio2019client, goetz2019active, cho2020client, chen2020optimal, ribero2020communication}. Such methods can yield faster convergence than random sampling by carefully selecting the clients that participate at each round, based on quantities such as the client loss. However, such approaches typically require the server to be able to choose which clients participate in a cohort. In practice, cohort selection in cross-device federated learning is often governed by client availability, and is not controlled by the server~\citep{bonawitz2019towards, paulik2021federated}. In this work we instead focus on the impact of size of the cohort, assuming the cohort is sampled at random.

\section{Preliminaries}\label{sec:preliminaries}

Federated optimization methods often aim to minimize a weighted average of client loss functions:
\begin{equation}\label{eq:objective_fn}
    \min_x f(x) := \sum_{k=1}^K p_kf_k(x),
\end{equation}
where $K$ is total number of clients, the $p_k$ are client weights satisfying $p_k \geq 0$, and $f_k$ is the loss function of client $k$. For practical reasons, $p_k$ is often set to the number of examples in client $k$'s local dataset~\citep{mcmahan2017aistats, li2018federated}.

To solve \eqref{eq:objective_fn}, each client in a sampled cohort could send $\nabla f_k(x)$ to the server, and the server could then apply (mini-batch) \sgd. This approach is referred to as \fedsgd~\citep{mcmahan2017aistats}. This requires communication for every model update, which may not be desirable in communication-limited settings. To address this, \citet{mcmahan2017aistats} propose \fedavg, in which clients perform multiple epochs of local training, potentially reducing the number of communication rounds needed for convergence.

We focus on a more general framework, \fedopt, introduced by \citet{reddi2021adaptive} that uses both client and server optimization. At each round, the server sends its model $x$ to a cohort of clients $C$ of size $M$. Each client $c_k \in C$ performs $E$ epochs of training using mini-batch \sgd with client learning rate $\eta_c$, producing a local model $x_k$. Each client $k \in C$ then communicates their \emph{client update} $\Delta_k$ to the server, where $\Delta_k:= x_k - x$ is the difference between the client's local model and the server model. The server computes a weighted average $\Delta$ of the client updates, and updates its own model via
\begin{equation}\label{eq:server_update}
x' = \serveropt(x, \eta_s, \Delta) \, ,
\end{equation}
where $\serveropt(x, \eta_s, g)$ is some first-order optimizer, $\eta_s$ is the server learning rate, and $g$ is a gradient estimate. For example, if \serveropt is \sgd, then $\serveropt(x, \eta_s, g) = x-\eta_sg$. The $\Delta$ in \eqref{eq:server_update} is referred to as a \textbf{pseudo-gradient}~\citep{reddi2021adaptive}. While $\Delta$ may not be an unbiased estimate of $\nabla f$, it can serve a somewhat comparable role (though as we show in \cref{sec:diagnosis}, there are important distinctions). Full pseudo-code of \fedopt is given in \cref{alg:fedopt}.

\setlength{\textfloatsep}{10pt}
\begin{algorithm}[h]
    \begin{algorithmic}
	\caption{\fedopt framework}
	\label{alg:fedopt}
	\STATE {\bf Input:}  $M$, $T$ $E$, $x^1$, $\eta_c$, $\eta_s$, \serveropt, $\{p_k\}_{k=1}^K$
	\FOR  {$t=1, \cdots, T$}
		\STATE The server selects a cohort $C_t$ of $M$ clients uniformly at random, without replacement.
		\STATE The server sends $x^t$ to all clients in $C_t$.
		\STATE Each client $k \in C_t$ performs $E$ epochs of mini-batch \sgd on $f_k$ with step-size $\eta_c$.
		\STATE After training, each $k \in C_t$ has a local model $x_k^t$ and sends $\Delta_k^t = x^t - x_k^t$ to the server.
		\STATE The server computes a pseudo-gradient $\Delta^t$ and updates its model via
		\[
		\Delta^t = \dfrac{\sum_{k \in C_t} p_k\Delta_k^t}{\sum_{k \in C_t} p_k},~~~x^{t+1} = \serveropt(x_t, \eta_s, \Delta^t).
		\]
    \ENDFOR
	\end{algorithmic}
\end{algorithm}

\cref{alg:fedopt} generalizes a number of federated learning algorithms, including \fedavg~\citep{mcmahan2017aistats}, \fedavgm~\citep{hsu2019measuring}, \fedadagrad~\citep{reddi2021adaptive}, and \fedadam~\citep{reddi2021adaptive}. These are the cases where \serveropt is \sgd, \sgd with momentum, Adagrad~\citep{mcmahan2010adaptive, duchi2011adaptive}, and Adam~\citep{kingma2014adam}, respectively. \fedsgd is realized when \serveropt is \sgd, $\eta_c = 1$, $E = 1$, and each client performs full-batch gradient descent.

\subsection{Experimental Setup}\label{sec:experiment_setup}

We aim to understand how the cohort size $M$ impacts the performance of \cref{alg:fedopt}. In order to study this, we perform a wide-ranging empirical evaluation using various special cases of \cref{alg:fedopt} across multiple datasets, models, and tasks. We discuss the key facets of our experiments below.

\textbf{Datasets, models, and tasks.} We use four datasets: CIFAR-100~\citep{krizhevsky2009learning}, EMNIST~\citep{cohen2017emnist}, Shakespeare~\citep{caldas2018leaf}, and Stack Overflow~\citep{stackoverflow}. For CIFAR-100, we use the client partitioning proposed by \citet{reddi2021adaptive}. The other three datasets have natural client partitions that we use. For EMNIST, the handwritten characters are partitioned by their author. For Shakespeare, speaking lines in Shakespeare plays are partitioned by their speaker. For Stack Overflow, posts on the forum are partitioned by their author.  The number of clients and examples in the training and test sets are given in \cref{table:datasets}.

\begin{table}[ht]
    \caption{Dataset statistics.}
    \label{table:datasets}
    \begin{center}
    \begin{small}
    \begin{sc}
    \begin{tabular}[t]{@{}lrrrr@{}}    
        \toprule
        Dataset & Train Clients & Train Examples & Test Clients & Test Examples \\
        \midrule
        CIFAR-100 & 500 & 50,000 & 100 & 10,000 \\
        EMNIST & 3,400 & 671,585 & 3,400 & 77,483\\
        Shakespeare & 715 & 16,068 & 715 & 2,356\\
        Stack Overflow & 342,477 & 135,818,730 & 204,088 & 16,586,035 \\
        \bottomrule
    \end{tabular}
    \end{sc}
    \end{small}
    \end{center}
\end{table}

For CIFAR-100, we train a ResNet-18, replacing batch normalization layers with group normalization (as proposed and empirically validated in federated settings by \citet{hsieh2019non}). For EMNIST, we train a convolutional network with two convolutional layers, max-pooling, dropout, and two dense layers. For Shakespeare, we train an RNN with two LSTM layers to perform next-character-prediction. For Stack Overflow, we perform next-word-prediction using an RNN with a single LSTM layer. For full details on the models and datasets, see Appendix \ref{appendix:models_and_datasets}.

\textbf{Algorithms.} We implement many special cases of \cref{alg:fedopt}, including \fedsgd, \fedavg, \fedavgm, \fedadagrad, and \fedadam. We also develop two novel methods: \fedlars and \fedlamb, which are the special cases of \cref{alg:fedopt} where \serveropt is LARS~\citep{you2017large} and Lamb~\citep{you2019large}, respectively. See \cref{sec:better_methods} for the motivation and full details of these algorithms.

\textbf{Implementation and tuning.} Unless otherwise specified, in \cref{alg:fedopt} clients perform $E = 1$ epochs of training with mini-batch \sgd. Their batch size is fixed per-task. We set $p_k$ to be the number of examples in client $k$'s dataset. We tune learning rates for all algorithms and models using a held-out validation set: We perform $T = 1500$ rounds of training with $M = 50, E = 1$ for each algorithm and model, varying $\eta_c, \eta_s$ over $\{10^i~|~-3 \leq i \leq 1\}$ and select the values that maximize the average validation performance over 5 random trials. All other hyperparameters (such as momentum) are fixed. For more details, see \cref{appendix:experiment_details}. We provide open-source implementations of all simulations in TensorFlow Federated~\citep{tff}\footnote{\url{https://github.com/google-research/federated/tree/bcb0dadb438280fcecf733db8f894d5c645a49a9/large_cohort}}. All experiments were conducted using clusters of multi-core CPUs, though our results are independent of wall-clock time and amount of compute resources.

\textbf{Presentation of results.} We apply the algorithms above to the tasks listed above with varying cohort sizes. For brevity, we present only a fraction of our results, selecting representative experiments to illustrate large-cohort training phenomena. The full set of experimental results can be found in \cref{appendix:full_results}. We run 5 random trials for each experiment, varying the model initialization and which clients are sampled per round. In all subsequent figures, dark lines indicate the mean across the 5 trials, and shaded regions indicate one standard deviation above and below the mean.

\section{Large-Cohort Training Challenges}\label{sec:challenges}

In this section we explore challenges that exist when using large cohorts in federated learning. While some of these challenges mirror issues in large-batch training, others are unique to federated settings. While we provide concrete recommendations for mitigating some of these challenges, our discussion is generally centered around introducing and exploring these challenges in the context of federated learning.

\subsection{Catastrophic Training Failures}\label{sec:catastrophe}

We first discuss a practical issue unique to large-cohort training. Due to data heterogeneity, the server model $x$ may be misaligned with some client's loss $f_k$, in which case $\nabla f_k(x)$ can blow up and lead to optimization problems. This issue is exacerbated by large cohorts, as we are more likely to sample misaligned clients. To demonstrate this, we applied \fedavg to EMNIST with varying cohort sizes $M$, using learning rates tuned for $M = 10$. For each $M$, we performed 5 random trials and recorded whether a \emph{catastrophic training failure} occurred, in which the training accuracy decreased by a factor of at least $1/2$ in a single round.

\begin{figure}[ht]
\centering
\begin{subfigure}{0.24\textwidth}
    \centering
    \includegraphics[width=1\linewidth]{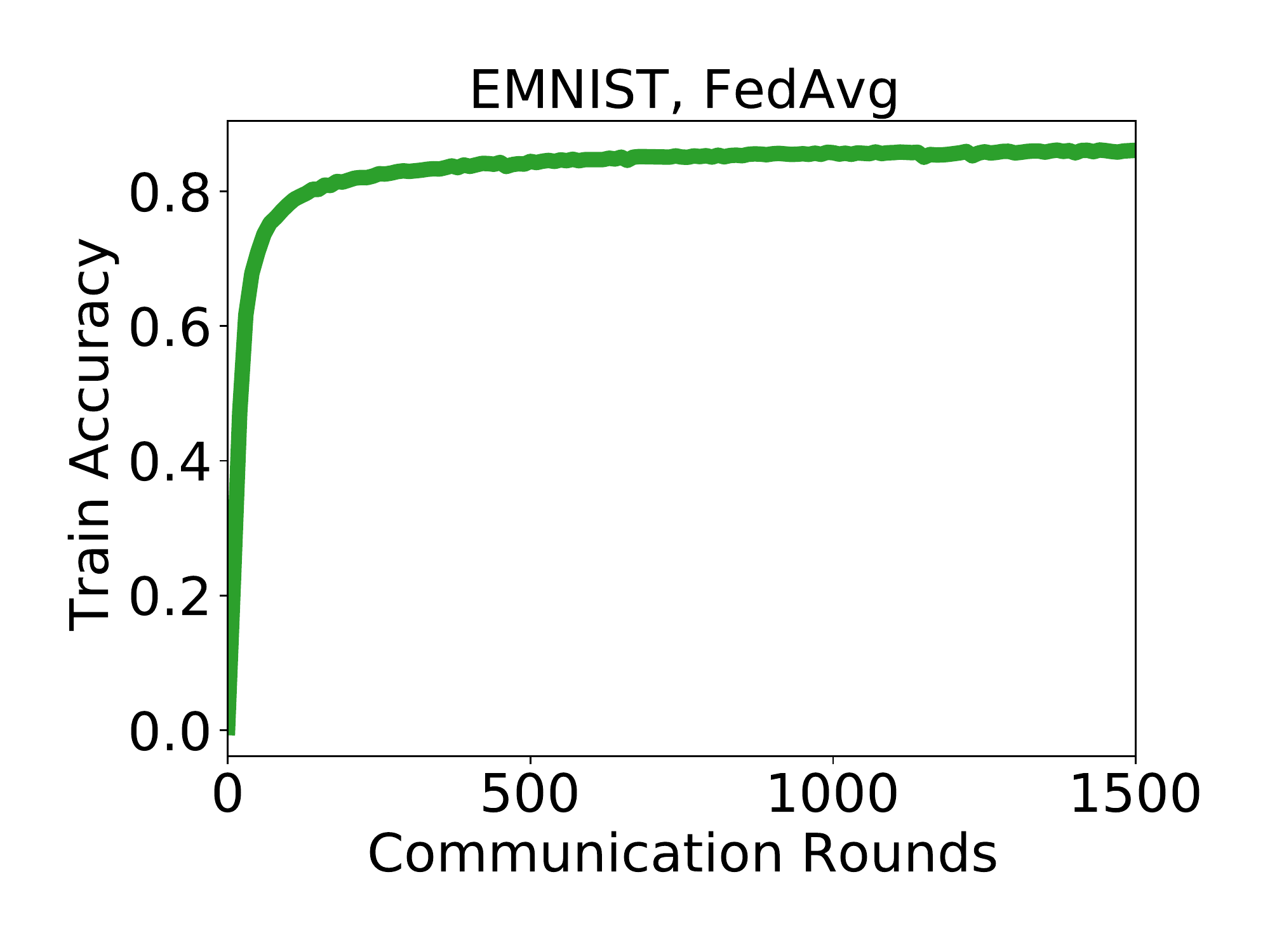}
\end{subfigure}%
\begin{subfigure}{0.24\textwidth}
    \centering
    \includegraphics[width=1\linewidth]{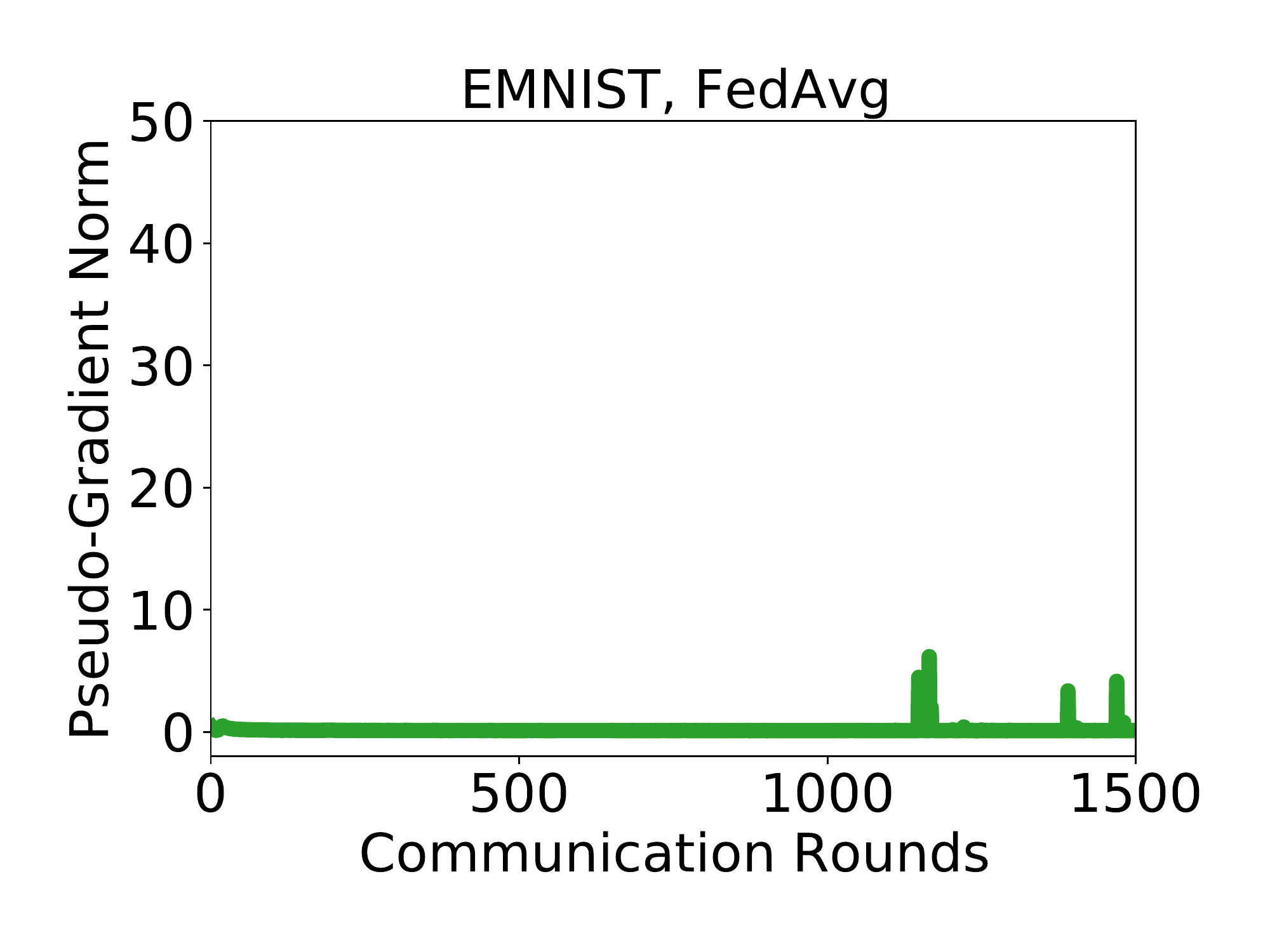}
\end{subfigure}%
\begin{subfigure}{0.24\textwidth}
    \centering
    \includegraphics[width=1\linewidth]{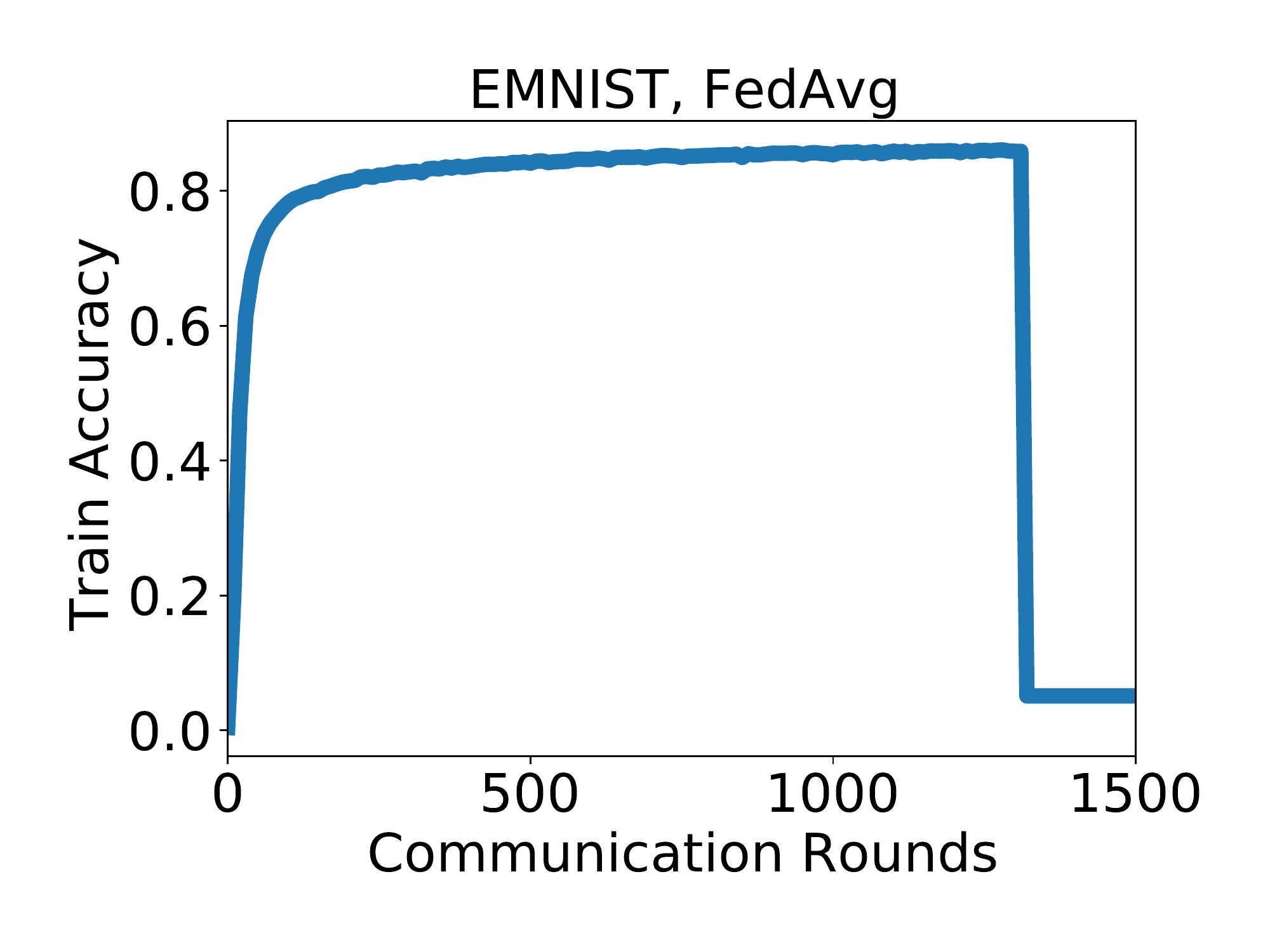}
\end{subfigure}%
\begin{subfigure}{0.24\textwidth}
    \centering
    \includegraphics[width=1\linewidth]{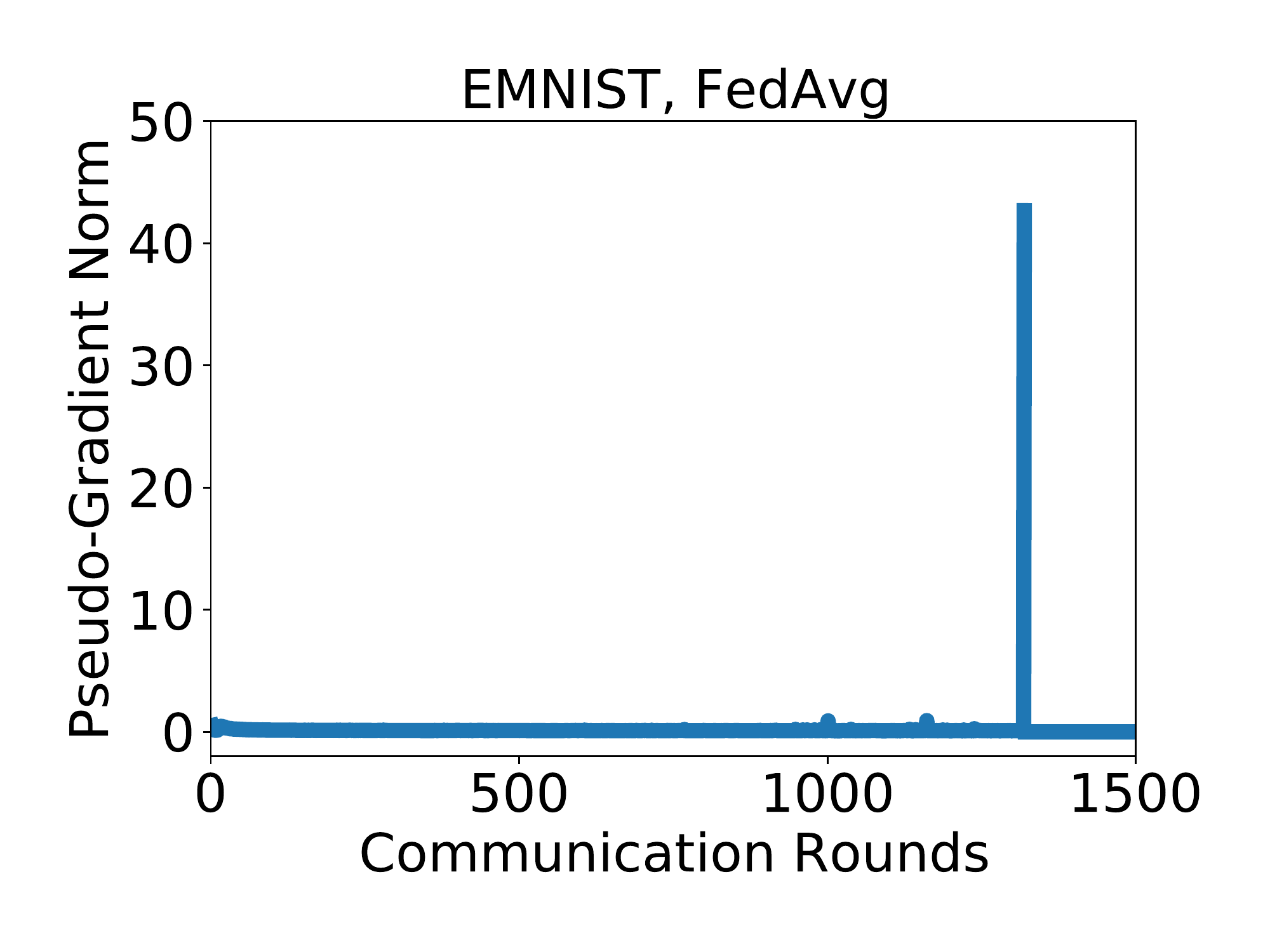}
\end{subfigure}%

\caption{Applying \fedavg to EMNIST with cohort size $200$. We plot the train accuracy and norm of the pseudo-gradient for a trial that ran successfully ({\color{OliveGreen}\textbf{left}}), and one that experienced a catastrophic training failure ({\color{NavyBlue}\textbf{right}}). The trials differed only in which clients were randomly sampled each round.}
\label{fig:catastrophic_failure_example}
\end{figure}

The failure rate increased from 0\% for $M = 10$ to 80\% for $M = 800$. When failures occurred, we consistently saw a spike in the norm of the pseudo-gradient $\Delta$ (see \cref{fig:catastrophic_failure_example}). In order to prevent this spike, we apply clipping to the client updates. We use the \emph{adaptive clipping} method of \citep{andrew2019differentially}. While this technique was originally designed for training with differential privacy, we found that it greatly improved the stability of large-cohort training. Applying \fedavg to EMNIST with adaptive clipping, no catastrophic training failures occurred for any cohort size. We use adaptive clipping in all subsequent experiments. For more details, see \cref{appendix:adaptive_clipping}.

\subsection{Diminishing Returns}\label{sec:diminishing_returns}

\begin{figure}[ht]
\centering
\begin{subfigure}{0.24\textwidth}
    \centering
    \includegraphics[width=1\linewidth]{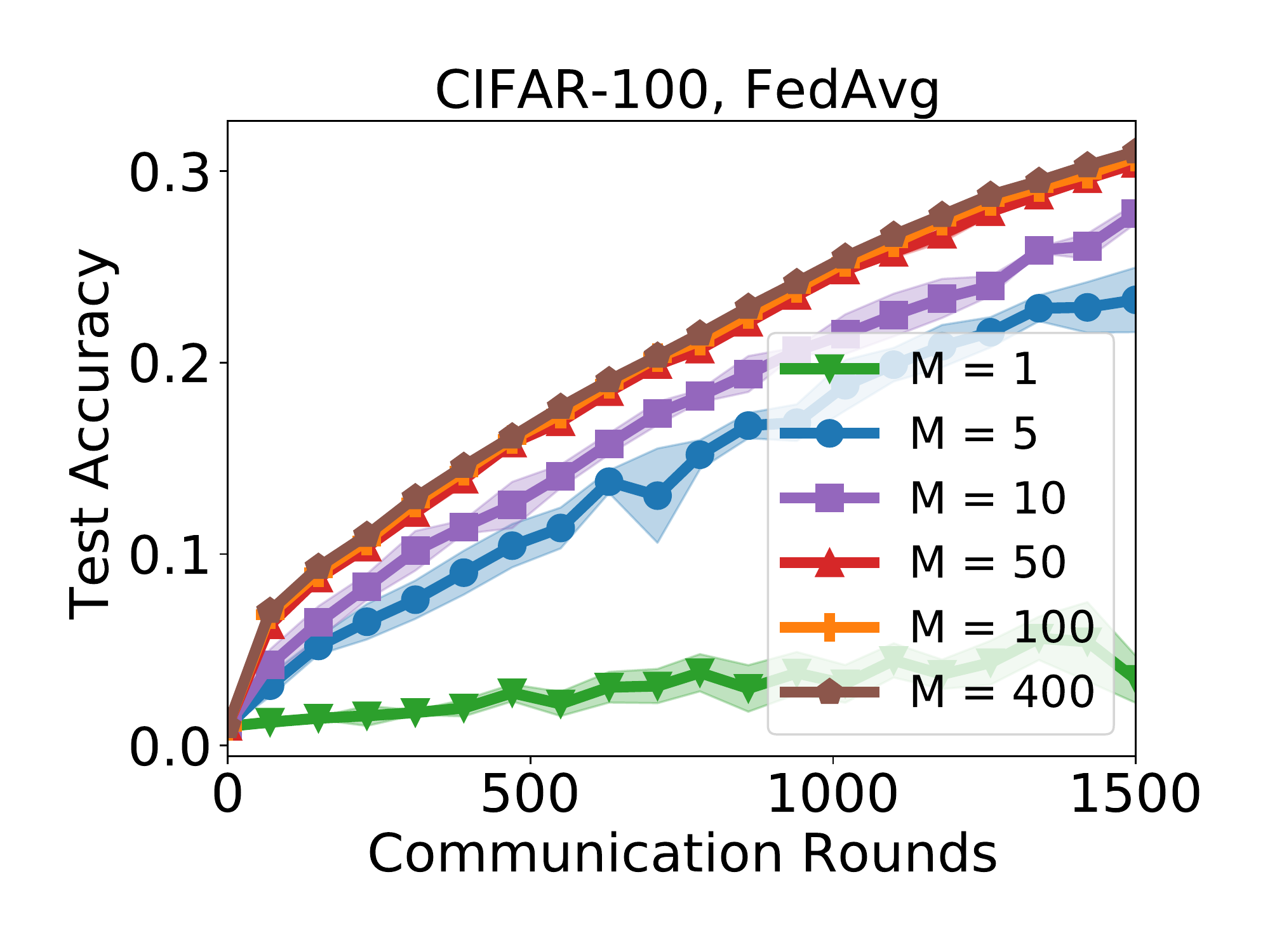}
\end{subfigure}%
\begin{subfigure}{0.24\textwidth}
    \centering
    \includegraphics[width=1\linewidth]{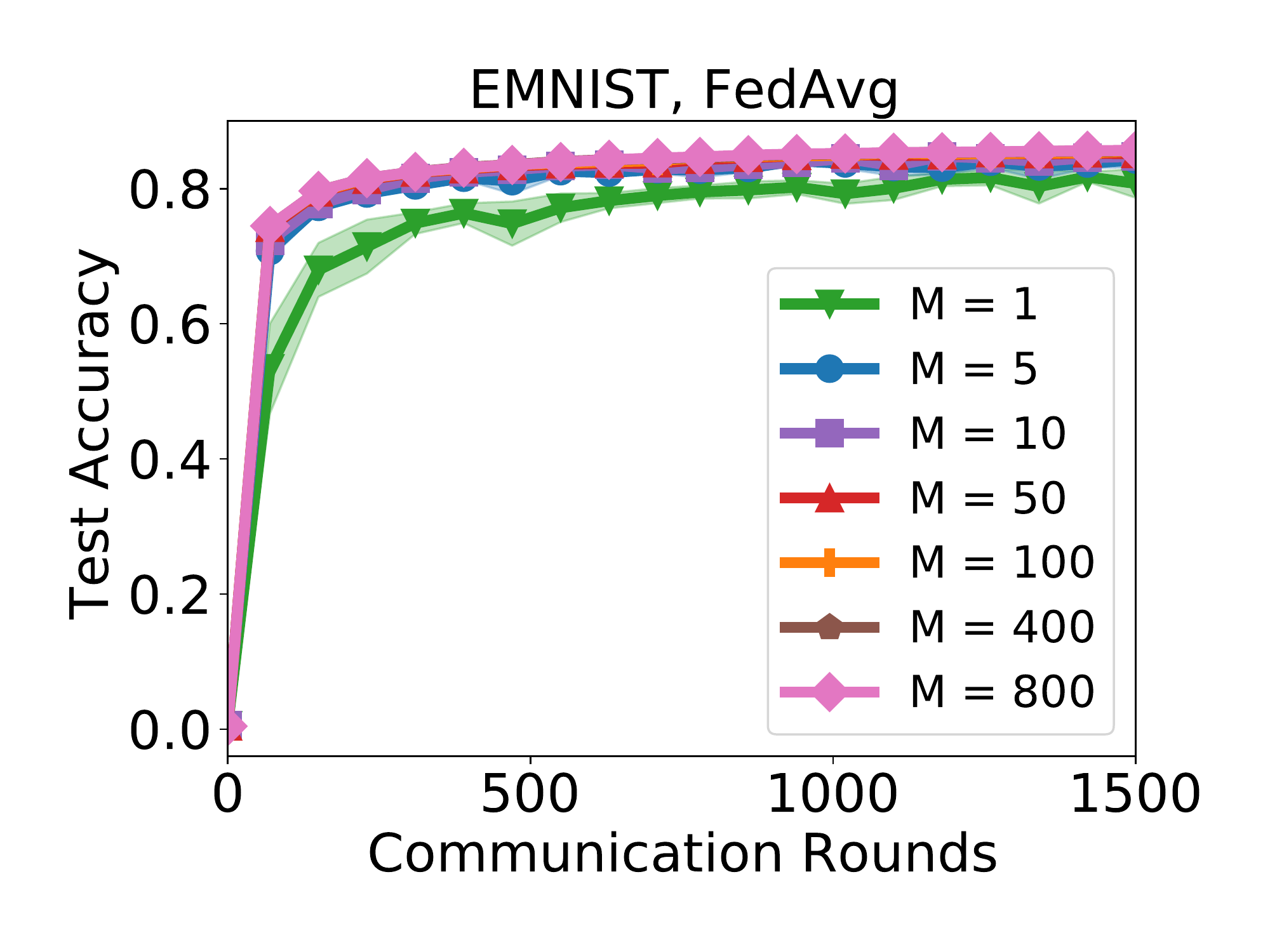}
\end{subfigure}%
\begin{subfigure}{0.24\textwidth}
    \centering
    \includegraphics[width=1\linewidth]{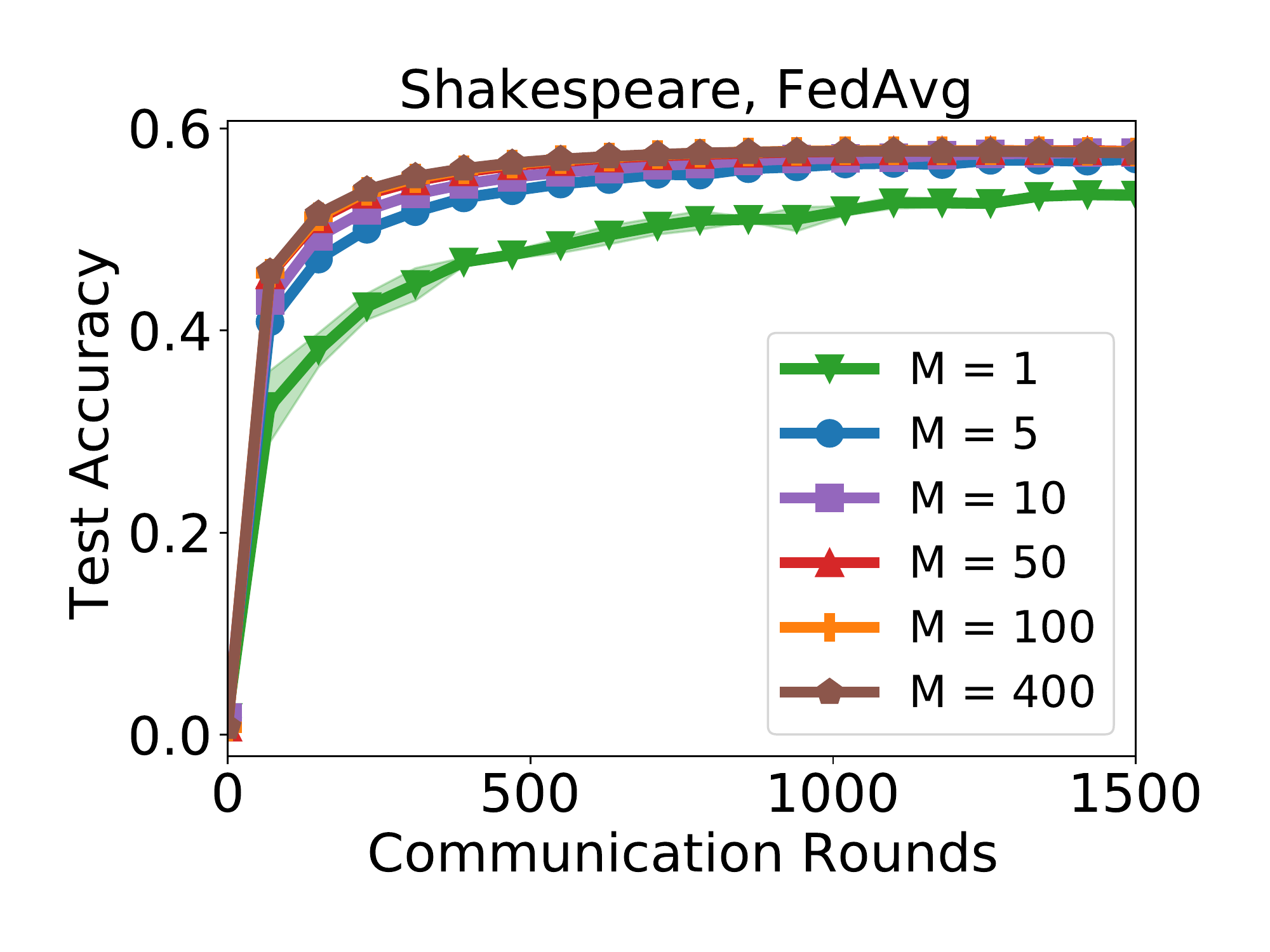}
\end{subfigure}%
\begin{subfigure}{0.24\textwidth}
    \centering
    \includegraphics[width=1\linewidth]{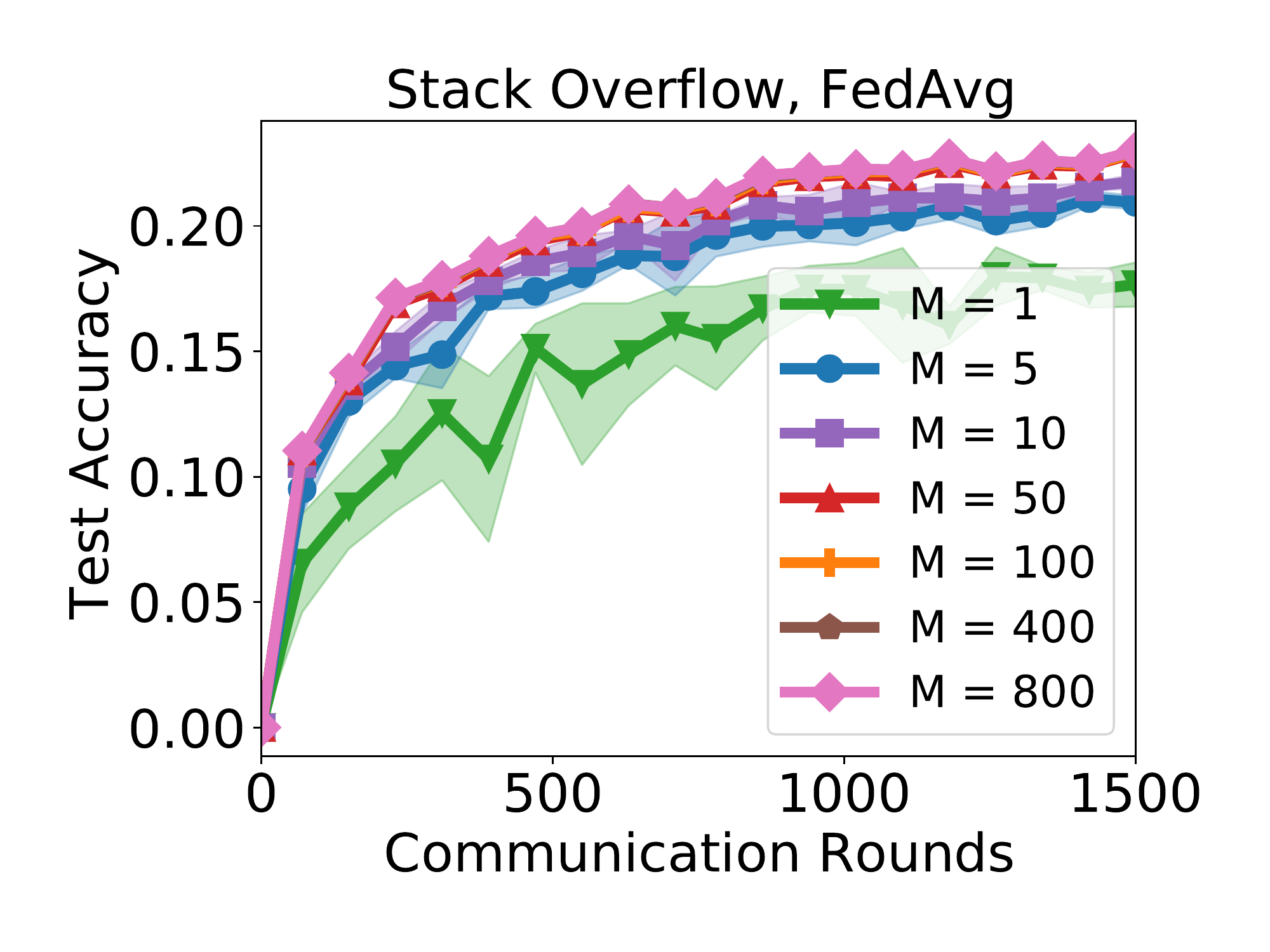}
\end{subfigure}
\begin{subfigure}{0.24\textwidth}
    \centering
    \includegraphics[width=1\linewidth]{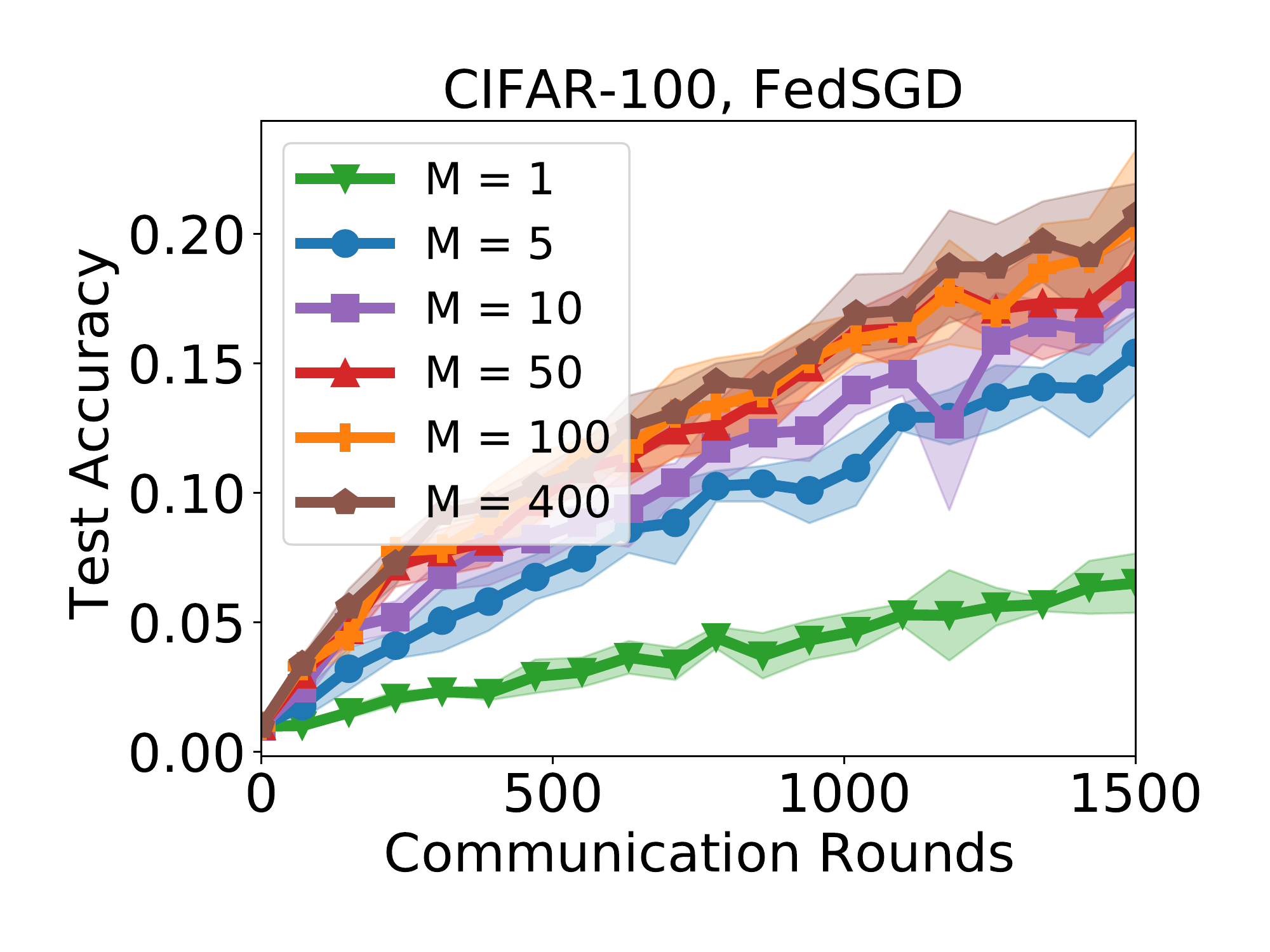}
\end{subfigure}%
\begin{subfigure}{0.24\textwidth}
    \centering
    \includegraphics[width=1\linewidth]{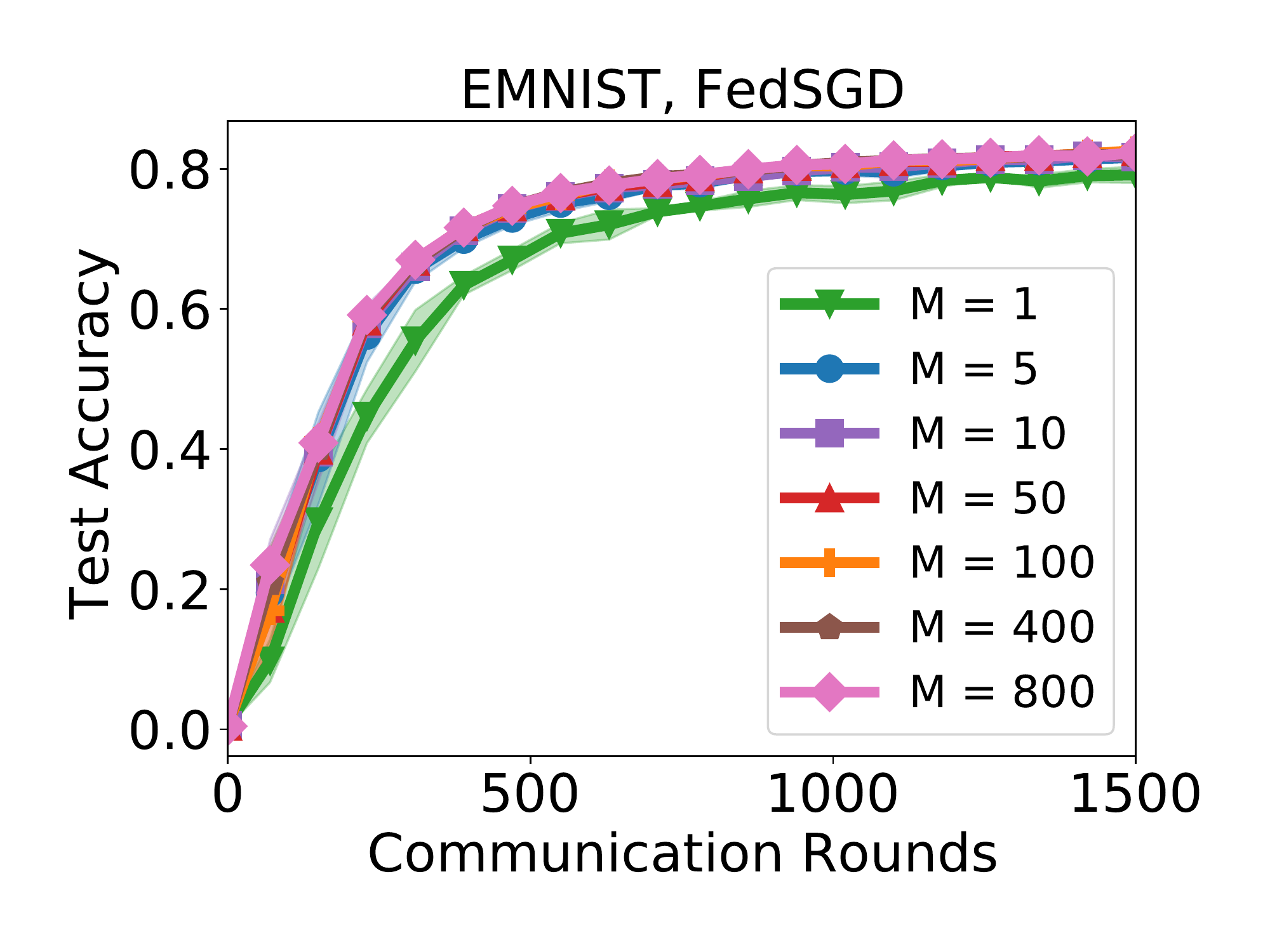}
\end{subfigure}%
\begin{subfigure}{0.24\textwidth}
    \centering
    \includegraphics[width=1\linewidth]{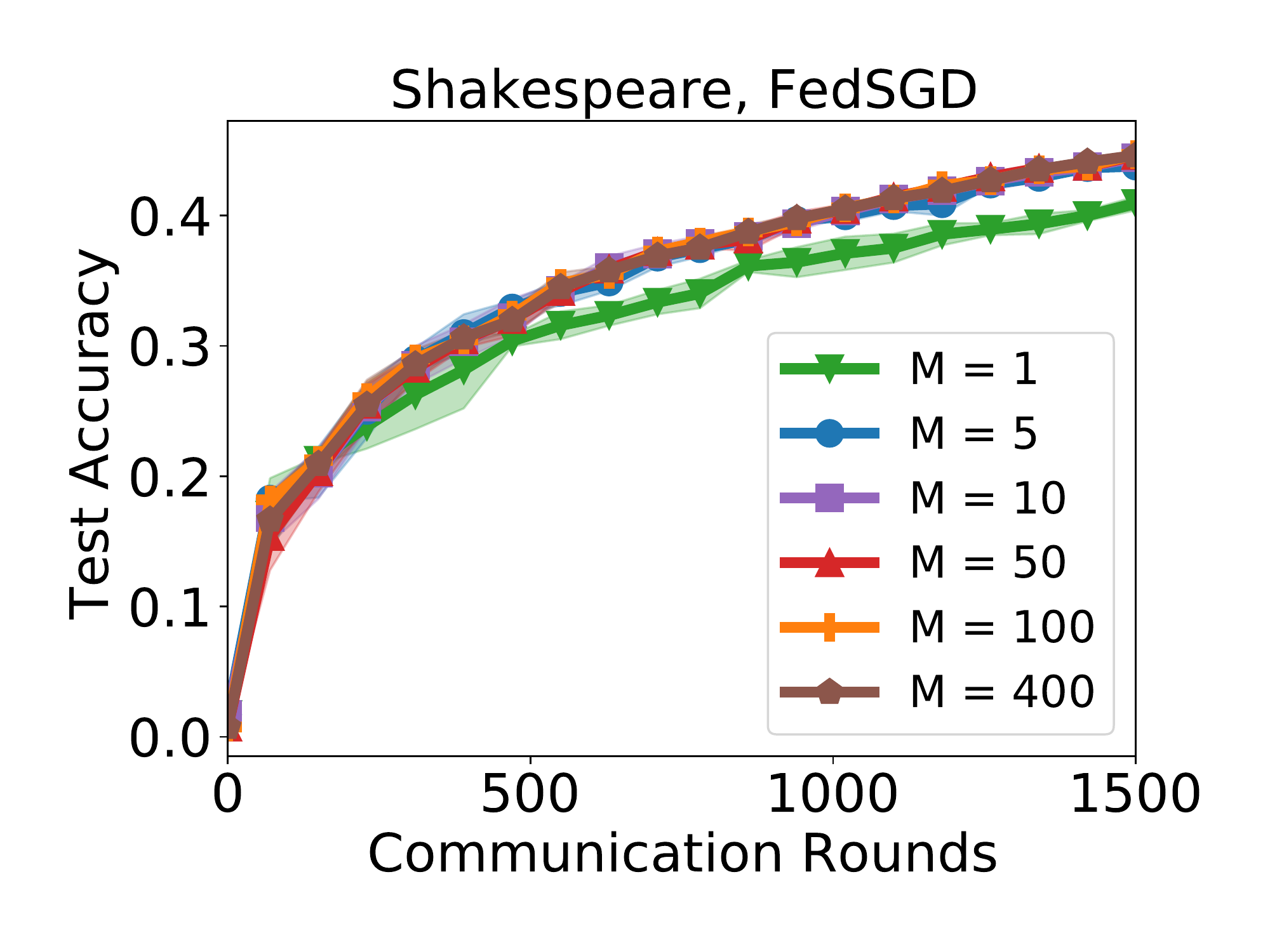}
\end{subfigure}%
\begin{subfigure}{0.24\textwidth}
    \centering
    \includegraphics[width=1\linewidth]{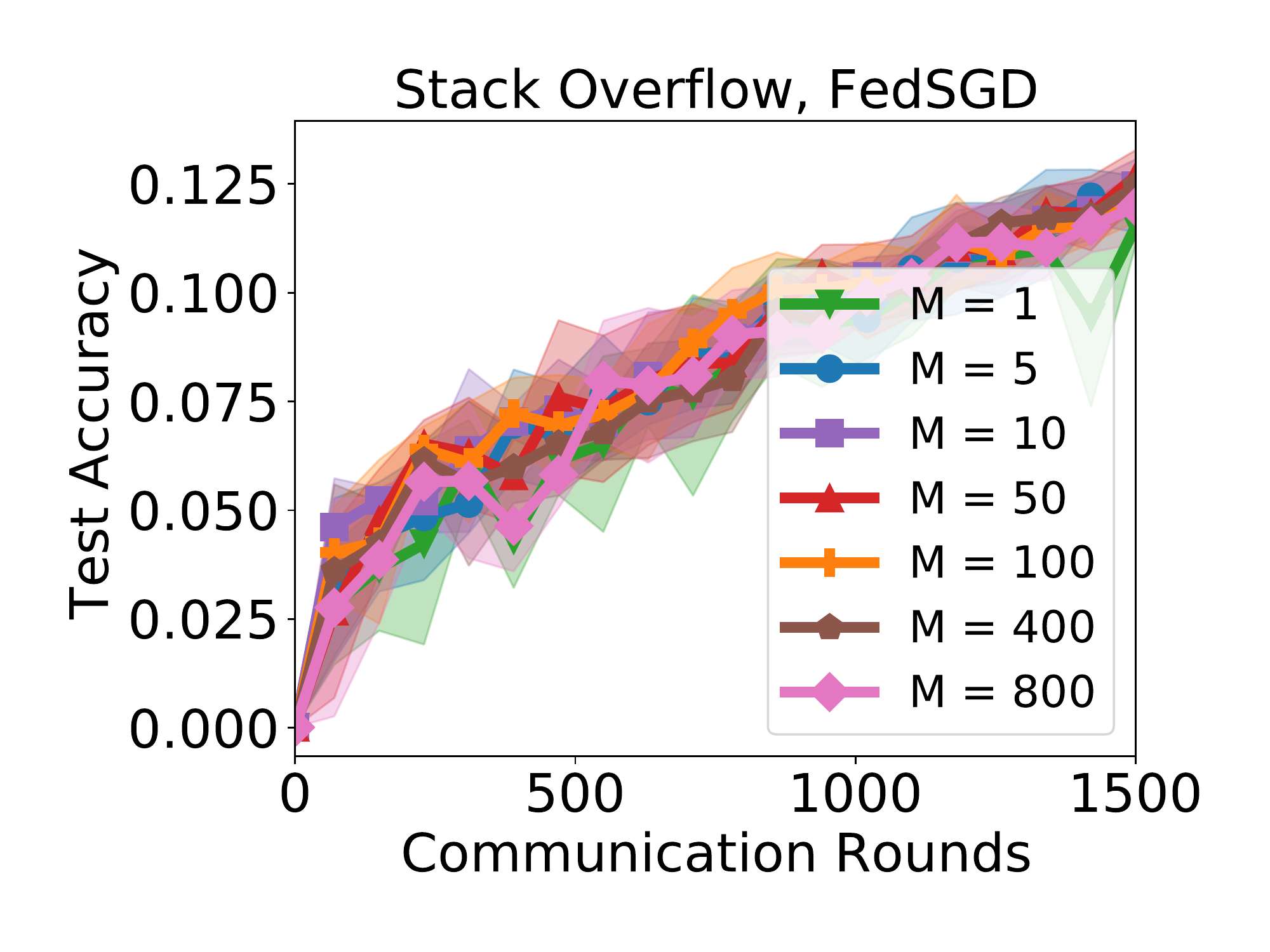}
\end{subfigure}%
\caption{Test accuracy of \fedavg (top) and \fedsgd (bottom), for various cohort sizes $M$, over the course of 1500 communication rounds.}
\label{fig:fedavg_and_fedsgd}
\end{figure}

In this section, we show that increasing $M$ in \cref{alg:fedopt} can lead to improved convergence, but that such improvements diminish with $M$.
To demonstrate this, we plot the test accuracy of \fedavg and \fedsgd across multiple tasks, for varying cohort sizes $M$ in \cref{fig:fedavg_and_fedsgd}.

We see that convergence benefits do not scale linearly with cohort size. While increasing $M$ from $1$ to $10$ can significantly improve convergence, there is generally a threshold after which point increasing $M$ incurs little to no change in convergence. This threshold is typically between $M = 10$ and $M = 50$. Interestingly, this seems to be true for both tasks, even though $M = 50$ represents 10\% of the training clients for CIFAR-10, but only approximately 0.015\% of the training clients for Stack Overflow. We see comparable results for EMNIST and Shakespeare, as well as for other optimizers, including \fedadam and \fedadagrad. See \cref{appendix:test_accuracy_versus_rounds} for the full results.
In short, we see that increasing $M$ alone can lead to \emph{diminishing returns}, or even no returns in terms of convergence. This mirrors issues of diminishing returns in large-batch training~\citep{dean2012large,mccandlish2018empirical,golmant2018computational,shallue2019measuring}.

\begin{figure}[ht]
\centering
\begin{subfigure}{0.24\textwidth}
    \centering
    \includegraphics[width=1\linewidth]{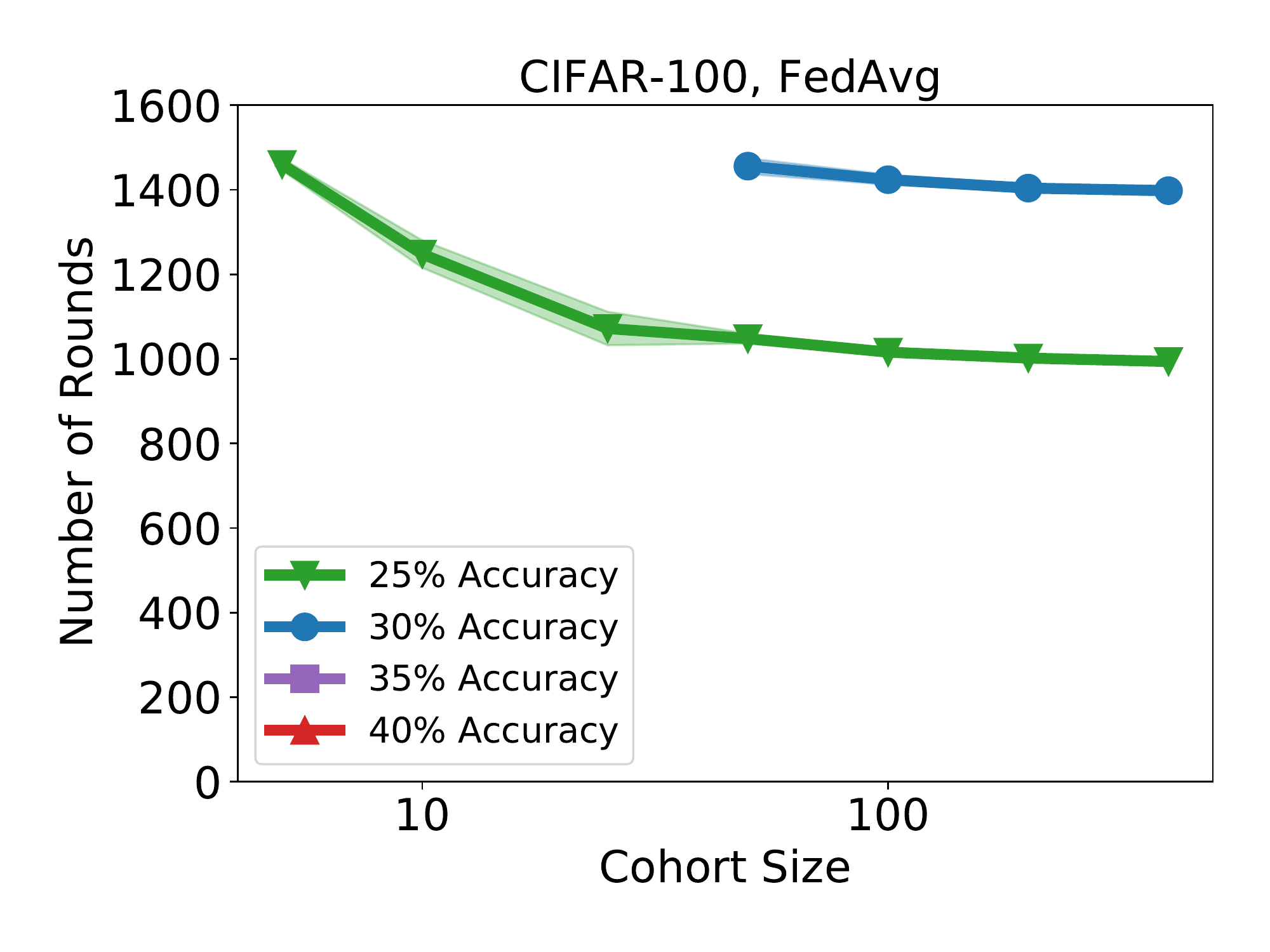}
\end{subfigure}%
\begin{subfigure}{0.24\textwidth}
    \centering
    \includegraphics[width=1\linewidth]{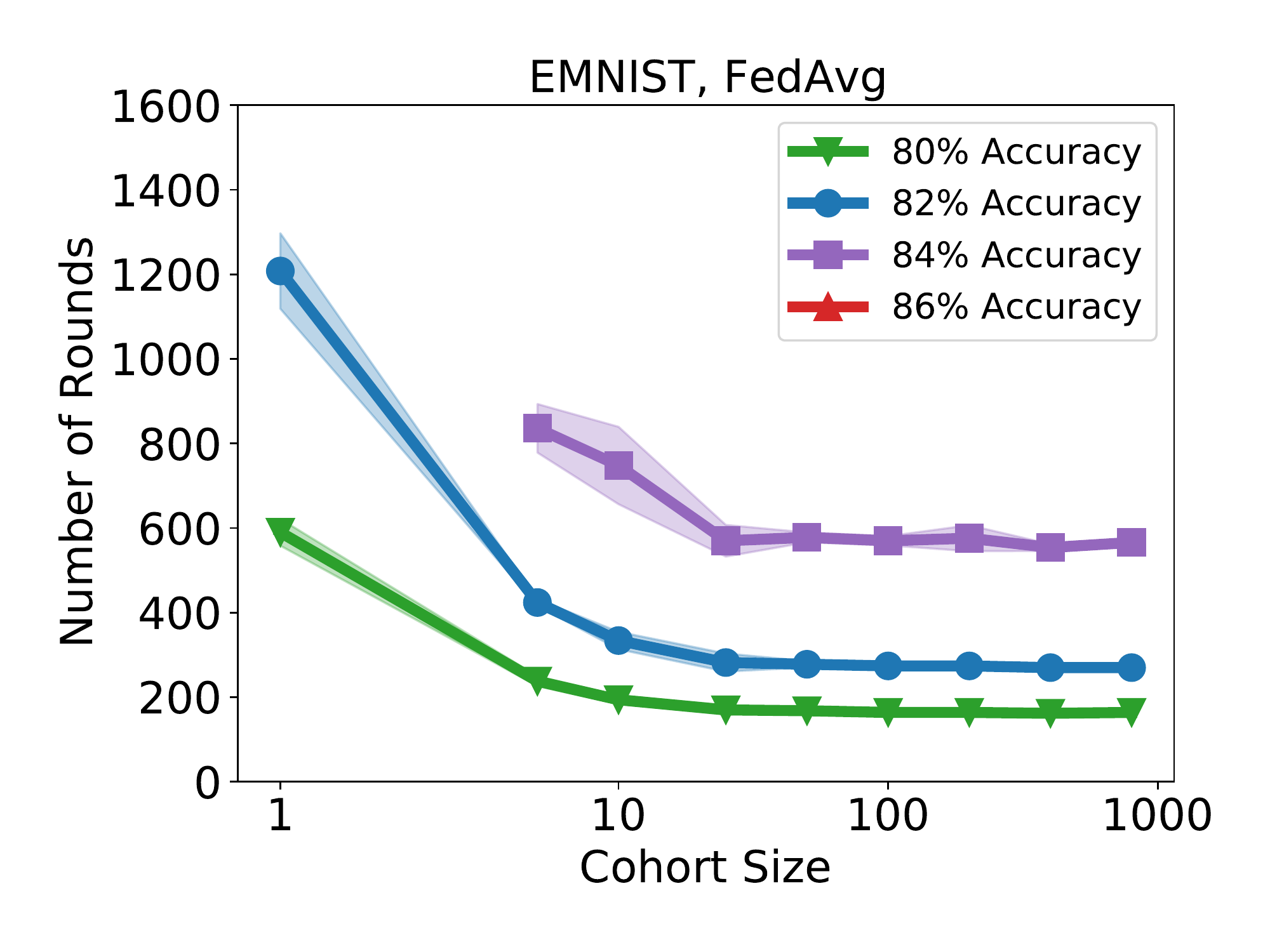}
\end{subfigure}%
\begin{subfigure}{0.24\textwidth}
    \centering
    \includegraphics[width=1\linewidth]{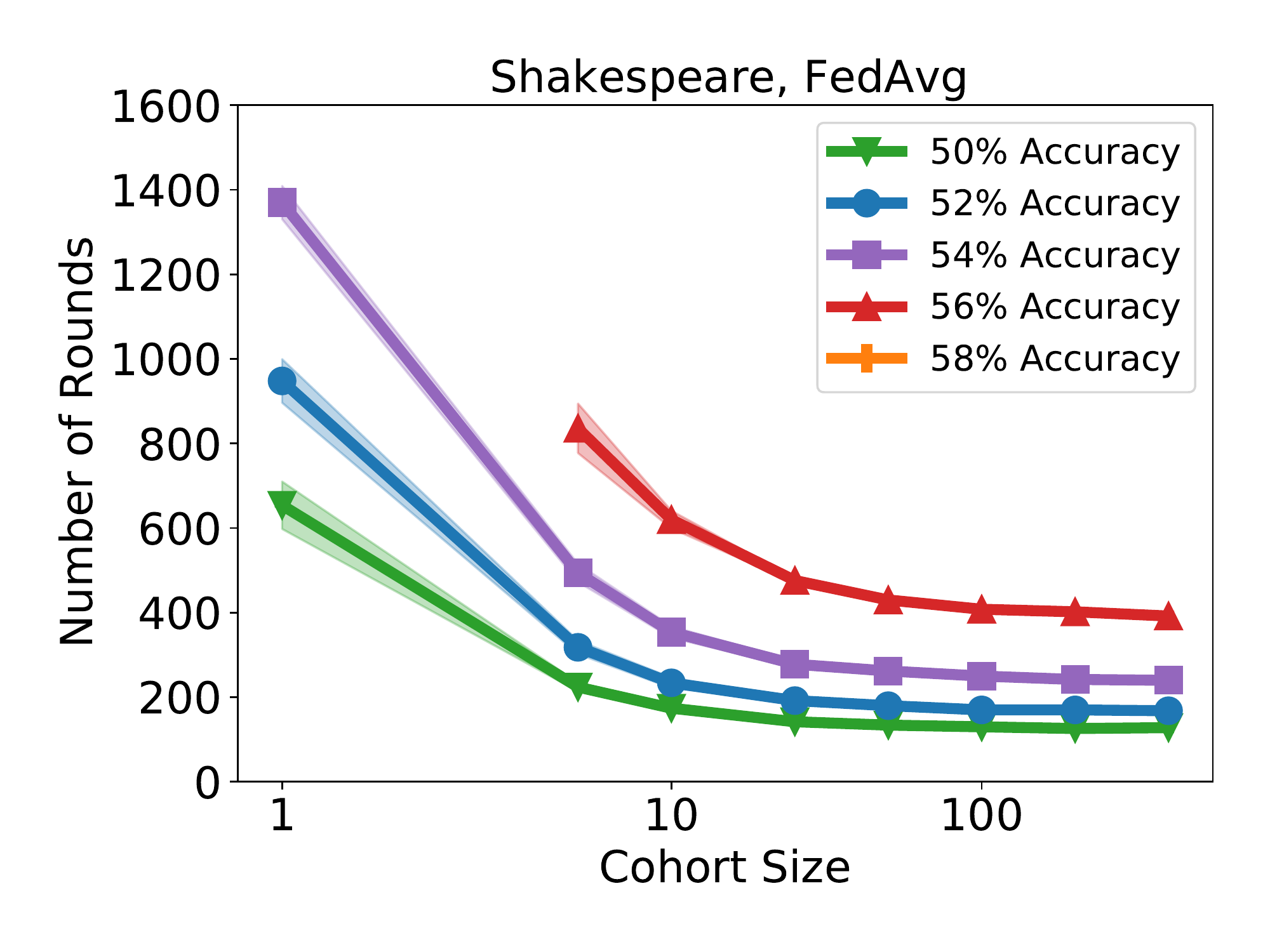}
\end{subfigure}%
\begin{subfigure}{0.24\textwidth}
    \centering
    \includegraphics[width=1\linewidth]{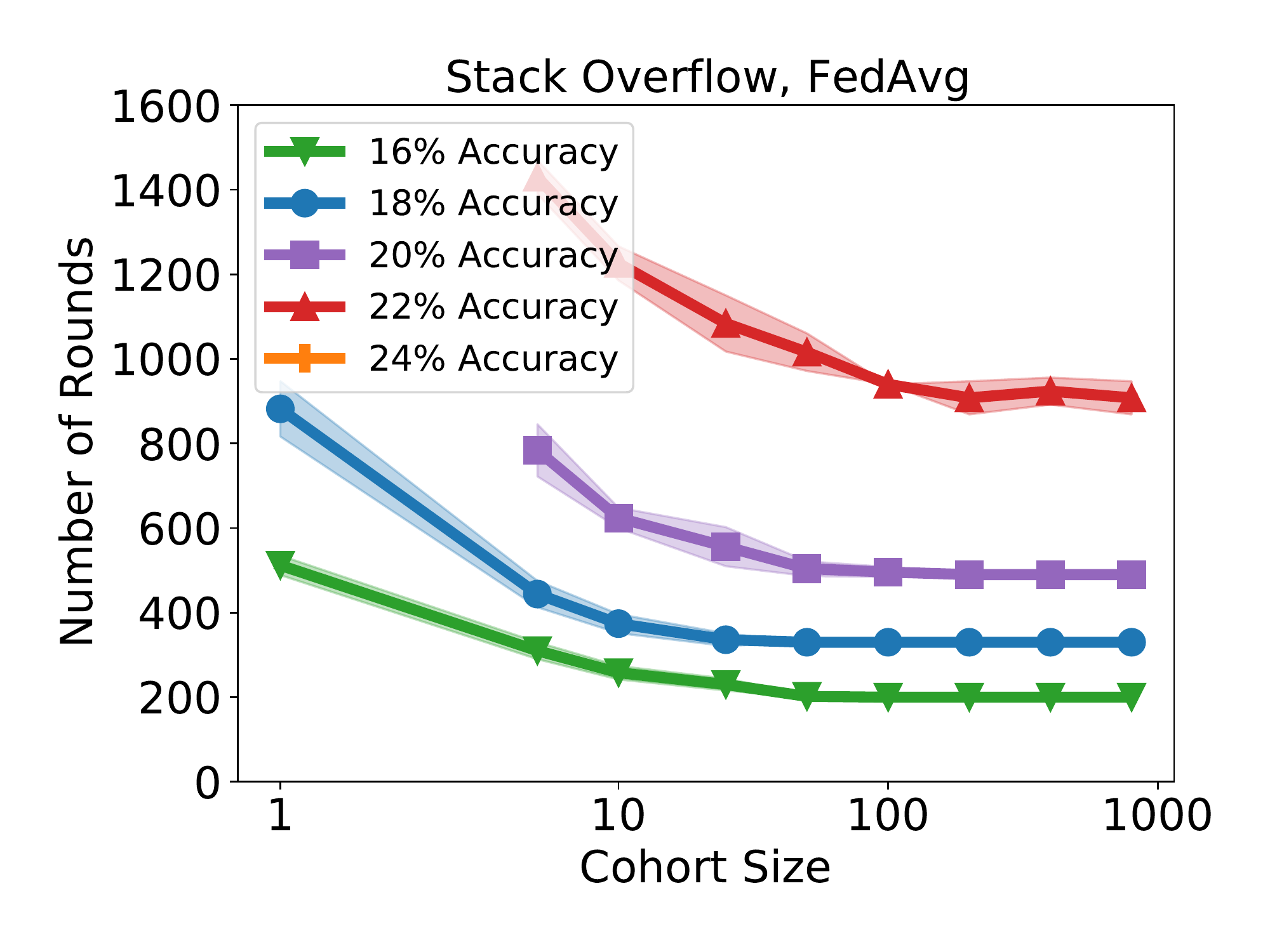}
\end{subfigure}
\caption{Number of communication rounds for \fedavg to obtain certain test accuracy thresholds. The $x$-axis denotes the cohort size.}
\label{fig:main_fedavg_rounds_to_test_accuracy}
\end{figure}

\begin{figure}[ht]
\centering
\begin{subfigure}{0.24\textwidth}
    \centering
    \includegraphics[width=1\linewidth]{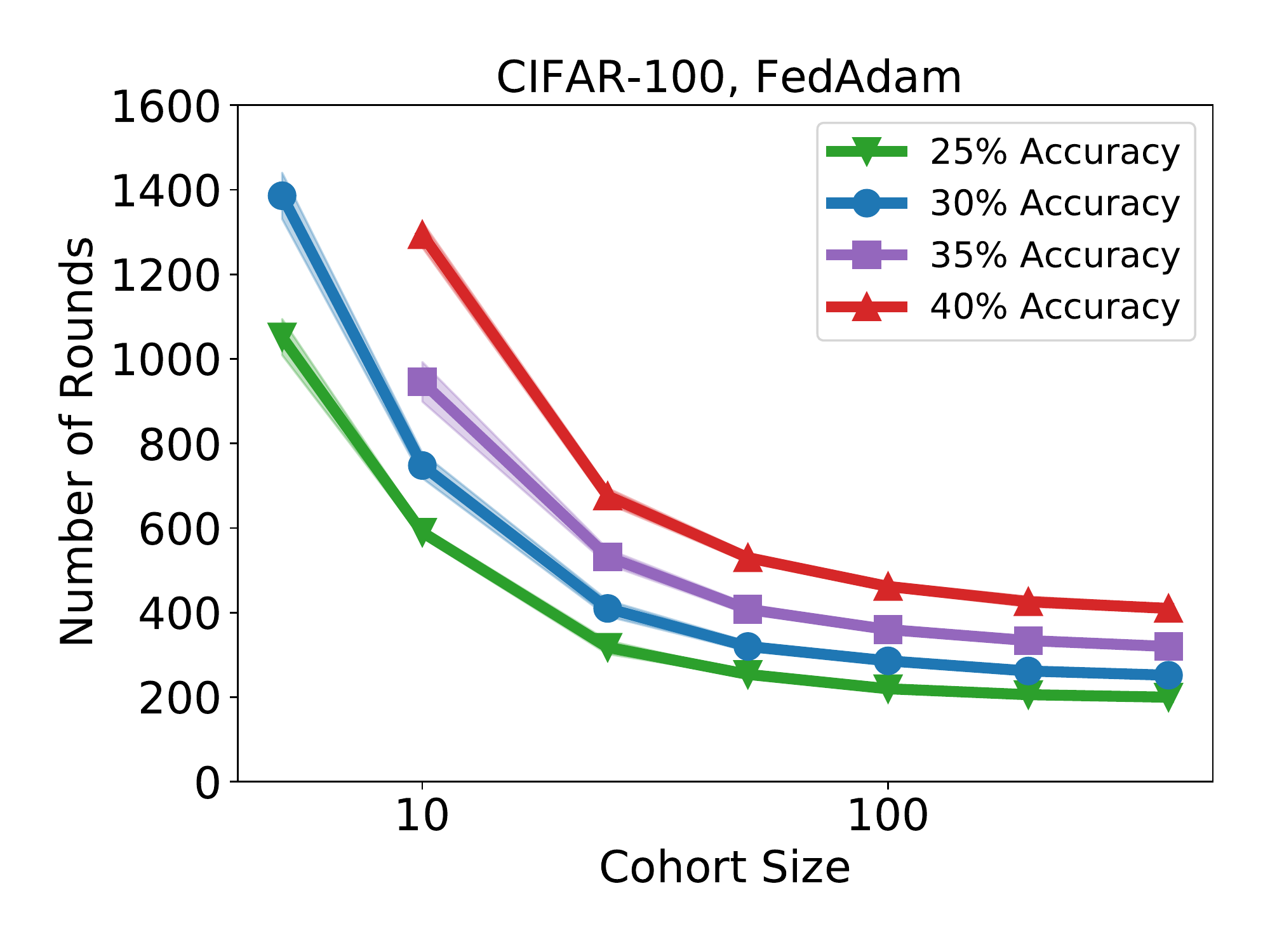}
\end{subfigure}%
\begin{subfigure}{0.24\textwidth}
    \centering
    \includegraphics[width=1\linewidth]{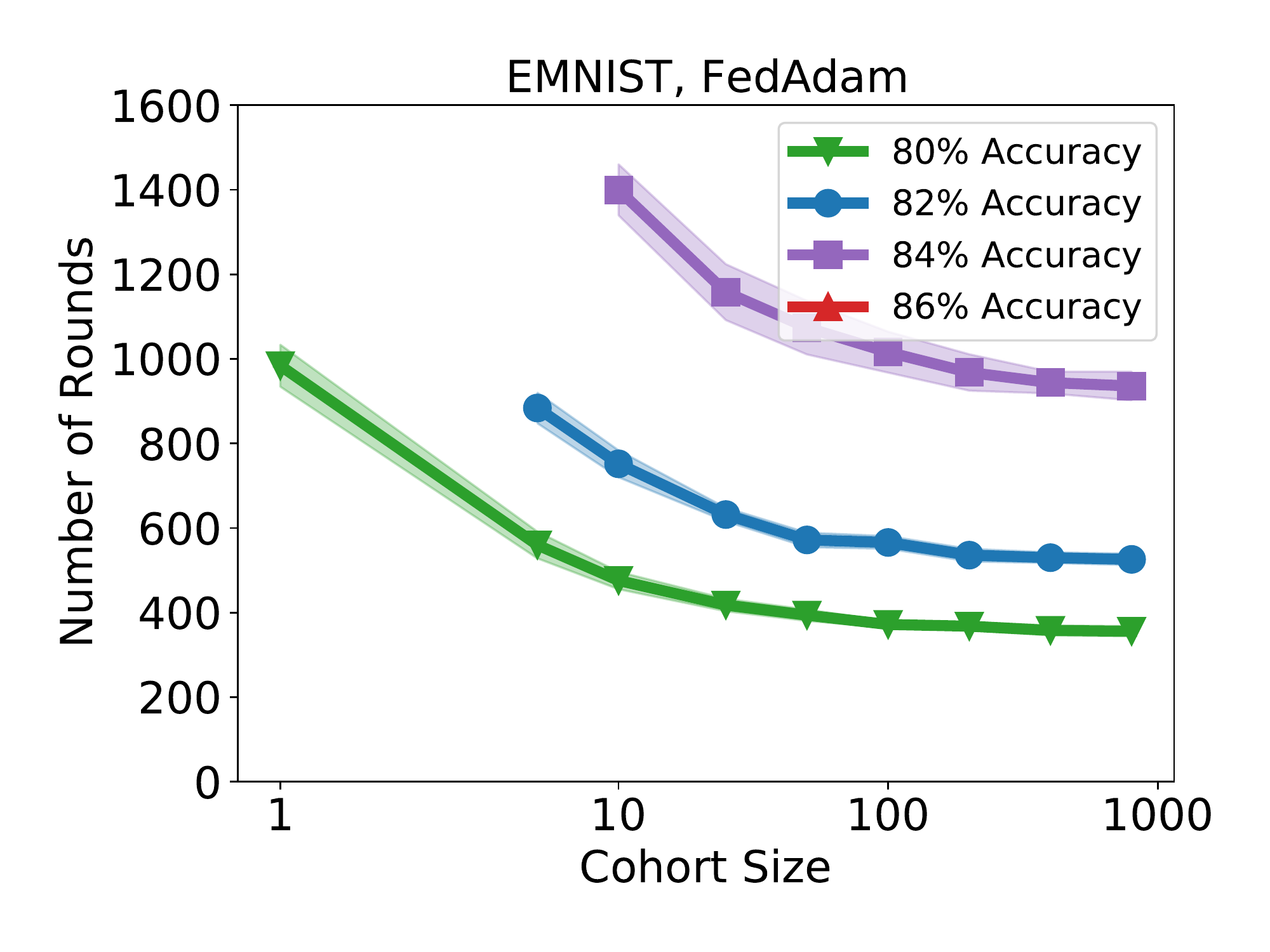}
\end{subfigure}%
\begin{subfigure}{0.24\textwidth}
    \centering
    \includegraphics[width=1\linewidth]{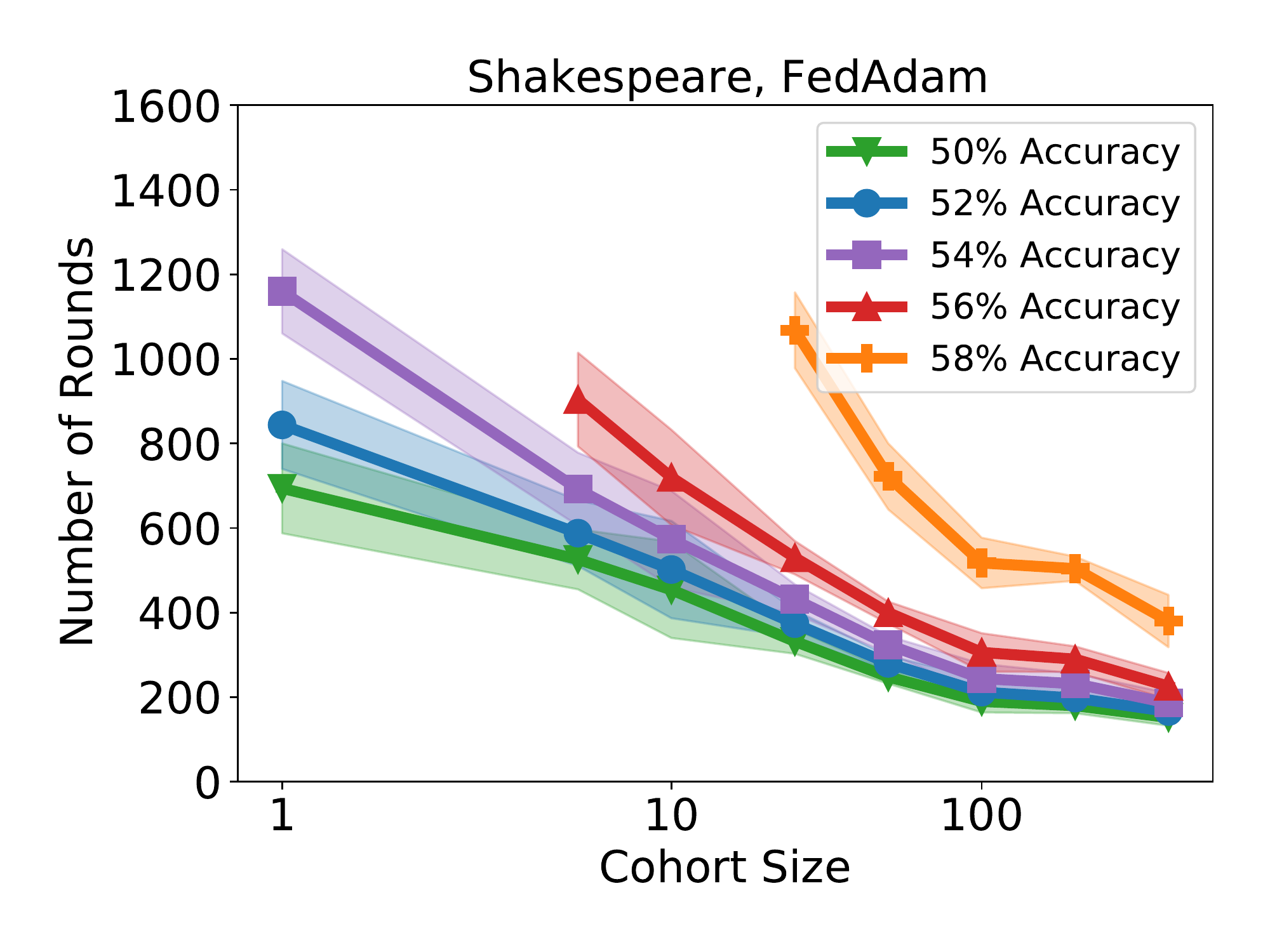}
\end{subfigure}%
\begin{subfigure}{0.24\textwidth}
    \centering
    \includegraphics[width=1\linewidth]{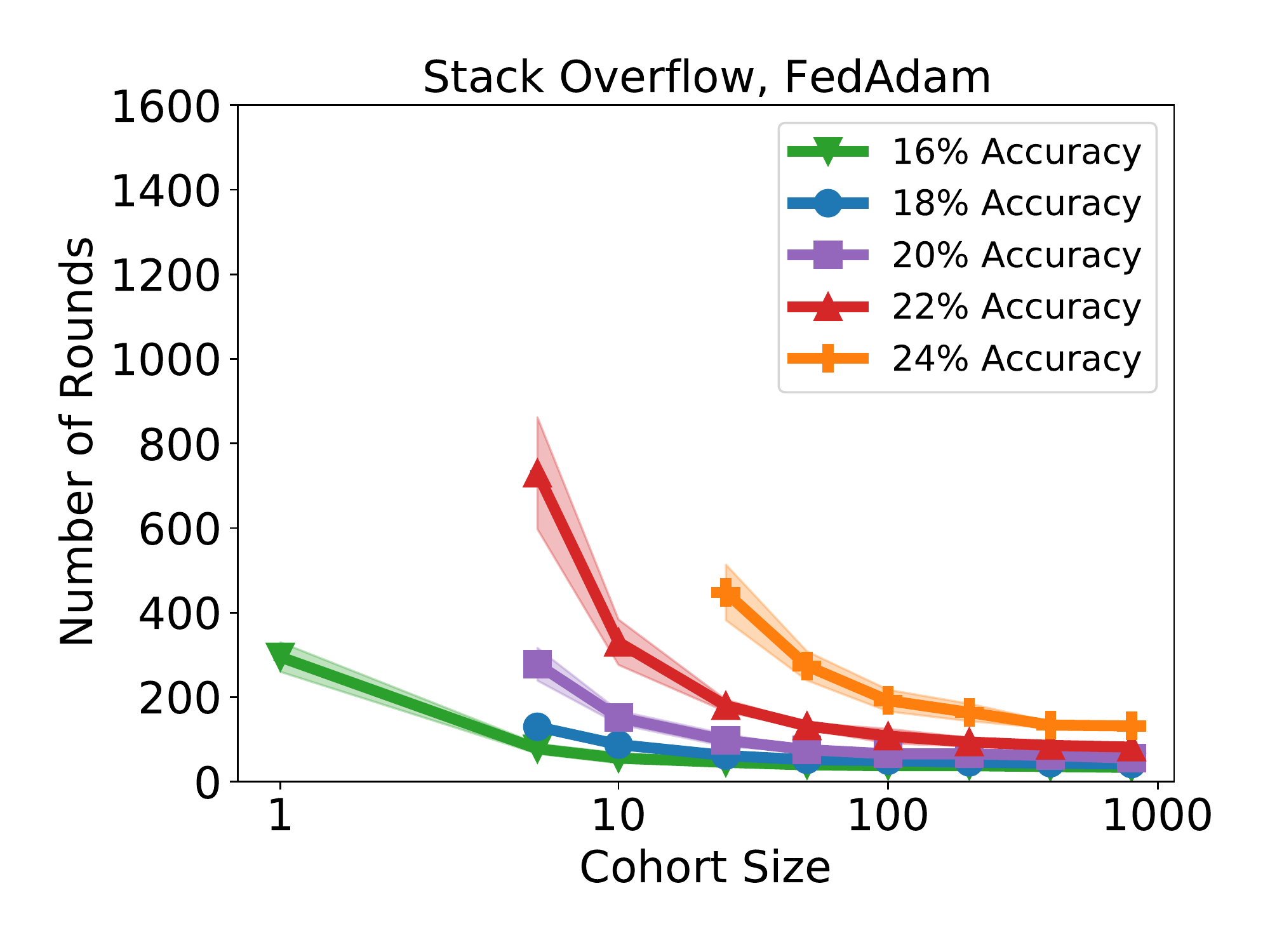}
\end{subfigure}
\caption{Number of communication rounds for \fedadam to obtain certain test accuracy thresholds. The $x$-axis denotes the cohort size.}
\label{fig:main_fedadam_rounds_to_test_accuracy}
\end{figure}

In fact, such issues are not unique to \fedavg and \fedsgd. More generally, we find that for all optimization methods we tried, the benefits of increasing the cohort size do not scale linearly. Instead, they saturate for large enough cohort sizes. To exemplify this, we plot the number of rounds needed by \fedavg and \fedadam to reach certain test accuracy thresholds in Figures \ref{fig:main_fedavg_rounds_to_test_accuracy} and \ref{fig:main_fedadam_rounds_to_test_accuracy}. We see that while we see nearly linear speedups when increasing the cohort size from $M = 1$ to a moderate value, we do not see linear speedups past this point. That being said, these figures demonstrate that large-cohort training does converge faster, and can reach accuracy values unobtainable by small cohort-training in communication-limited settings. For results for other optimizers, see \cref{appendix:cohort_size_speedups}

\subsection{Generalization Failures}\label{subsec:generalization_failures}

\begin{figure}[ht]
\centering
\begin{subfigure}{0.24\textwidth}
    \centering
    \includegraphics[width=1\linewidth]{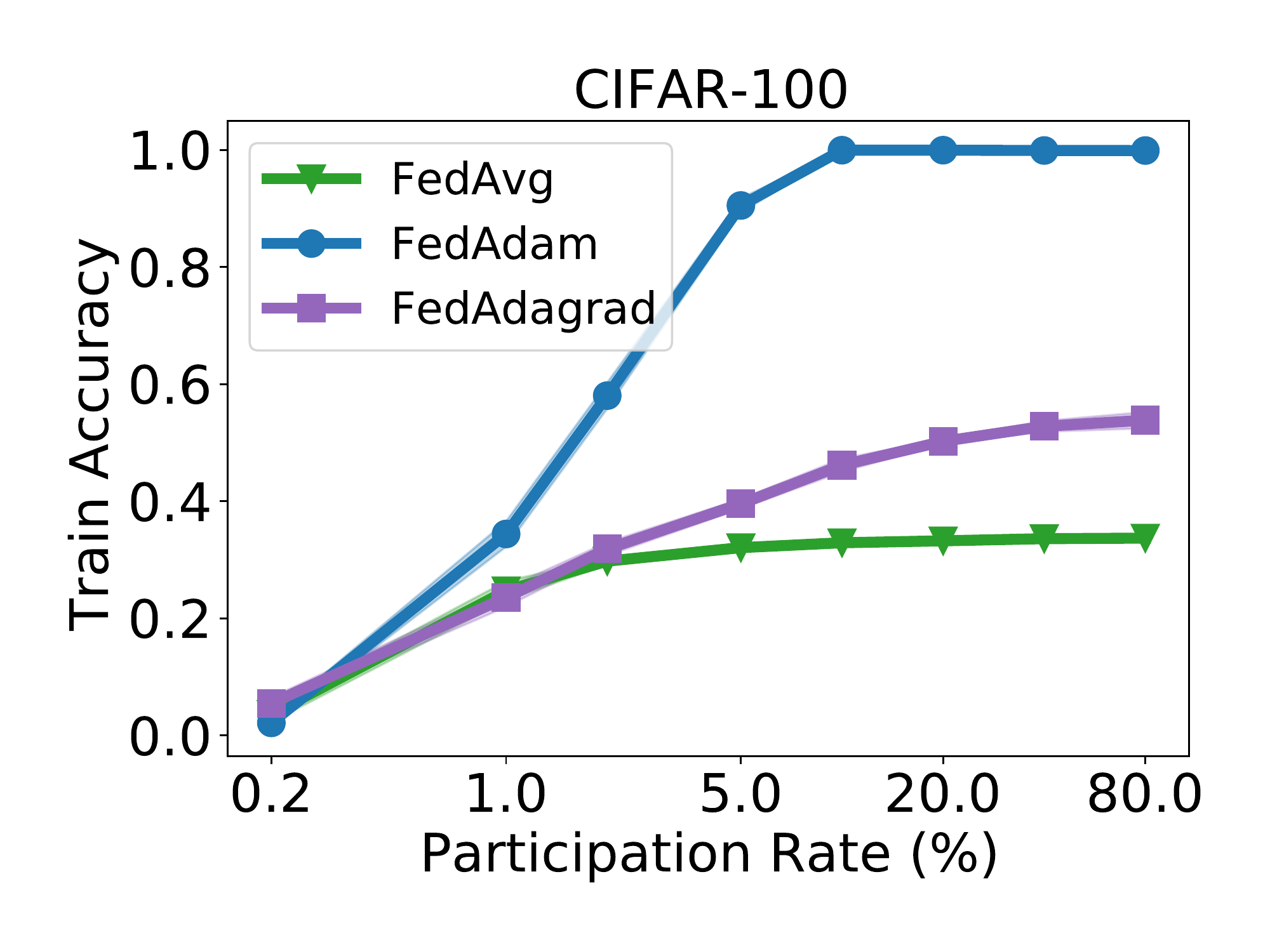}
\end{subfigure}%
\begin{subfigure}{0.24\textwidth}
    \centering
    \includegraphics[width=1\linewidth]{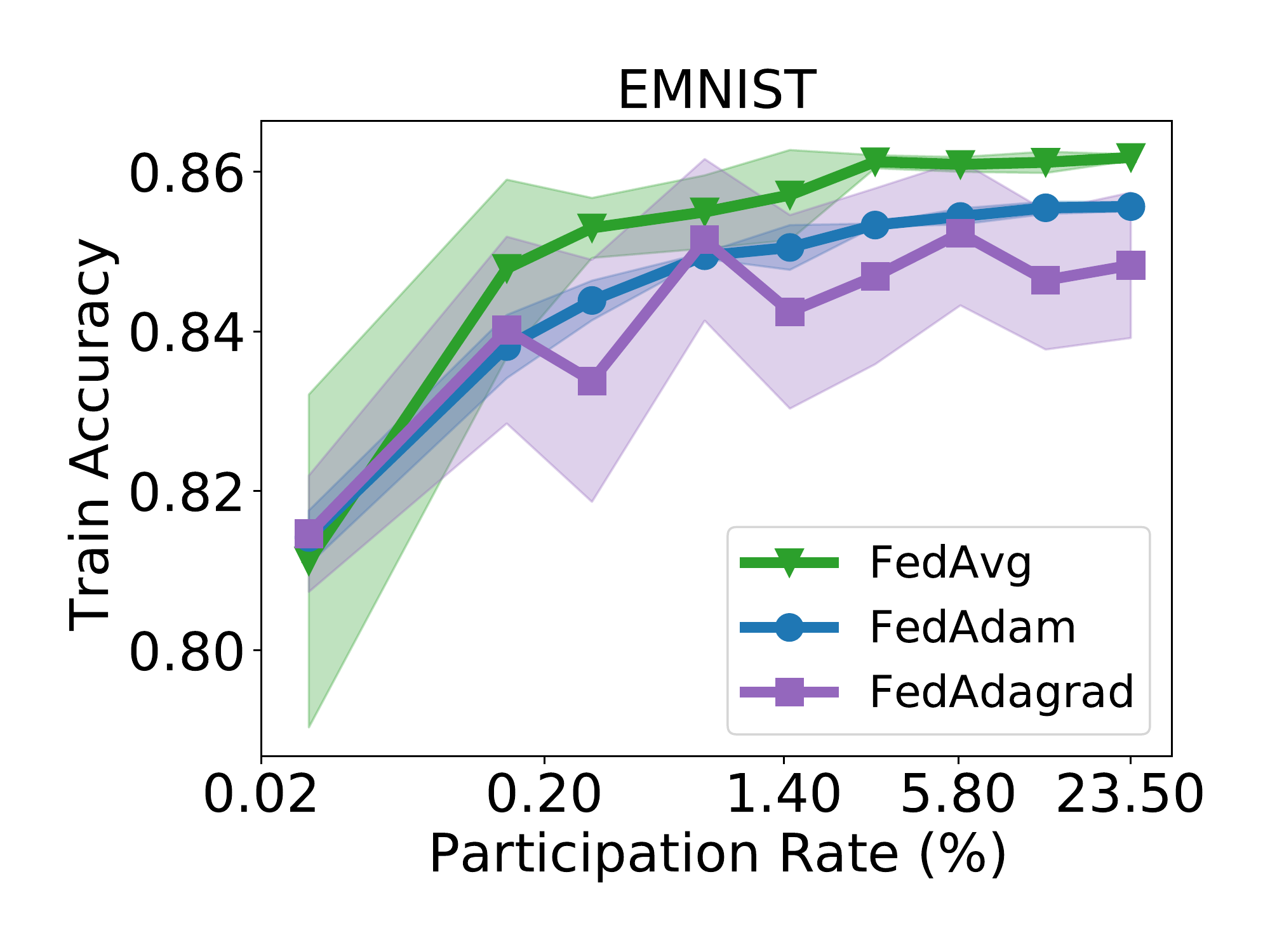}
\end{subfigure}%
\begin{subfigure}{0.24\textwidth}
    \centering
    \includegraphics[width=1\linewidth]{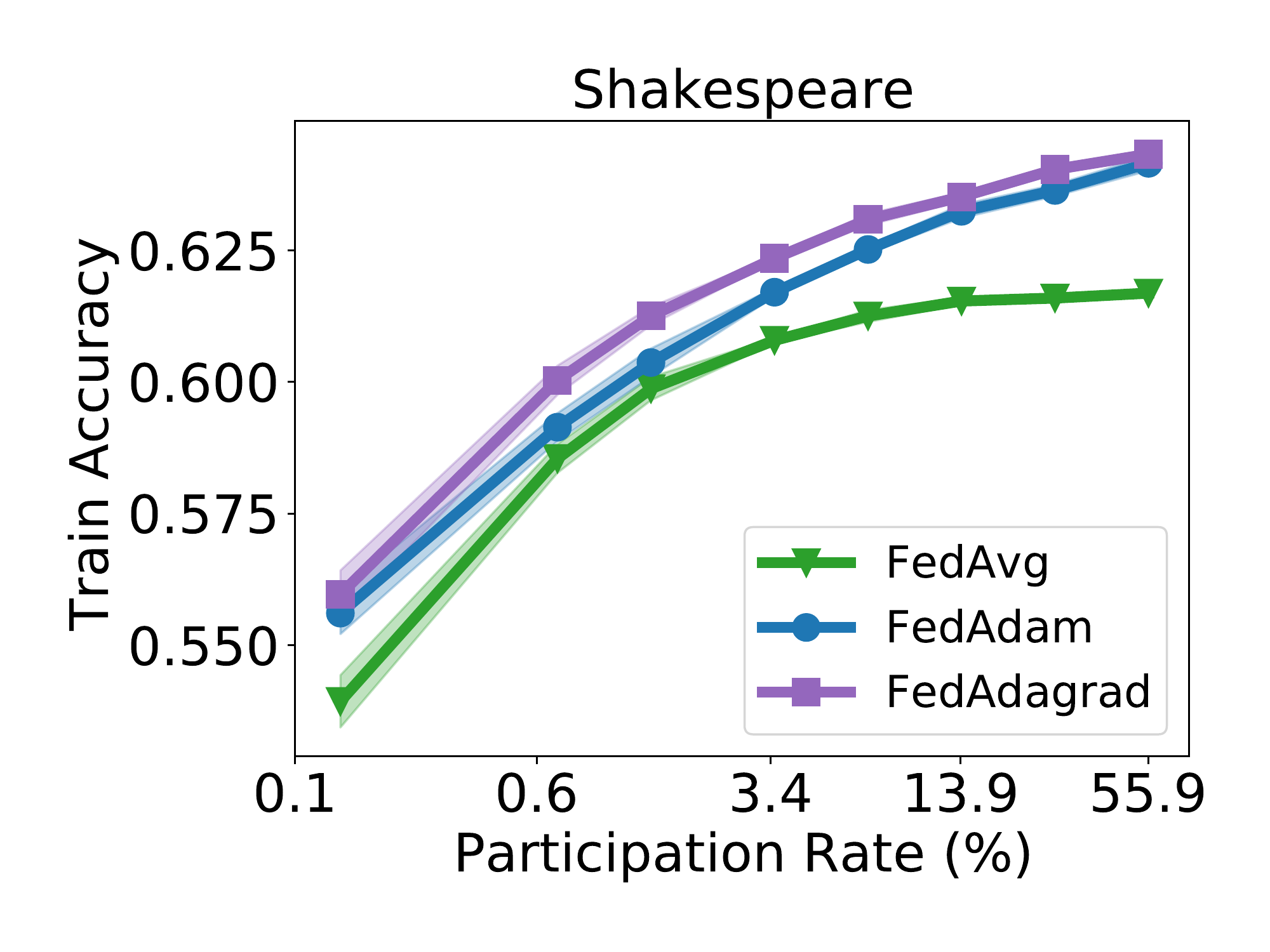}
\end{subfigure}
\begin{subfigure}{0.24\textwidth}
    \centering
    \includegraphics[width=1\linewidth]{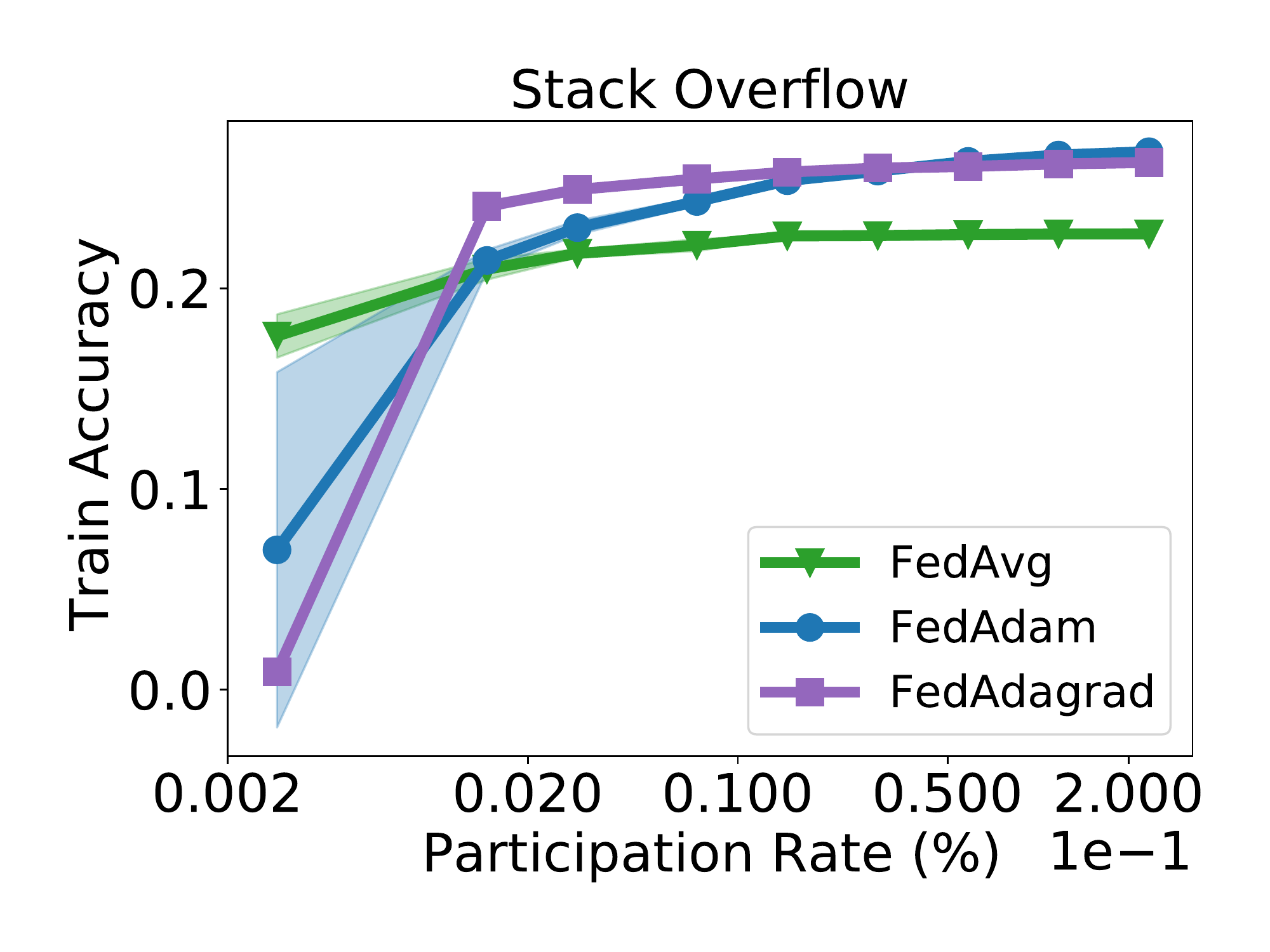}
\end{subfigure}
\begin{subfigure}{0.24\textwidth}
    \centering
    \includegraphics[width=1\linewidth]{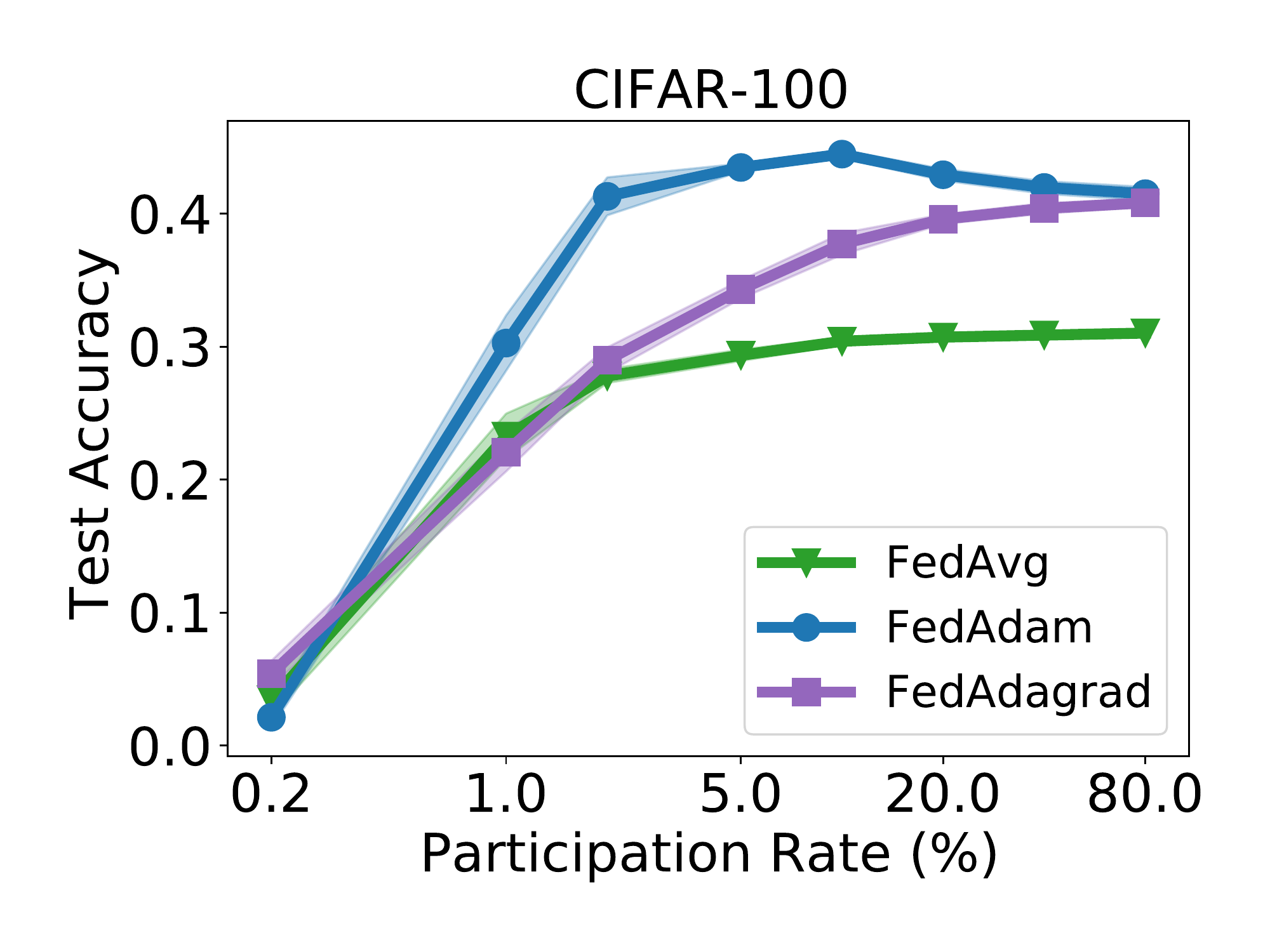}
\end{subfigure}%
\begin{subfigure}{0.24\textwidth}
    \centering
    \includegraphics[width=1\linewidth]{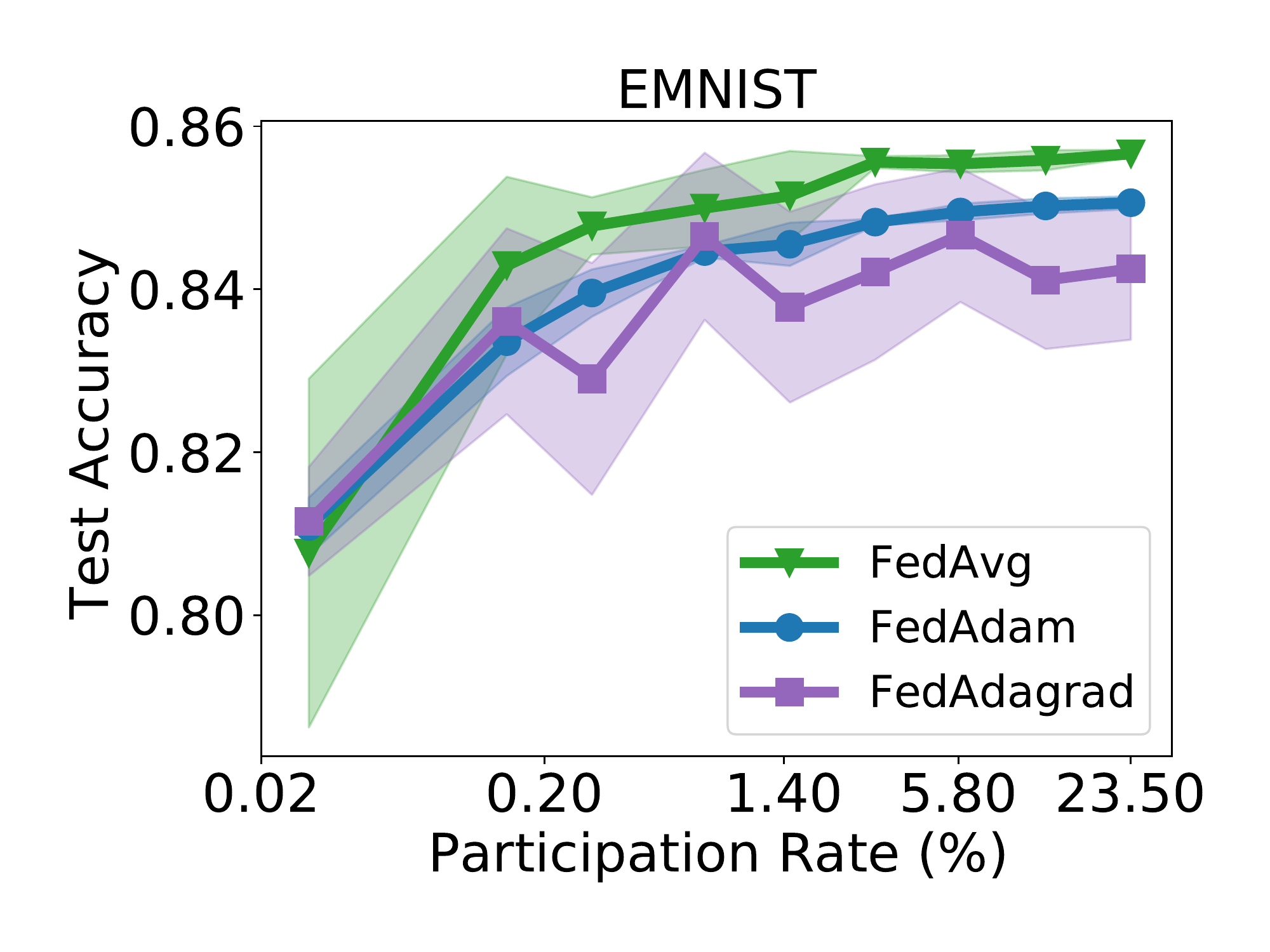}
\end{subfigure}%
\begin{subfigure}{0.24\textwidth}
    \centering
    \includegraphics[width=1\linewidth]{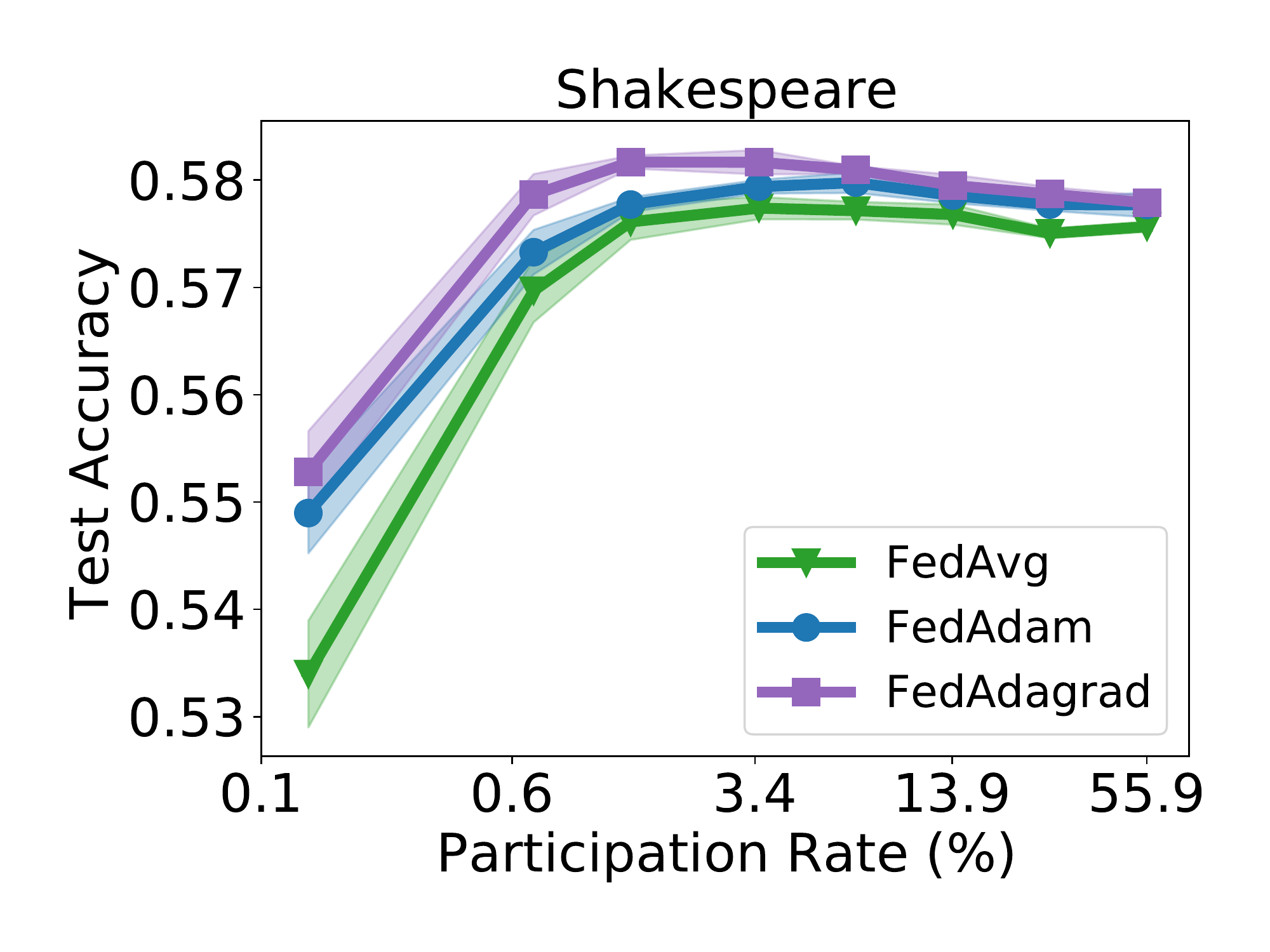}
\end{subfigure}
\begin{subfigure}{0.24\textwidth}
    \centering
    \includegraphics[width=1\linewidth]{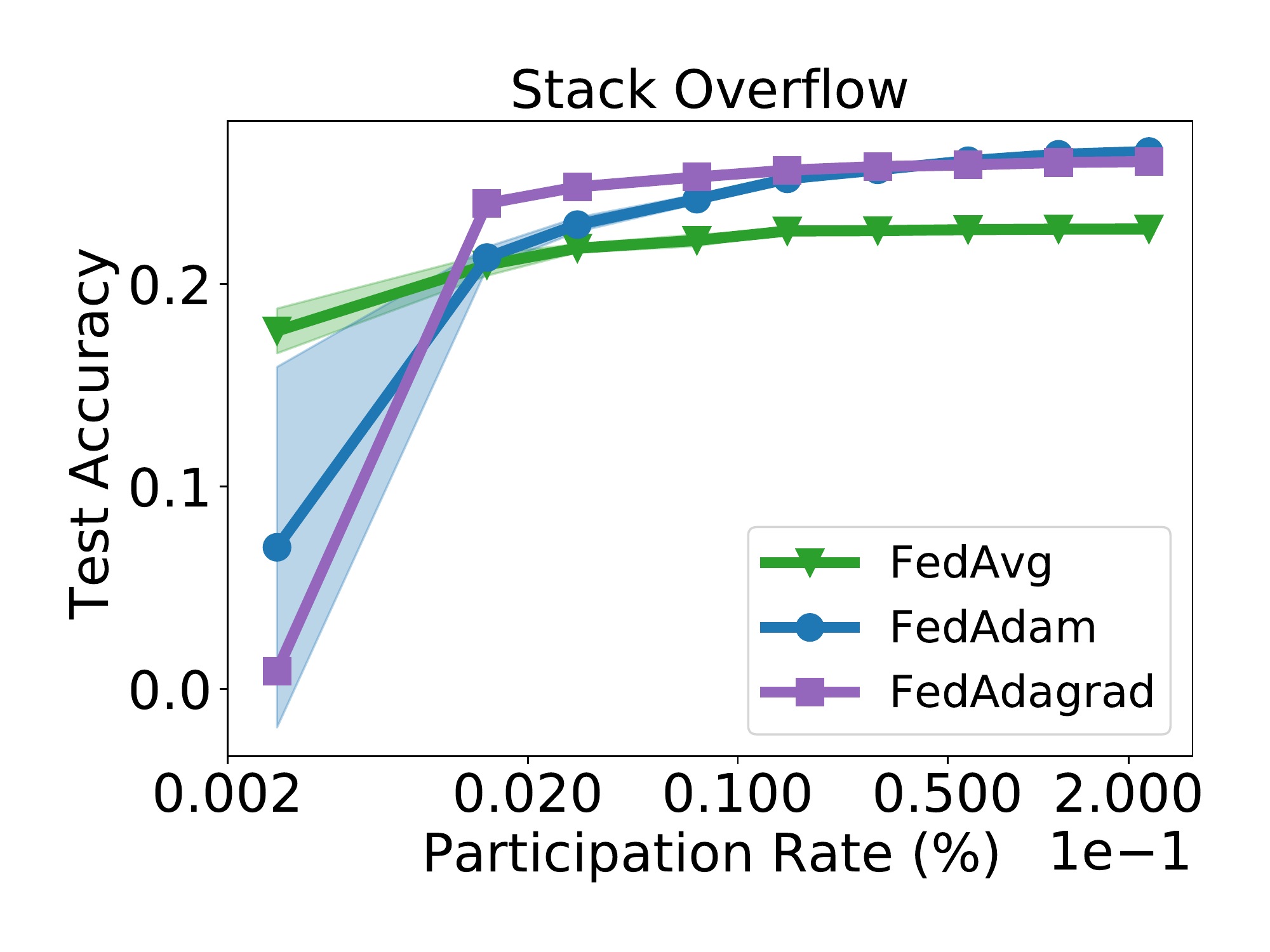}
\end{subfigure}
\caption{The train accuracy (top) and test accuracy (bottom) of \fedavg, \fedadam, and \fedadagrad on various tasks after training for 1500 communication rounds, for varying cohort sizes. The $x$-axis denotes the percentage of training clients in each cohort.}
\label{fig:main_accuracy_versus_cohort_size}
\end{figure}

Large-batch centralized optimization methods have repeatedly been shown to converge to models with worse generalization ability than models found by small-batch methods~\citep{keskar2017iclr,hoffer2017train,you2017large,masters2018revisiting,lin2019don,lin2020extrapolation}. Given the parallels between batch size in centralized learning and cohort size in FL, this raises obvious questions about whether similar issues occur in FL. In order to test this, we applied \fedavg, \fedadam, and \fedadagrad with different cohort sizes to various models. In \cref{fig:main_accuracy_versus_cohort_size} we plot the train and test accuracy of our models after $T = 1500$ communication rounds of \fedavg, \fedadam, and \fedadagrad.

We find that generalization issues do occur in FL. For example, consider \fedadam on the CIFAR-100 task. While it attains roughly the same training accuracy for $M \in \{50, 100, 200, 400\}$, we see that the larger cohorts uniformly lead to worse generalization. This resembles the findings of \citet{keskar2017iclr}, who show that generalization issues of large-batch training can occur even though the methods reach similar training losses. However, generalization issues do not occur uniformly. It is often optimizer-dependent (as in CIFAR-100) and does not occur on the EMNIST and Stack Overflow datasets. Notably, CIFAR-100 and Shakespeare have many fewer clients overall. Thus, large-cohort training may reduce generalization, especially when the cohort size is large compared to the total number of clients.

\subsection{Fairness Concerns} 
\begin{figure}[ht]
\centering
\begin{subfigure}{0.24\textwidth}
    \centering
    \includegraphics[width=1\linewidth]{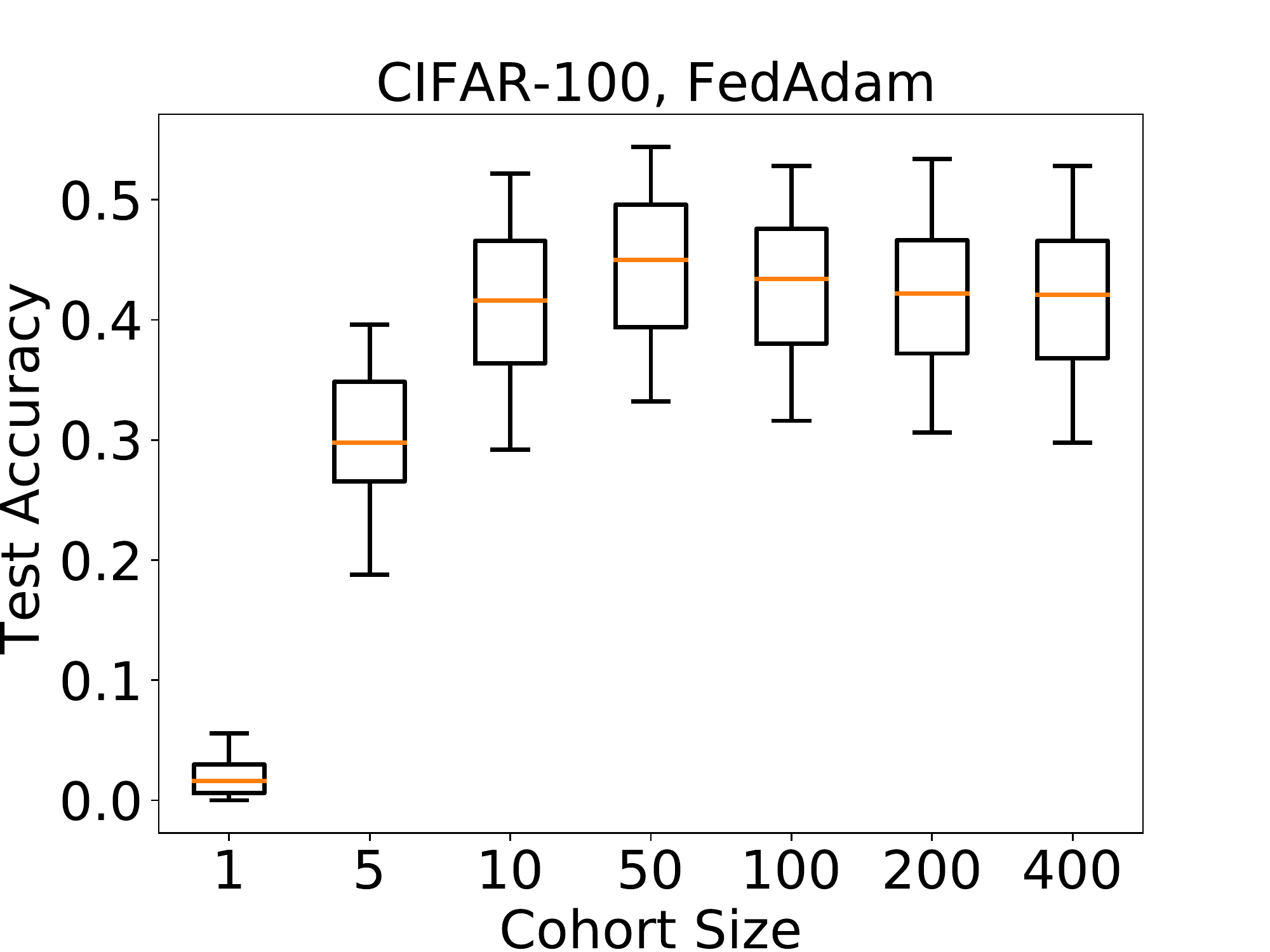}
\end{subfigure}%
\begin{subfigure}{0.24\textwidth}
    \centering
    \includegraphics[width=1\linewidth]{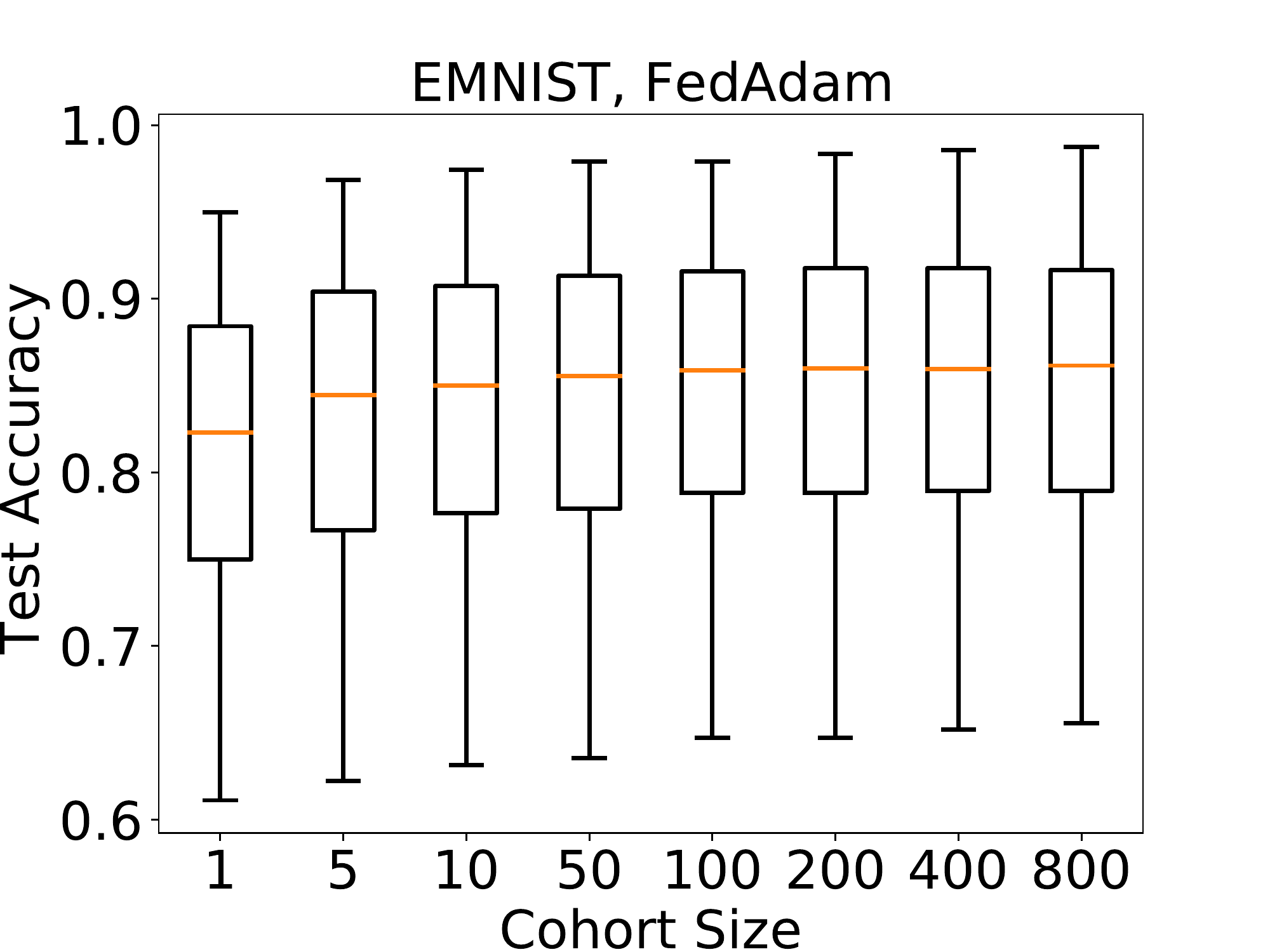}
\end{subfigure}%
\begin{subfigure}{0.24\textwidth}
    \centering
    \includegraphics[width=1\linewidth]{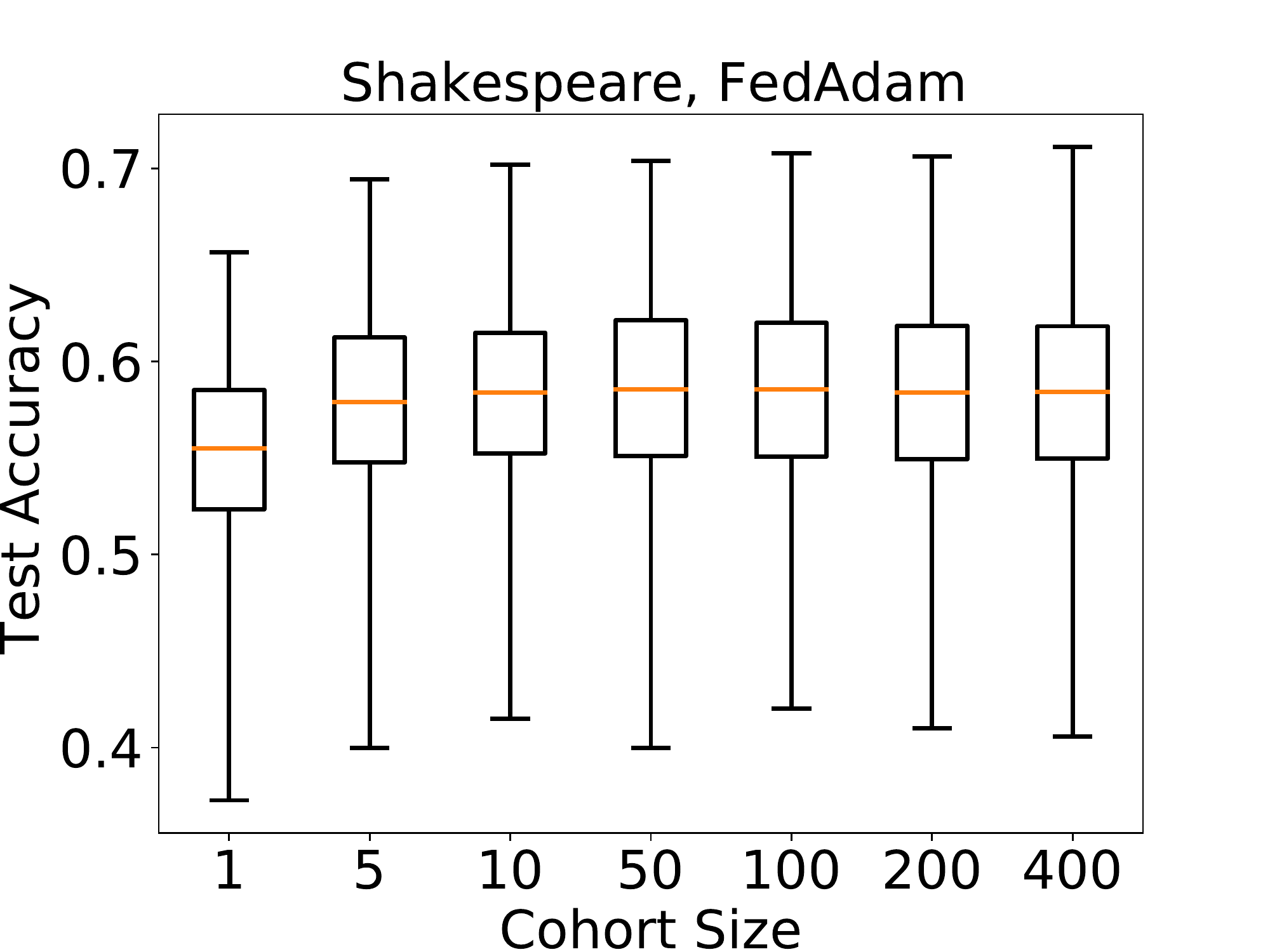}
\end{subfigure}%
\begin{subfigure}{0.24\textwidth}
    \centering
    \includegraphics[width=1\linewidth]{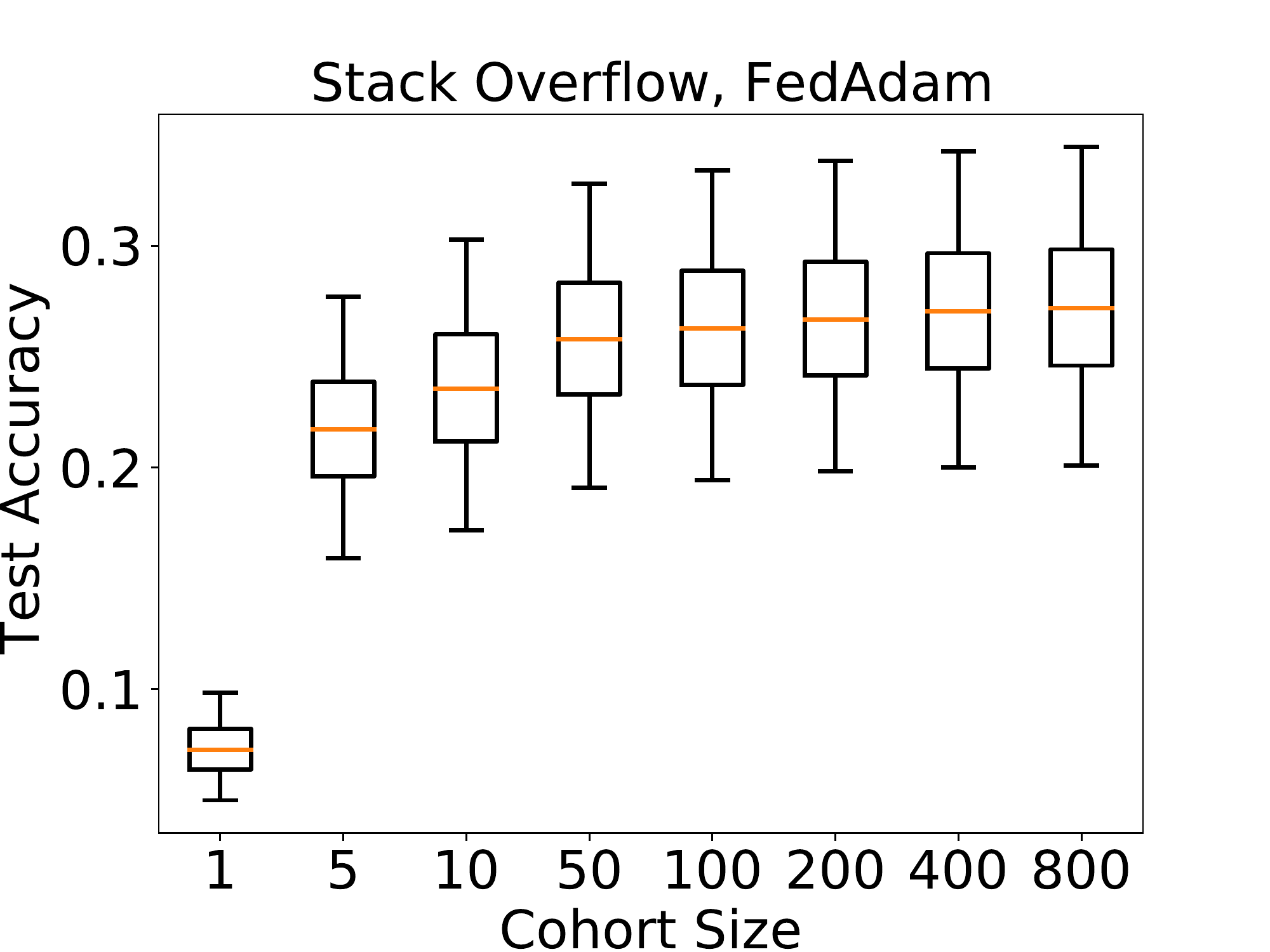}
\end{subfigure}%
\caption{Accuracy of \fedadam after training for 1500 communication rounds using varying cohort sizes and tasks. The box plots show the 5th, 25th, 50th, 75th, and 95th percentiles of accuracy across test clients.}
\label{fig:client_accuracies}
\end{figure}

One critical issue in FL is fairness across clients, as minimizing \eqref{eq:objective_fn} may disadvantage some clients~\citep{mohri2019agnostic,li2019fair}. Intuitively, large-cohort training methods may be better suited for ensuring fairness, since a greater fraction of the population is allowed to contribute to the model at each round. As a coarse measure of fairness, we compute percentiles of accuracy of our trained models across test clients. Fairer algorithms should lead to higher accuracy values for smaller percentiles. The percentiles for \fedadam on each task are given in \cref{fig:client_accuracies}.

We find that the cohort size seems to affect all percentiles in the same manner. For example, on CIFAR-100, $M = 50$ performs better for smaller percentiles and larger percentiles than larger $M$. This mirrors the CIFAR-100 generalization failures from \cref{subsec:generalization_failures}. By contrast, for Stack Overflow we see increases in all percentiles as we increase $M$. While the accuracy gains are only slight, they are consistent across percentiles. This suggests a connection between the fairness of a federated training algorithm and the fraction of test clients participating at every round. Notably, increasing $M$ seems to have little effect on the spread between percentiles (such as the difference between the 75th and 25th percentiles) beyond a certain point. See \cref{appendix:fairness} for more results.

\subsection{Decreased Data Efficiency}\label{sec:data_efficiency}

\begin{figure}[ht]
\centering
\begin{subfigure}{0.24\textwidth}
    \centering
    \includegraphics[width=1\linewidth]{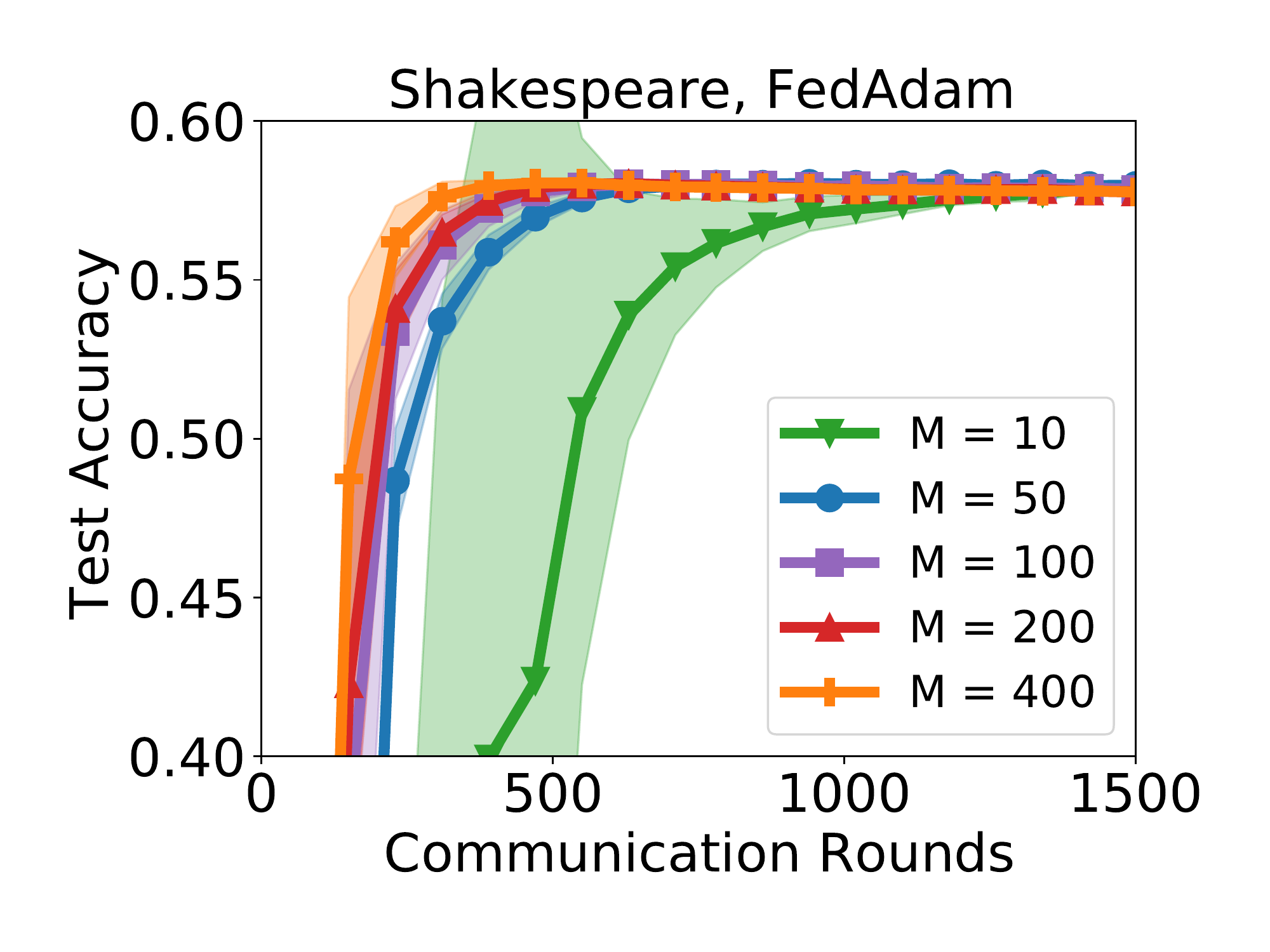}
\end{subfigure}%
\begin{subfigure}{0.24\textwidth}
    \centering
    \includegraphics[width=1\linewidth]{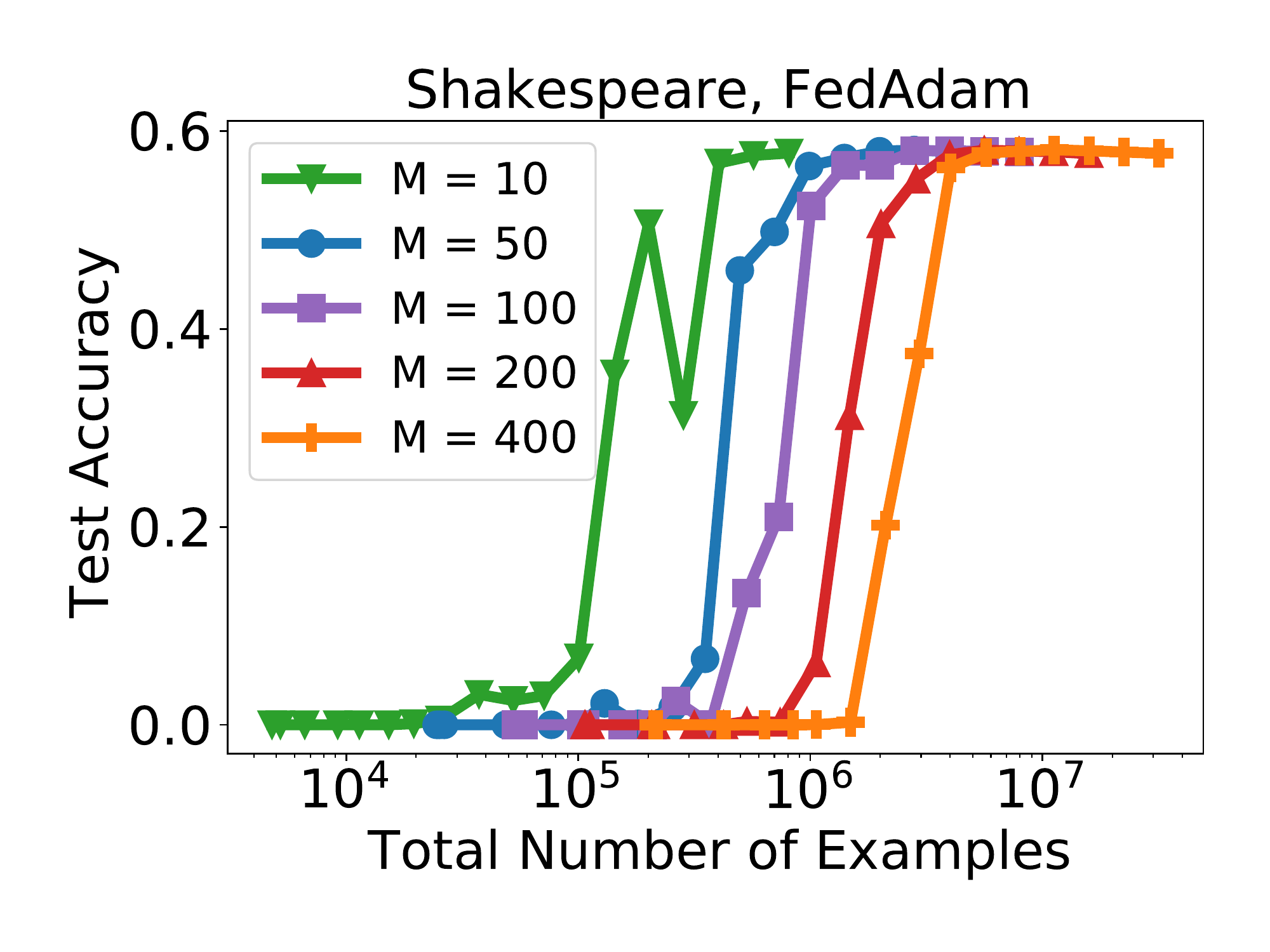}
\end{subfigure}%
\begin{subfigure}{0.24\textwidth}
    \centering
    \includegraphics[width=1\linewidth]{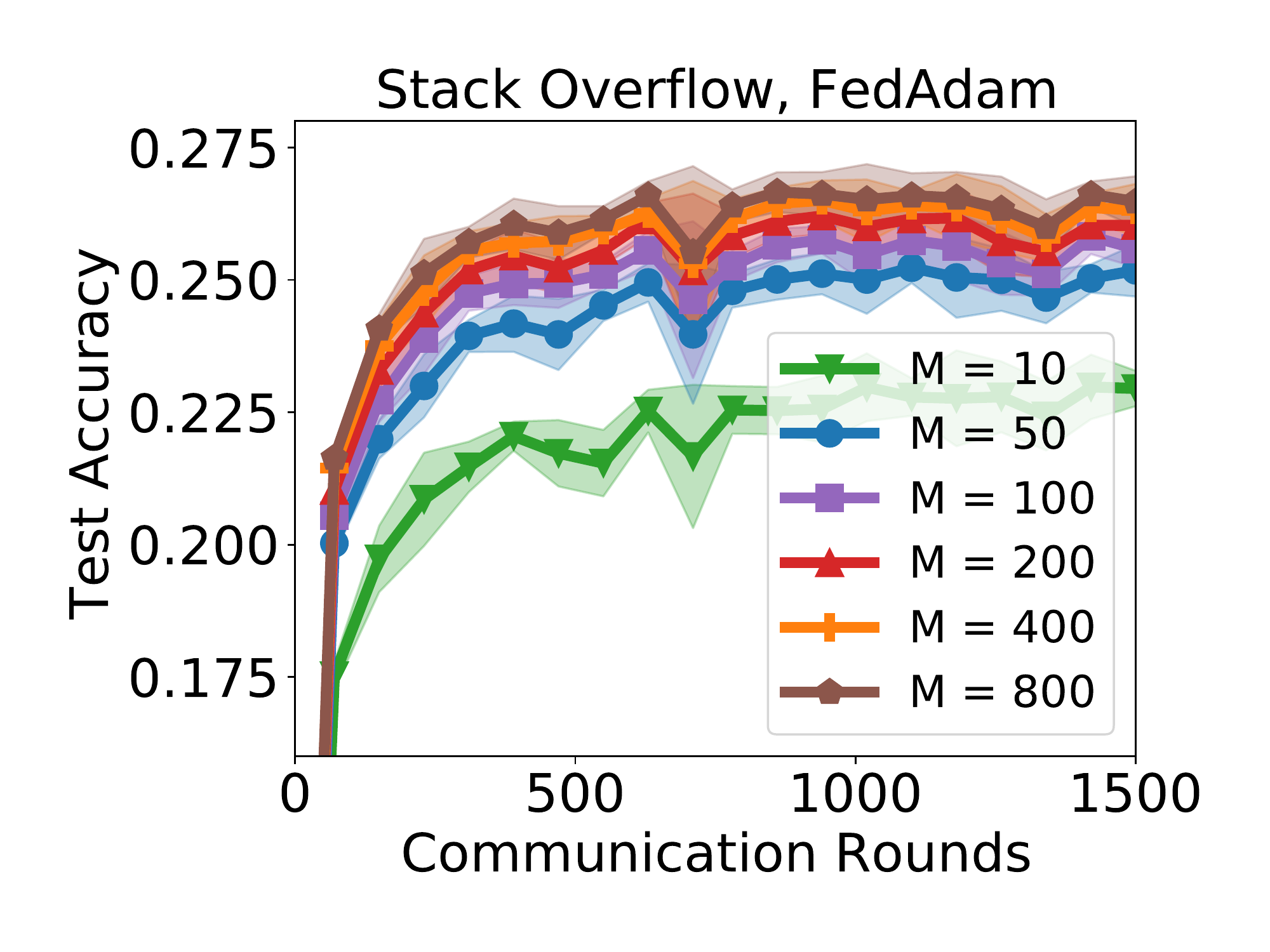}
\end{subfigure}%
\begin{subfigure}{0.24\textwidth}
    \centering
    \includegraphics[width=1\linewidth]{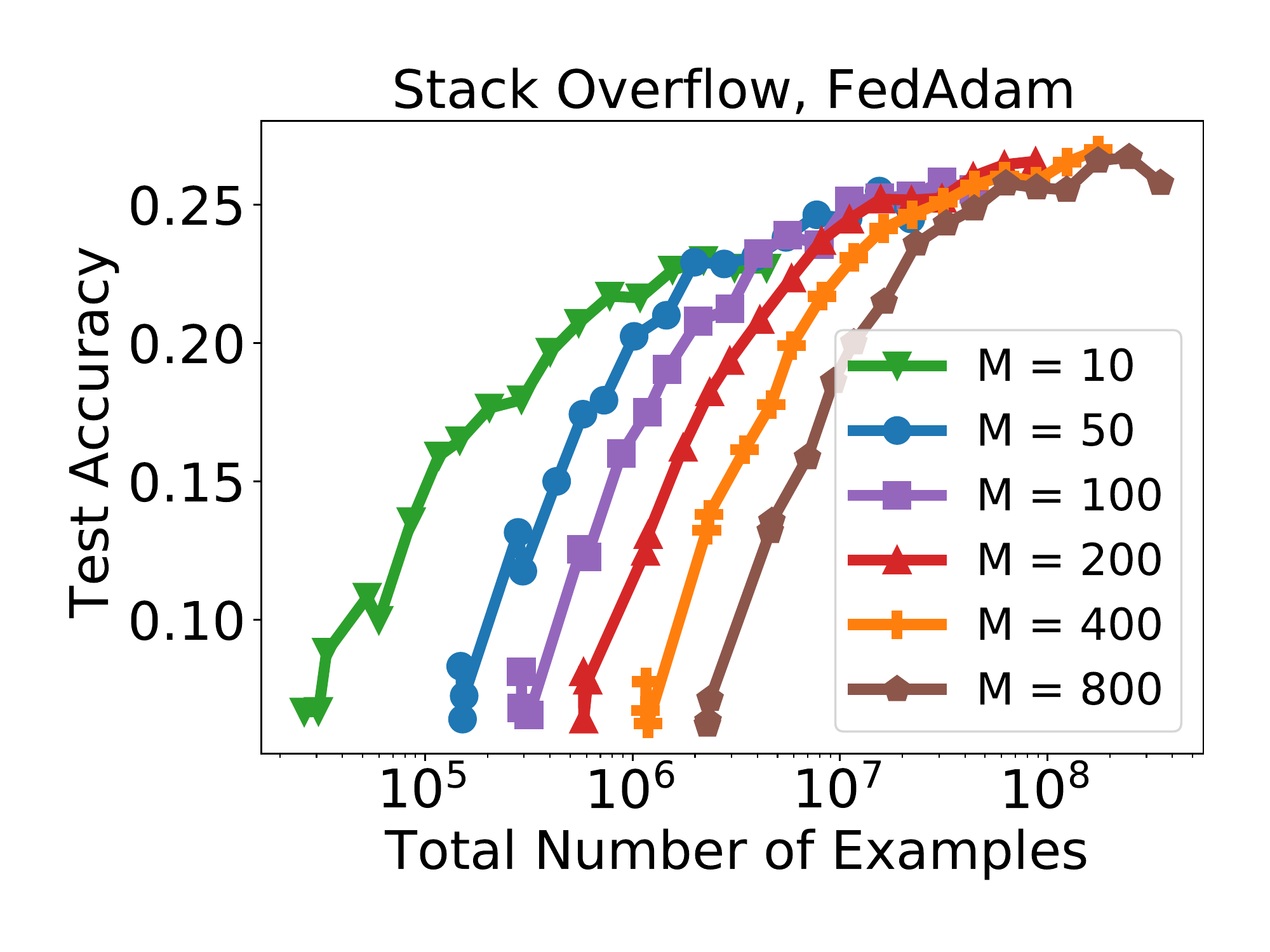}
\end{subfigure}%
\caption{Test accuracy of \fedadam on Shakespeare (left) and Stack Overflow (right) with various cohort sizes. We plot versus the number of communication rounds and the number of examples processed in total.}
\label{fig:accuracy_per_example}
\end{figure}

Despite issues such as diminishing returns and generalization failures, federated optimization methods can see some benefit from larger cohorts. Large-cohort training, especially with adaptive optimizers, often leads to faster convergence to given accuracy thresholds. For example, in \cref{fig:accuracy_per_example}, we see that the number of rounds \fedadam requires to reach certain accuracy thresholds generally decreases with the cohort size.

While it is tempting to say that large-cohort methods are ``faster'', this ignores the practical costs of large-cohort training. Completing a single communication round often requires more resources with larger cohorts. To showcase this, we also plot the accuracy of \fedadam with respect to the number of examples seen in \cref{fig:accuracy_per_example}. This measures the data-efficiency of large-cohort training, and shows that large cohort-training requires significantly more examples per unit-accuracy.

While data-inefficiency also occurs in large-batch training~\citep{mccandlish2018empirical}, it is especially important in federated learning. Large-cohort training faces greater limitations on parallelizability due to data-sharing constraints. Worse, in realistic cross-device settings client compute times can scale super-linearly with their amount of data, so clients with more data are more likely to become \emph{stragglers}~\citep{bonawitz2019towards}. This straggler effect means that data-inefficient algorithms may require longer training times. To demonstrate this, we show in \cref{appendix:stragglers} that under the probabilistic straggler runtime model from \citep{lee2017speeding}, large-cohort training can require significantly more compute time to converge.

\section{Diagnosing Large-Cohort Challenges}\label{sec:diagnosis}

We now examine the challenges in \cref{sec:challenges}, and provide partial explanations for their occurrence. One of the key differences between \fedavg and \fedsgd is what the pseudo-gradient $\Delta$ in \eqref{eq:objective_fn} represents. In \fedsgd, $\Delta$ is a stochastic gradient estimate (\ie $\E[\Delta] = \nabla f$, where the expectation is over all randomness in a given communication round). For special cases of \cref{alg:fedopt} where clients perform multiple local training steps, $\Delta$ is not an unbiased estimator of $\nabla f$~\citep{malinovsky2020local, pathak2020fedsplit, charles2021convergence}. While increasing the cohort size should reduce the variance of $\Delta$ as an estimator of $\E[\Delta]$, it is unclear what this quantity represents.

\begin{figure}[ht]
\centering
\begin{subfigure}{0.24\textwidth}
    \centering
    \includegraphics[width=1\linewidth]{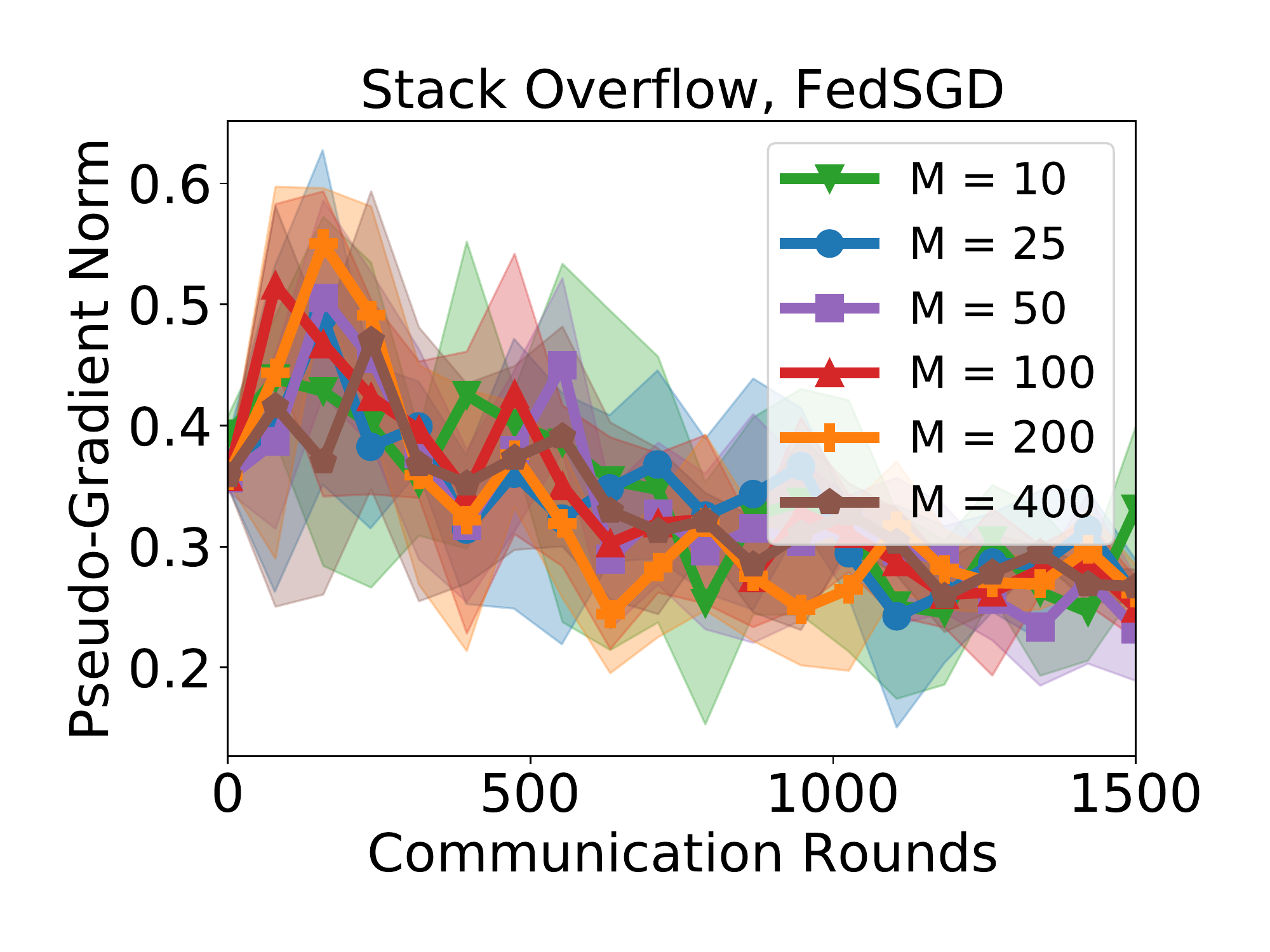}
    \caption{}
    \label{fig:pseudo_gradient_norm_a}
\end{subfigure}%
\begin{subfigure}{0.24\textwidth}
    \centering
    \includegraphics[width=1\linewidth]{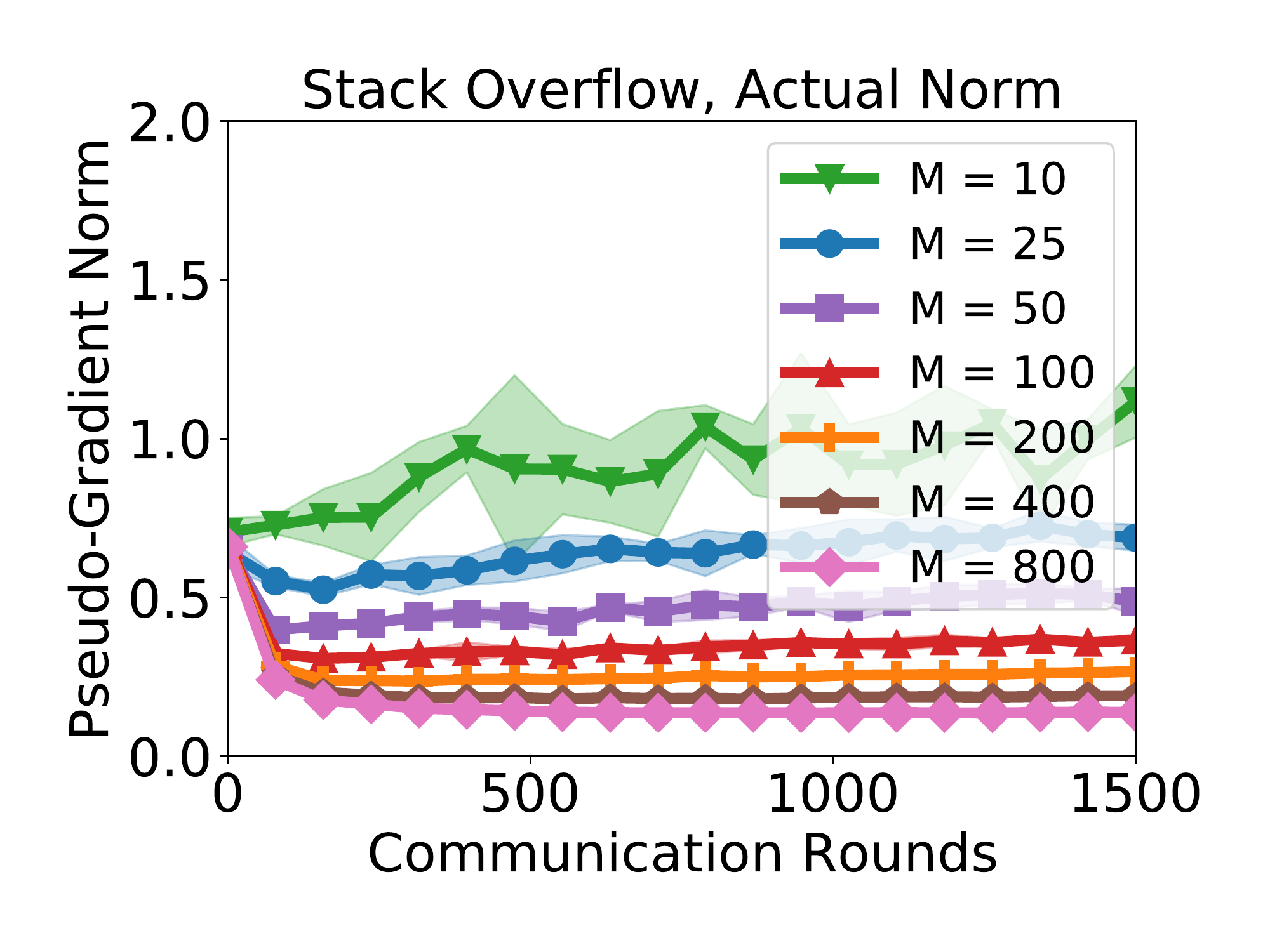}
    \caption{}
    \label{fig:pseudo_gradient_norm_b}
\end{subfigure}%
\begin{subfigure}{0.24\textwidth}
    \centering
    \includegraphics[width=1\linewidth]{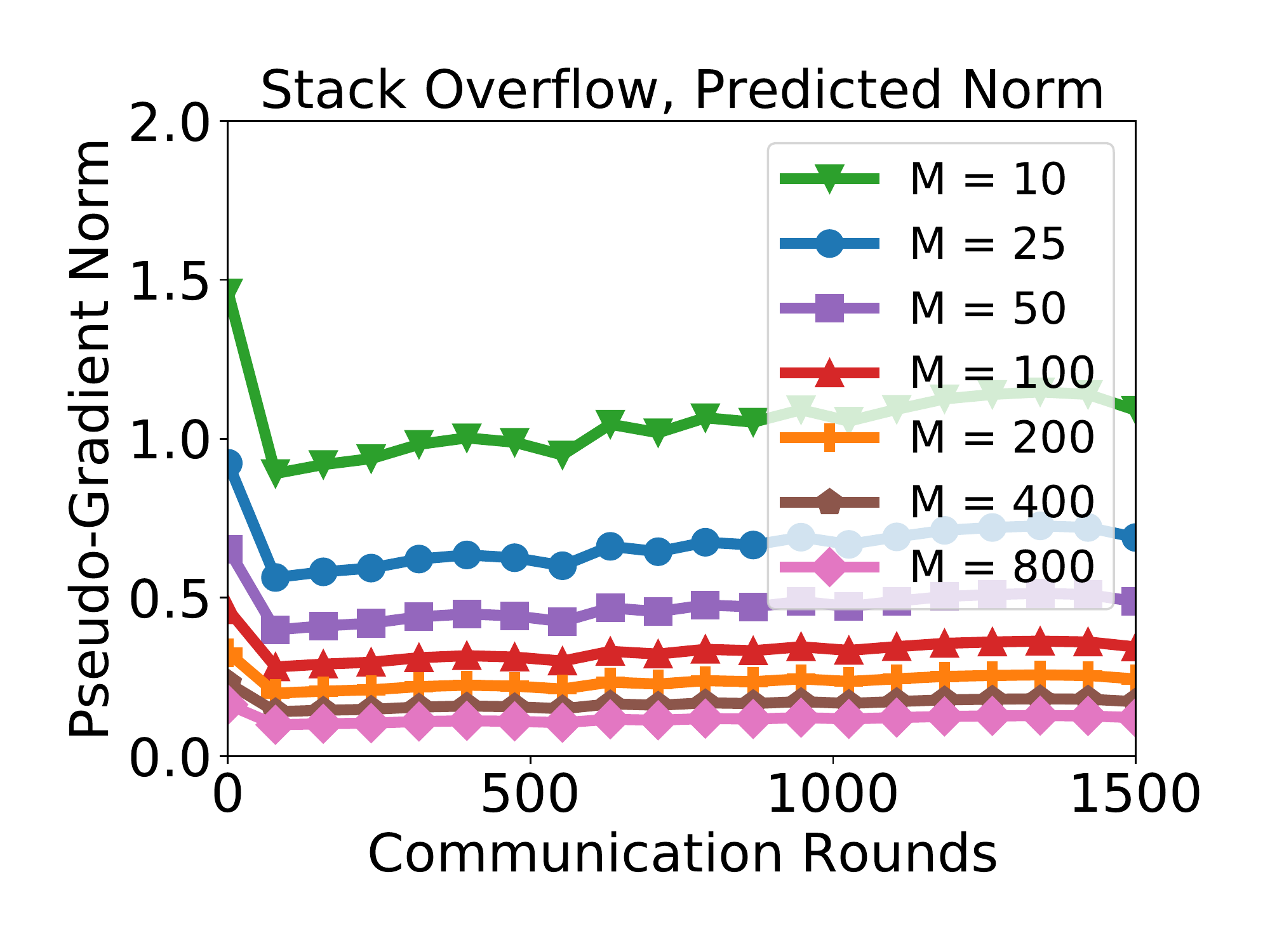}
    \caption{}
    \label{fig:pseudo_gradient_norm_c}
\end{subfigure}%
\begin{subfigure}{0.24\textwidth}
    \centering
    \includegraphics[width=1\linewidth]{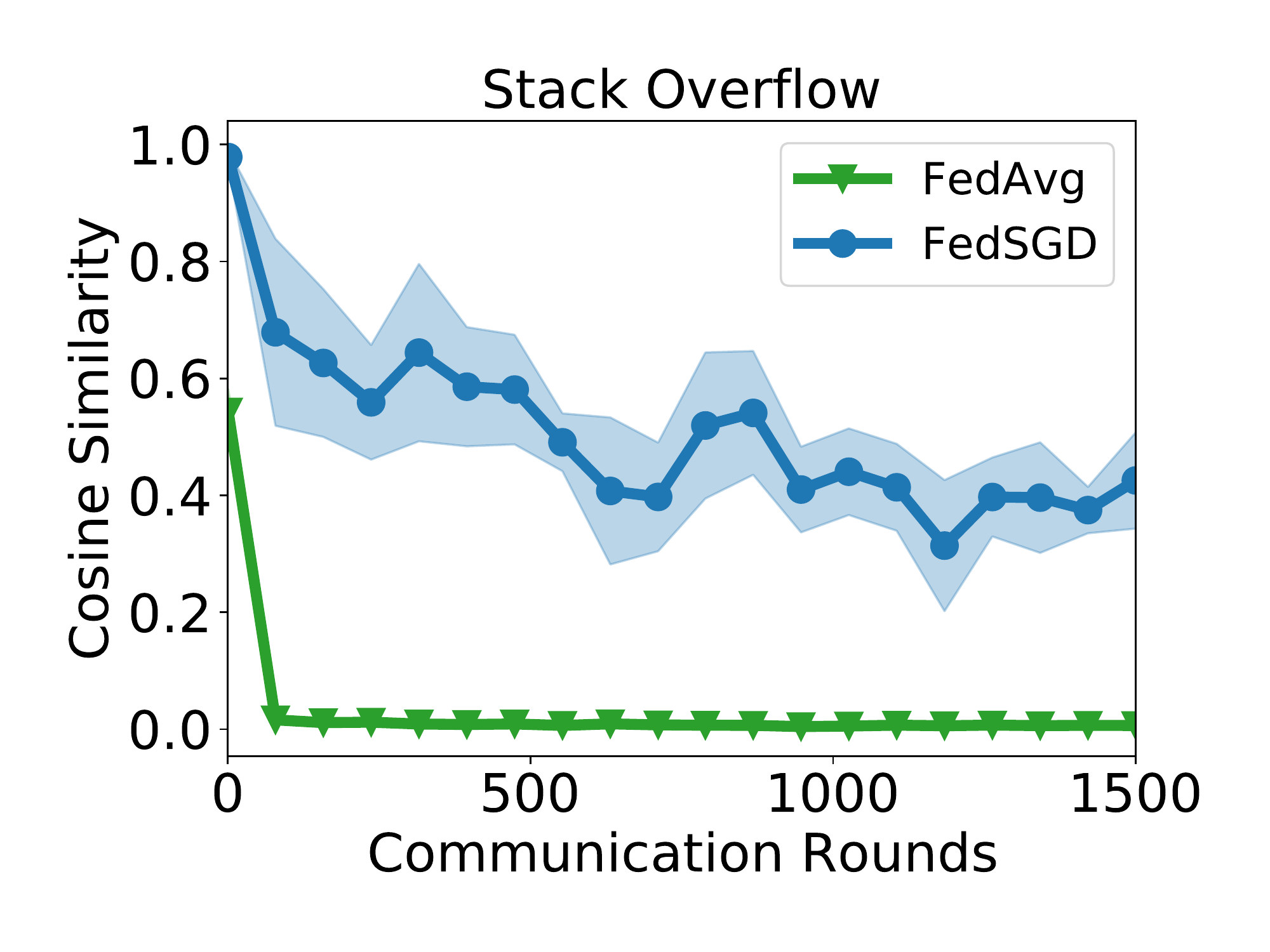}
    \caption{}
    \label{fig:pseudo_gradient_norm_d}
\end{subfigure}%
\caption{The pseudo-gradient norm of \fedsgd (a) and \fedavg (b) on Stack Overflow with varying cohort sizes $M$. We also plot the predicted norm for \fedavg using an inverse square root scaling rule relative to $M = 50$ (c) and the average cosine similarity of client updates for $M = 50$ (d).}
\label{fig:pseudo_gradient_norm}
\end{figure}

To better understand $\Delta$, we plot its norm on Stack Overflow in Figures \ref{fig:pseudo_gradient_norm_a} and \ref{fig:pseudo_gradient_norm_b}. For \fedsgd, $\|\Delta\|$ decreases slightly with $M$, but has high variance. By contrast, for \fedavg larger cohorts lead to smaller norms with little overlap. The decrease in norm obeys an inverse square root rule: Let $\Delta_1, \Delta_2$ be pseudo-gradients at some round for cohort sizes $M_1, M_2$. For \fedavg, $\|\Delta_1\|/\|\Delta_2\| \approx \sqrt{M_2/M_1}$. We use this rule to predict pseudo-gradient norms for \fedavg in \cref{fig:pseudo_gradient_norm_c}. After a small number of rounds, we obtain a remarkably good approximation. To explain this, we plot the average cosine similarity between client updates $\Delta_k^t$ at each round in \cref{fig:pseudo_gradient_norm_d}, with $M = 50$. For \fedavg, the client updates are on average almost orthogonal. This explains \cref{fig:pseudo_gradient_norm_b}, as $\Delta$ is an average of nearly orthogonal vectors. As we show in \cref{appendix:pseudo_gradient_norms}, similar results hold for other tasks and optimizers.

\textbf{Implications for large-cohort training.} This near-orthogonality of client updates is key to understanding the challenges in \cref{sec:challenges}. The diminishing returns in \cref{sec:diminishing_returns} occur in part because increasing $M$ leads to smaller updates. This also sheds light on \cref{sec:data_efficiency}: In large-cohort training, we take an average of many nearly-orthogonal vectors, so each client's examples contribute little. The decreasing pseudo-gradient norms in \cref{fig:pseudo_gradient_norm_c} also highlights an advantage of methods such as \fedadam and \fedadagrad: Adaptive server optimizers employ a form of normalization that makes them somewhat scale-invariant, compensating for this norm reduction.

\section{Designing Better Methods}\label{sec:better_methods}

We now explore an initial set of approaches aimed at improving large-cohort training, drawing inspiration where possible from large-batch training. Our solutions are designed to provide simple baselines for improving large-cohort training. In particular, our methods and experiments are intended to serve as a useful reference for future work in the area, not to fully solve the challenges of large-cohort training.

\subsection{Learning Rate Scaling}
One common technique for large-batch training is to scale the learning rate according to the batch size. Two popular scaling methods are square root scaling~\citep{krizhevsky2014one} and linear scaling~\citep{goyal2017accurate}. While such techniques have had clear empirical benefit in centralized training, there are many different ways that they could be adapted to federated learning. For example, in \cref{alg:fedopt}, the client and server optimization both use learning rates that could be scaled.

We consider the following scaling method for large-cohort training: We fix the client learning rate, and scale the server learning rate with the cohort size. Such scaling may improve convergence by compensating for the pseudo-gradient norm reduction in \cref{fig:pseudo_gradient_norm}. We use square root and linear scaling rules: Given a learning rate $\eta_s$ tuned for $M$, for $M' \geq M$ we use a learning rate $\eta_s'$ where
\begin{equation}\label{eq:server_lr_scaling}
\eta_s' = \frac{\sqrt{M'}}{\sqrt{M}}\eta_s\text{ (square root scaling)~~~OR~~~}\eta_s' = \frac{M'}{M}\eta_s\text{ (linear scaling)}.
\end{equation}
We also use a version of the warmup strategy from \citep{goyal2017accurate}. For the first $W$ communication rounds, we linearly increase the server learning rate from $\eta_s$ to $\eta_s'$. In our experiments, we set $W = 100$ and use a reference server learning rate $\eta_s$ tuned for $M = 50$.

Our experiments show that server learning rate scaling rules have mixed efficacy in large-cohort training. Linear scaling is often too aggressive for federated learning, and caused catastrophic training failures beyond $M = 100$ even when using adaptive clipping. As we show in \cref{appendix:server_lr_scaling}, when applying \fedavg to EMNIST with the linear scaling rule, we found that catastrophic failures occurred for $M > 100$, with no real improvement in accuracy for $M = 100$.

\begin{figure}[ht]
\centering
\begin{subfigure}{0.24\textwidth}
    \centering
    \includegraphics[width=1\linewidth]{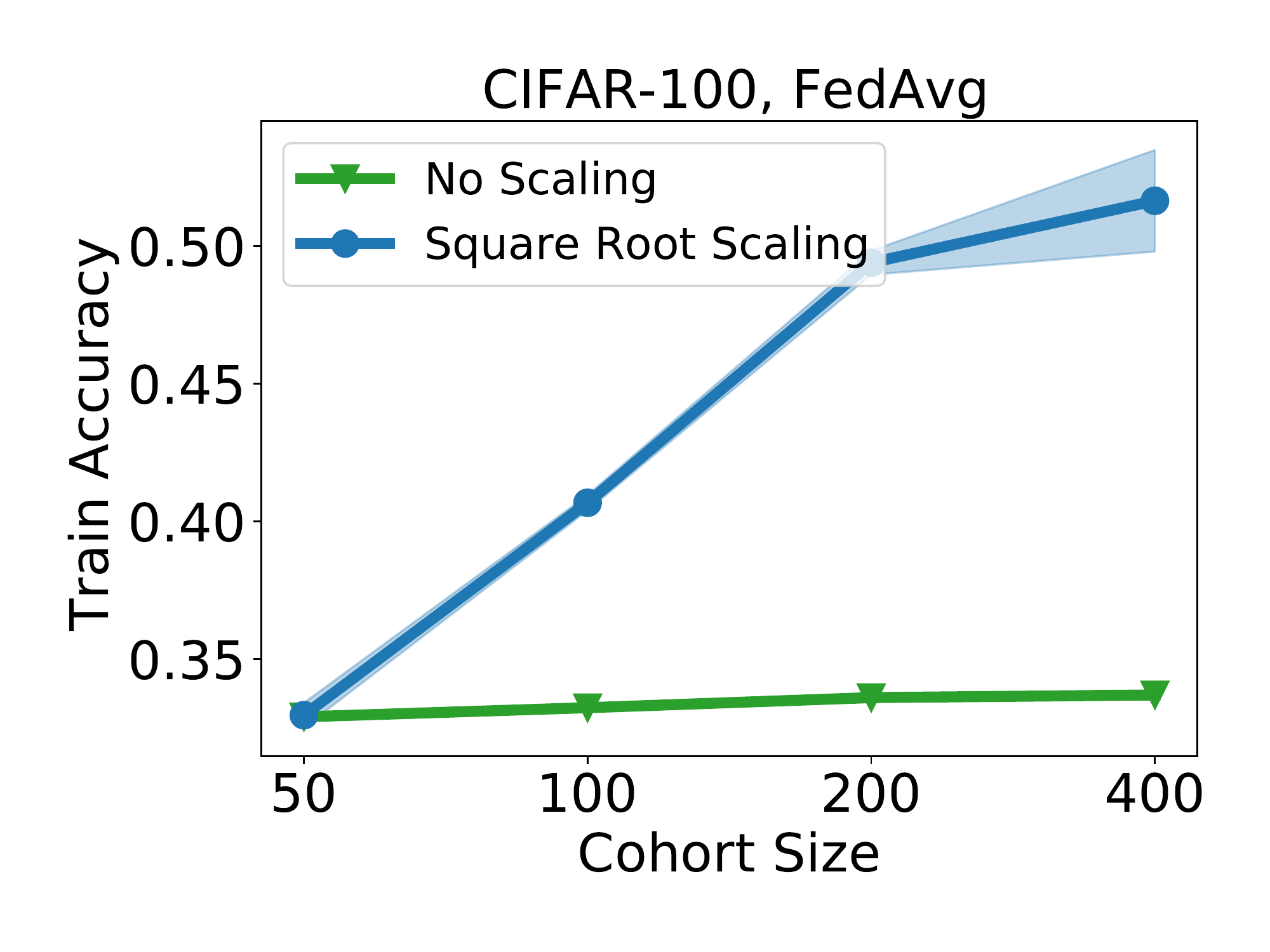}
\end{subfigure}%
\begin{subfigure}{0.24\textwidth}
    \centering
    \includegraphics[width=1\linewidth]{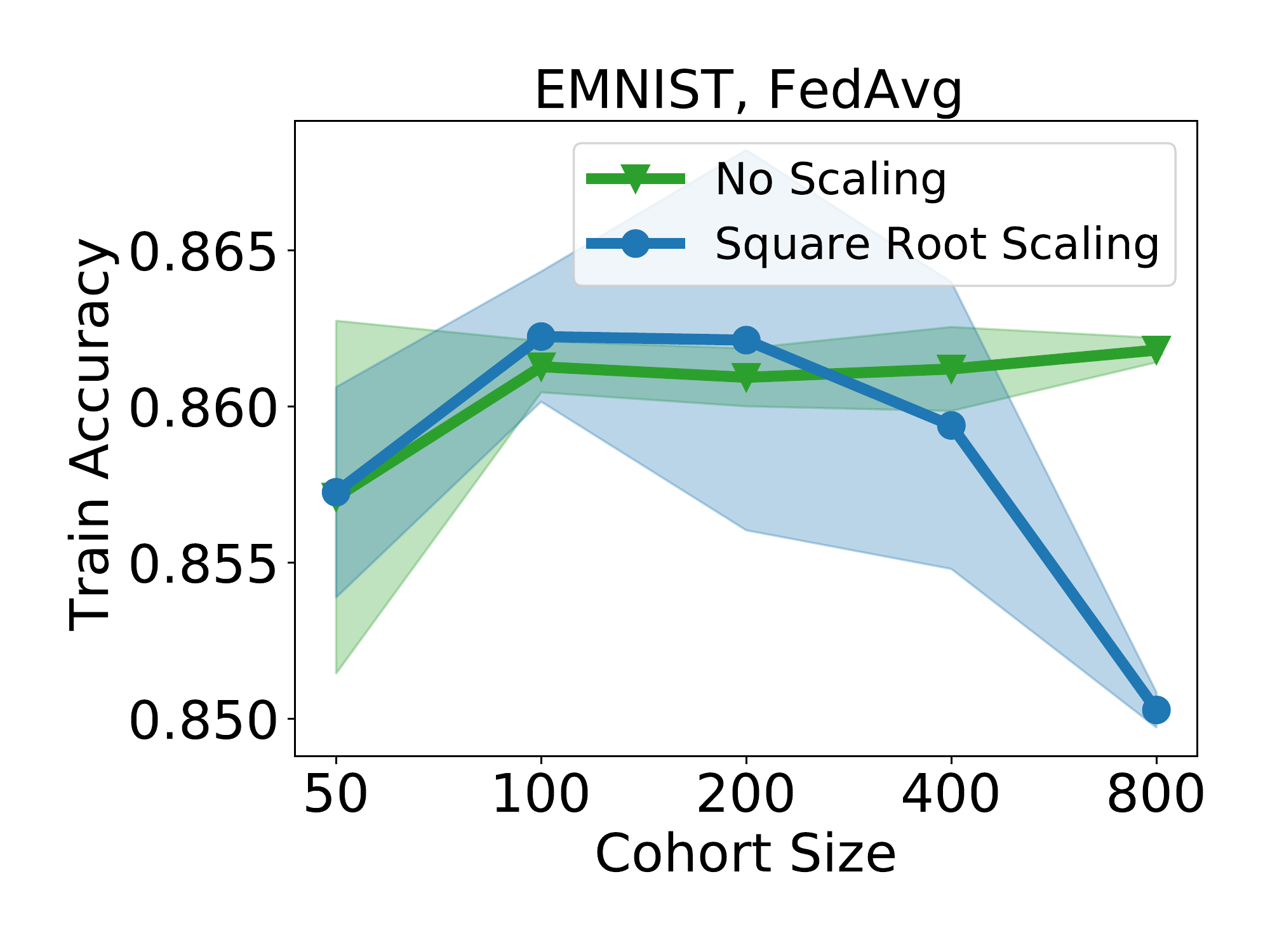}
\end{subfigure}%
\begin{subfigure}{0.24\textwidth}
    \centering
    \includegraphics[width=1\linewidth]{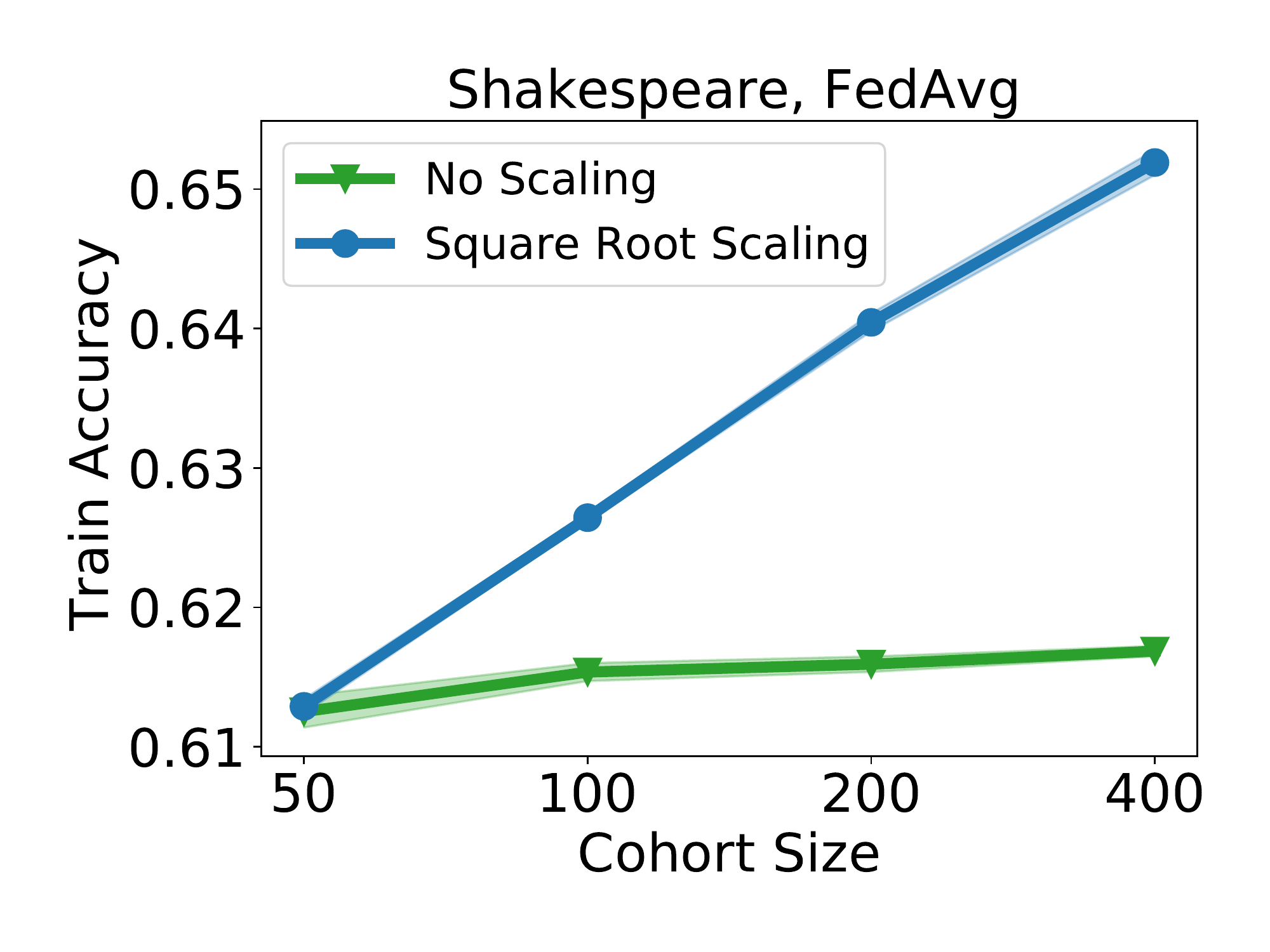}
\end{subfigure}%
\begin{subfigure}{0.24\textwidth}
    \centering
    \includegraphics[width=1\linewidth]{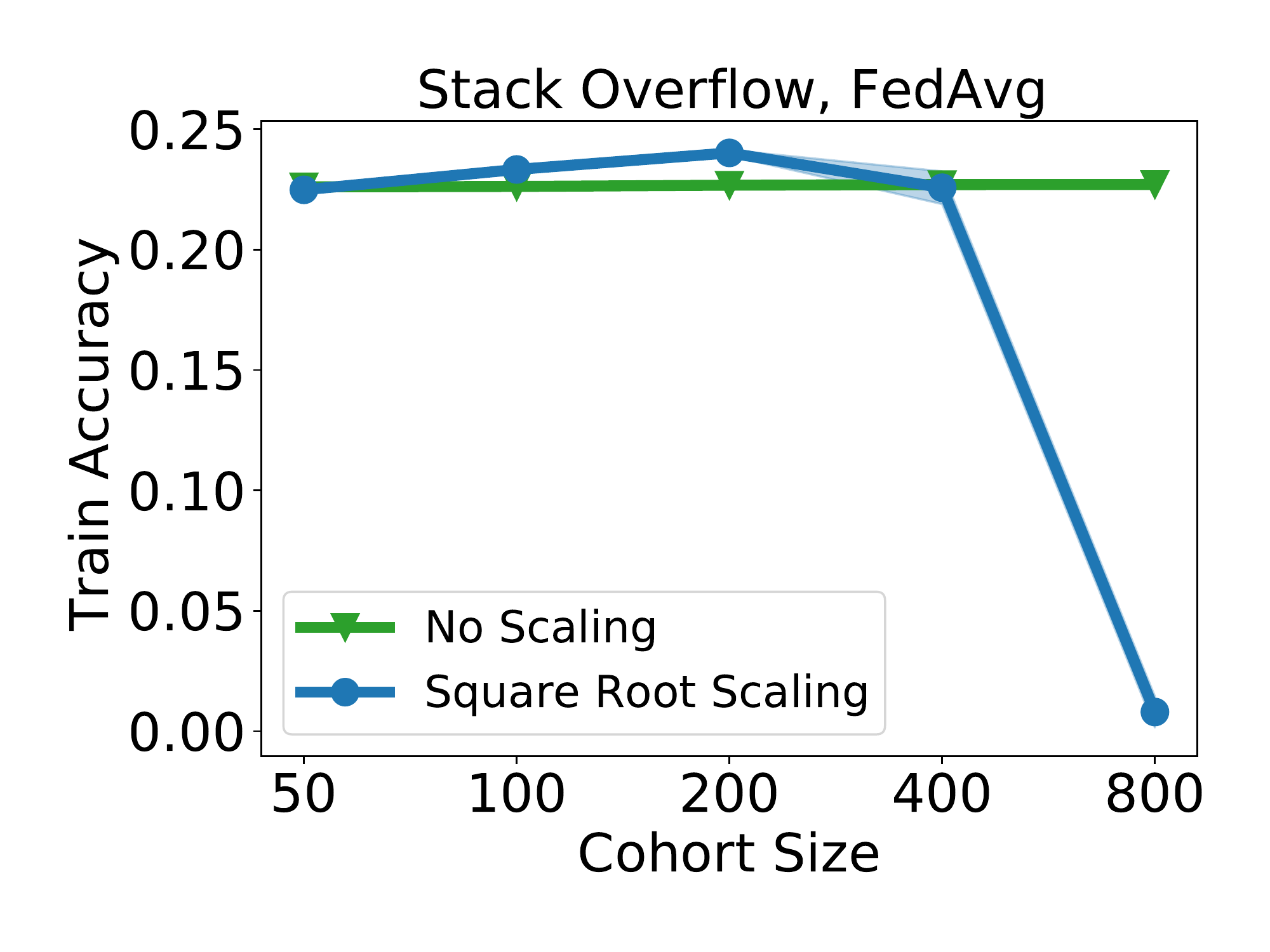}
\end{subfigure}
\begin{subfigure}{0.24\textwidth}
    \centering
    \includegraphics[width=1\linewidth]{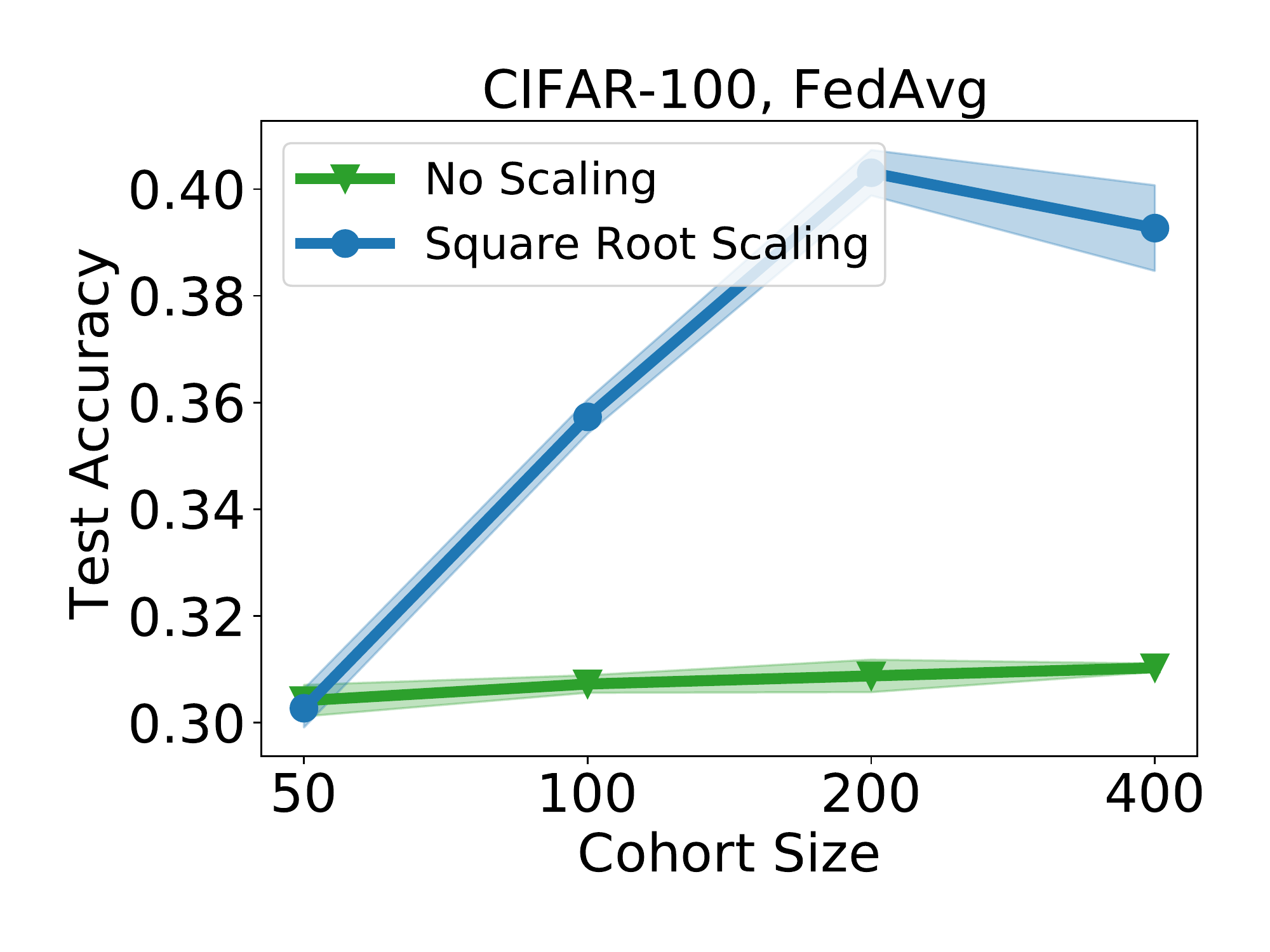}
\end{subfigure}%
\begin{subfigure}{0.24\textwidth}
    \centering
    \includegraphics[width=1\linewidth]{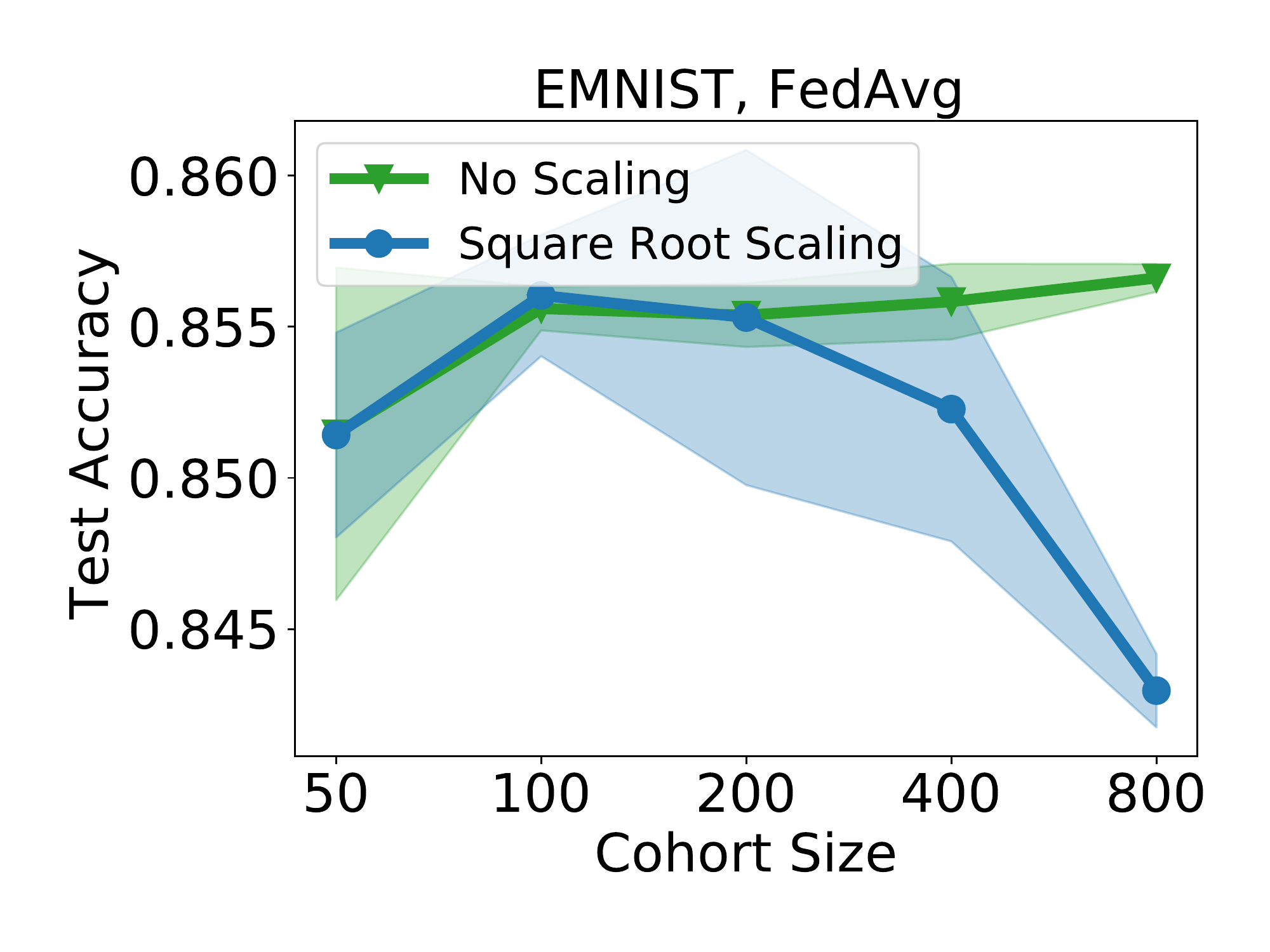}
\end{subfigure}%
\begin{subfigure}{0.24\textwidth}
    \centering
    \includegraphics[width=1\linewidth]{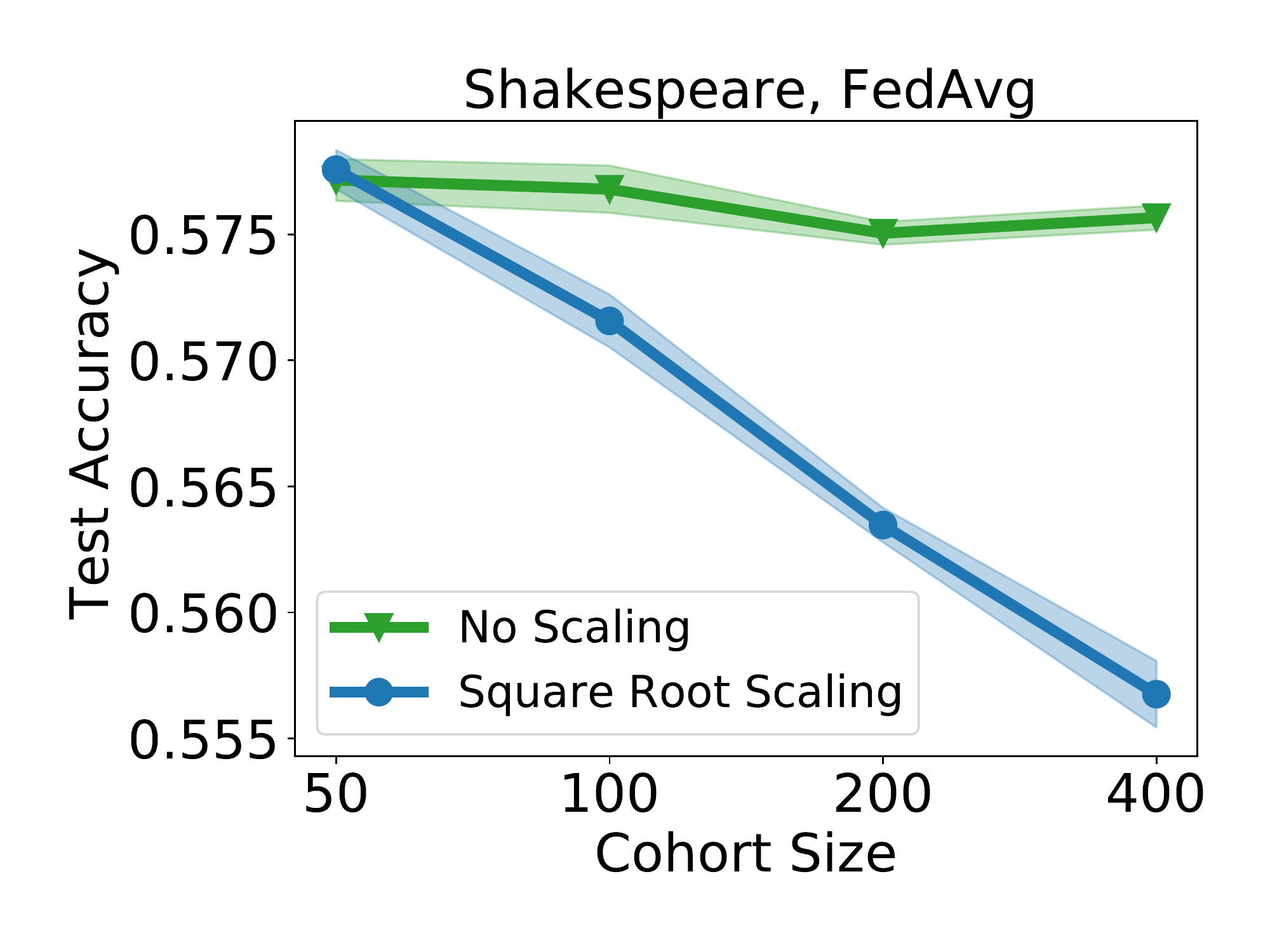}
\end{subfigure}%
\begin{subfigure}{0.24\textwidth}
    \centering
    \includegraphics[width=1\linewidth]{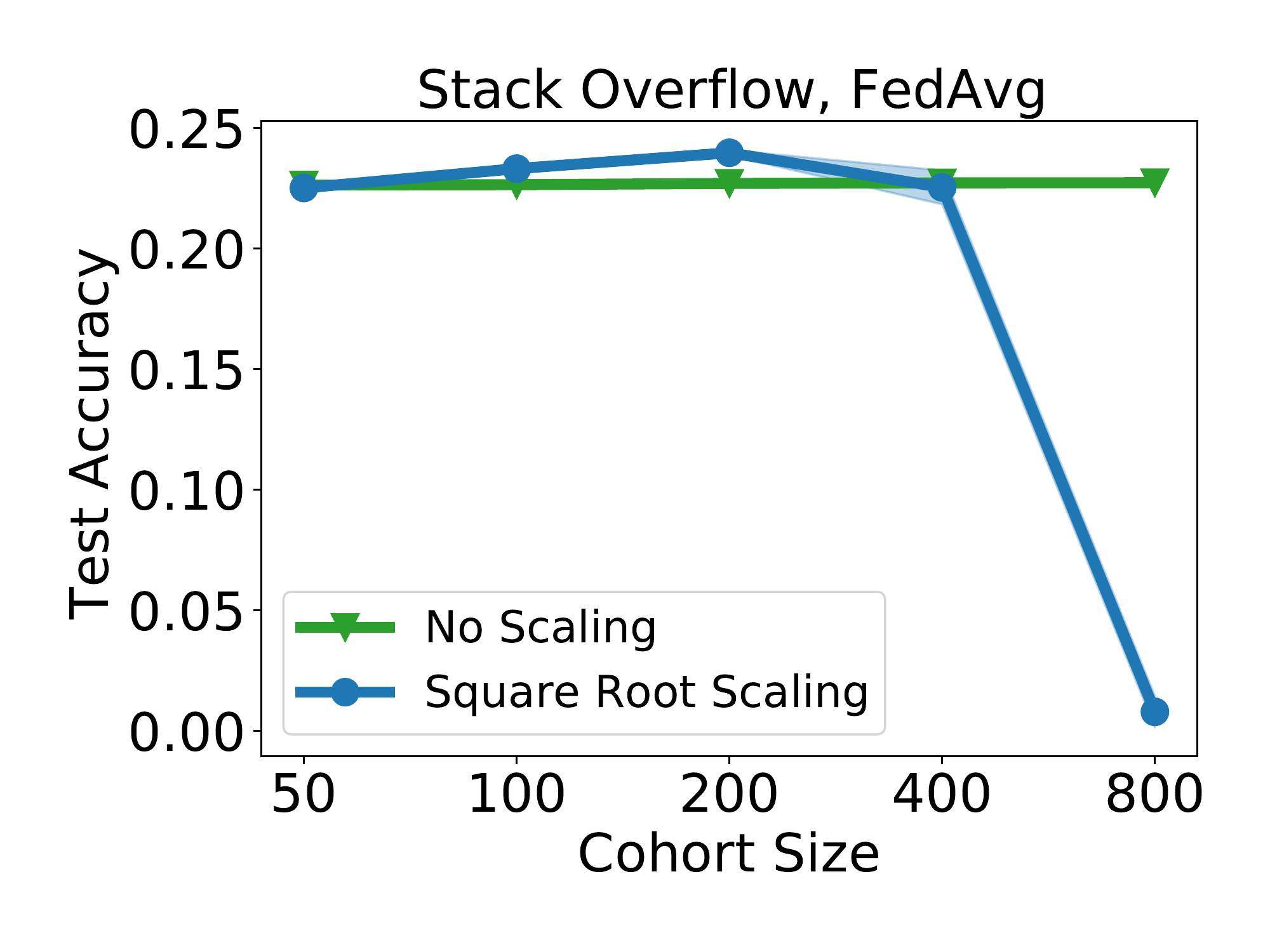}
\end{subfigure}
\caption{The train accuracy (top) and test accuracy (bottom) of \fedavg using square root scaling with warmup, versus no scaling. Results are recorded after 1500 rounds, for varying cohort sizes.}
\label{fig:main_accuracy_versus_cohort_size_scaling}
\end{figure}

By contrast, square root scaling did not cause catastrophic training failures. We plot the train and test accuracy of \fedavg using square root scaling in \cref{fig:main_accuracy_versus_cohort_size_scaling}. As we see, the performance varied across tasks. While it improved both train and test accuracy for CIFAR-100, for Shakespeare it improved train accuracy while reducing test accuracy. While we saw small improvements in accuracy for some cohort sizes on EMNIST and Stack Overflow, it degraded accuracy for the largest cohort sizes. In sum, we find that applying learning rate scaling at the server may not directly improve large-cohort training. 

\subsection{Layer-wise Adaptivity}
Another popular technique for large-batch training is \emph{layer-wise adaptivity}. Methods such as LARS~\citep{you2017large} and Lamb~\citep{you2019large} use layer-wise adaptive learning rates, which may allow the methods to train faster than \sgd with linear scaling and warmup in large-batch settings~\citep{you2017large, you2019large}. We propose two new federated versions of these optimizers, \fedlars and \fedlamb. These are special cases of \cref{alg:fedopt}, where the server uses LARS and Lamb, respectively. Given the difficulties of learning rate scaling above, \fedlars and \fedlamb may perform better in large-cohort settings.

\begin{figure}[ht]
\centering
\begin{subfigure}{0.24\textwidth}
    \centering
    \includegraphics[width=1\linewidth]{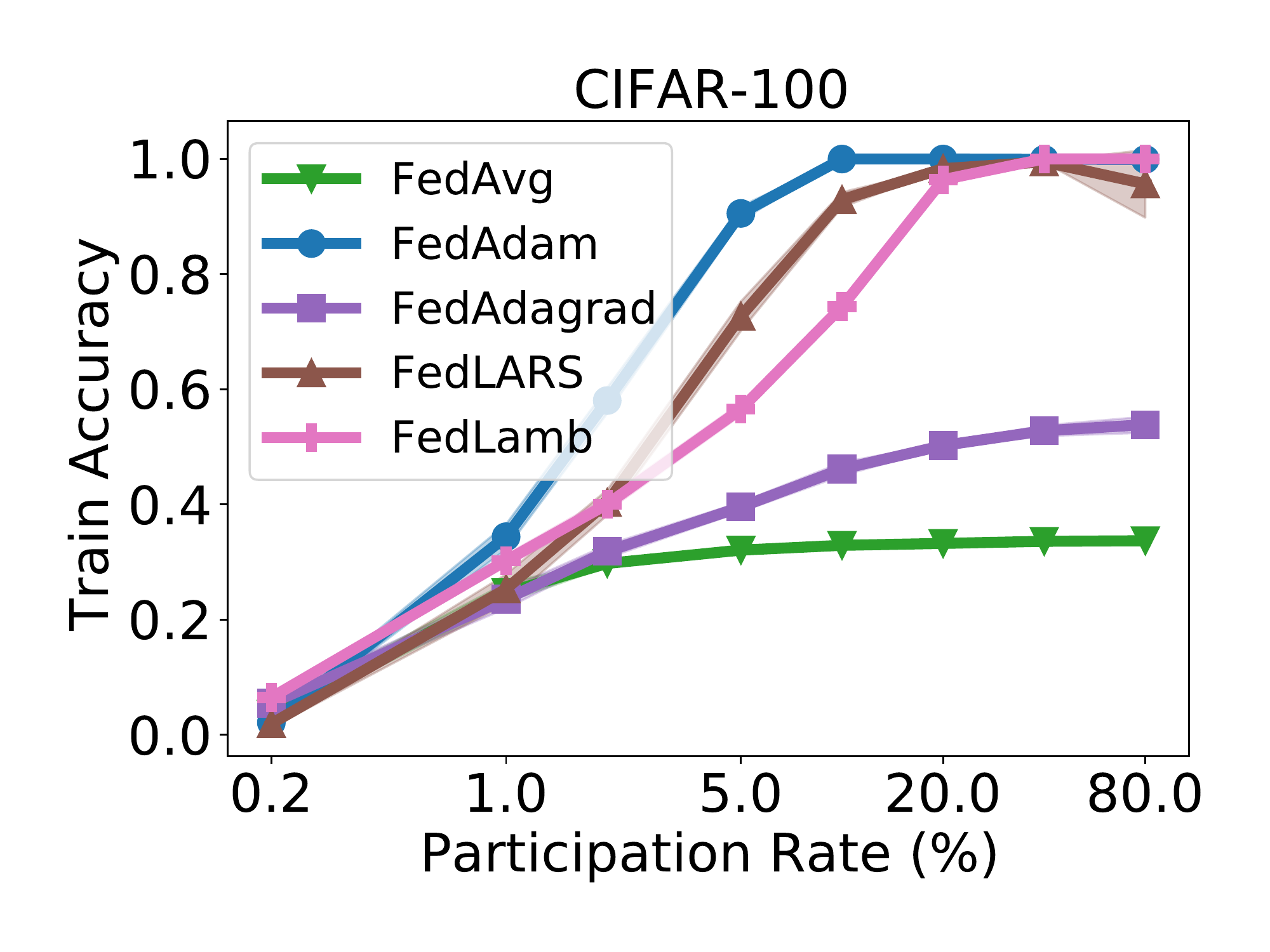}
\end{subfigure}%
\begin{subfigure}{0.24\textwidth}
    \centering
    \includegraphics[width=1\linewidth]{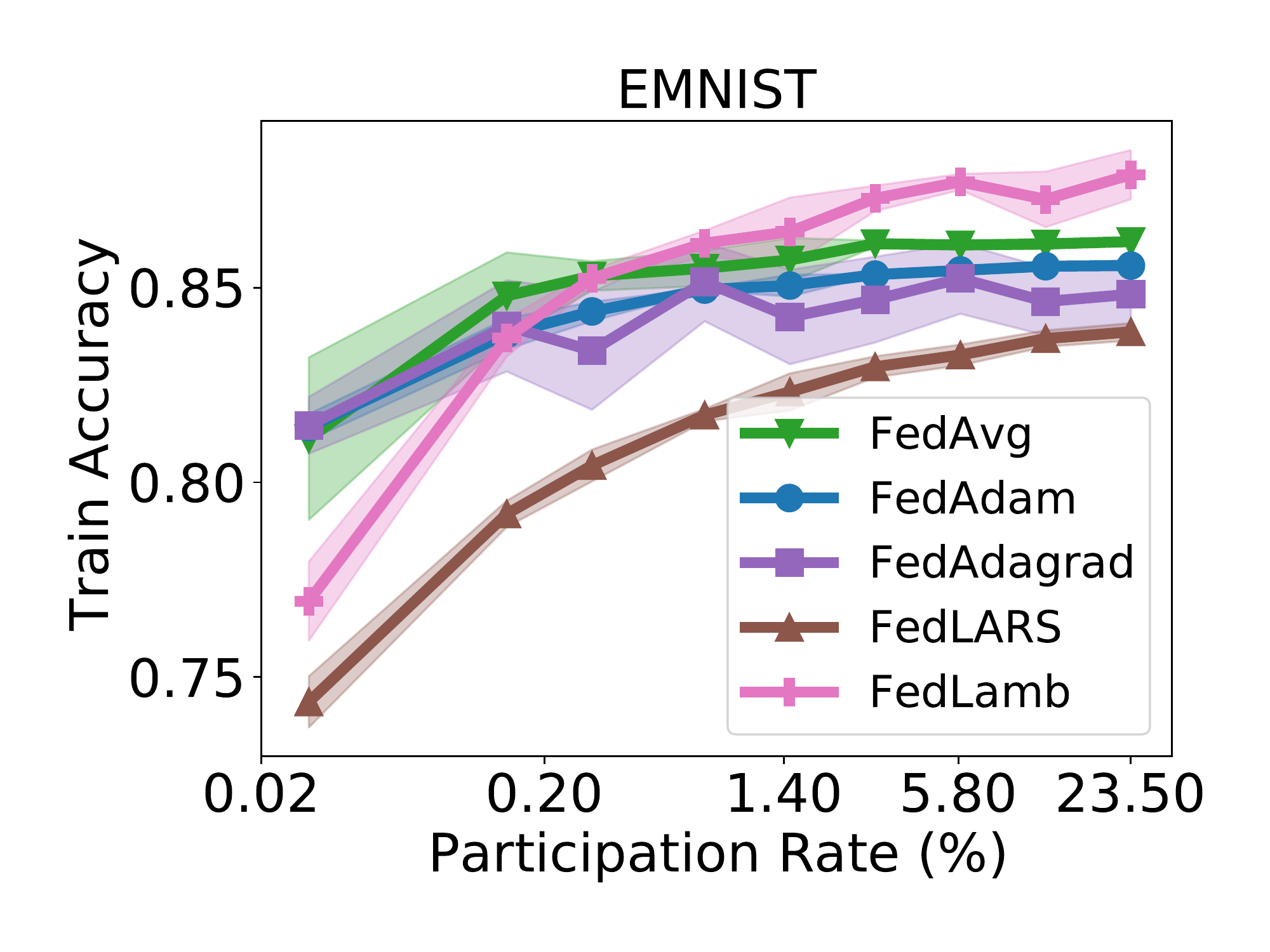}
\end{subfigure}%
\begin{subfigure}{0.24\textwidth}
    \centering
    \includegraphics[width=1\linewidth]{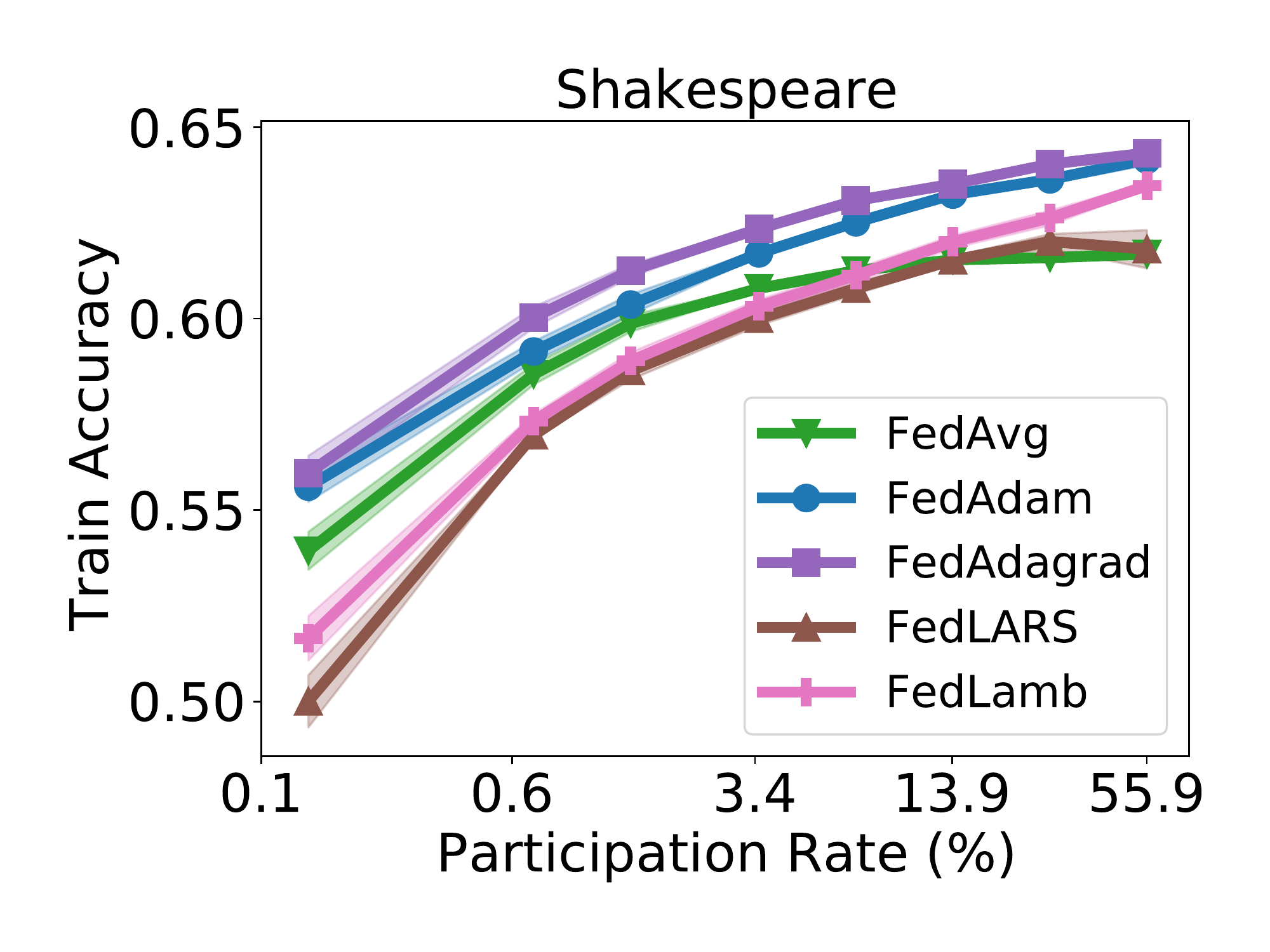}
\end{subfigure}%
\begin{subfigure}{0.24\textwidth}
    \centering
    \includegraphics[width=1\linewidth]{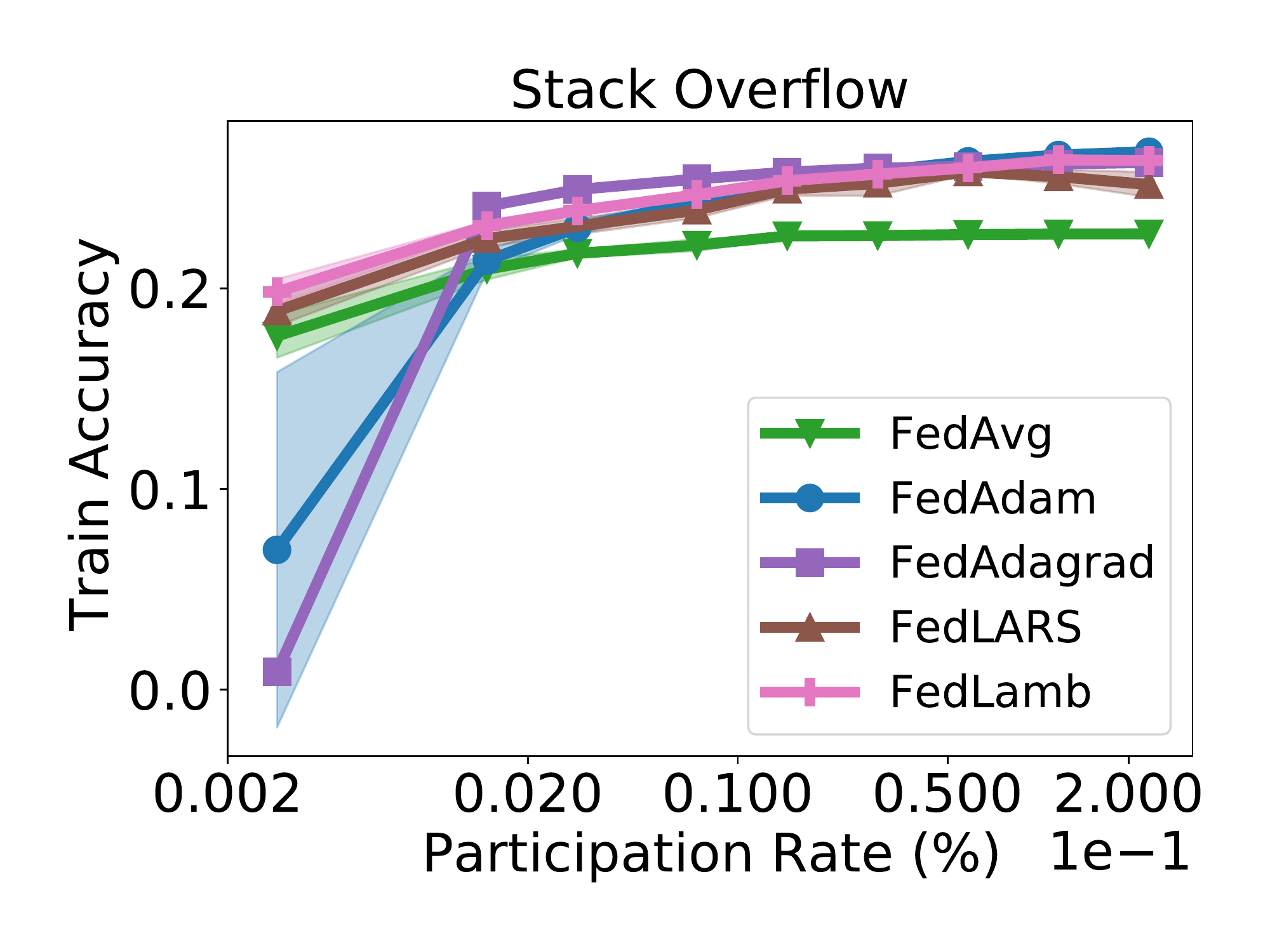}
\end{subfigure}
\begin{subfigure}{0.24\textwidth}
    \centering
    \includegraphics[width=1\linewidth]{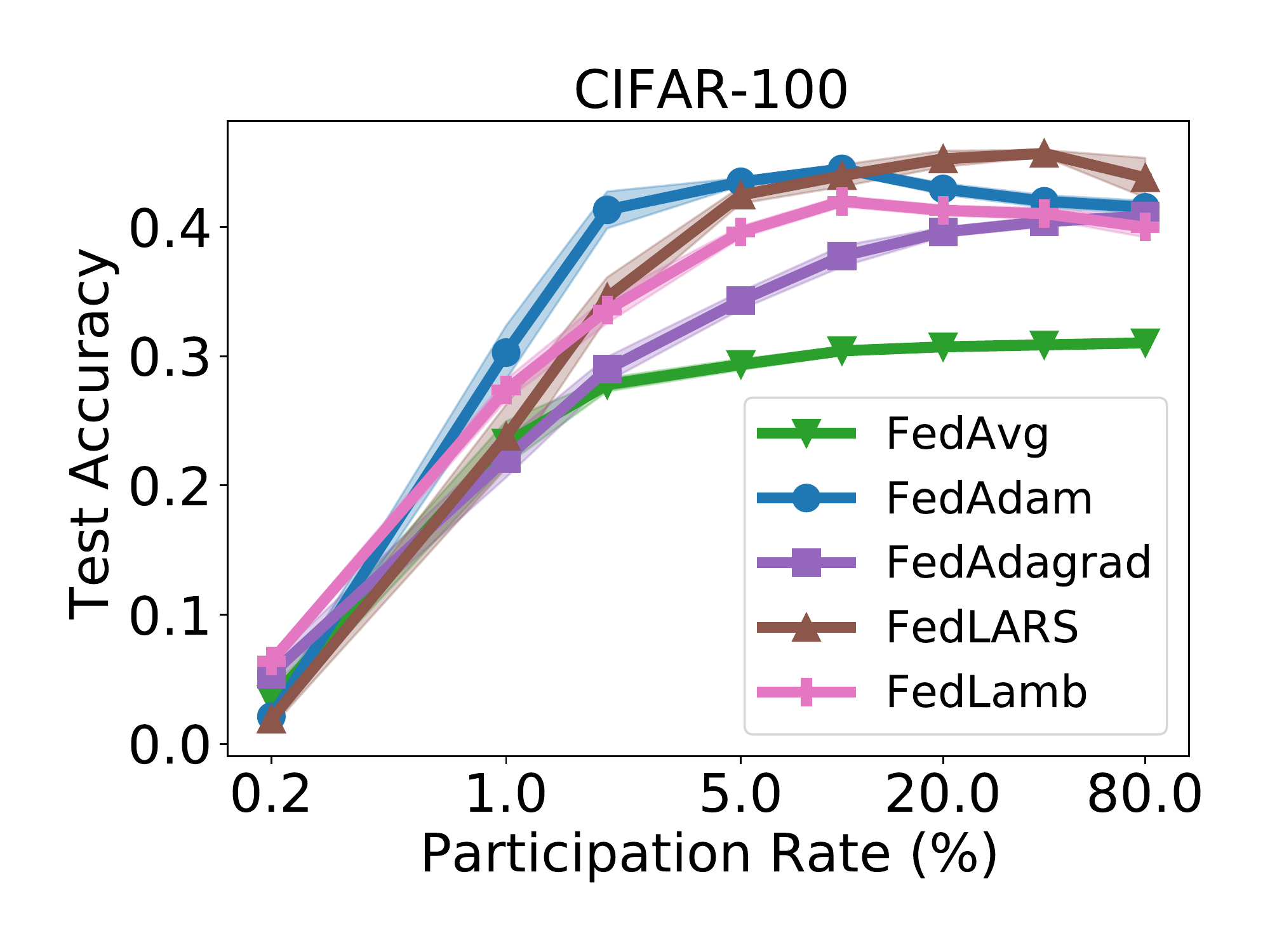}
\end{subfigure}%
\begin{subfigure}{0.24\textwidth}
    \centering
    \includegraphics[width=1\linewidth]{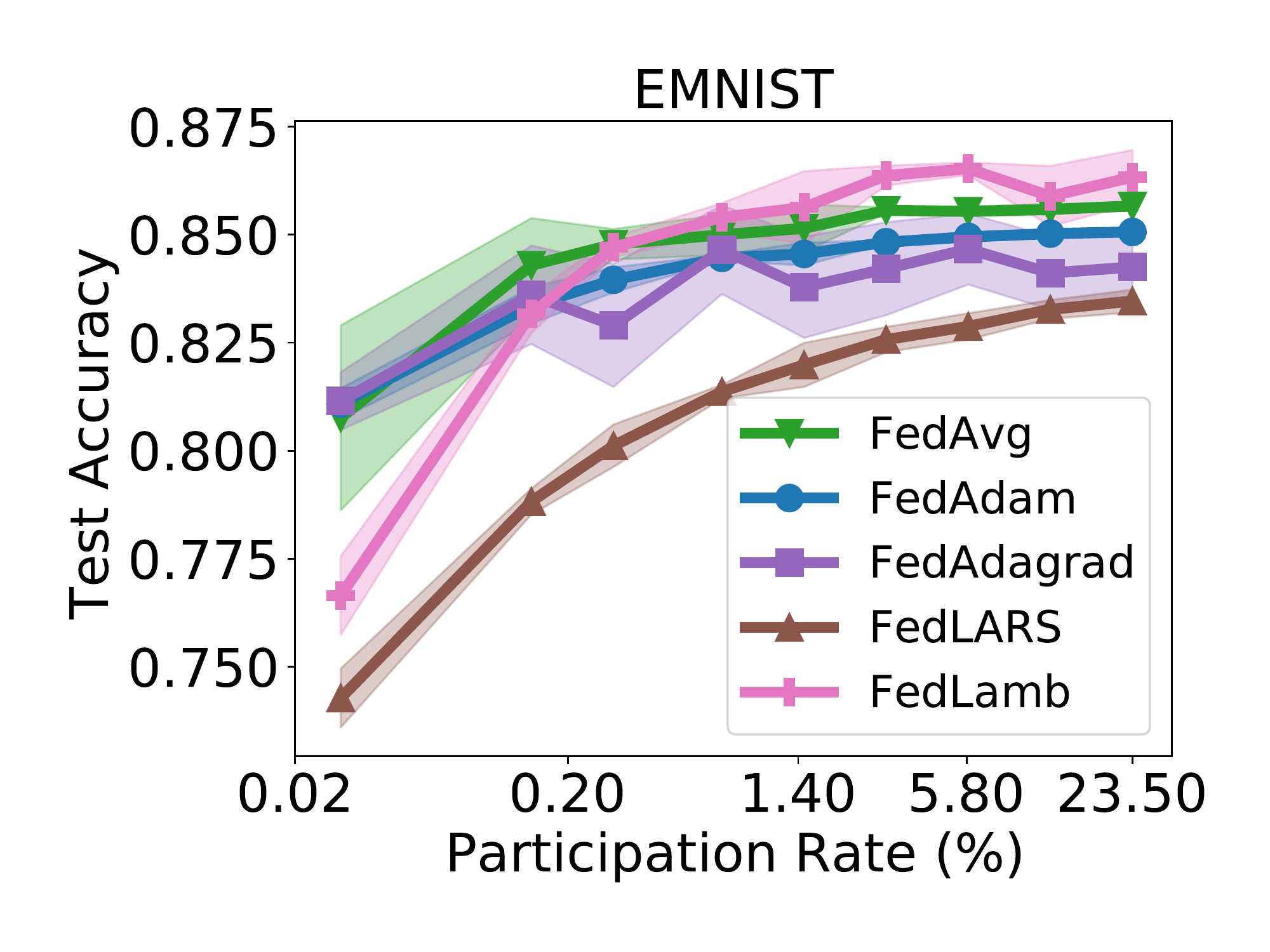}
\end{subfigure}%
\begin{subfigure}{0.24\textwidth}
    \centering
    \includegraphics[width=1\linewidth]{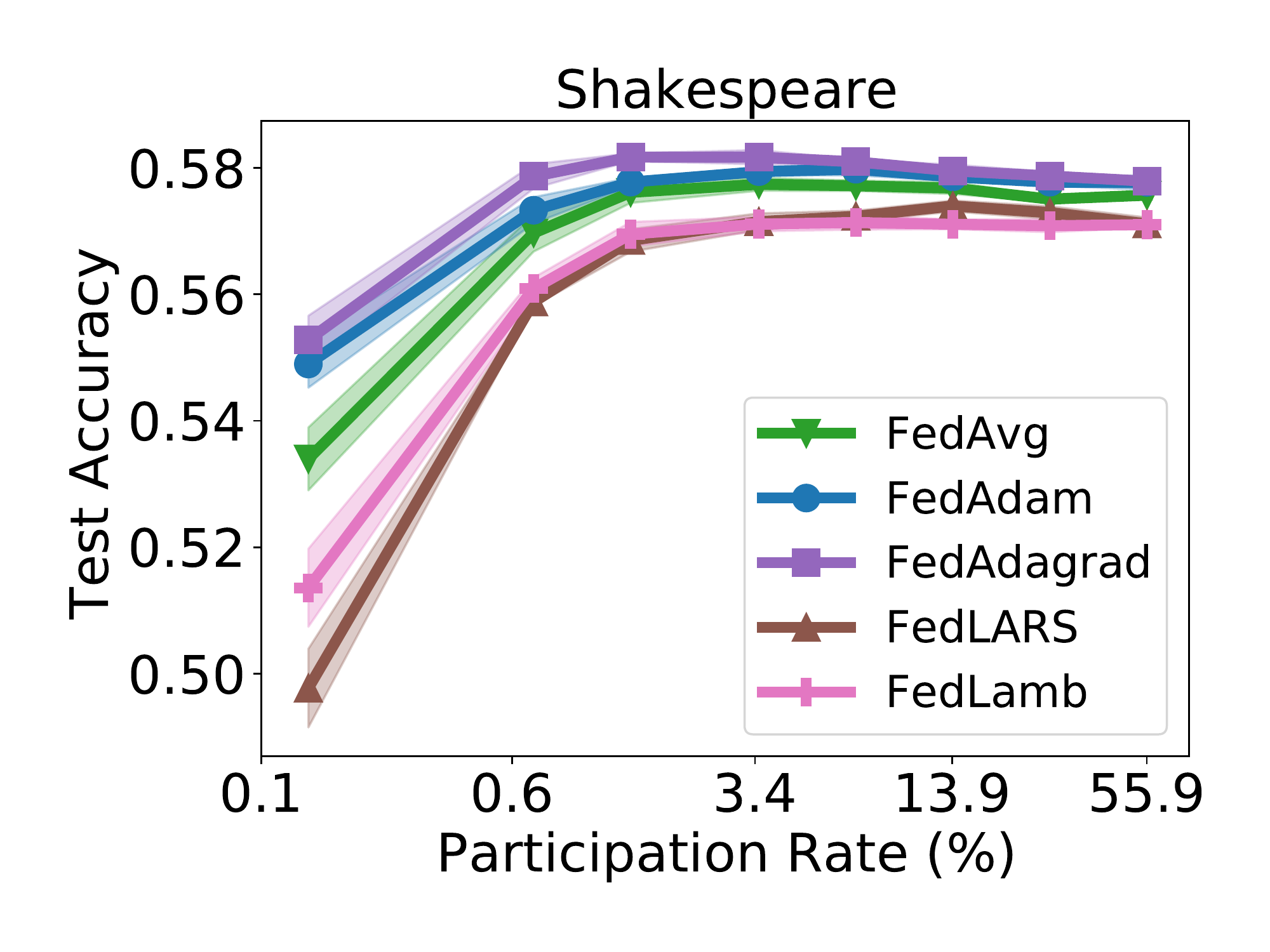}
\end{subfigure}%
\begin{subfigure}{0.24\textwidth}
    \centering
    \includegraphics[width=1\linewidth]{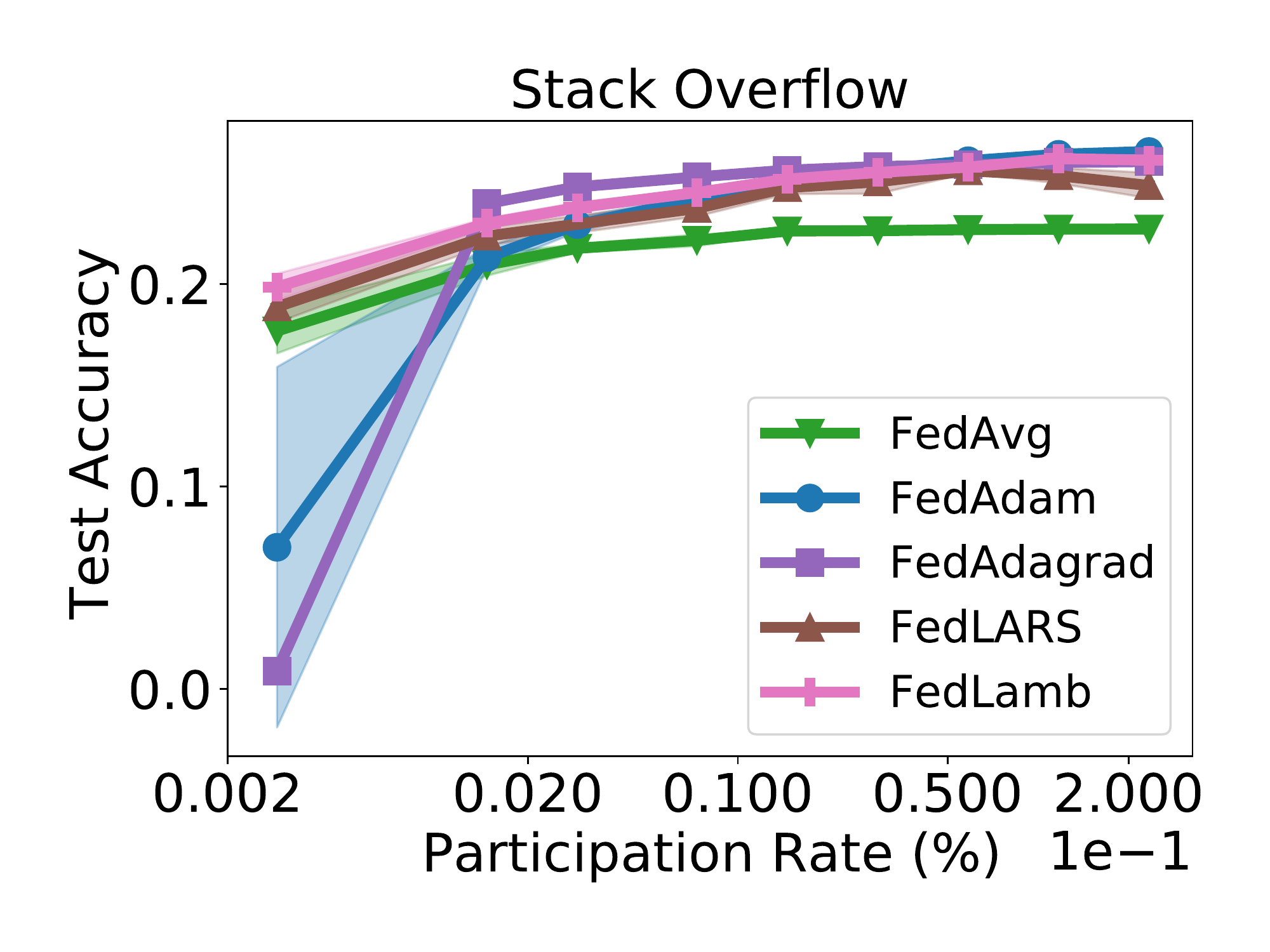}
\end{subfigure}
\caption{The train accuracy (top) and test accuracy (bottom) of various methods, including \fedlars and \fedlamb, after training for 1500 rounds, for varying cohort sizes and on varying tasks. The $x$-axis denotes percentage of training clients in each cohort.}
\label{fig:main_layerwise_accuracy_versus_cohort_size}
\end{figure}

In \cref{fig:main_layerwise_accuracy_versus_cohort_size} we present the train and test accuracy of various methods, including \fedlars and \fedlamb, for varying cohort sizes. In most cases, we see that \fedlamb performs comparably to \fedadam for large cohort sizes, but with slightly worse performance in intermediate stages. One notable exception is Stack Overflow, in which \fedlamb performs well even for $M = 1$. As in \cref{subsec:generalization_failures}, \fedlamb sees an eventual drop in test accuracy for $M > 100$. \fedlars has decidedly mixed performance. While it performs well on CIFAR-10, it does not do well on EMNIST or Shakespeare. While federated layer-wise adaptive algorithms can be better than coordinate-wise adaptive algorithms on certain datasets in some large-cohort settings, our results do not indicate that they are universally better.

\subsection{Dynamic Cohort Sizes} As we saw in \cref{sec:data_efficiency}, large-cohort training can reduce data efficiency. Part of this stems from the fact that larger cohorts may help very little for smaller accuracy thresholds (see \cref{fig:accuracy_per_example}). In order to improve data efficiency, we may be able to use smaller cohorts in earlier optimization stages, and increase the cohort size over time. This technique is parallel to ``dynamic batch size'' techniques used in large-batch training~\citep{smith2017don}. In order to test the efficacy of such techniques in large-cohort training, we start with an initial cohort size of $M = 50$ and double the size every 300 rounds up to $M=800$ (or the maximum population size if smaller). This results in doubling the cohort size a maximum of 4 times over the $1500$ rounds of training we perform. We plot the results for \fedavg and \fedadam on CIFAR-100 and Stack Overflow in \cref{fig:cohort_doubling_experiments}. See \cref{appendix:dyanmic_cohort_sizes} for results on all tasks. 

\begin{figure}[ht]
\centering
\begin{subfigure}{0.24\textwidth}
    \centering
    \includegraphics[width=1\linewidth]{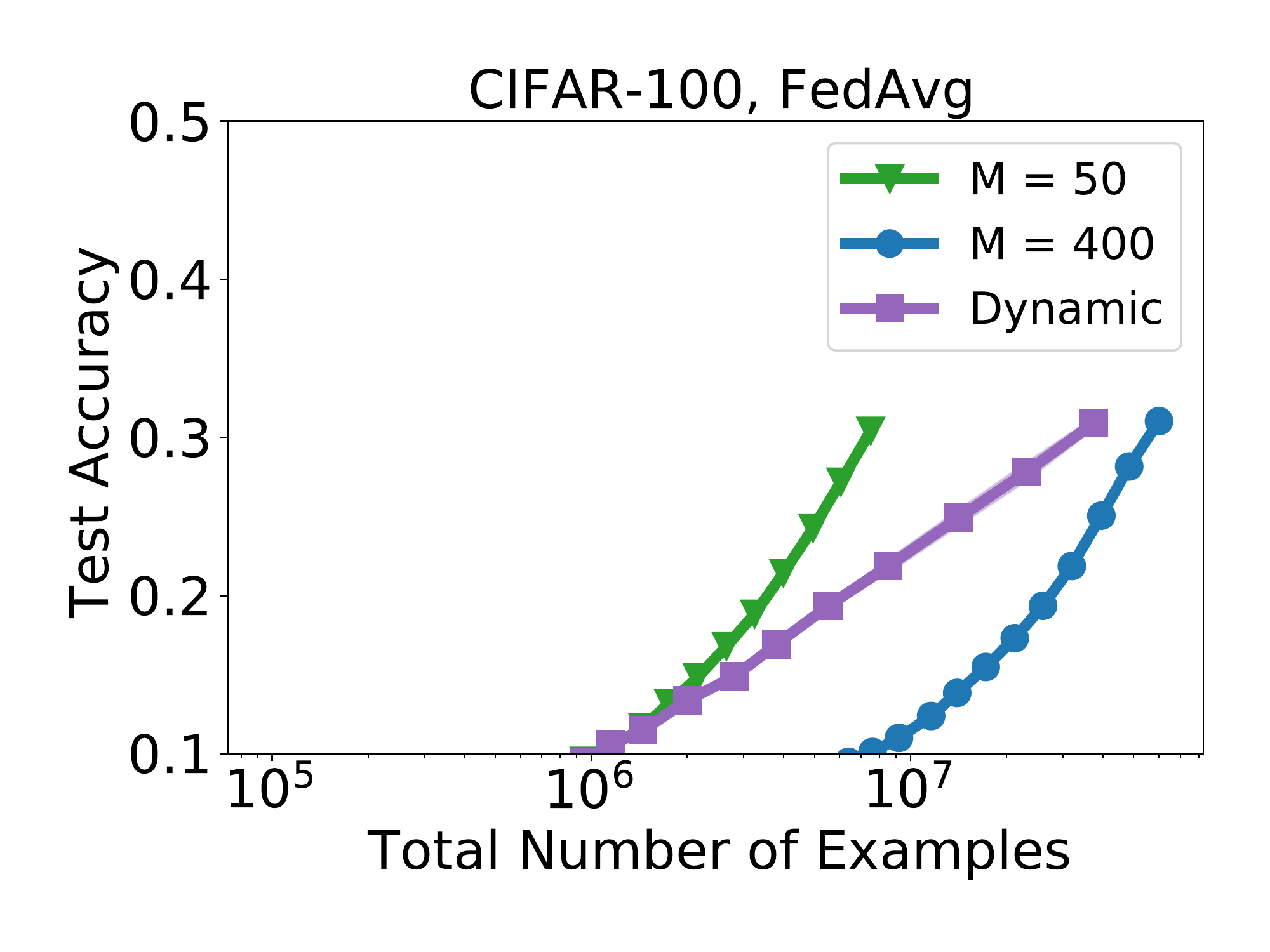}
\end{subfigure}%
\begin{subfigure}{0.24\textwidth}
    \centering
    \includegraphics[width=1\linewidth]{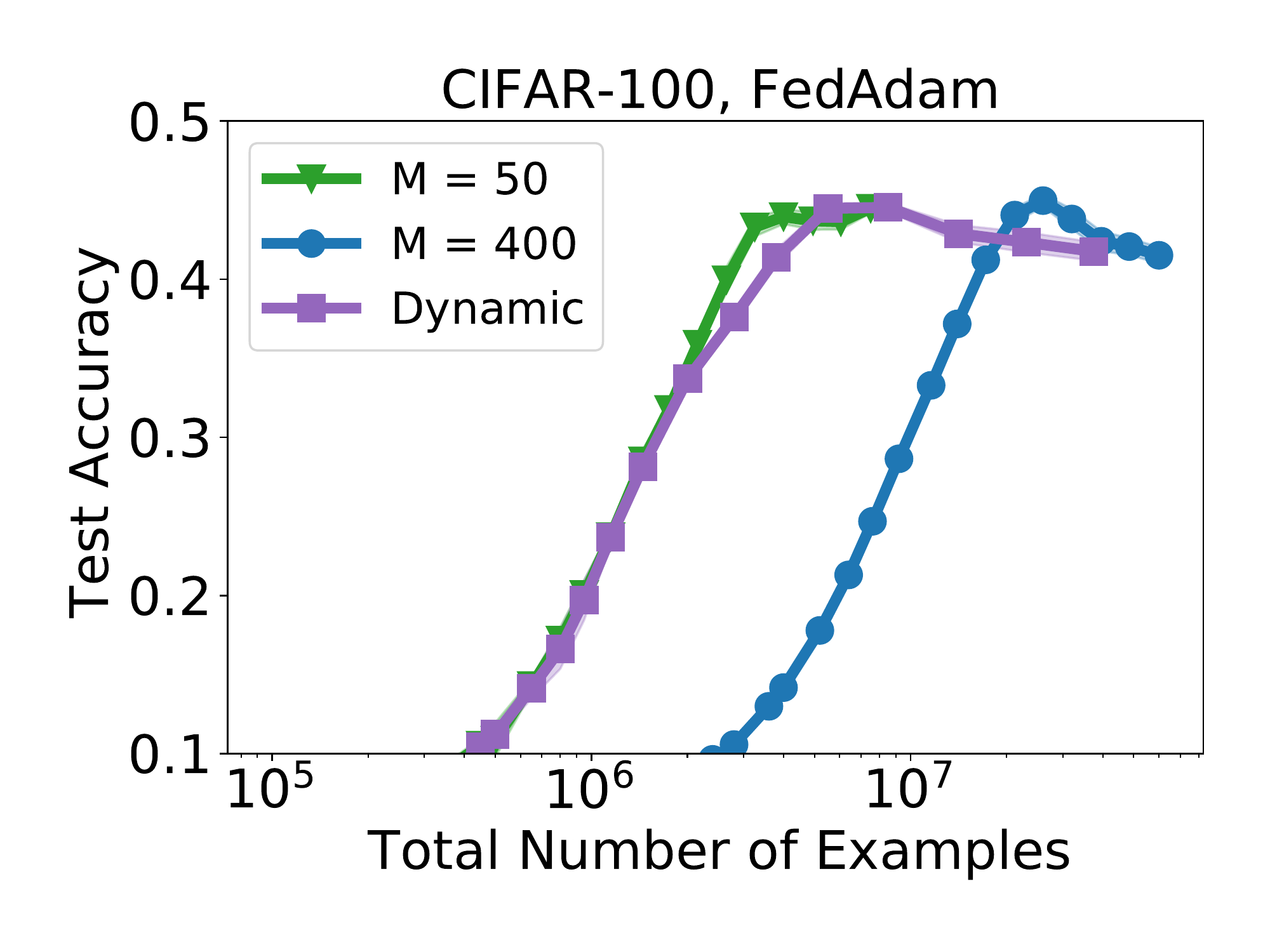}
\end{subfigure}%
\begin{subfigure}{0.24\textwidth}
    \centering
    \includegraphics[width=1\linewidth]{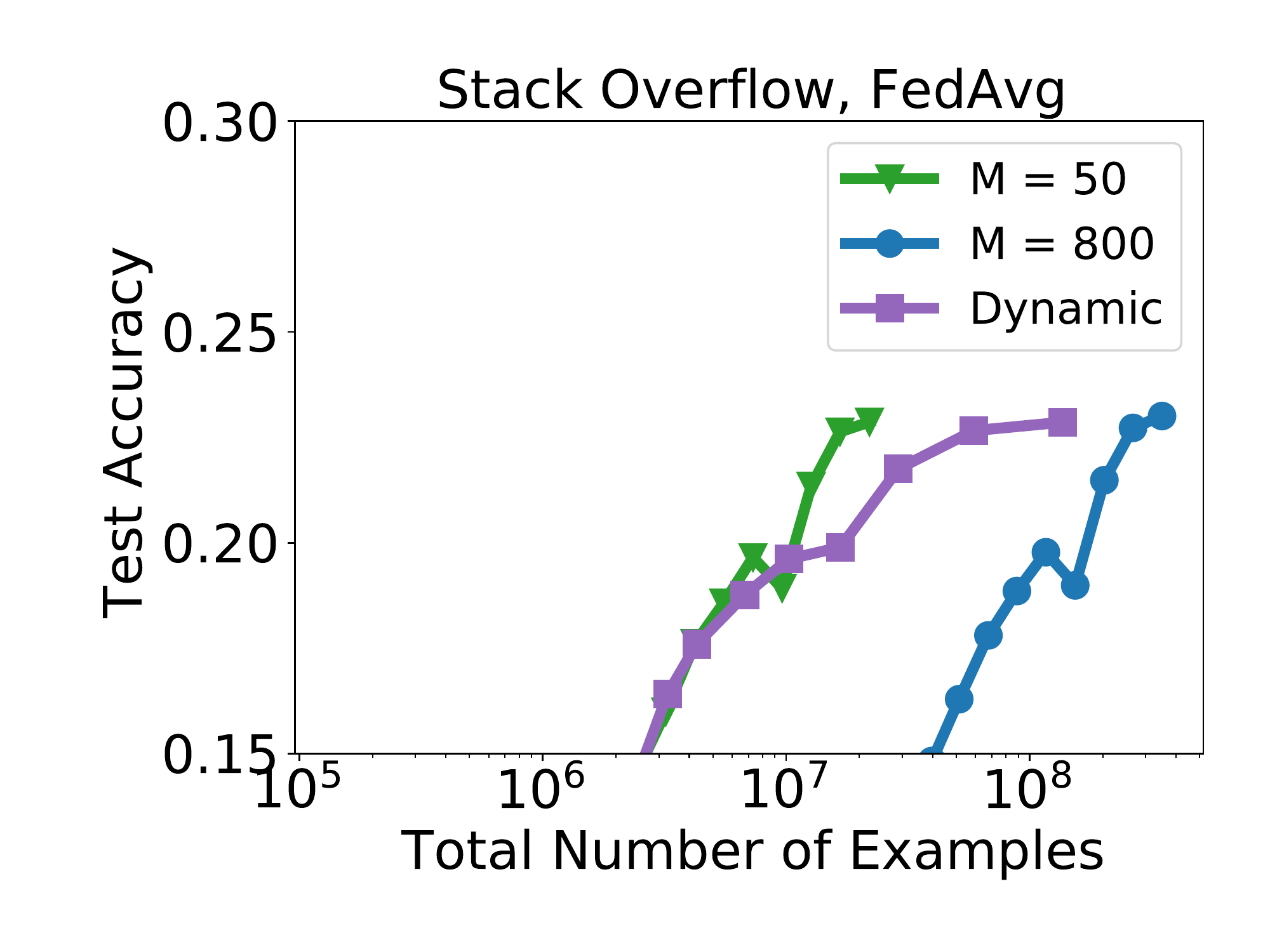}
\end{subfigure}
\begin{subfigure}{0.24\textwidth}
    \centering
    \includegraphics[width=1\linewidth]{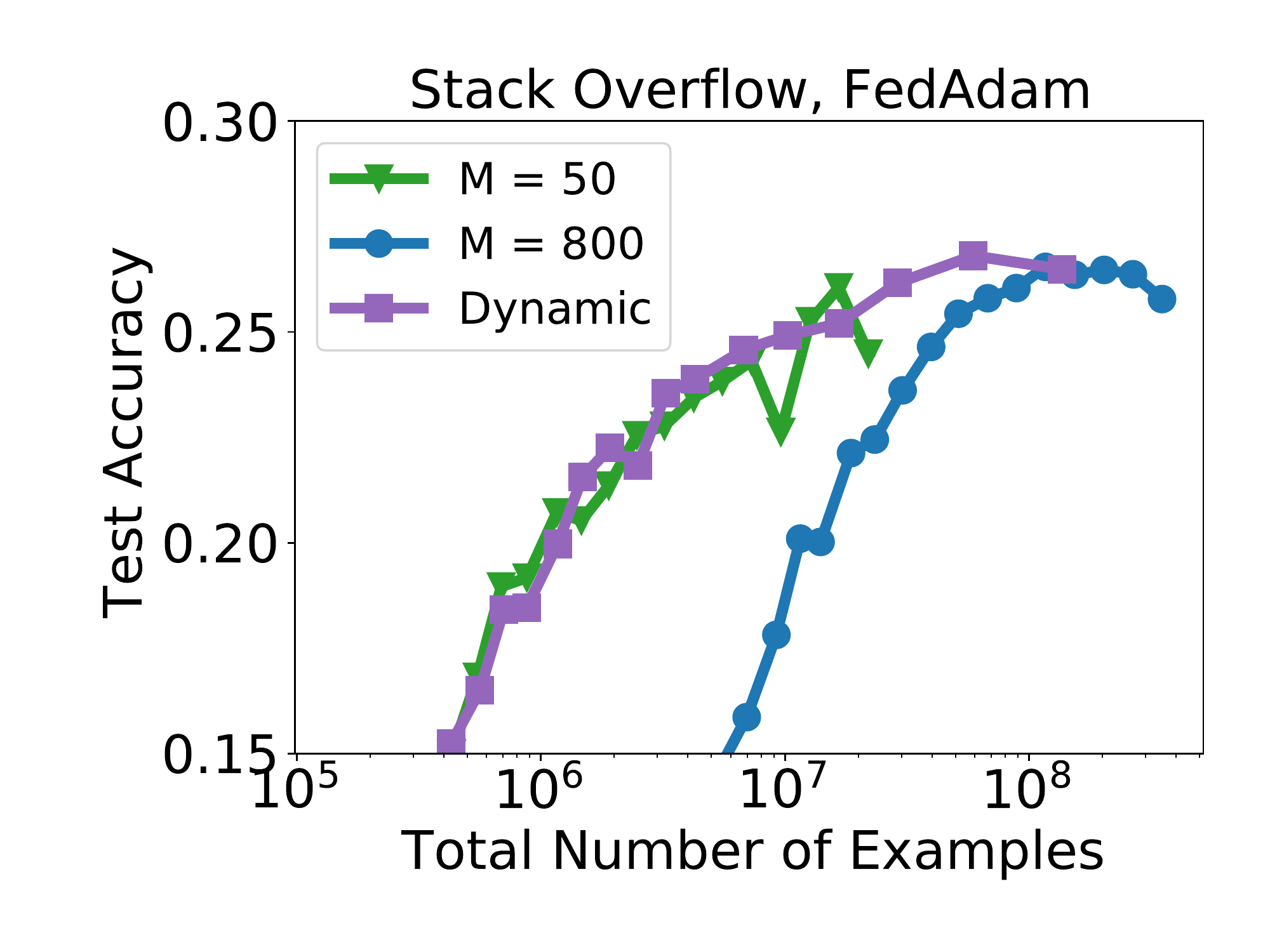}
\end{subfigure}%
\caption{Test accuracy of \fedavg and \fedadam on Shakespeare (left) and Stack Overflow (right), with respect to the total number of examples processed, using fixed and dynamic cohort sizes.}
\label{fig:cohort_doubling_experiments}
\end{figure}

This dynamic strategy attains data efficiency closer to a fixed cohort size of $M = 50$, while still obtaining a final accuracy closer to having used a large fixed cohort size. While our initial findings are promising, we note two important limitations. First, the accuracy of the dynamic strategy is bounded by the minimum and maximum cohort size used; It never attains a better accuracy than $M = 800$. Second, the doubling strategy still faces the generalization issues discussed in \cref{subsec:generalization_failures}.

\subsection{Normalized \fedavg}

While the methods above show promise in resolving some of the issues of large-cohort training, they also introduce extra hyperparameters (such as what type of learning rate scaling to use, or how often to double the cohort size). Notably, hyperparameter tuning can be difficult in federated learning, especially cross-device federated learning~\citep{kairouz2019advances}. Even adaptive methods like \fedadam introduce a number of new hyperparameters that can be challenging to contend with. We are therefore motivated to design a large-cohort training method that does not introduce any new hyperparameters.

\begin{figure}[ht!]
\centering
\begin{subfigure}{0.24\textwidth}
    \centering
    \includegraphics[width=1\linewidth]{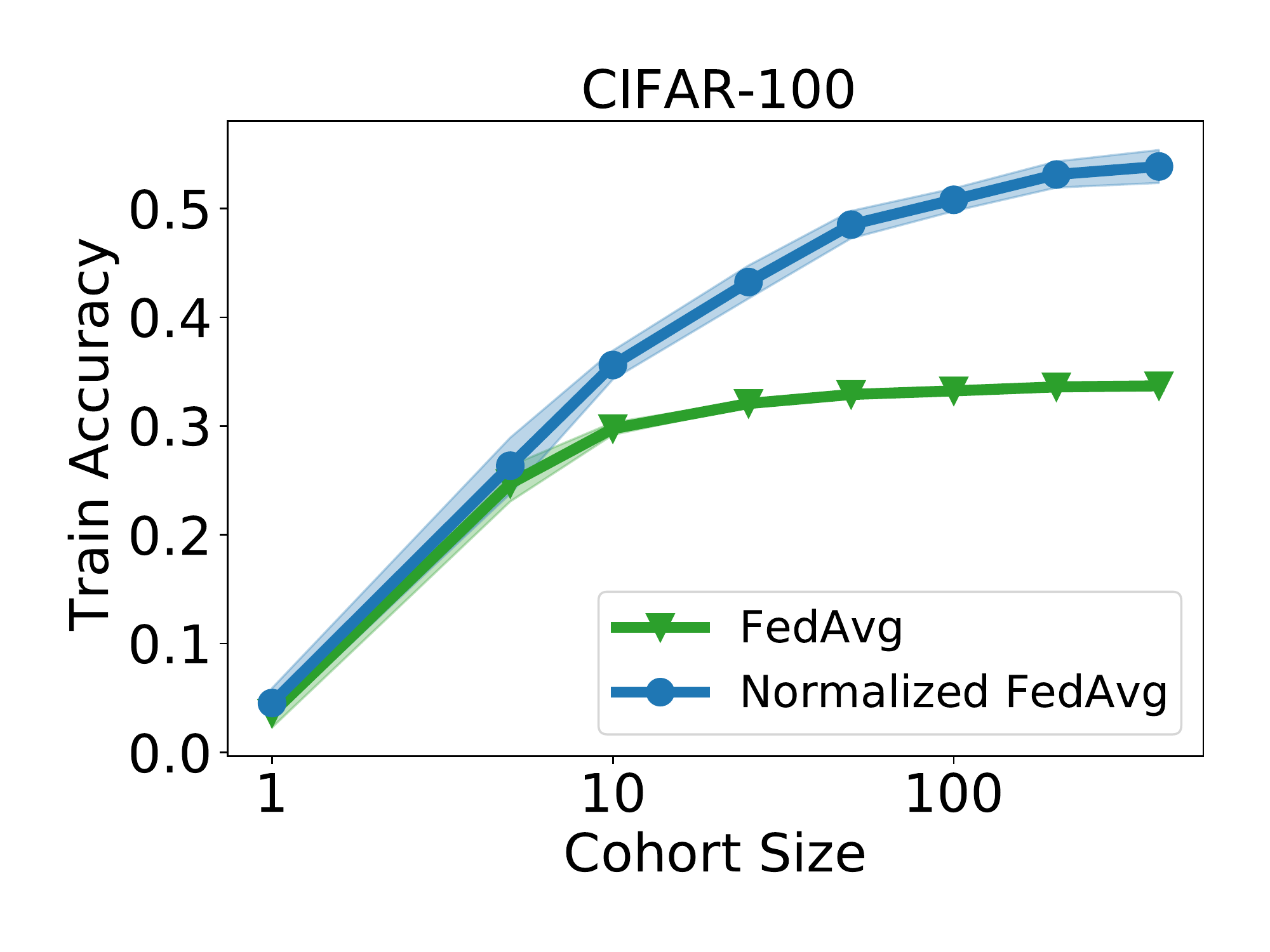}
\end{subfigure}%
\begin{subfigure}{0.24\textwidth}
    \centering
    \includegraphics[width=1\linewidth]{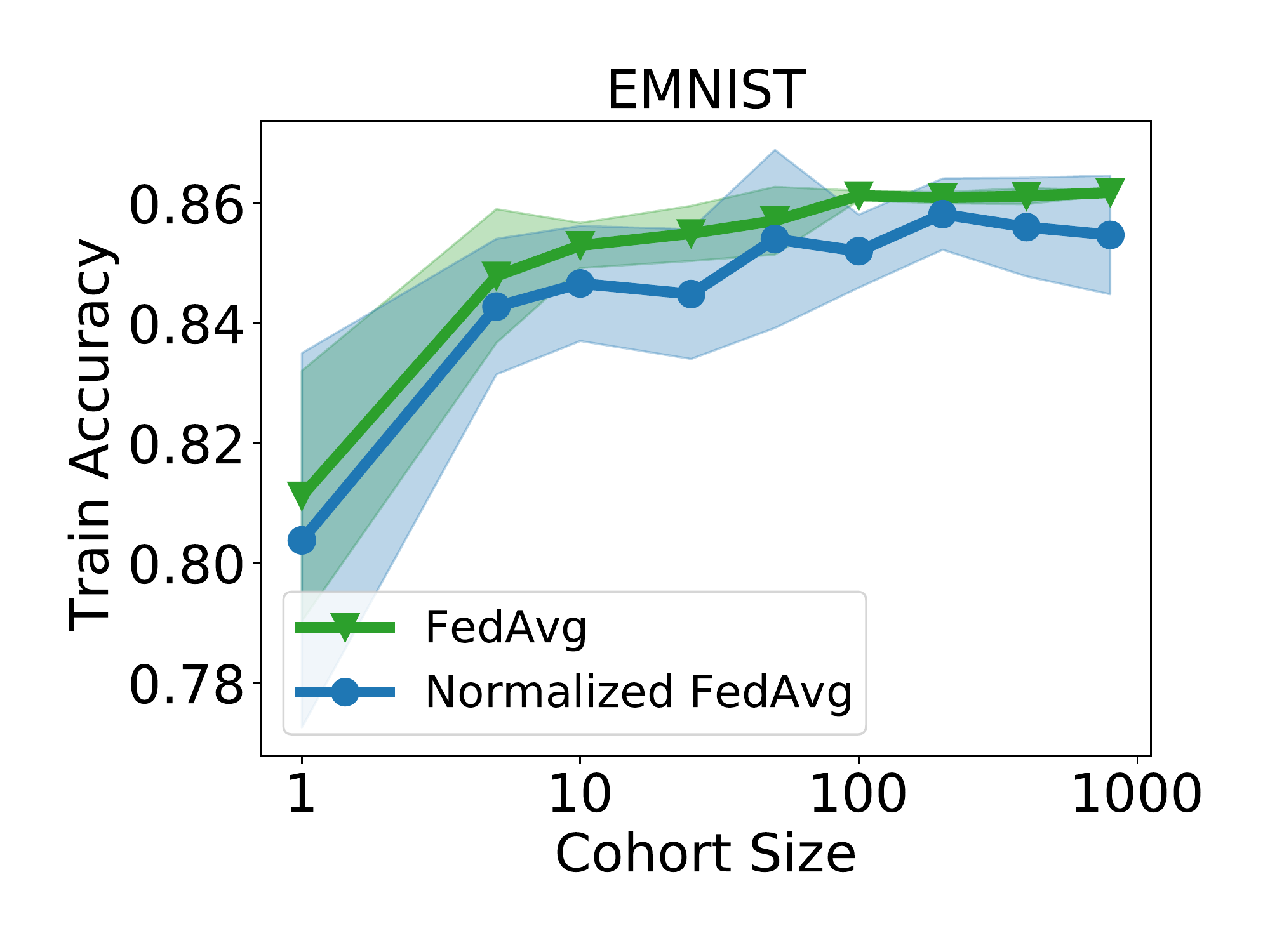}
\end{subfigure}%
\begin{subfigure}{0.24\textwidth}
    \centering
    \includegraphics[width=1\linewidth]{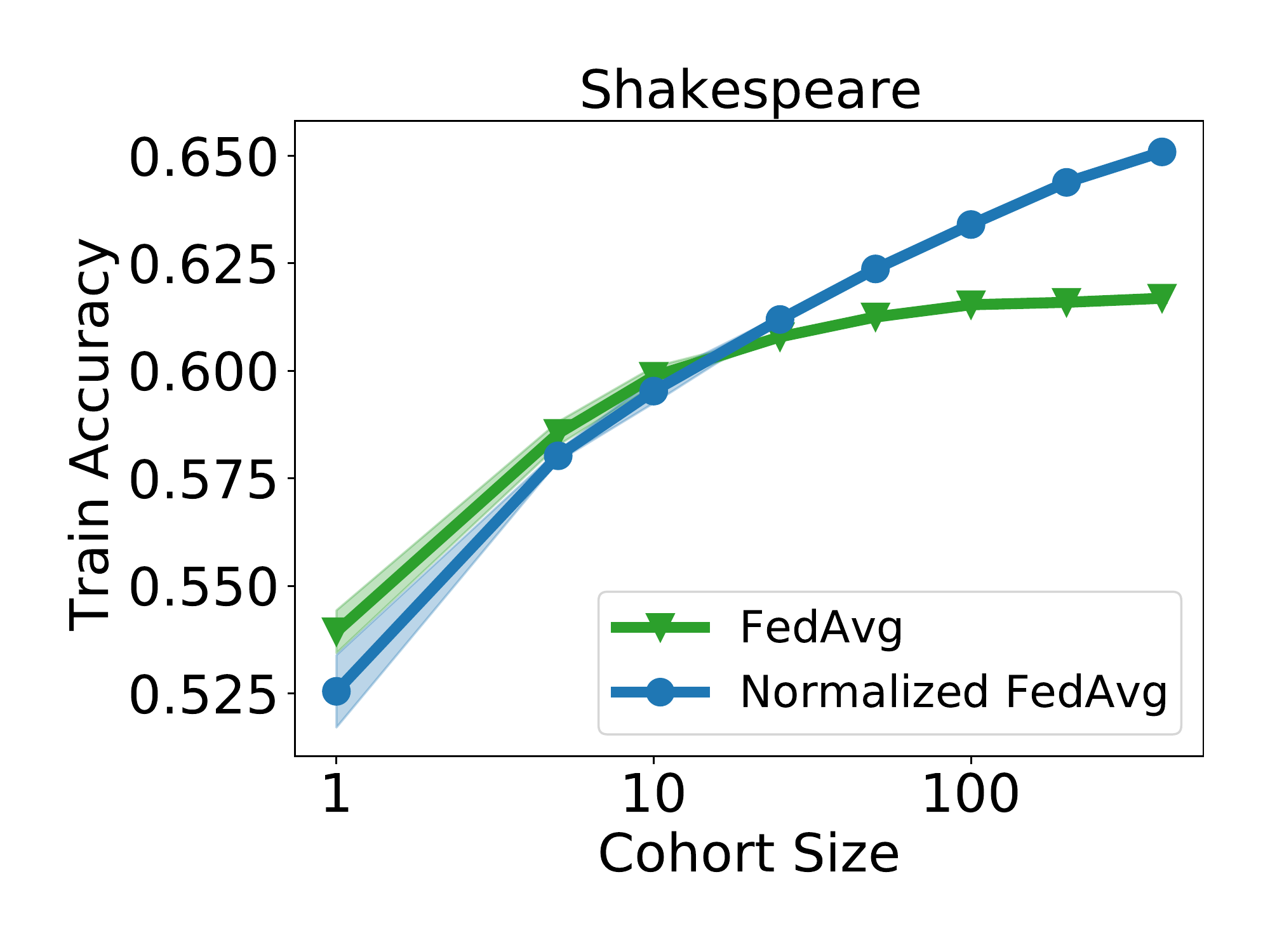}
\end{subfigure}%
\begin{subfigure}{0.24\textwidth}
    \centering
    \includegraphics[width=1\linewidth]{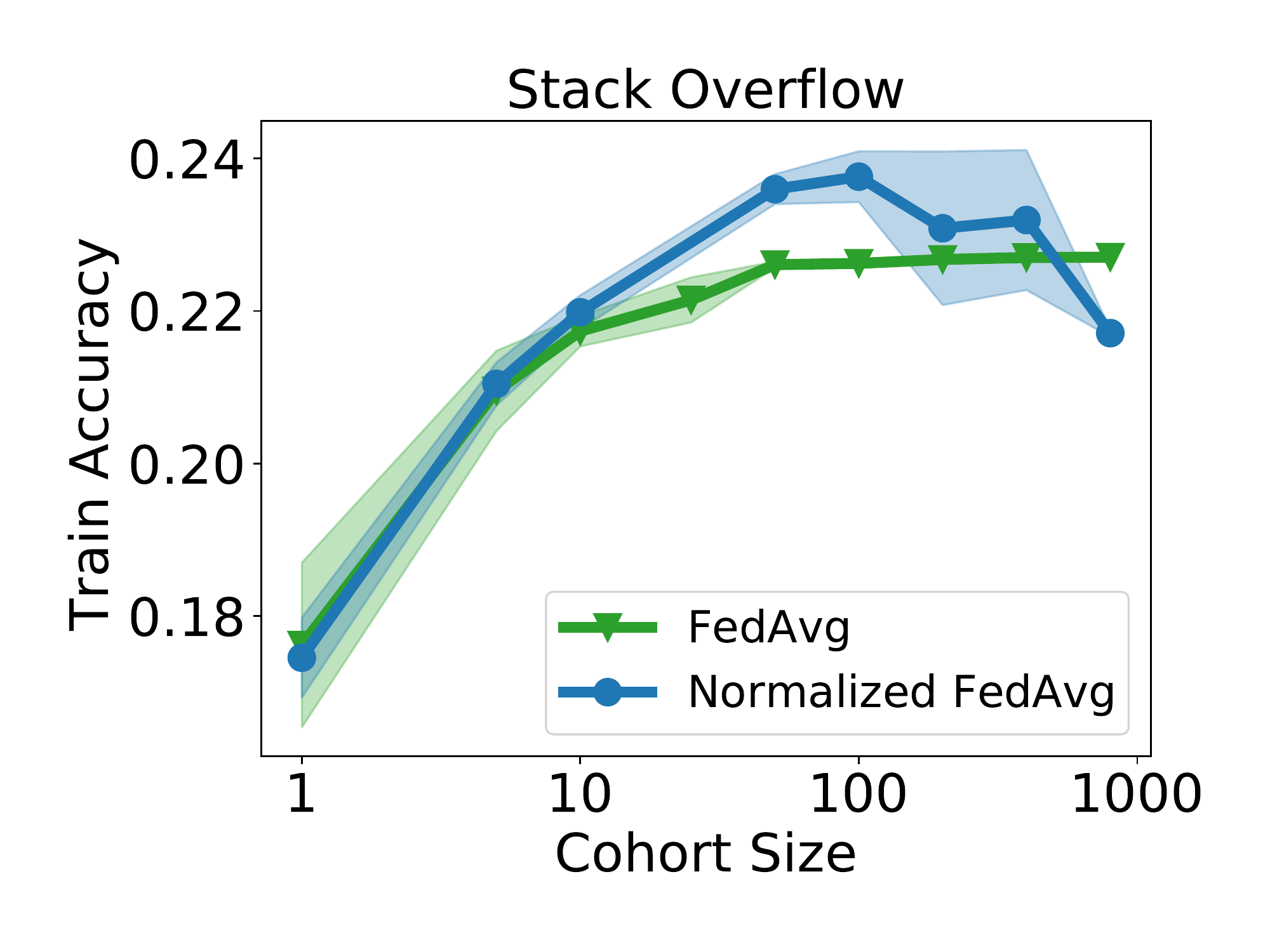}
\end{subfigure}
\begin{subfigure}{0.24\textwidth}
    \centering
    \includegraphics[width=1\linewidth]{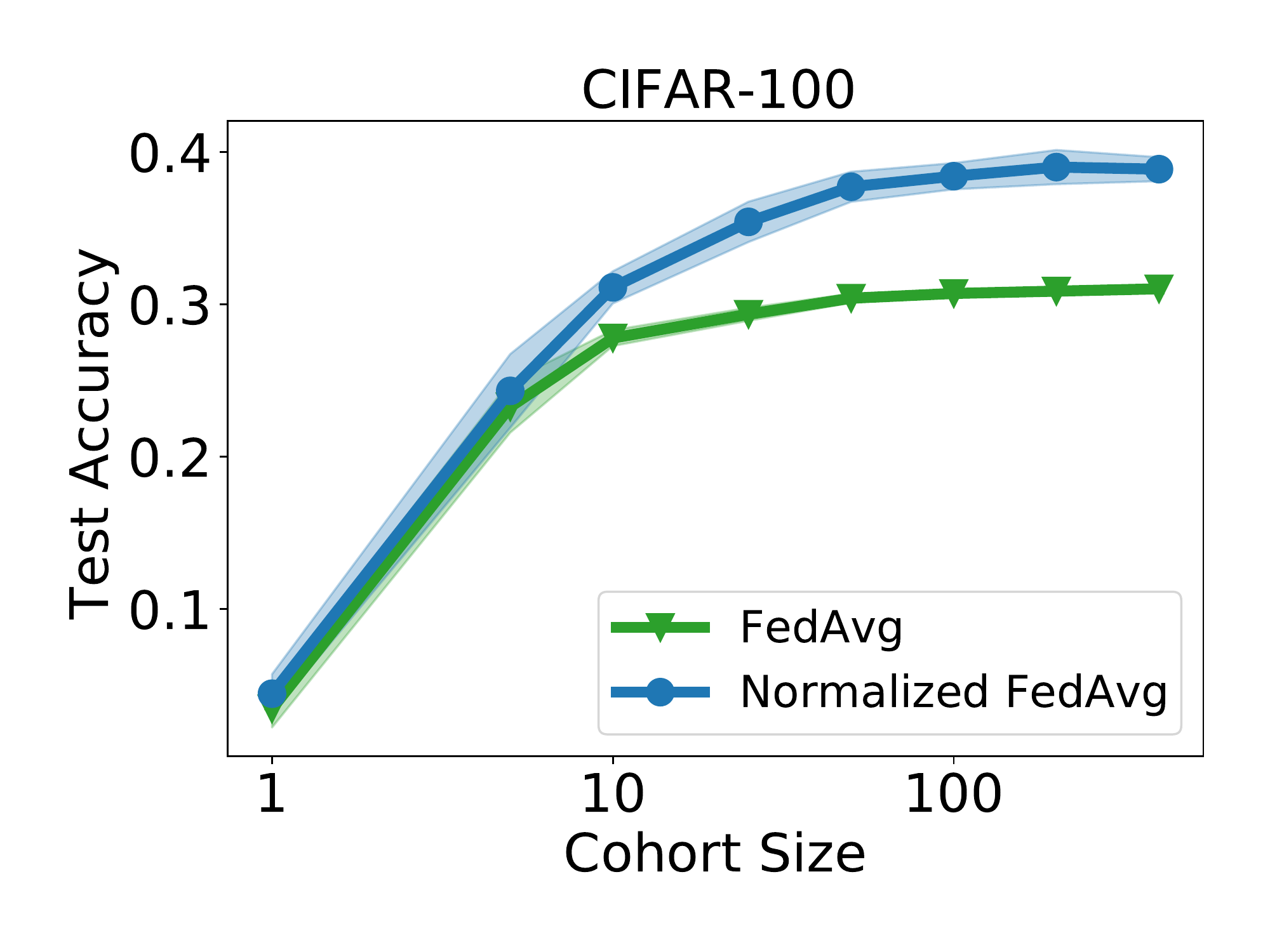}
\end{subfigure}%
\begin{subfigure}{0.24\textwidth}
    \centering
    \includegraphics[width=1\linewidth]{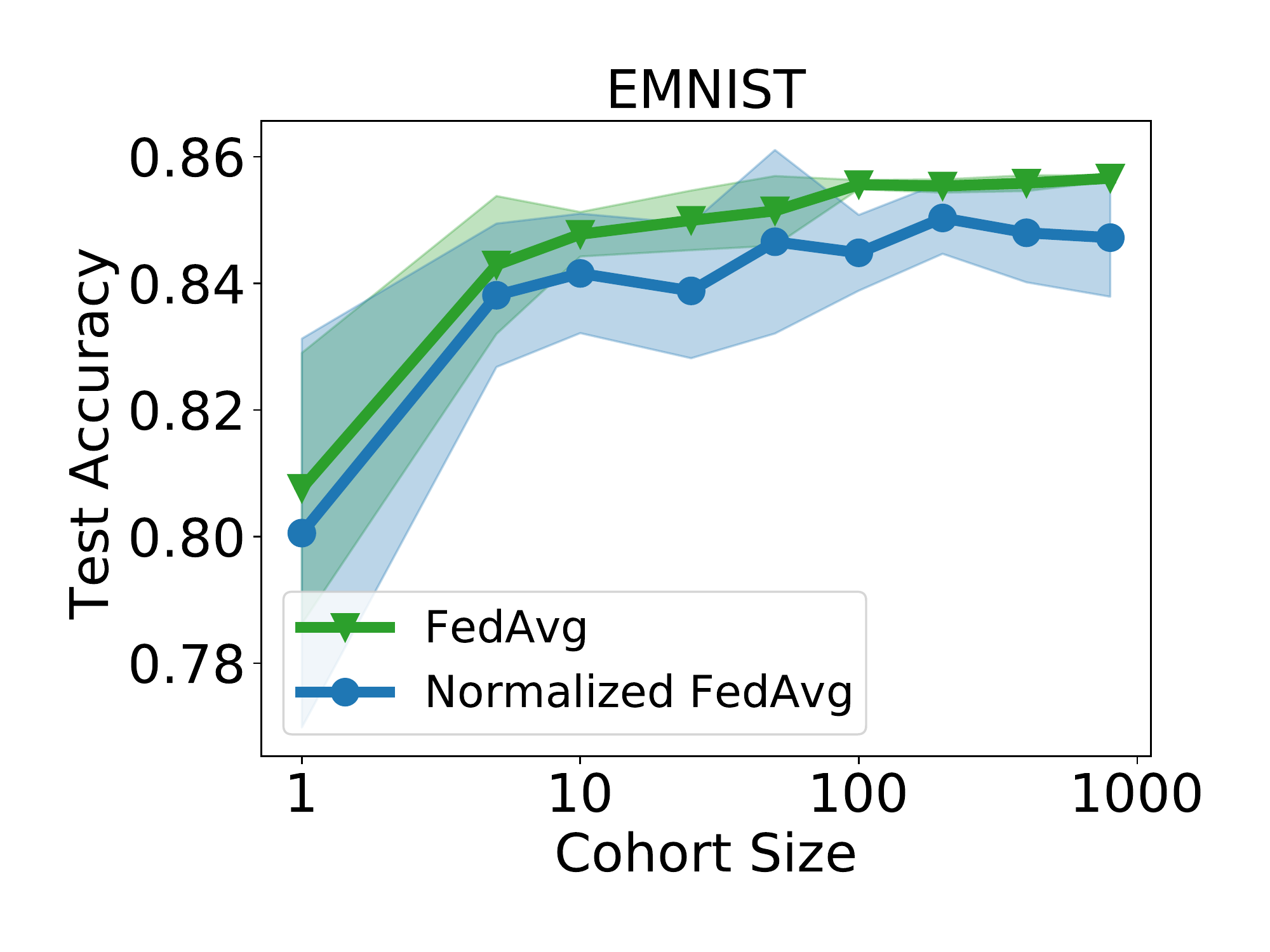}
\end{subfigure}%
\begin{subfigure}{0.24\textwidth}
    \centering
    \includegraphics[width=1\linewidth]{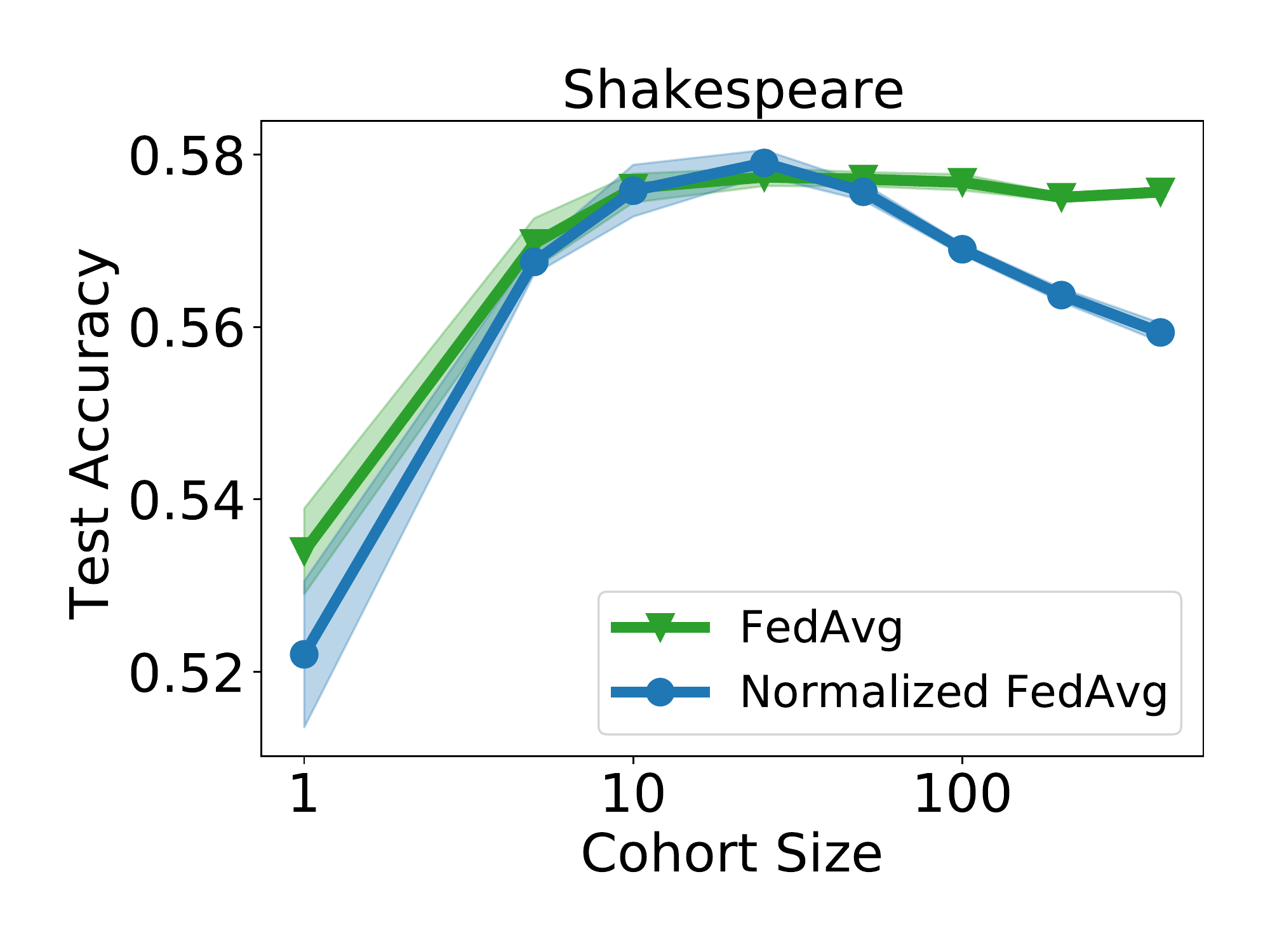}
\end{subfigure}%
\begin{subfigure}{0.24\textwidth}
    \centering
    \includegraphics[width=1\linewidth]{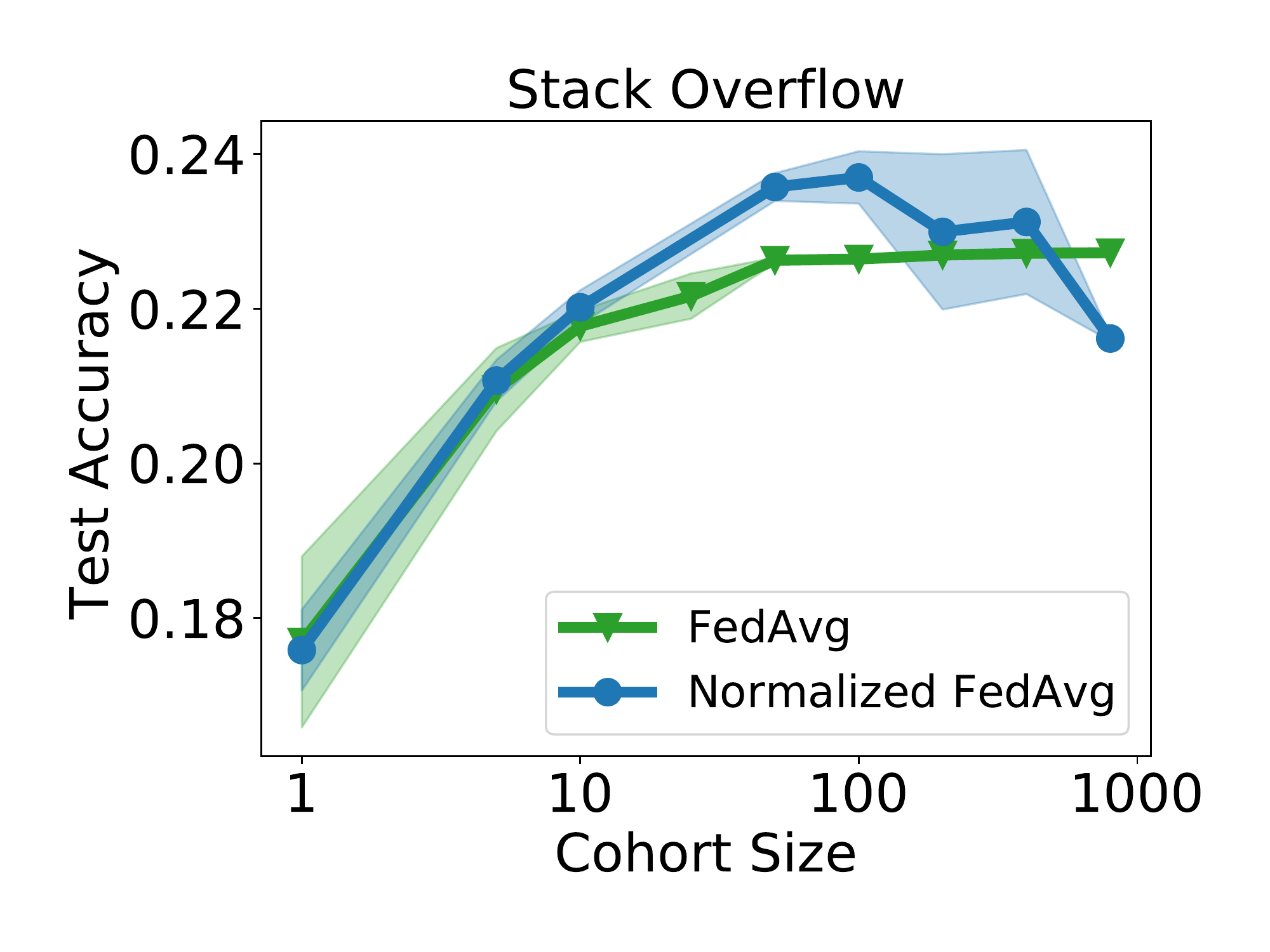}
\end{subfigure}
\caption{The train accuracy (top) and test accuracy (bottom) of \fedavg and the normalized variant of \fedavg, after training for 1500 communication rounds. Results are given for various cohort sizes and tasks}
\label{fig:main_normalized_fedavg}
\end{figure}

Recall that in \cref{sec:diagnosis}, we showed that for \fedavg, the client updates ($\Delta_k^t$ in \cref{alg:fedopt} and \cref{alg:fedopt_with_clipping}) are nearly orthogonal in expectation. By averaging nearly orthogonal updates in large-cohort training, we get a server pseudo-gradient $\Delta^t$ that is close to zero, meaning that the server does not make much progress at each communication round. In order to compensate for this without having to tune a learning rate scaling strategy, we propose a simple variant of \fedavg that tries to account for this near-orthogonality of client updates. Rather than applying \sgd to the server pseudo-gradient (as in \cref{alg:fedopt}), we apply \sgd to the normalized server pseudo-gradient. That is, the server updates its model via
\[
x' = x-\eta_s\frac{\Delta}{\|\Delta\|_2}.
\]
This is a kind of federated analog of normalized \sgd methods used for centralized learning~\citep{nacson2019convergence}. It introduces no new hyperparameters with respect to \cref{alg:fedopt}. To test this method, which we refer to as normalized \fedavg, we present its training and test accuracy versus cohort size in \cref{fig:main_normalized_fedavg}. Notably, we do not re-tune any learning rates. We simply use the same learning rates tuned for (unnormalized) \fedavg.

We find that for most cohort sizes and on most tasks, normalized \fedavg achieves better training accuracy for larger cohorts. Thus, this helps mitigate the diminishing returns issue in \cref{sec:diminishing_returns}. We note two important exceptions: for EMNIST, the normalized \fedavg is slightly worse for all cohort sizes. For Stack Overflow, it obtains worse training accuracy for the largest cohort size. However, we see significant improvements on CIFAR-100 and all but the largest cohort sizes for Stack Overflow. We believe that the method therefore exhibits promising results, and may be improved in future work.

\subsection{Changing the Number of Local Steps}\label{sec:number_of_steps}
While the cohort size has clear parallels to batch size, it is not the only factor determining the number of examples seen per round in \cref{alg:fedopt}. The number of client epochs $E$ and the client batch size also affect this. To study this ``effective batch size'' in FL, we fix the client batch size, and investigate how the cohort size and number of local steps simultaneously impact the performance of \fedavg.

\begin{figure}[ht!]
\centering
\begin{subfigure}{0.45\textwidth}
    \centering
    \includegraphics[width=1\linewidth]{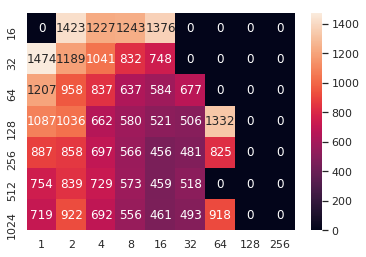}
\end{subfigure}%
\begin{subfigure}{0.45\textwidth}
    \centering
    \includegraphics[width=1\linewidth]{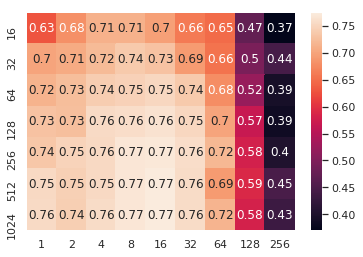}
\end{subfigure}%
\caption{The number of rounds for to reach a test accuracy of $70\%$ (left) and the test accuracy after 1500 rounds (right). Results are for \fedavg on EMNIST with varying numbers of local steps ($x$-axis) and cohort sizes ($y$-axis).
}
\label{fig:main_local_steps_emnist}
\end{figure}

In particular, we fix a local batch size of $1$ and vary the cohort size over $\{16, 32, \dots, 1024\}$. We vary the number of local steps over $\{1, 2, 4,\dots, 256\}$. We plot the number of rounds needed for convergence, and the final test accuracy in Figure \ref{fig:main_local_steps_emnist}. By construction, each square on an anti-diagonal corresponds to the same number of examples per round.

In the left figure, we see that if we fix the cohort size, then increasing the number of local steps can accelerate convergence, but only up to a point, after which catastrophic training failures occur. By contrast, if we have convergence for some number of local steps and cohort size, convergence occurs for all cohort sizes. Similarly, we see in the right hand figure that increasing the number of local steps can drastically reduce generalization, more so than increasing the cohort size. In essence, we see that the number of local steps obeys many of the same issues outlined in \cref{sec:challenges}. Therefore, correctly tuning the number of local steps in unison with the cohort size may be critical to ensuring good performance of large-cohort methods.

\section{Limitations and Future Work}\label{sec:future_work}

In this work we explore the benefits and limitations of large-cohort training in federated learning. As discussed in Sections \ref{sec:data_efficiency} and \ref{sec:better_methods}, focusing on the number of communication rounds often obscures the data efficiency of a method. This in turn impacts many metrics important to society, such as total energy consumption or total carbon emissions. While we show that large-cohort training can negatively impact such metrics by reducing data-efficiency (see \cref{sec:data_efficiency} and \cref{appendix:stragglers}), a more specialized focus on these issues is warranted. Similarly, we believe that an analysis of fairness in large-cohort settings going beyond \cref{subsec:generalization_failures} would be beneficial. 

Future work also involves connecting large-cohort training to other important aspects of federated learning, and continuing to explore connections with growing lines of work in large-batch training. In particular, we wish to see whether noising strategies, especially differential privacy mechanisms, can help overcome the generalization issues of large-cohort training. Personalization may also help mitigate issues of generalization and fairness. Finally, although not a focus of our work, we note that some of the findings above may extend to cross-silo settings, especially if communication restrictions require subsampling clients.

\bibliographystyle{plainnat}
\bibliography{bibliography}

\appendix
\section{Full Experimental Details}\label{appendix:experiment_details}

\subsection{Datasets and Models}\label{appendix:models_and_datasets}

We use four datasets throughout our work: CIFAR-100~\citep{krizhevsky2009learning}, the federated extended MNIST dataset (EMNIST)~\citep{cohen2017emnist}, the Shakespeare dataset~\citep{caldas2018leaf}, and the Stack Overflow dataset~\citep{stackoverflow}. The first two datasets are image datasets, the second two are language datasets. All datasets are publicly available. We specifically use the versions available in TensorFlow Federated~\citep{tff}, which gives a federated structure to all four datasets. Below we discuss the specifics of the dataset and classification task, as well as the model used to perform classification.

\paragraph{CIFAR-100} The CIFAR-100 dataset is a computer vision dataset consisting of $32 \times 32 \times 3$ images with 100 possible labels. While this dataset does not have a natural partition among clients, a federated version was created by \citet{reddi2021adaptive} using hierarchical latent Dirichlet allocation to enforce moderate amounts of heterogeneity among clients. This partitioning among clients was based on Pachinko allocation~\citep{li2006pachinko}. Note that under this partitioning, each client typically has only a subset of the 100 possible labels. The dataset has 500 training clients and 100 test clients, each with 100 examples in their local dataset.

We train a ResNet-18~\citep{he2016deep} on this dataset, where we replace all batch normalization layers with group normalization layers~\citep{wu2018group}. The use of group norm over batch norm in federated learning was first advocated by \citet{hsieh2019non}, who showed that this helped improve classification accuracy in the presence of heterogeneous clients. We specifically use group normalization layers with two groups. We perform small amounts of data augmentation and preprocessing for each train and test sample. We first centrally crop each image $(24, 24, 3)$. We then normalize the pixel values according to their mean and standard deviation.

\paragraph{EMNIST} The EMNIST dataset consists of images hand-written alphanumeric characters. Each image consists of $28 \times 28$ gray-scale pixel values. There are 62 total alphanumeric characters represented in the dataset. The images are partitioned among clients according to their author. The dataset has 3,400 clients, who have both train and test datasets. The dataset has natural heterogeneity stemming from the writing style of each person. We train a convolutional network on the dataset (the same one used by \citet{reddi2021adaptive}). The network uses two convolutional layers (each with $3\times 3$ kernels and strides of length 1), followed by a max pooling layer using dropout with $p = 0.25$, a dense layer with 128 units and dropout with $p = 0.5$, and a final dense output layer.

\paragraph{Shakespeare} The Shakespeare dataset is derived from the benchmark designed by \citet{caldas2018leaf}. The dataset corpus is the collected works of William Shakespeare, and the clients correspond to roles in Shakespeare's plays with at least two lines of dialogue. To eliminate confusion, \emph{character} here will refer to alphanumeric characters (such as the letter \textit{q}) and symbols such as punctuation, while we will use \emph{client} to denote the various roles in plays (such as Macbeth). There are a total of 715 clients, whose lines are partitioned between train and test datasets.

We split each client's lines into sequences of 80 characters, padding if necessary. We use a vocabulary size of 90, where 86 characters are contained in Shakespeare's work, and the remaining 4 are beginning and end of line tokens, padding tokens, and out-of-vocabulary tokens. We perform next-character prediction on the clients' dialogue using a recurrent neural network (RNN)~\citep{mikolov2010recurrent}. We use the same model as \citet{reddi2021adaptive}. The RNN takes as input a sequence of 80 characters, embeds it into a learned 8-dimensional space, and passes the embedding through 2 LSTM layers~\citep{hochreiter1997long}, each with 256 units. Finally, we use a softmax output layer with 80 units, where we try to predict a sequence of 80 characters formed by shifting the input sequence over by one. Therefore, our output dimension is $80\times 90$. We compute loss using cross-entropy loss.

\paragraph{Stack Overflow} Stack Overflow is a language dataset consisting of question and answers from the Stack Overflow site. The questions and answers also have associated metadata, including tags. Each client corresponds to a user. The specific train/validation/test split from~\citep{stackoverflow} has 342,477 train clients, 38,758 validation clients, and 204,088 test clients. Notably, the train clients only have examples from before 2018-01-01 UTC, while the test clients only have examples from after 2018-01-01 UTC. The validation clients have examples with no date restrictions, and all validation examples are held-out from both the test and train sets.

We perform next-word prediction on this dataset. We restrict each client to the first 1000 sentences in their dataset (if they contain this many, otherwise we use the full dataset). We also perform padding and truncation to ensure that each sentence has 20 words. We then represent the sentence as a sequence of indices corresponding to the 10,000 most frequently used words, as well as indices representing padding, out-of-vocabulary words, the beginning of a sentence, and the end of a sentence. We perform next-word-prediction on these sequences using an a recurrent neural network (RNN)~\citep{mikolov2010recurrent} that embeds each word in a sentence into a learned 96-dimensional space. It then feeds the embedded words into a single LSTM layer~\citep{hochreiter1997long} of hidden dimension 670, followed by a densely connected softmax output layer. Note that this is the same model used by \citet{reddi2021adaptive}. The metric used in the main body is the accuracy over the 10,000-word vocabulary; it does not include padding, out-of-vocab, or beginning or end of sentence tokens when computing the accuracy.

\subsection{Implementation and Hyperparameters}

We implement the previously proposed methods of \fedavg, \fedsgd, \fedavgm, \fedadam, \fedadagrad, as well as two novel methods, \fedlars and \fedlamb. All implementations are special cases of \cref{alg:fedopt}. In all cases, clients use mini-batch \sgd with batch size $B$. For \fedsgd, the batch size $B$ of a client is set to the size of its local dataset (so that the client only takes a single step). For all other optimizers, we fix $B$ at a per-task level (see \cref{table:batch_sizes}). Note that we use larger batch sizes for datasets where clients have more examples, like Stack Overflow. Except for the experiments in \cref{sec:number_of_steps}, we set $E = 1$ throughout.

\begin{table}[ht]
    \centering
    \caption{Batch sizes used for each for all algorithms (except for \fedsgd) on each dataset.}
    \label{table:batch_sizes}
    \begin{center}
    \begin{small}
    \begin{sc}
    \begin{tabular}[t]{@{}cc@{}}
    \toprule
    Dataset & Batch Size \\
    \midrule
    CIFAR-100 & 20 \\
    EMNIST & 20 \\
    Shakespeare & 4 \\
    Stack Overflow & 32 \\
    \bottomrule
    \end{tabular}
    \end{sc}
    \end{small}
    \end{center}
\end{table}

For the actual implementation of the algorithms above, all methods (except for \fedsgd) differ only in the choice of \serveropt in \cref{alg:fedopt}. For \fedsgd, in addition to having clients use full-batch \sgd (as mentioned above), the client learning rate is set to be $\eta_c = 1$ in order to allow \cref{alg:fedopt} to recover the version of \fedsgd proposed by \citet{mcmahan2017aistats}. For all other algorithms, we present the choice of \serveropt and relevant hyperparameters (except for learning rates, see Section \ref{appendix:learning_rates}) in Table \ref{table:server_opt}. Note that here we use the notation from \citep{kingma2014adam}, where $\beta_1$ refers to a first-moment momentum parameter, $\beta_2$ refers to a second-moment momentum parameter, and $\epsilon$ is a numerical stability constant used in adaptive methods. Note that for all adaptive methods, we set their initial accumulators to be 0.

\begin{table}[ht]
    \centering
    \caption{Hyperparameters and implementation details for all algorithms, relative to \cref{alg:fedopt}. Here, $\beta_1$ denotes a first-moment momentum parameter, $\beta_2$ denotes a second-moment momentum parameter, and $\epsilon$ is a value used for numerical stability purposes in adaptive methods.}
    \label{table:server_opt}
    \begin{center}
    \begin{tabular}[t]{@{}cccccc@{}}
    \toprule
    \textsc{Algorithm} & \serveropt & $\beta_1$ & $\beta_2$ & $\epsilon$\\
    \midrule
    \fedavg~\citep{mcmahan2017aistats} & \sgd & 0 & N/A & N/A \\
    \fedavgm~\citep{hsieh2019non} & \sgd & 0.9 & N/A & N/A \\
    \fedadagrad~\citep{reddi2021adaptive} & Adagrad~\citep{duchi2011adaptive} & N/A & N/A & 0.001 \\
    \fedadam~\citep{reddi2021adaptive} & Adam~\citep{kingma2014adam} & 0.9 & 0.99 & 0.001 \\
    \fedlars & LARS~\citep{you2017large} & 0.9 & N/A & 0.001 \\
    \fedlamb & Lamb~\citep{you2019large} & 0.9 & 0.99 & 0.001 \\
    \bottomrule
    \end{tabular}
    \end{center}
\end{table}

\subsection{Adaptive Clipping}\label{appendix:adaptive_clipping} 

As exemplified in \cref{fig:catastrophic_failure_example}, catastrophic training failures can occur when the server pseudo-gradient $\Delta^t$ is too large, which occurs more frequently for larger cohort sizes. To mitigate this issue, we use the adaptive clipping method proposed by \citet{andrew2019differentially}. While we encourage the reader to see this paper for full details and motivation, we give a brief overview of the method below.

Recall that in \cref{alg:fedopt}, $\Delta^t$ is an average of client updates $\Delta^t_k$. Thus, $\Delta^t$ can only be large if some client update is also large. In order to prevent this norm blow-up, we clip the client updates before averaging them. Rather than send $\Delta_k^t$ to the server, for a clipping level $\rho > 0$, the clients send $h(\Delta_k^t, \rho)$ where
\[
h(v, \rho) = \begin{cases}
                v, & \text{ if }\|v\| \leq \rho \\
                \dfrac{\rho v}{\|v\|}, &\text{ if }\|v\| > \rho.
            \end{cases}
\]

Instead of fixing $\rho$ a priori, we use the adaptive method proposed by \citet{andrew2019differentially}. In this method, the clipping level varies across rounds, and is adaptively updated via a geometric update rule, where the goal is for $\rho$ to estimate some norm percentile $q \in [0, 1]$. Notably, \citet{andrew2019differentially} show that the clipping level can be learned in a federated manner that is directly compatible with \cref{alg:fedopt}. At each round $t$, let $\rho^t$ be the clipping level (intended to estimate the $q$th percentile of norms across clients), and let $C_t$ be the cohort of clients selected. Each client $k \in C_t$ computes their local model update $\Delta_k^t$ in the same manner as in \cref{alg:fedopt}. Instead of sending $\Delta_k^t$ to the server, the client instead sends their clipped update $h(\Delta_k^t, \rho^t)$ to the server, along with $b_k^t := \mathbb{I}[\|\Delta_k^t\| \leq \rho^t]$, where $\mathbb{I}[A]$ denotes the indicator function of an event $A$. The server then computes:
\[
\Delta^t = \dfrac{\sum_{k \in C_t} p_kh(\Delta_k^t)}{\sum_{k \in C_t} p_k},~~~b^t = \dfrac{1}{|C_t|}\sum_{k \in C_t} b_k^t.
\]

That is, $\Delta^t$ is a weighted average of the clipped client updates, and $b^t$ is the fraction of unclipped client updates that did not exceed the clipping threshold. The server then updates its global model as in \eqref{eq:server_update}, but it also updates its estimate of the $q$th norm percentile using a learning rate $\eta_a > 0$ via
\begin{equation}\label{eq:clip_update}
    \rho^{t+1} = \rho^t\exp(-\eta_{a}(b^t - q)).
\end{equation}

While \citet{andrew2019differentially} add noise in order to ensure that $\rho$ is learned in a differentially private manner, we do not use such noise. Full pseudo-code combining \cref{alg:fedopt} and the adaptive clipping mechanisms discussed above is given in \cref{alg:fedopt_with_clipping}.

\setlength{\textfloatsep}{10pt}
\begin{algorithm}[ht]
    \begin{algorithmic}
	\caption{\fedopt framework with adaptive clipping}
	\label{alg:fedopt_with_clipping}
	\STATE {\bf Input:}  $M$, $T$ $E$, $x^1$, $\eta_c$, $\eta_s$, $\eta_a$, $q$, $\rho^1$, \serveropt, $\{p_k\}_{k=1}^K$
	\FOR  {$t=1, \cdots, T$}
		\STATE The server selects a cohort $C_t$ of $M$ clients uniformly at random, without replacement.
		\STATE The server sends $x^t, \rho^t$ to all clients in $C_t$.
		\STATE Each client $k \in C_t$ updates $x^t$ for $E$ epochs of mini-batch \sgd with step-size $\eta_c$ on $f_k$.
		\STATE After training, each client has a local model $x_k^t$.
		\STATE Each client $k \in C_t$ computes $\Delta_k^t = x^t - x_k^t$ and $b_k^t = \mathbb{I}[\|\Delta_k^t\| \leq \rho^t].$
		\STATE Each client $k \in C_t$ computes
		\[
		h(\Delta_k^t) = \Delta_k^t\min\left\{1, \frac{\rho^t}{\|\Delta_k^t\|}\right\}.
		\]
		\STATE Each client $k \in C_t$ sends $h(\Delta_k^t)$ and $b_k^t$ to the server.
		\STATE The server computes a pseudo-gradient $\Delta^t$ and updates its model via
		\[
		\Delta^t = \dfrac{\sum_{k \in C_t} p_kh(\Delta_k^t)}{\sum_{k \in C_t} p_k},~~~x^{t+1} = \serveropt(x_t, \eta_s, \Delta^t).
		\]
		\STATE The server updates its clipping level via
		\[
		b^t = \dfrac{1}{|C_t|}\sum_{k \in C_t} b_k^t,~~~\rho^{t+1} = \rho^t\exp(-\eta_{a}(b^t - q)).
		\]
    \ENDFOR
	\end{algorithmic}
\end{algorithm}

\textbf{Usage and hyperparameters.} We use \cref{alg:fedopt_with_clipping} in all experiments (save for those in \cref{fig:catastrophic_failure_example}, which illustrate the potential failures that can occur if clipping is not used). For hyperparameters, we use a target percentile of $q = 0.8$, with an initial clipping level of $\rho_1 = 1$. In our geometric update rule, we use a learning rate of $\eta_a = 0.2$.

\subsection{Learning Rates and Tuning}\label{appendix:learning_rates}

For our experiments, we use client and server learning rates $\eta_s, \eta_c$ that are tuned a priori on a held-out validation dataset. We tune both learning rates over $\{10^i~|~-3 \leq i \leq 1\}$ for each algorithm and dataset, therefore resulting in 25 possible configurations for each pair. This tuning, like the experiments following it, is based on the algorithm implementations discussed above. In particular, the tuning also uses the adaptive clipping framework discussed in \cref{appendix:adaptive_clipping} and \cref{alg:fedopt_with_clipping}.

\begin{table}[ht]
    \centering
    \caption{Server learning rate $\eta_s$ used for each algorithm and dataset.}
    \label{table:server_learning_rates}
    \begin{center}
    \begin{tabular}[t]{@{}lcccc@{}}
    \toprule
    \textsc{Algorithm} & \multicolumn{4}{c}{\textsc{Dataset}} \\
    \cmidrule(r){2-5}
     & CIFAR-100 & EMNIST & Shakespeare & Stack Overflow \\
    \fedavg & 1 & 1 & 1 & 1 \\
    \fedavgm & 1 & 1 & 0.1 & 1 \\
    \fedadagrad & 0.01 & 0.1 & 0.1 & 10 \\
    \fedadam & 0.01 & 0.001 & 0.01 & 1 \\
    \fedlars & 0.01 & 0.001 & 0.01 & 0.01 \\
    \fedlamb & 0.001 & 0.01 & 0.01 & 0.01 \\
    \fedsgd & 0.1 & 0.1 & 1 & 10 \\
    \bottomrule
    \end{tabular}
    \end{center}
\end{table}

\begin{table}[ht]
    \centering
    \caption{Client learning rate $\eta_c$ used for each algorithm and dataset.}
    \label{table:client_learning_rates}
    \begin{center}
    \begin{tabular}[t]{@{}lcccc@{}}
    \toprule
    \textsc{Algorithm} & \multicolumn{4}{c}{\textsc{Dataset}} \\
    \cmidrule(r){2-5}
     & CIFAR-100 & EMNIST & Shakespeare & Stack Overflow \\
    \fedavg & 0.1 & 0.1 & 1 & 10 \\
    \fedavgm & 0.1 & 0.1 & 1 & 10 \\
    \fedadagrad & 0.1 & 0.001 & 10 & 10 \\
    \fedadam & 0.1 & 0.1 & 10 & 10 \\
    \fedlars & 0.1 & 0.1 & 10 & 1 \\
    \fedlamb & 0.01 & 0.1 & 10 & 10 \\
    \bottomrule
    \end{tabular}
    \end{center}
\end{table}

While Stack Overflow has an explicit validation set distinct from the test and train datasets~\citep{stackoverflow}, the other three datasets do not. In order to tune on these datasets, we randomly split the training clients (not the training examples!) into train and validation subsets according to an 80-20 split. We then use these federated datasets to perform held-out set tuning. We select the learning rates that have the best average validation performance after 1500 communication rounds with cohort size $M = 10$ over 5 random trials. A table of the resulting learning rates is given in Tables \ref{table:server_learning_rates} and \ref{table:client_learning_rates}. Note that there is no client learning rate for \fedsgd, as we must use $\eta_c = 1$ in \cref{alg:fedopt} in order to recover the version of \fedsgd in \citep{mcmahan2017aistats}. Note that we use the same learning rates for all cohort sizes.

\section{Full Experiment Results}\label{appendix:full_results}


\subsection{Test Accuracy Versus Communication Round}\label{appendix:test_accuracy_versus_rounds}

\begin{figure}[ht!]
\centering
\begin{subfigure}{0.24\textwidth}
      \centering
      \includegraphics[width=1\linewidth]{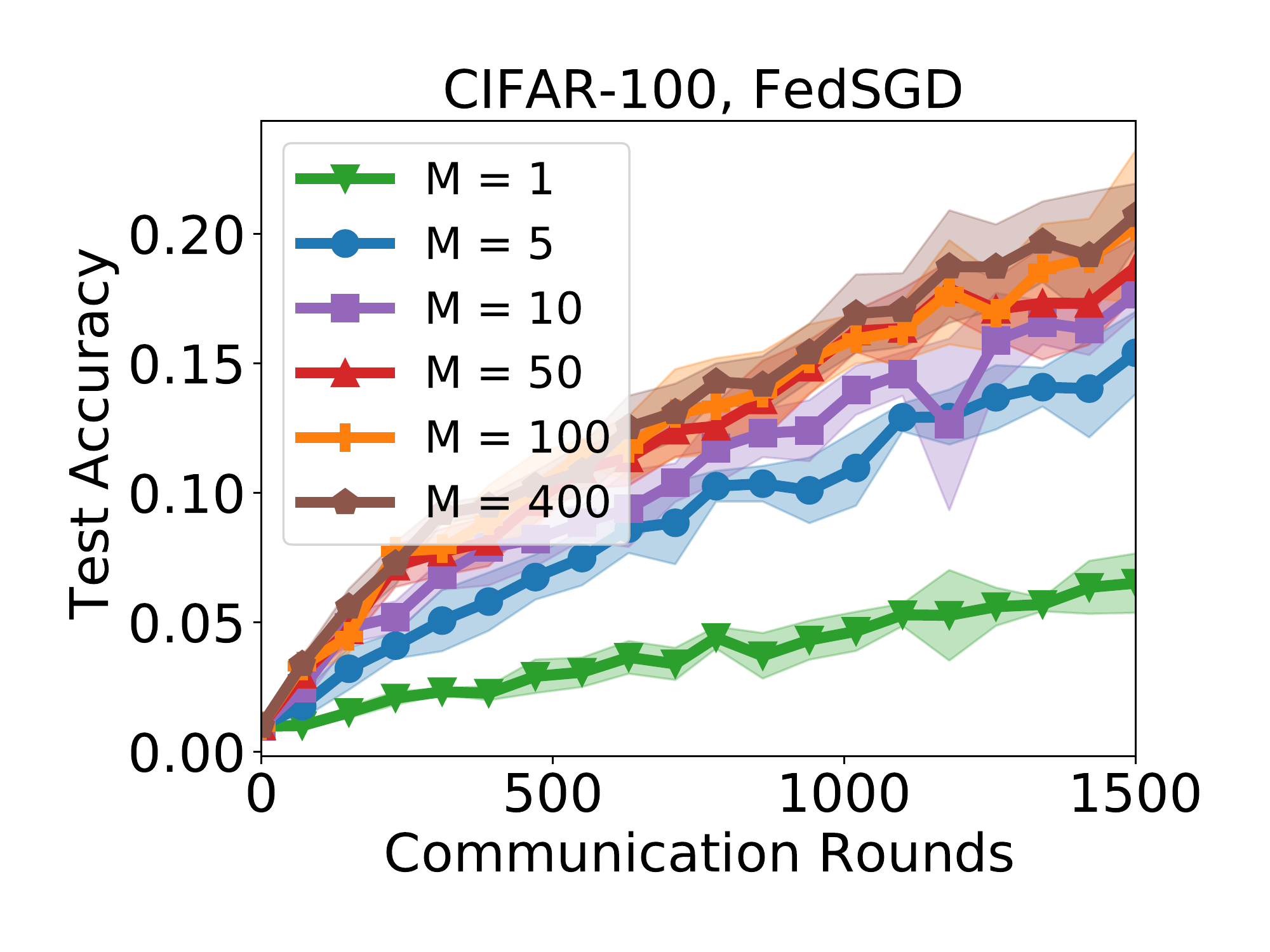}
\end{subfigure}%
\begin{subfigure}{0.24\textwidth}
      \centering
      \includegraphics[width=1\linewidth]{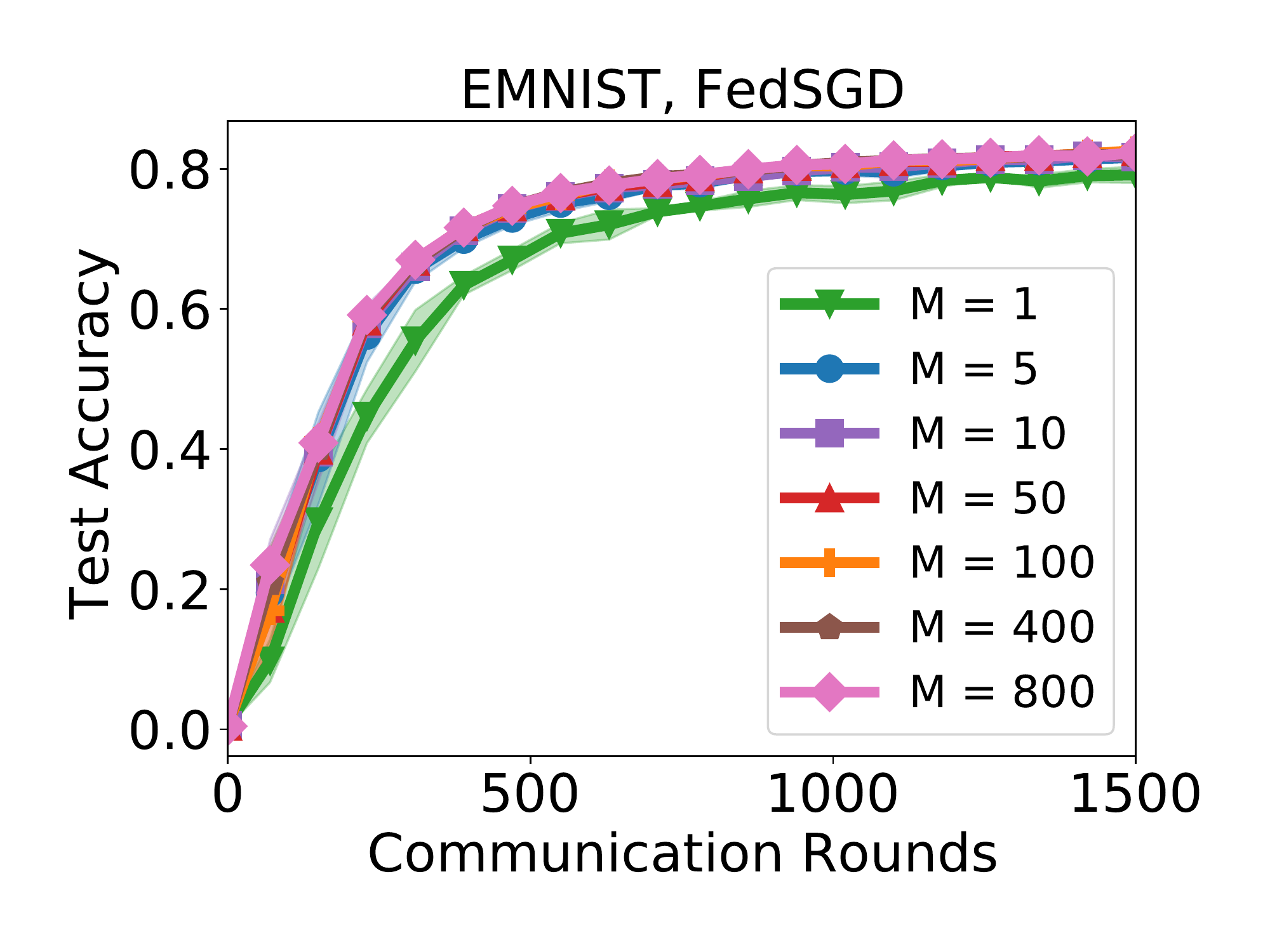}
\end{subfigure}%
\begin{subfigure}{0.24\textwidth}
      \centering
      \includegraphics[width=1\linewidth]{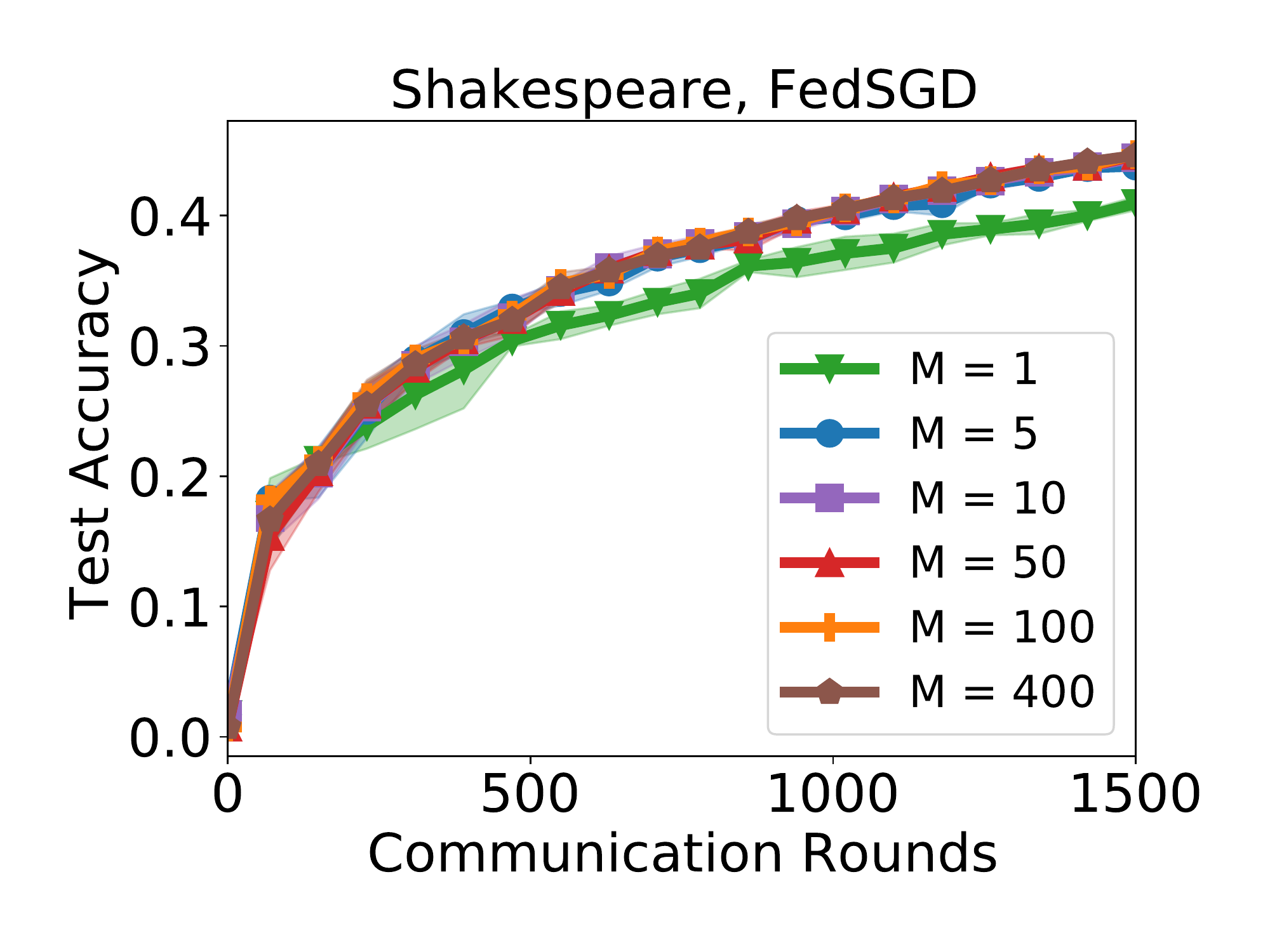}
\end{subfigure}%
\begin{subfigure}{0.24\textwidth}
      \centering
      \includegraphics[width=1\linewidth]{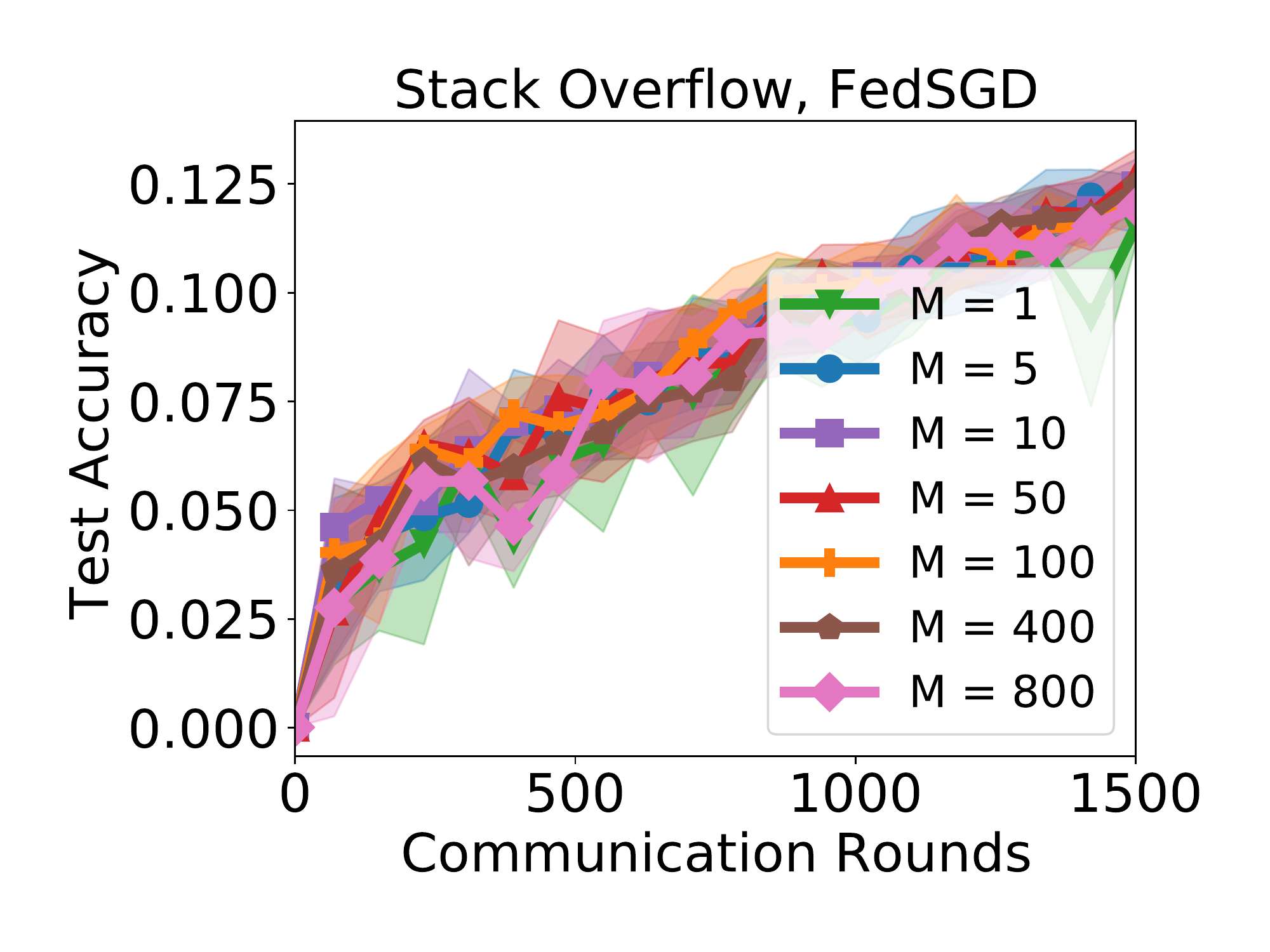}
\end{subfigure}%
\caption{Average test accuracy of \fedsgd versus the number of communication rounds, for various tasks and cohort sizes $M$.}
\label{fig:fedsgd_test_accuracy_versus_rounds}
\end{figure}

\begin{figure}[ht!]
\centering
\begin{subfigure}{0.24\textwidth}
      \centering
      \includegraphics[width=1\linewidth]{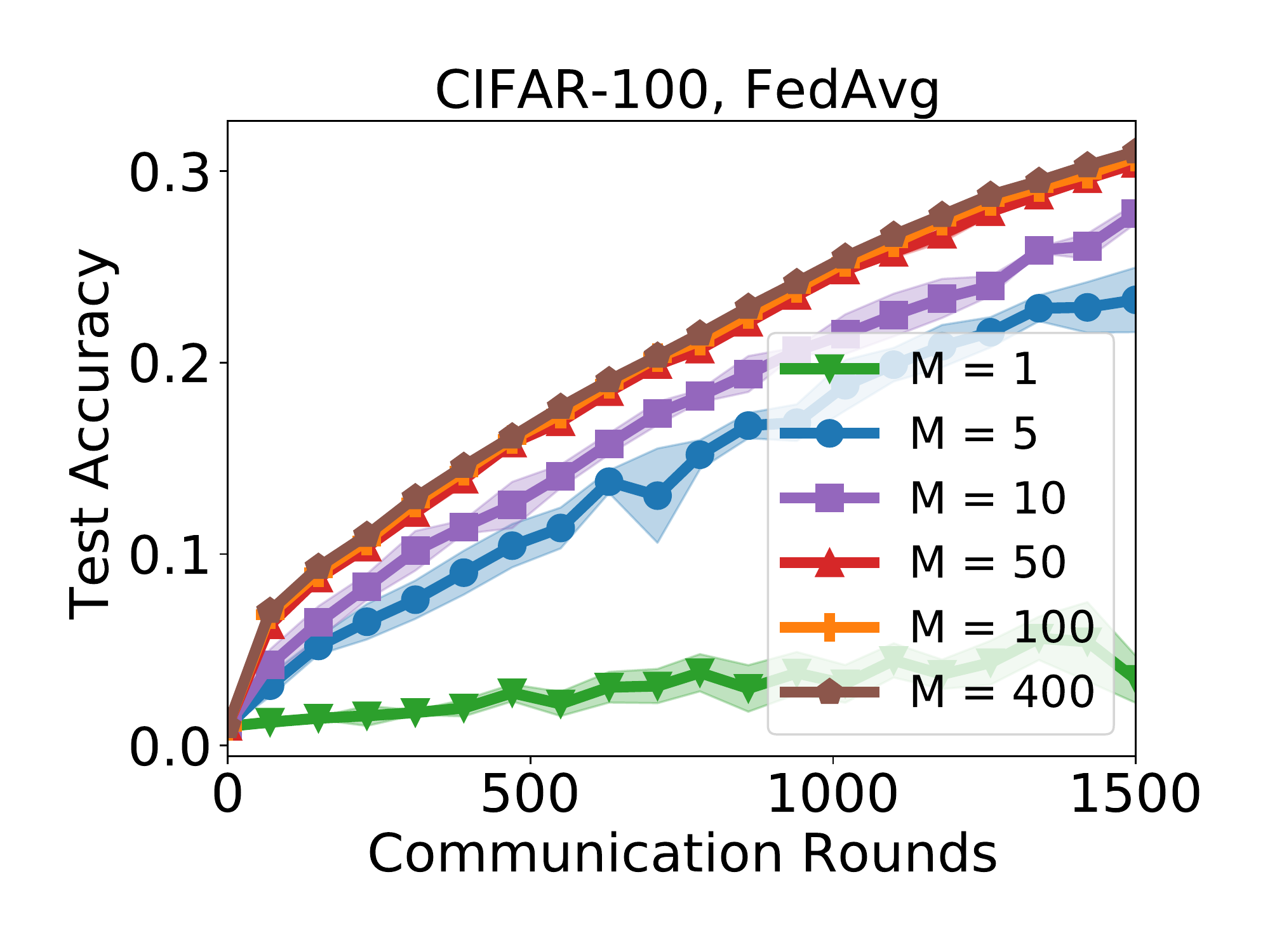}
\end{subfigure}%
\begin{subfigure}{0.24\textwidth}
      \centering
      \includegraphics[width=1\linewidth]{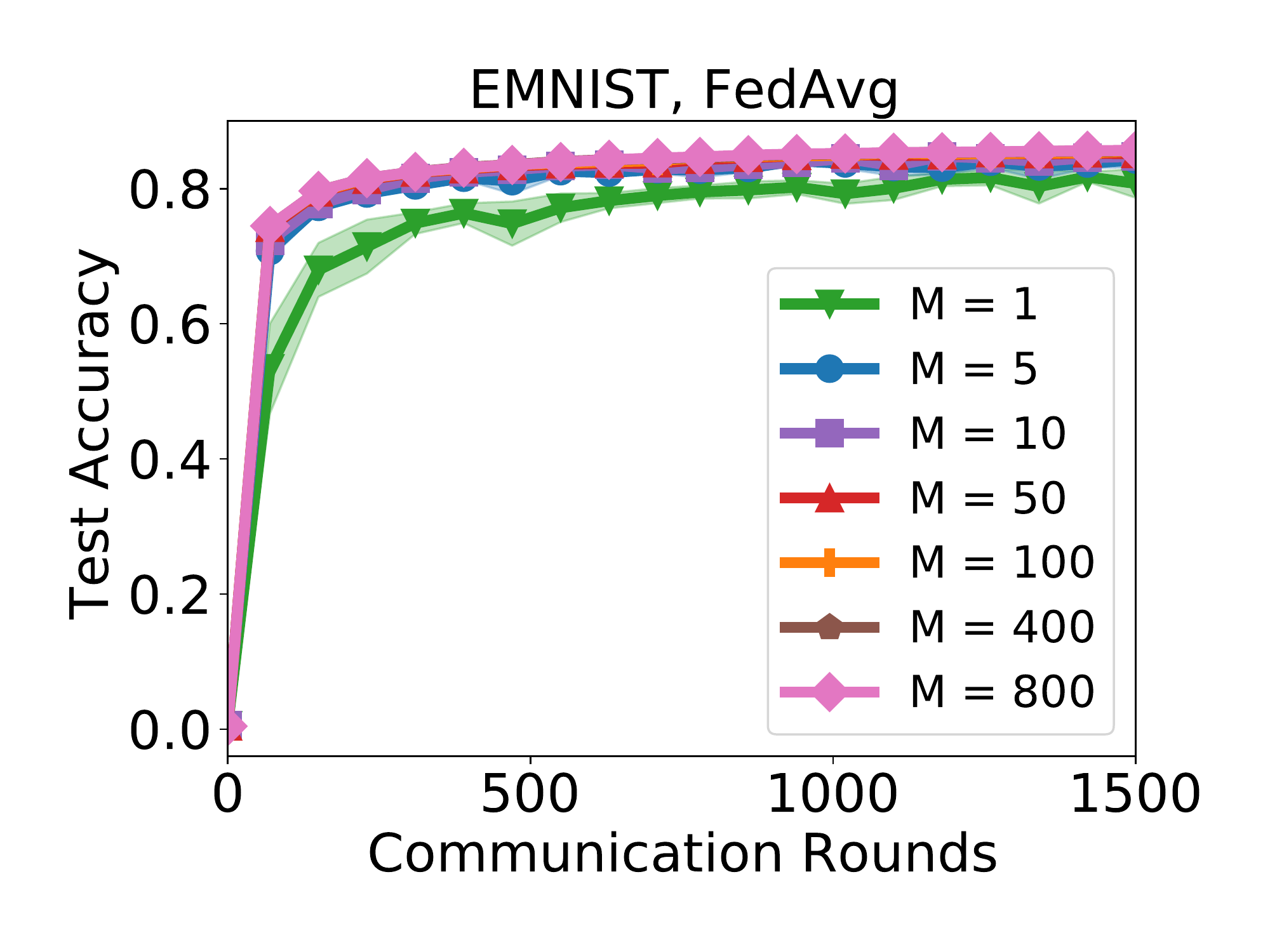}
\end{subfigure}%
\begin{subfigure}{0.24\textwidth}
      \centering
      \includegraphics[width=1\linewidth]{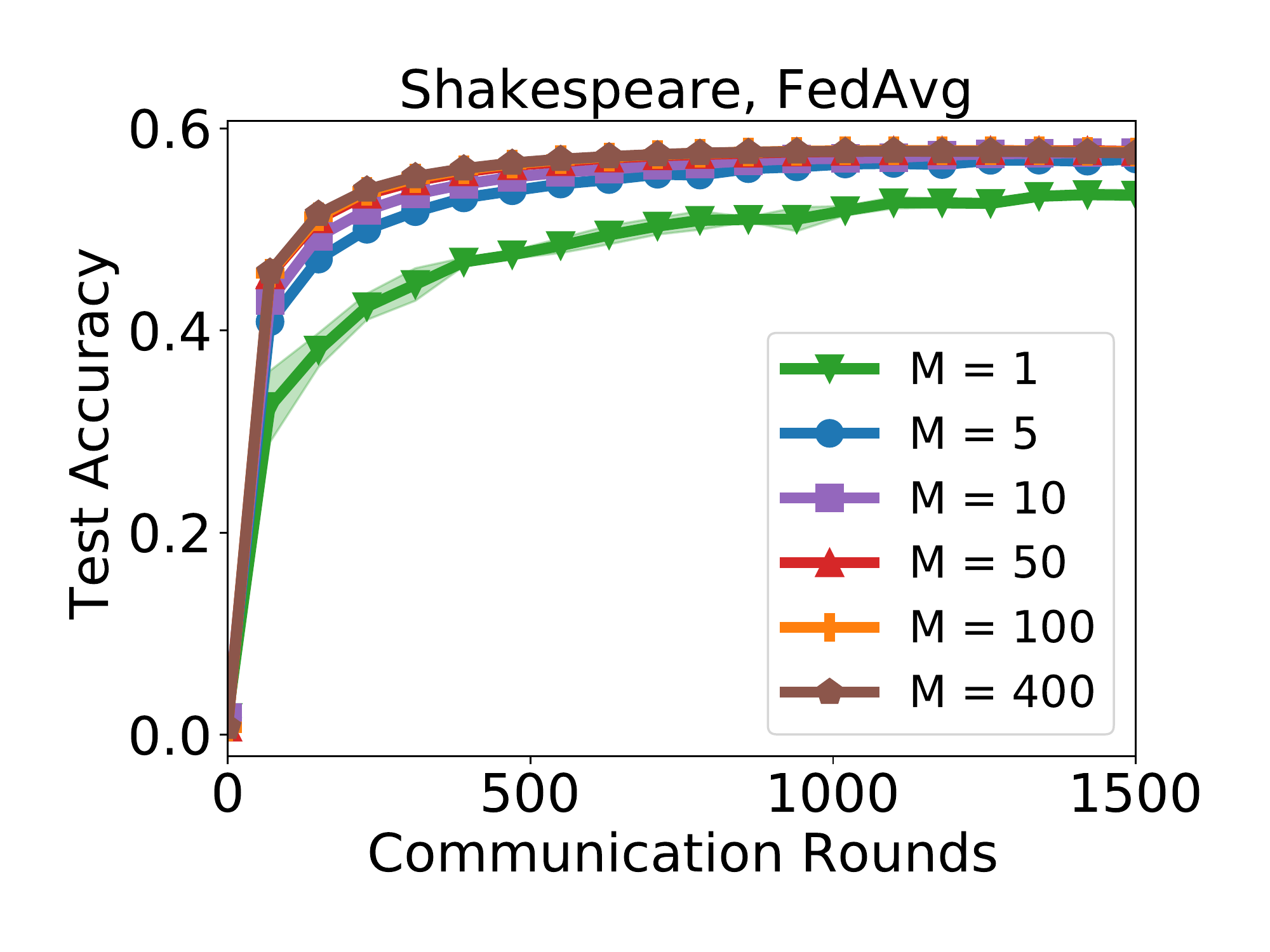}
\end{subfigure}%
\begin{subfigure}{0.24\textwidth}
      \centering
      \includegraphics[width=1\linewidth]{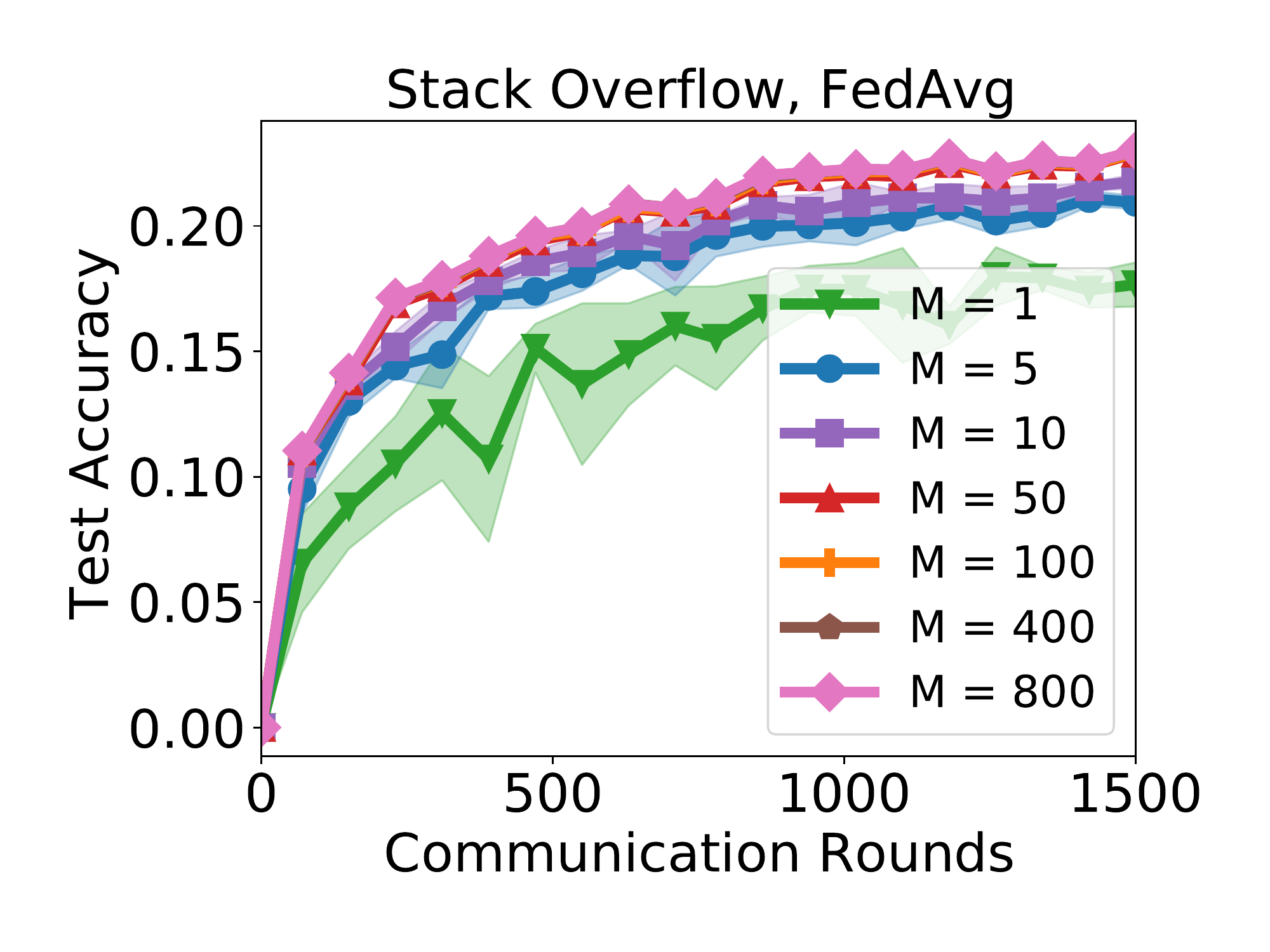}
\end{subfigure}%
\caption{Average test accuracy of \fedavg versus the number of communication rounds, for various tasks and cohort sizes $M$.}
\label{fig:fedavg_test_accuracy_versus_rounds}
\end{figure}

\begin{figure}[ht!]
\centering
\begin{subfigure}{0.24\textwidth}
      \centering
      \includegraphics[width=1\linewidth]{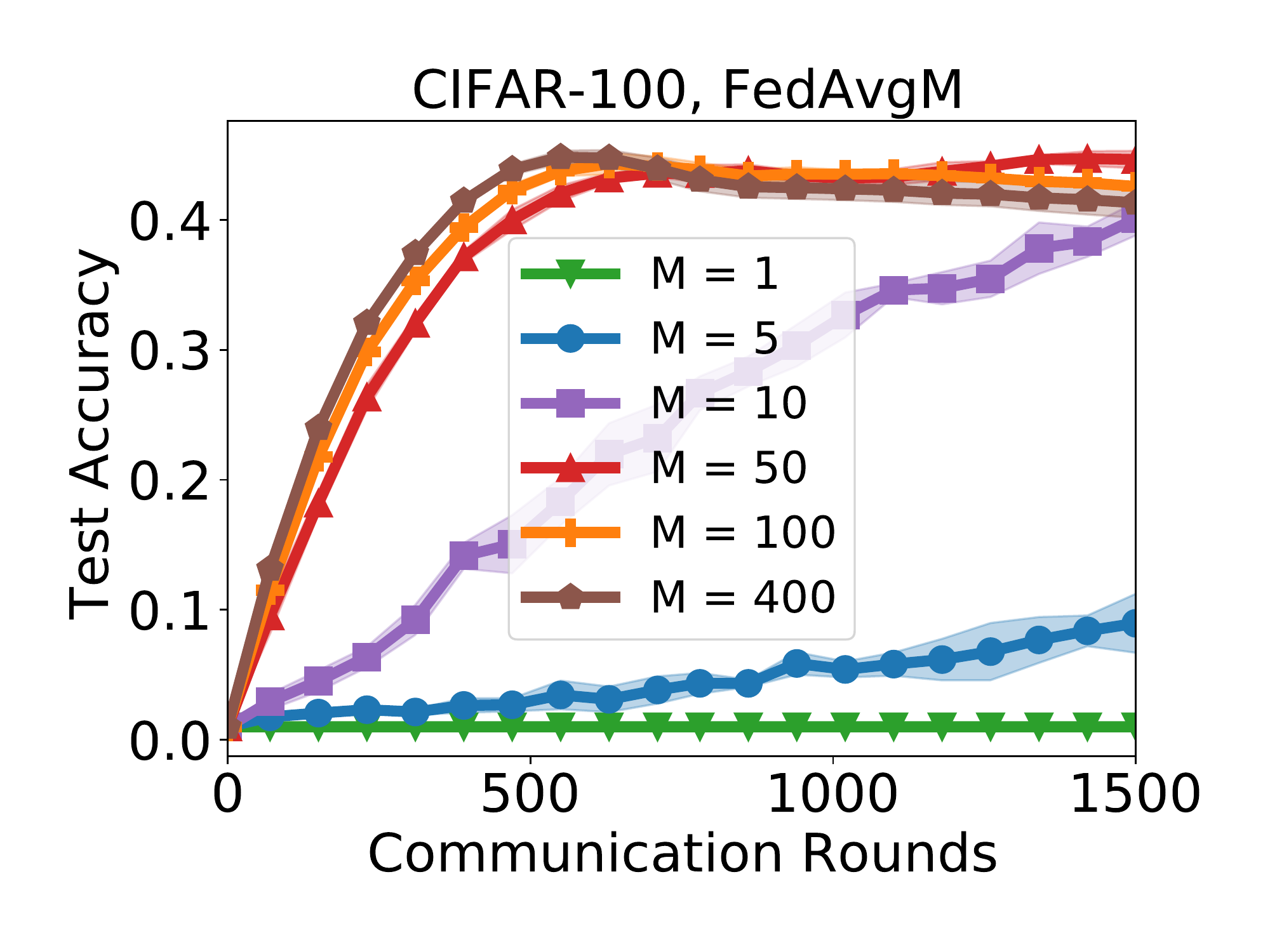}
\end{subfigure}%
\begin{subfigure}{0.24\textwidth}
      \centering
      \includegraphics[width=1\linewidth]{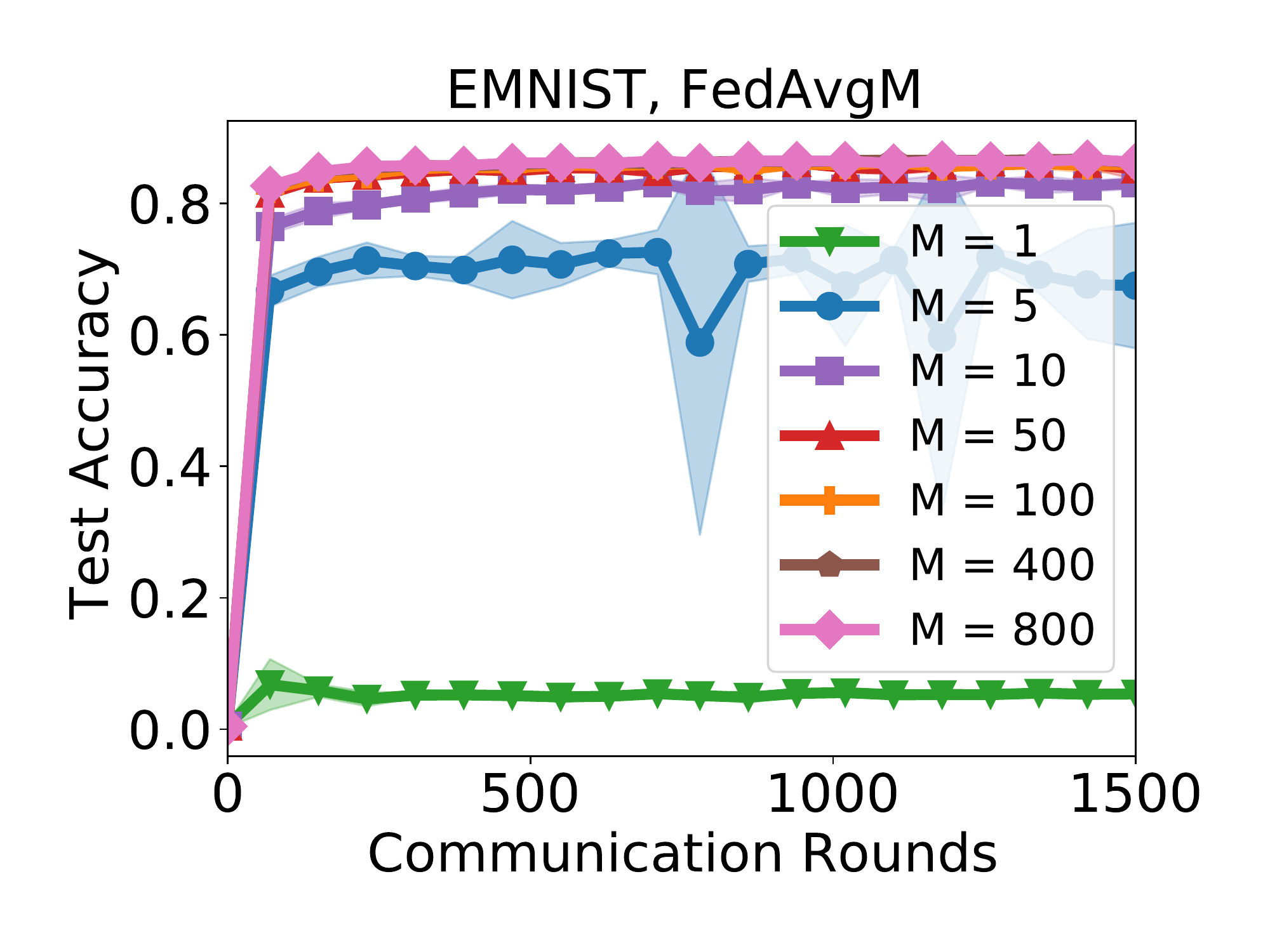}
\end{subfigure}%
\begin{subfigure}{0.24\textwidth}
      \centering
      \includegraphics[width=1\linewidth]{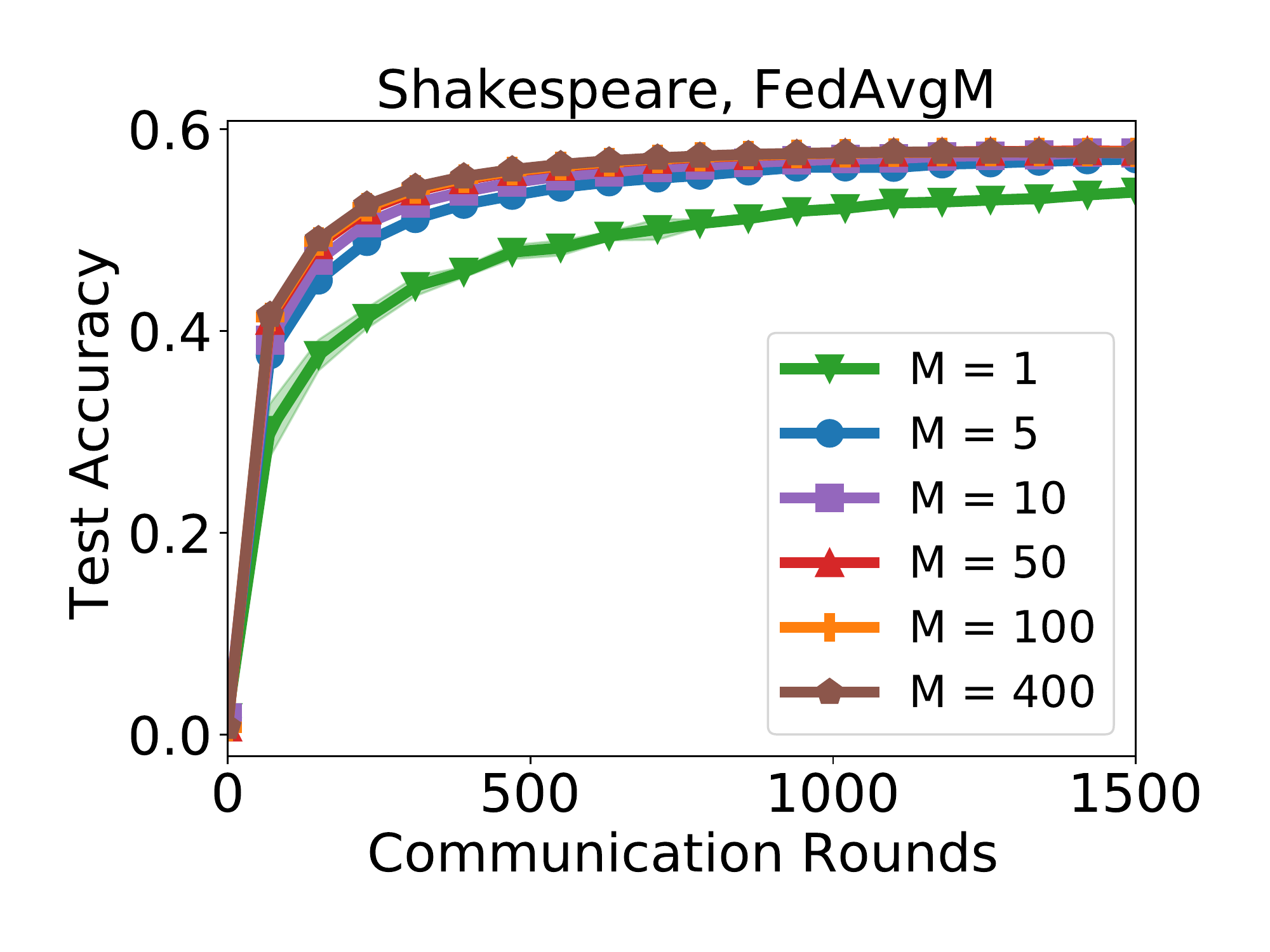}
\end{subfigure}%
\begin{subfigure}{0.24\textwidth}
      \centering
      \includegraphics[width=1\linewidth]{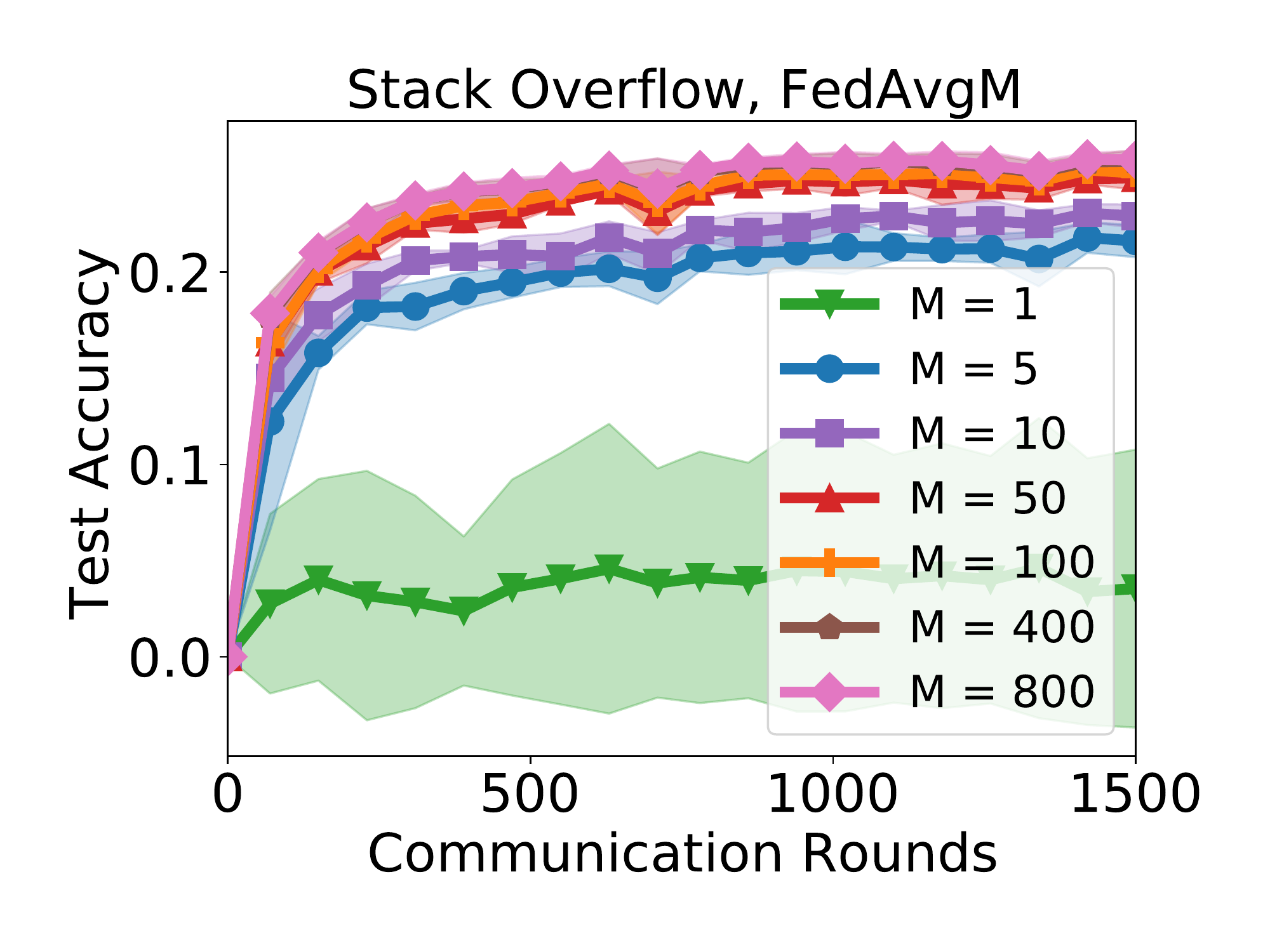}
\end{subfigure}%
\caption{Average test accuracy of \fedavgm versus the number of communication rounds, for various tasks and cohort sizes $M$.}
\label{fig:fedavgm_test_accuracy_versus_rounds}
\end{figure}

\begin{figure}[ht!]
\centering
\begin{subfigure}{0.24\textwidth}
      \centering
      \includegraphics[width=1\linewidth]{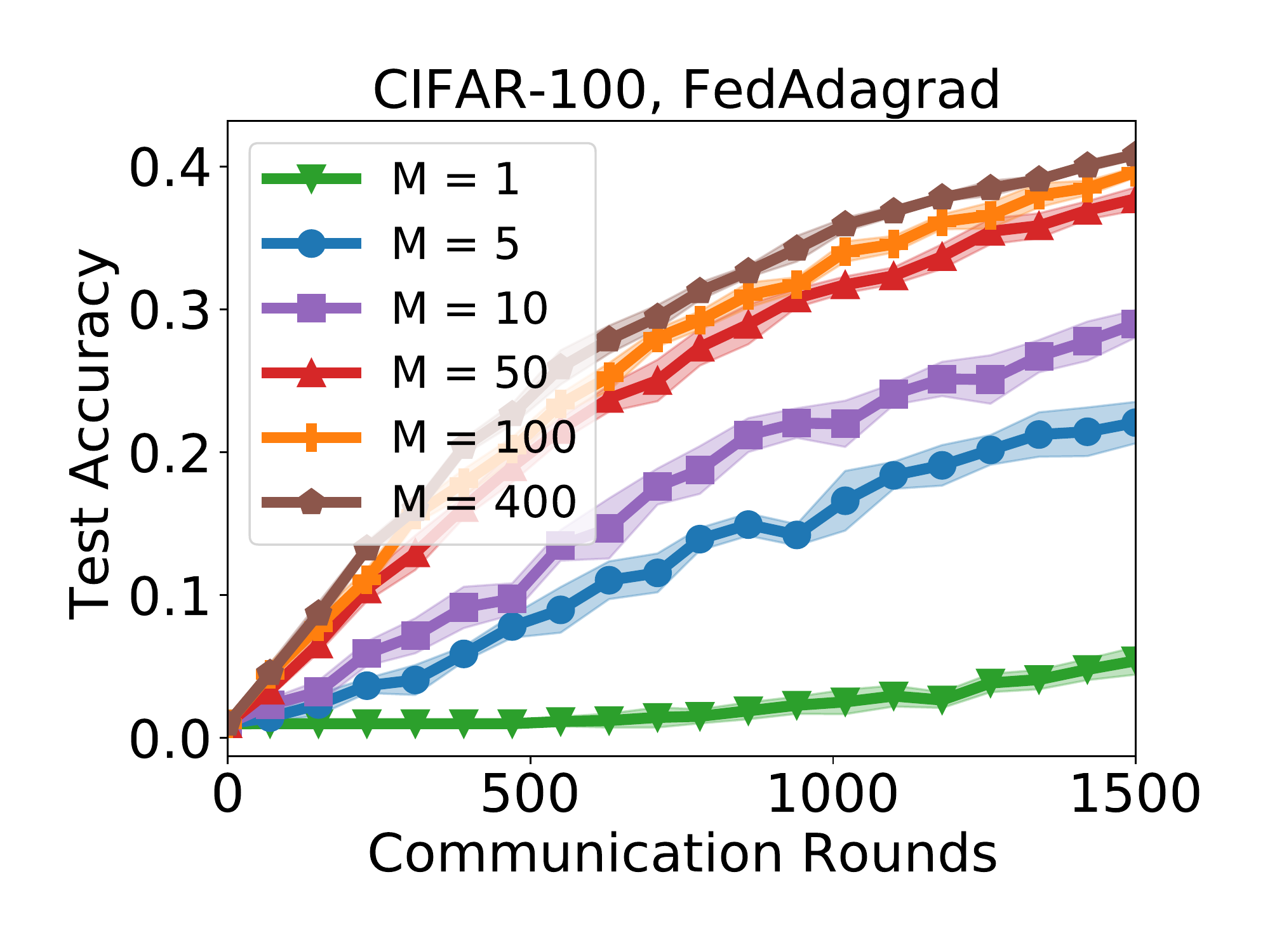}
\end{subfigure}%
\begin{subfigure}{0.24\textwidth}
      \centering
      \includegraphics[width=1\linewidth]{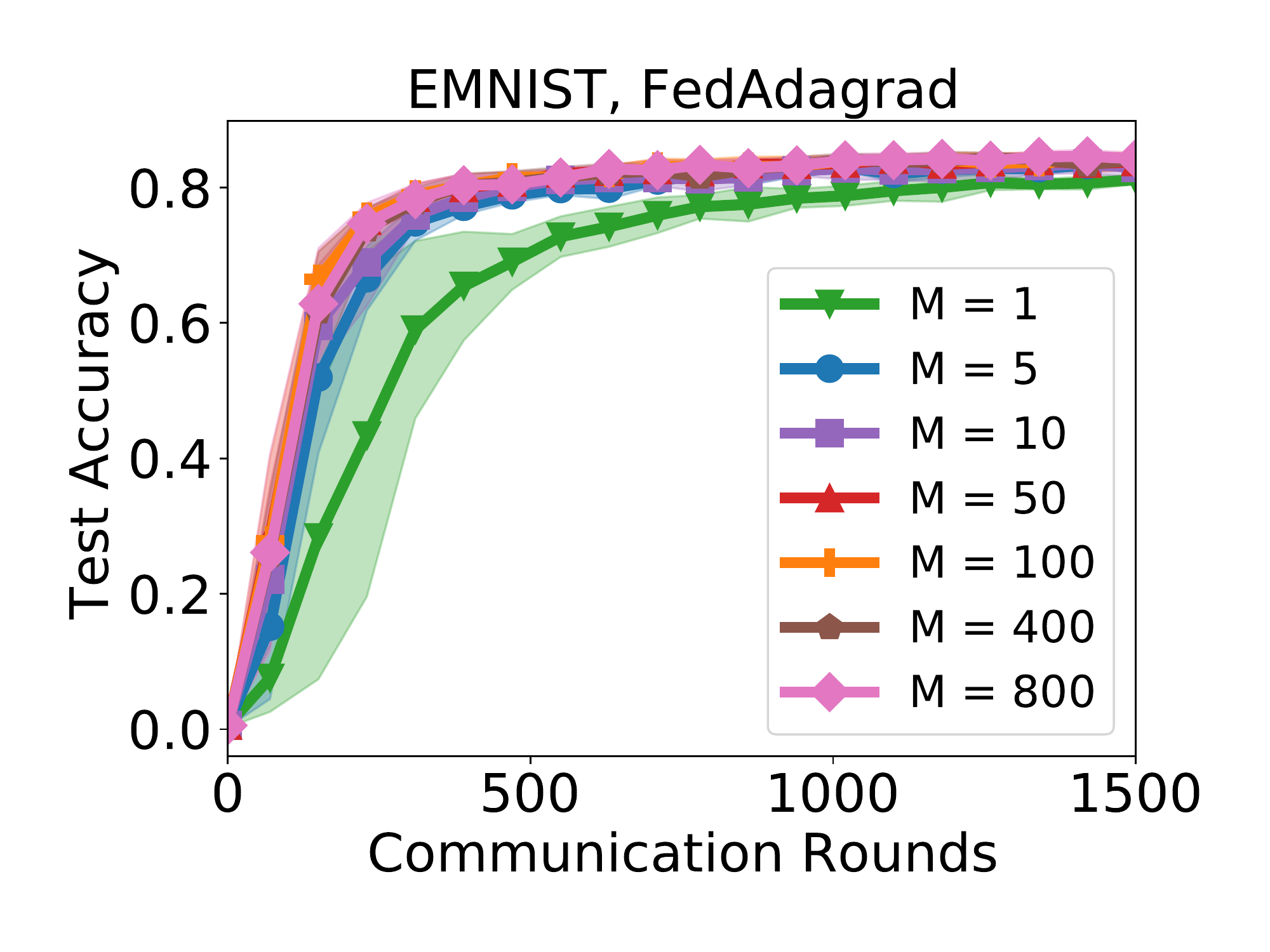}
\end{subfigure}%
\begin{subfigure}{0.24\textwidth}
      \centering
      \includegraphics[width=1\linewidth]{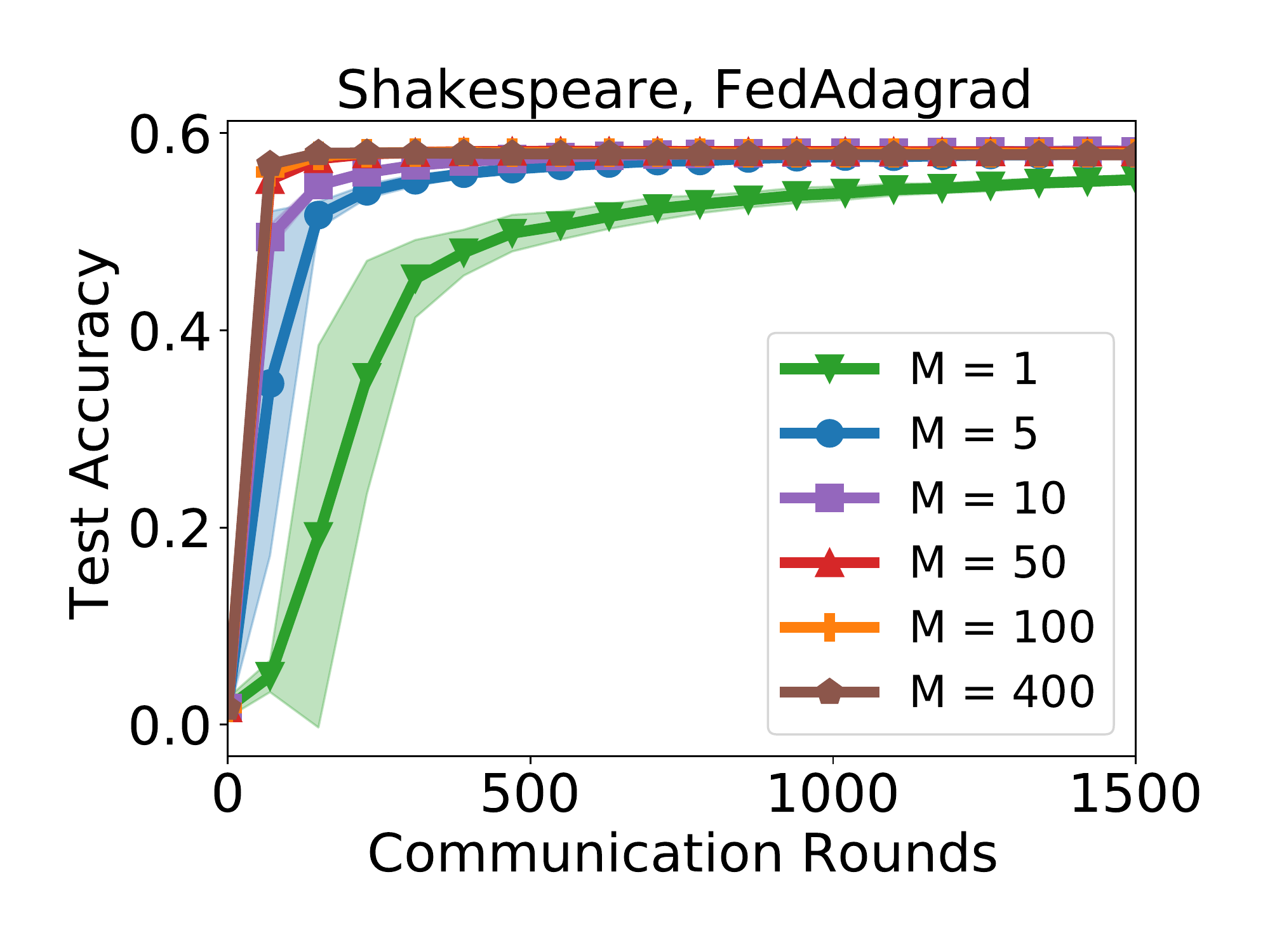}
\end{subfigure}%
\begin{subfigure}{0.24\textwidth}
      \centering
      \includegraphics[width=1\linewidth]{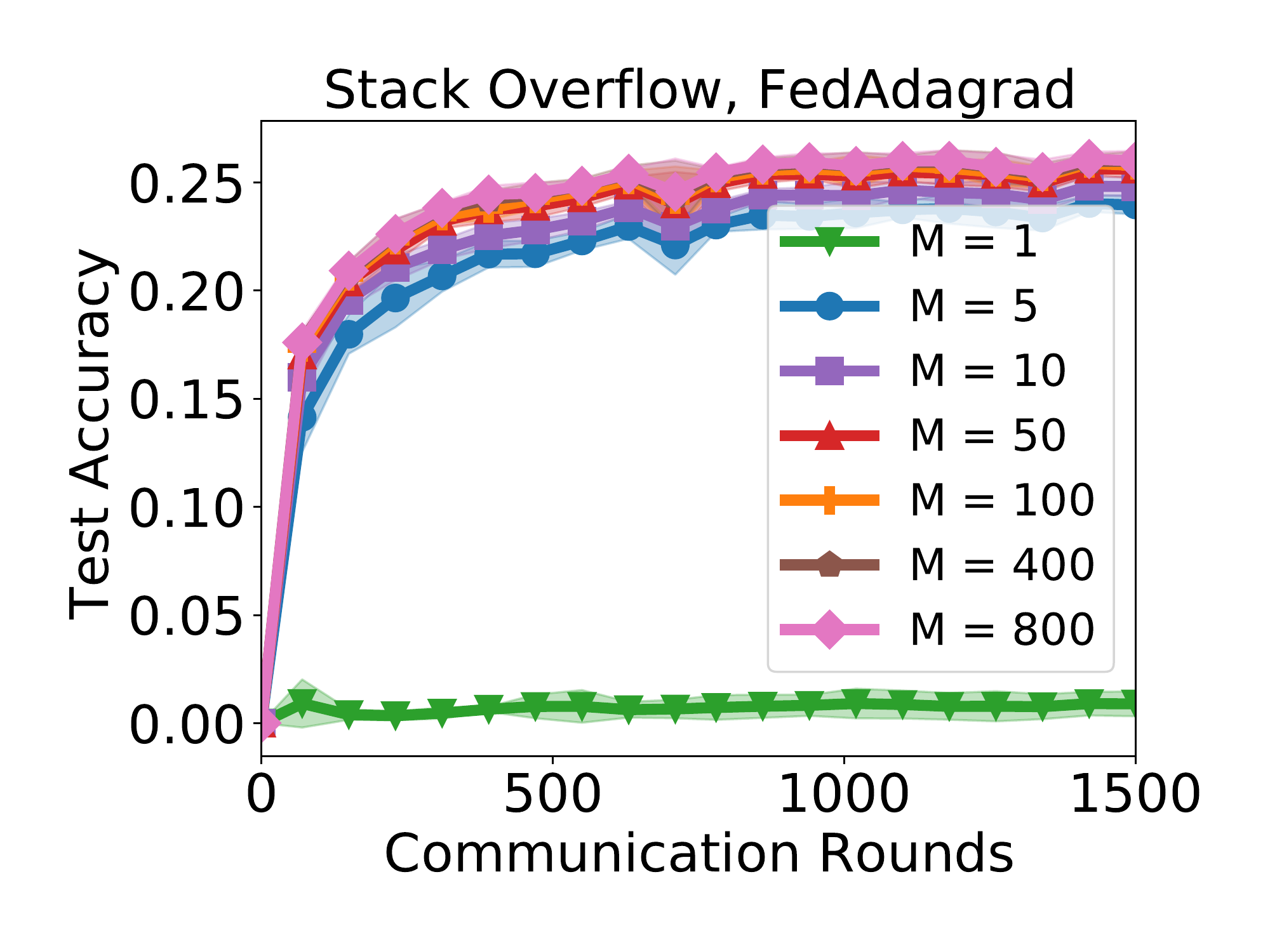}
\end{subfigure}%
\caption{Average test accuracy of \fedadagrad versus the number of communication rounds, for various tasks and cohort sizes $M$.}
\label{fig:fedadagrad_test_accuracy_versus_rounds}
\end{figure}

\begin{figure}[ht!]
\centering
\begin{subfigure}{0.24\textwidth}
      \centering
      \includegraphics[width=1\linewidth]{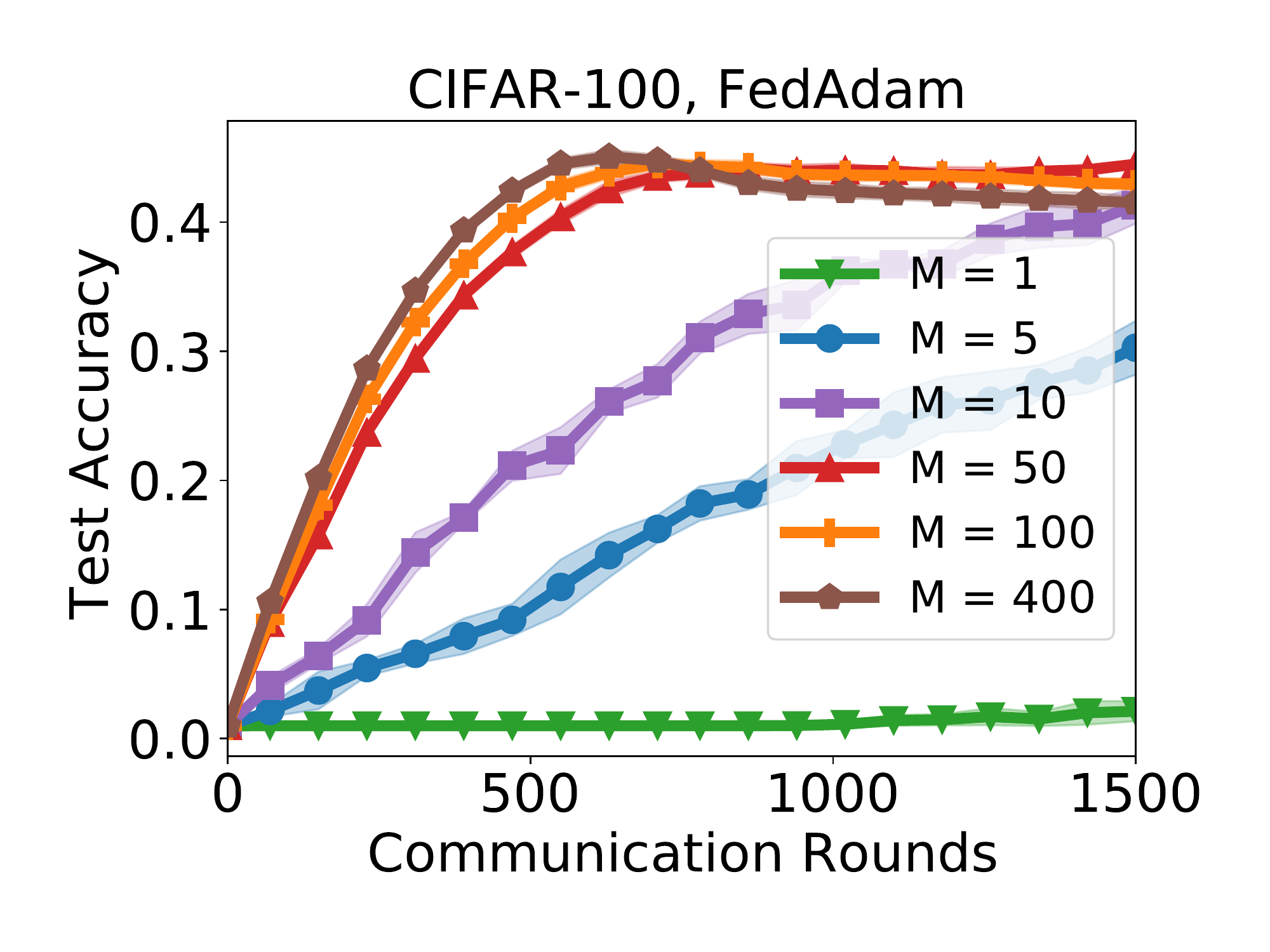}
\end{subfigure}%
\begin{subfigure}{0.24\textwidth}
      \centering
      \includegraphics[width=1\linewidth]{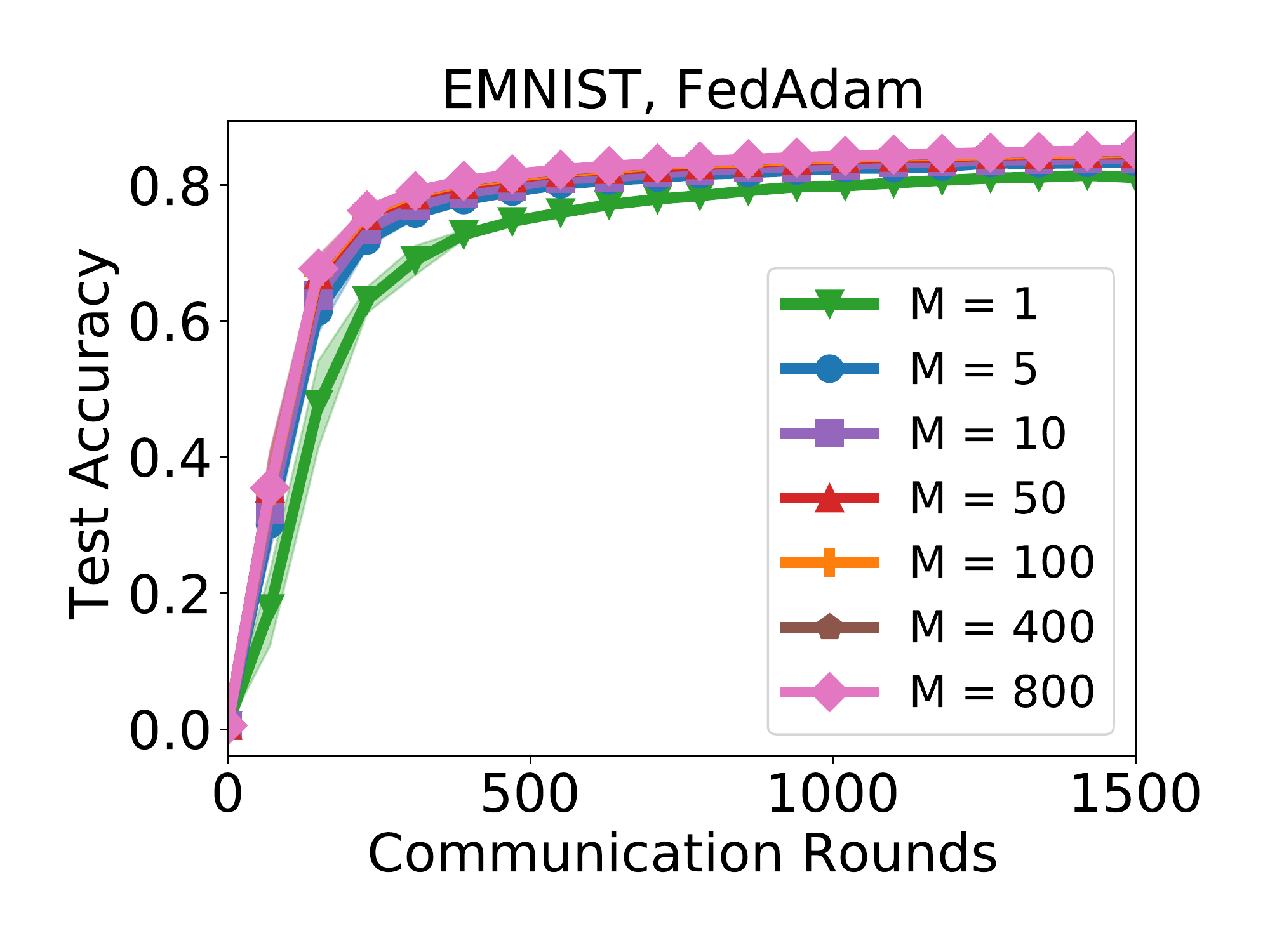}
\end{subfigure}%
\begin{subfigure}{0.24\textwidth}
      \centering
      \includegraphics[width=1\linewidth]{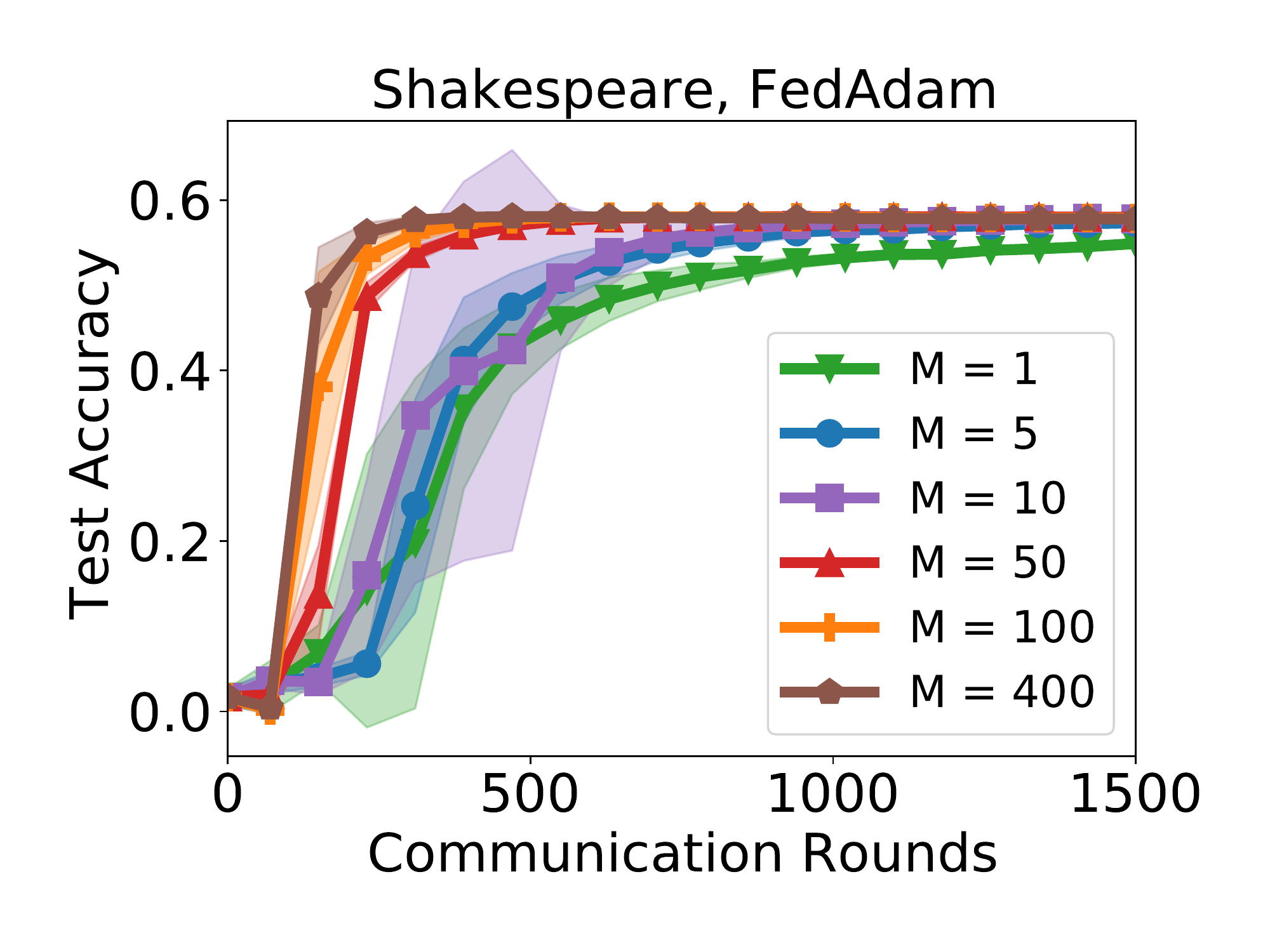}
\end{subfigure}%
\begin{subfigure}{0.24\textwidth}
      \centering
      \includegraphics[width=1\linewidth]{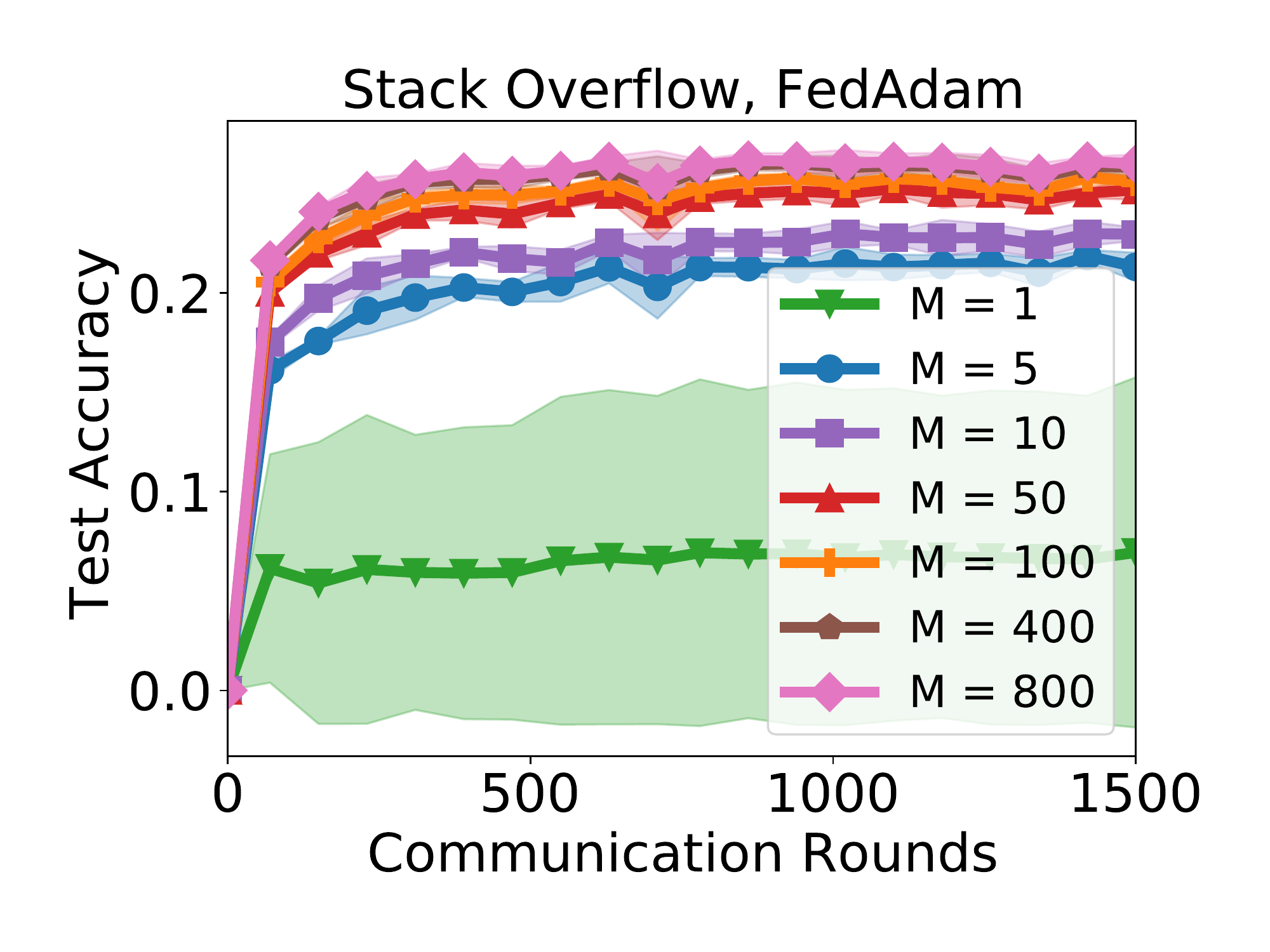}
\end{subfigure}%
\caption{Average test accuracy of \fedadam versus the number of communication rounds, for various tasks and cohort sizes $M$.}
\label{fig:fedadam_test_accuracy_versus_rounds}
\end{figure}

\begin{figure}[ht!]
\centering
\begin{subfigure}{0.24\textwidth}
      \centering
      \includegraphics[width=1\linewidth]{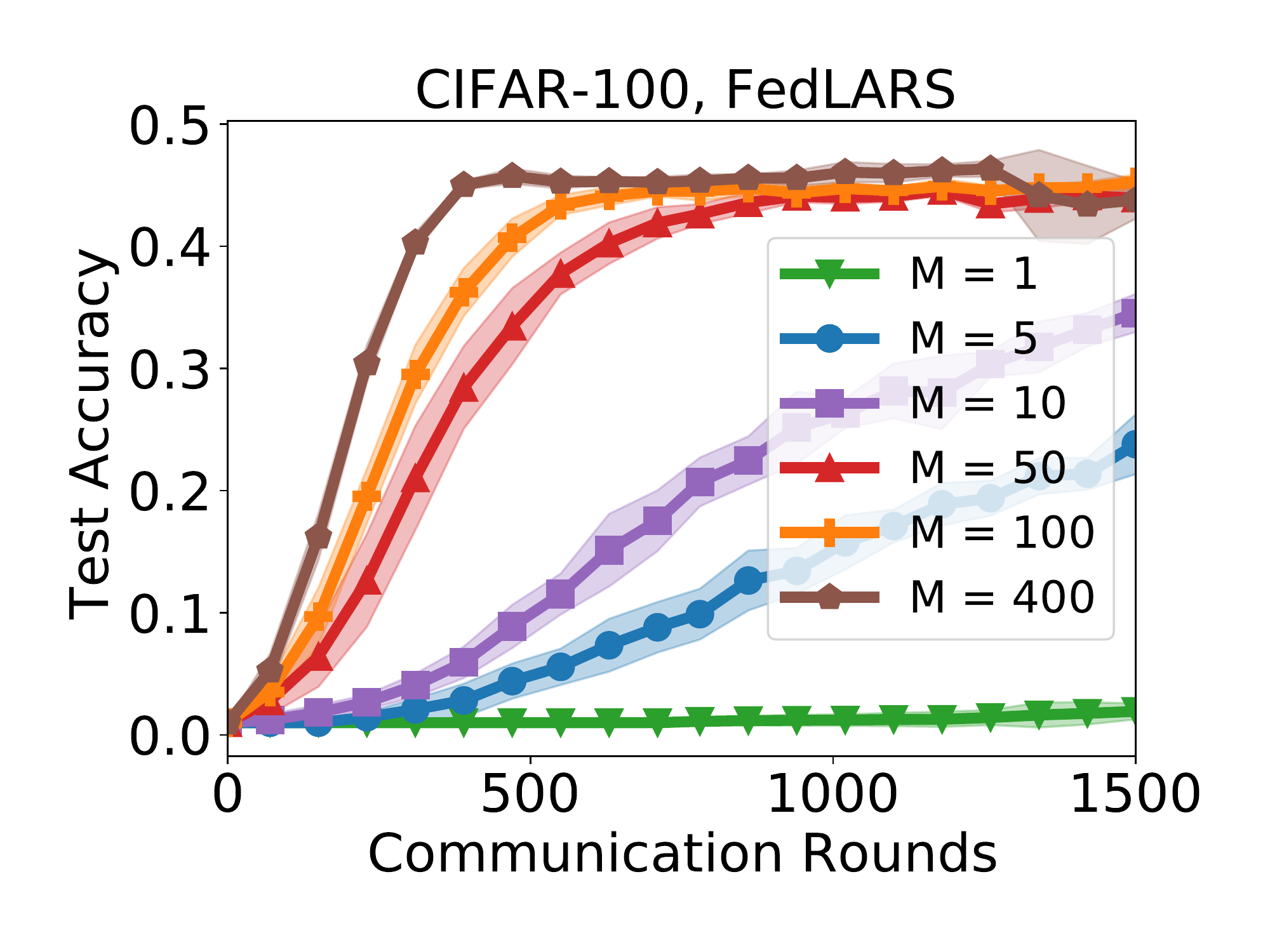}
\end{subfigure}%
\begin{subfigure}{0.24\textwidth}
      \centering
      \includegraphics[width=1\linewidth]{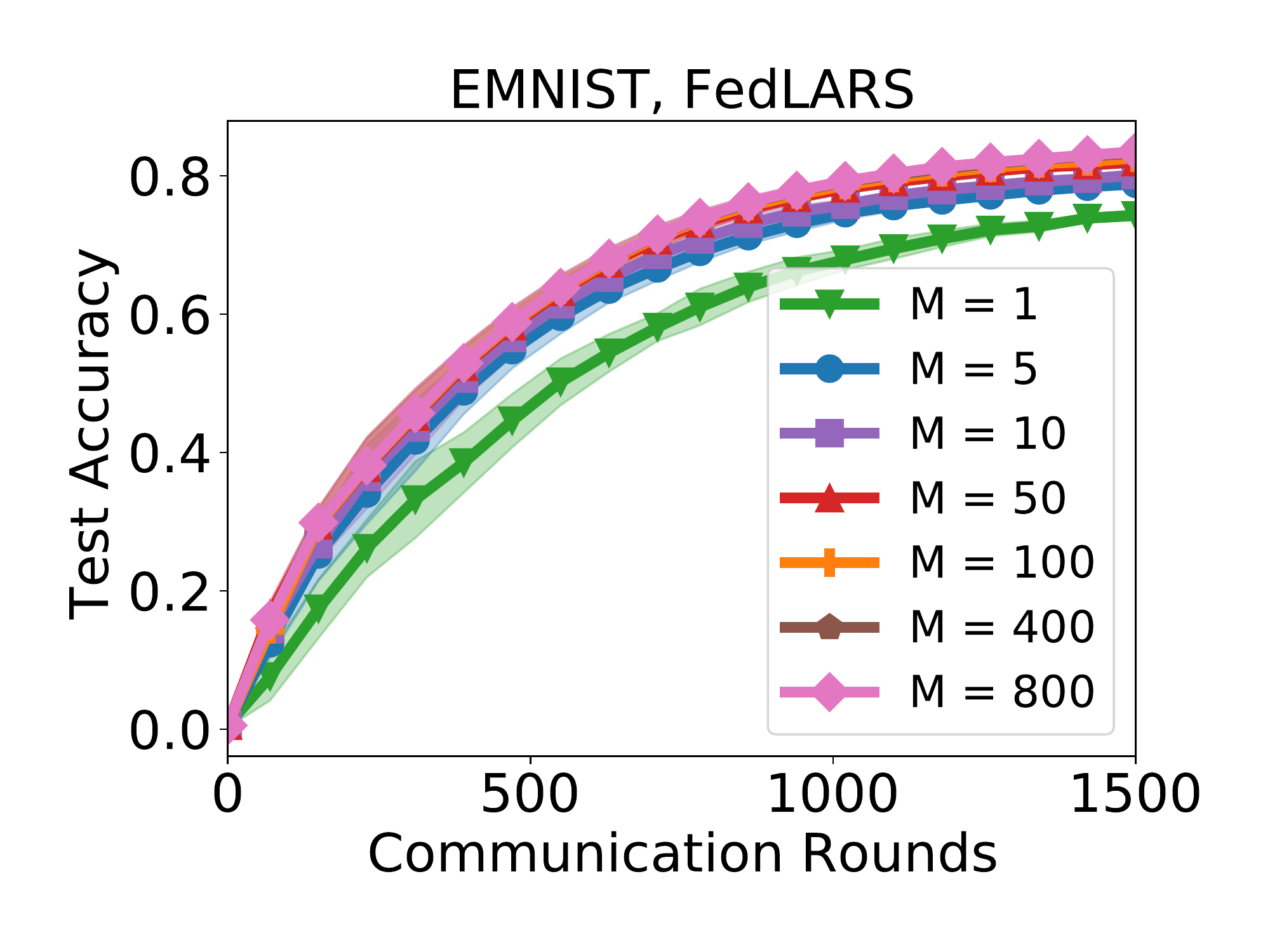}
\end{subfigure}%
\begin{subfigure}{0.24\textwidth}
      \centering
      \includegraphics[width=1\linewidth]{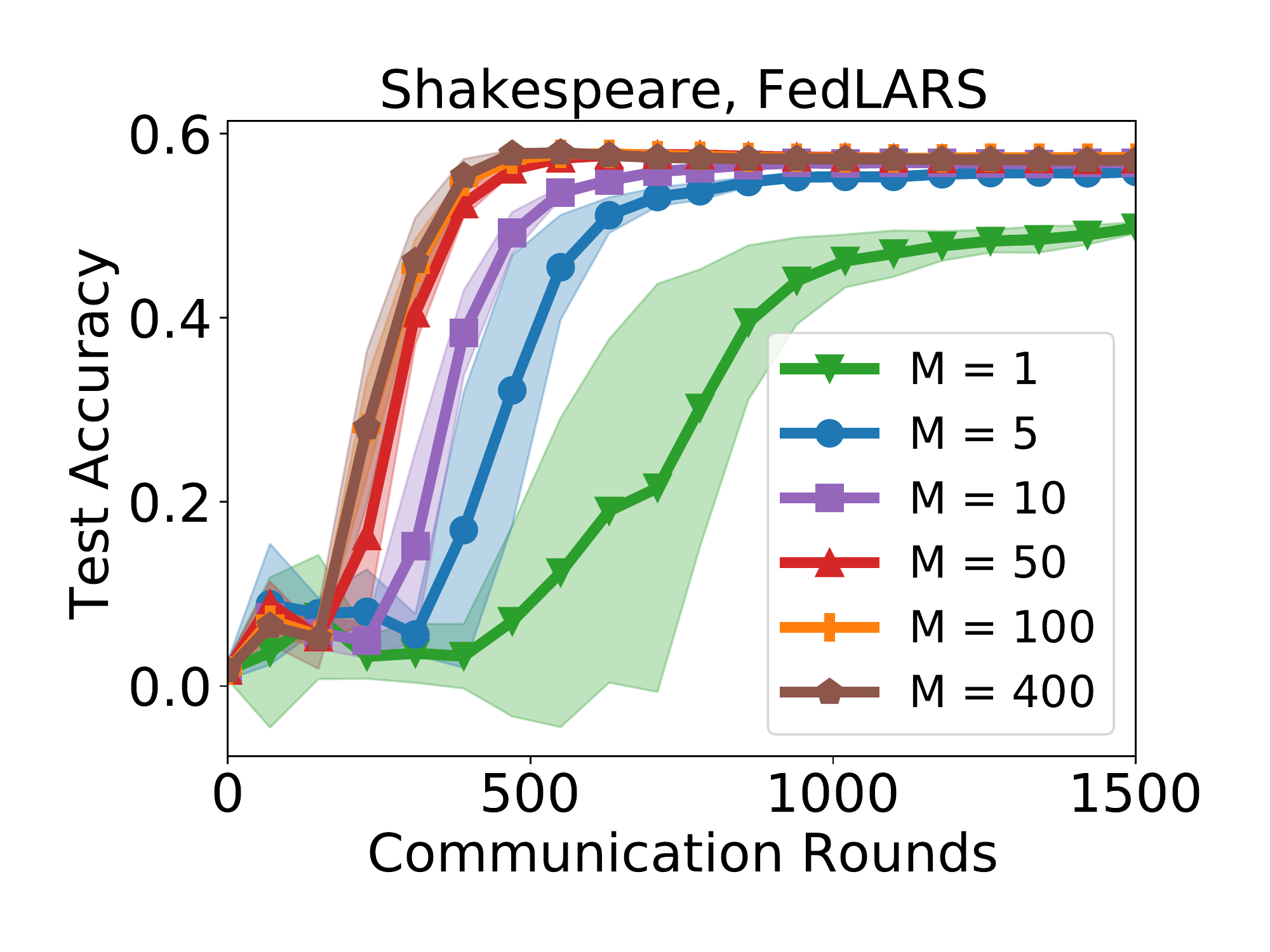}
\end{subfigure}%
\begin{subfigure}{0.24\textwidth}
      \centering
      \includegraphics[width=1\linewidth]{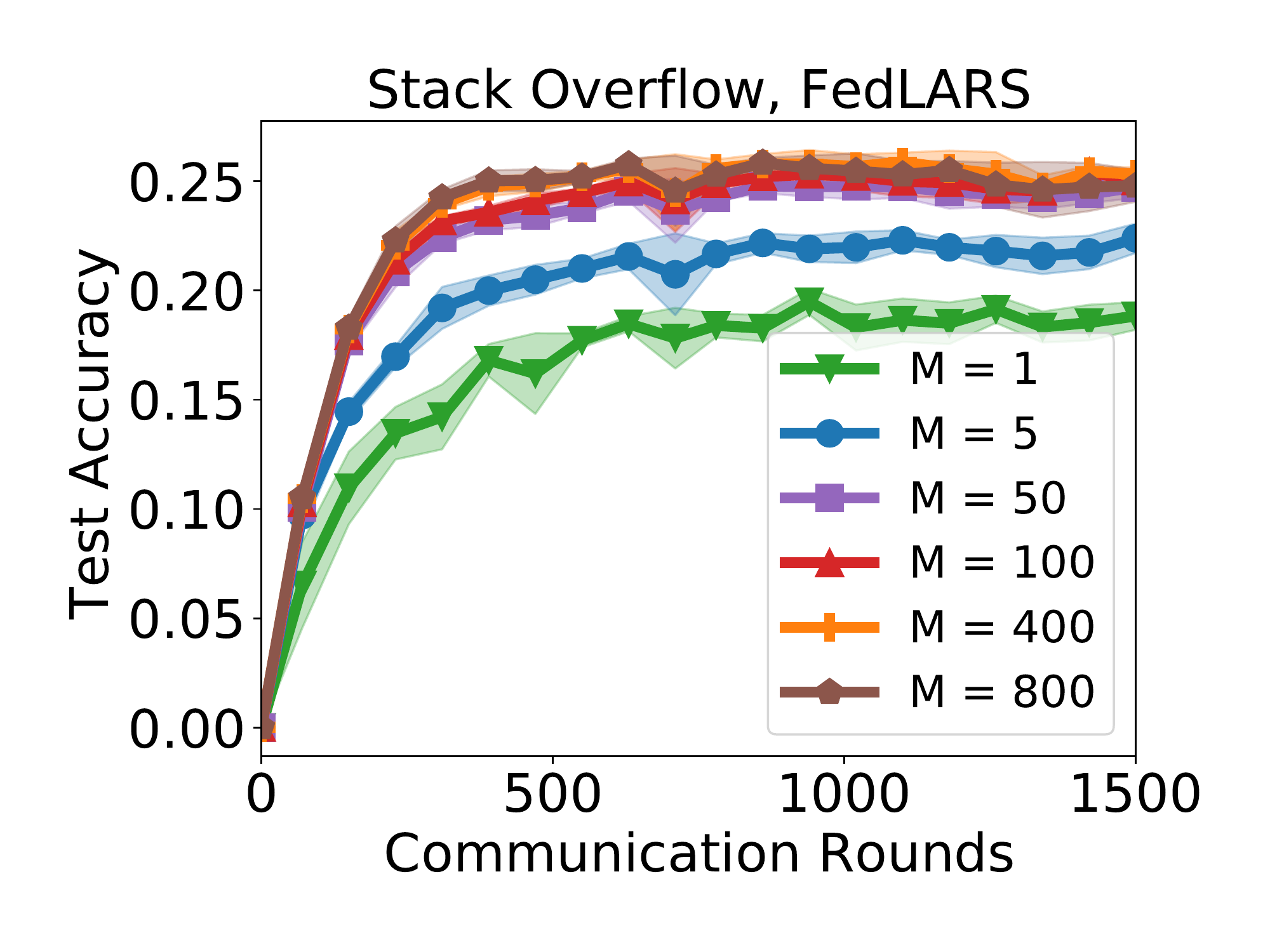}
\end{subfigure}%
\caption{Average test accuracy of \fedlars versus the number of communication rounds, for various tasks and cohort sizes $M$.}
\label{fig:fedlars_test_accuracy_versus_rounds}
\end{figure}

\begin{figure}[ht!]
\centering
\begin{subfigure}{0.24\textwidth}
      \centering
      \includegraphics[width=1\linewidth]{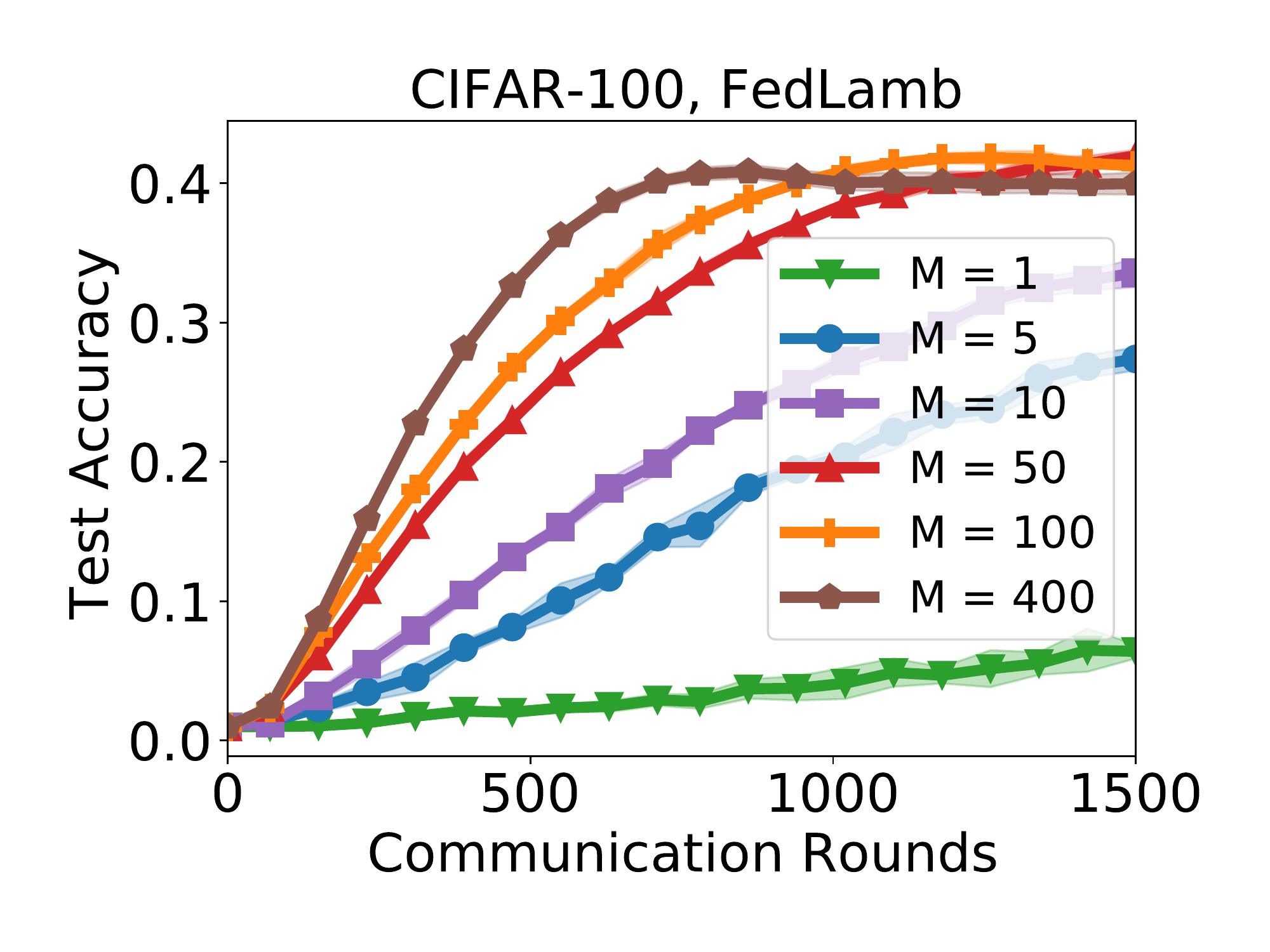}
\end{subfigure}%
\begin{subfigure}{0.24\textwidth}
      \centering
      \includegraphics[width=1\linewidth]{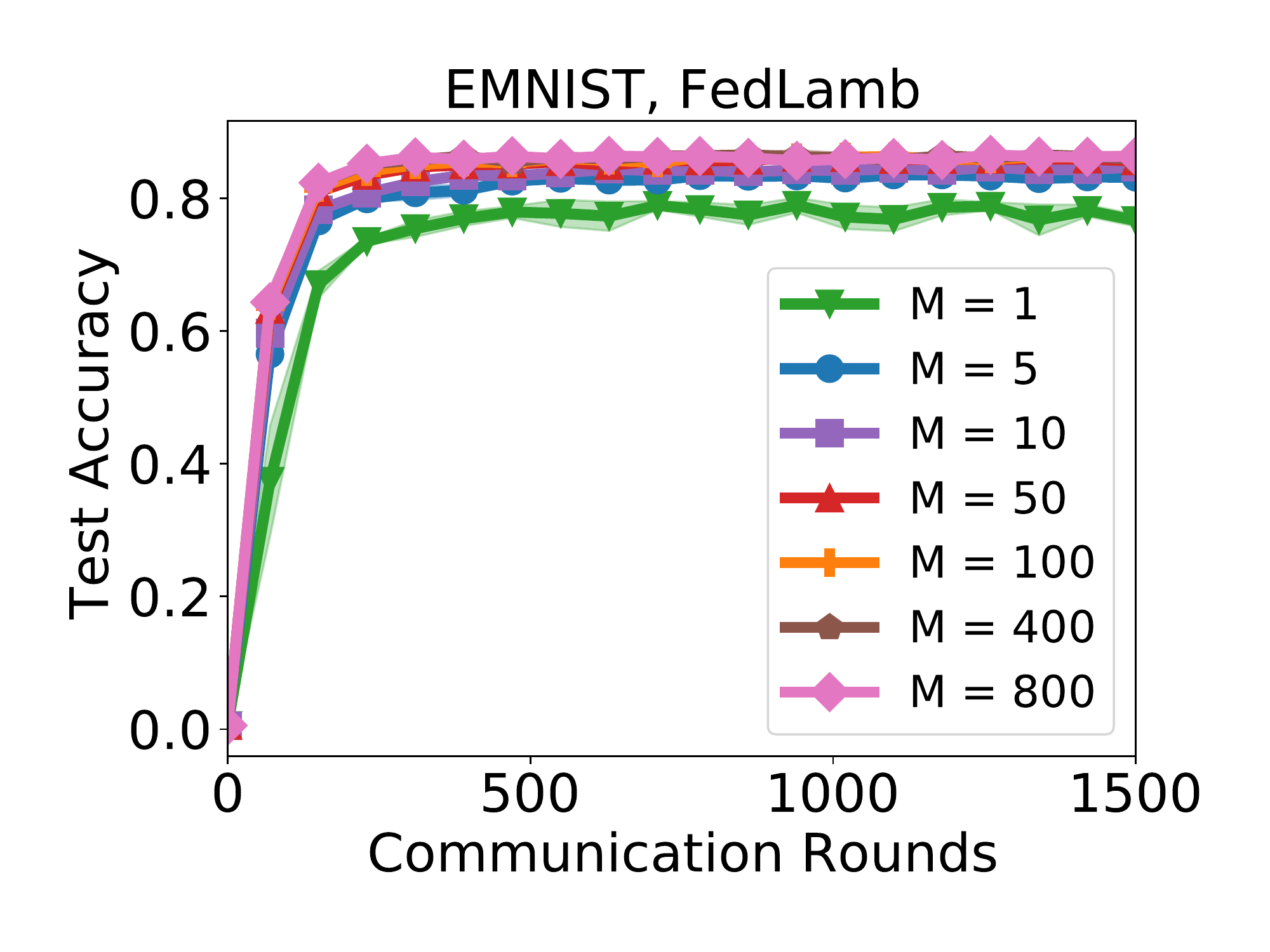}
\end{subfigure}%
\begin{subfigure}{0.24\textwidth}
      \centering
      \includegraphics[width=1\linewidth]{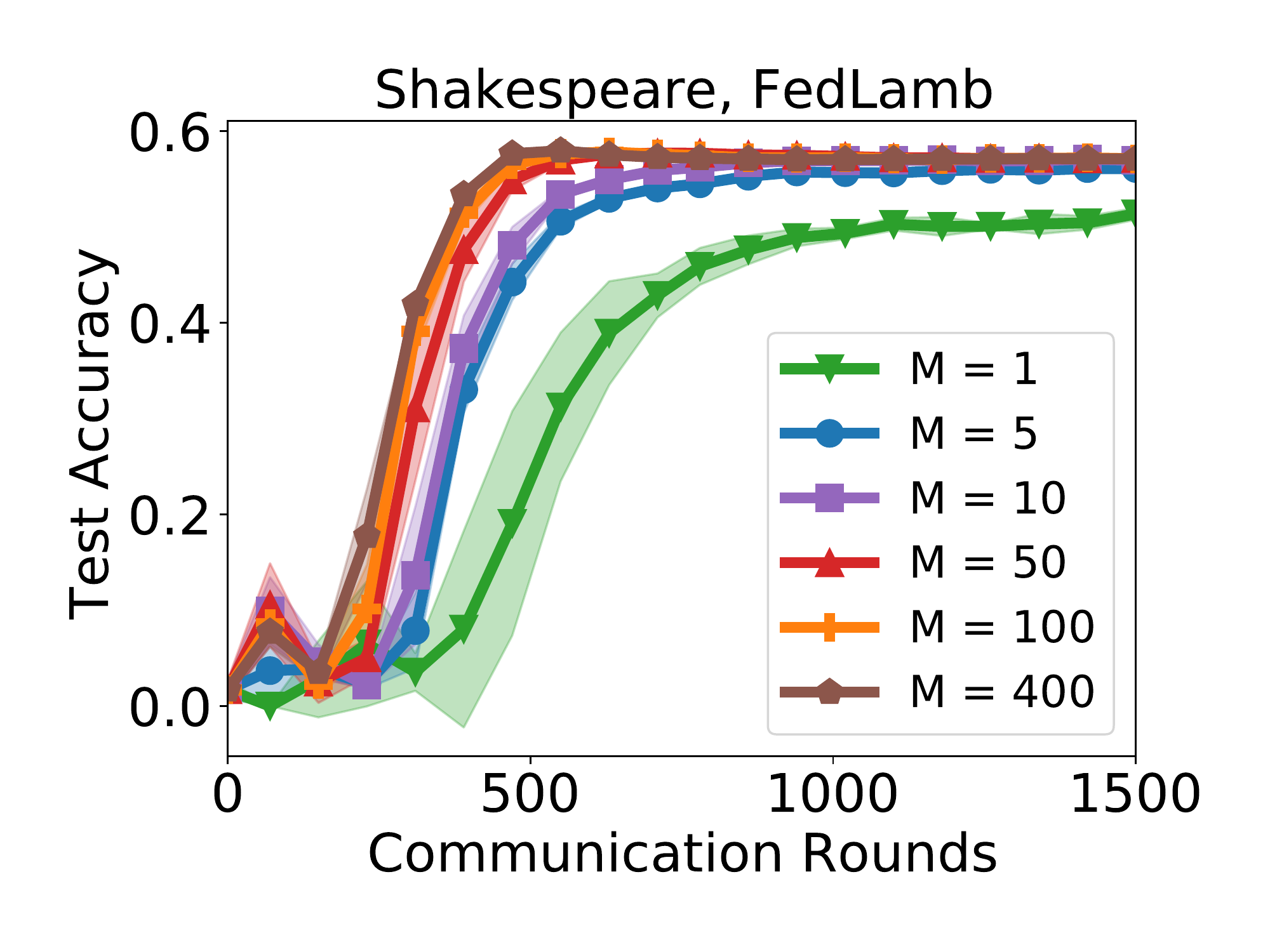}
\end{subfigure}%
\begin{subfigure}{0.24\textwidth}
      \centering
      \includegraphics[width=1\linewidth]{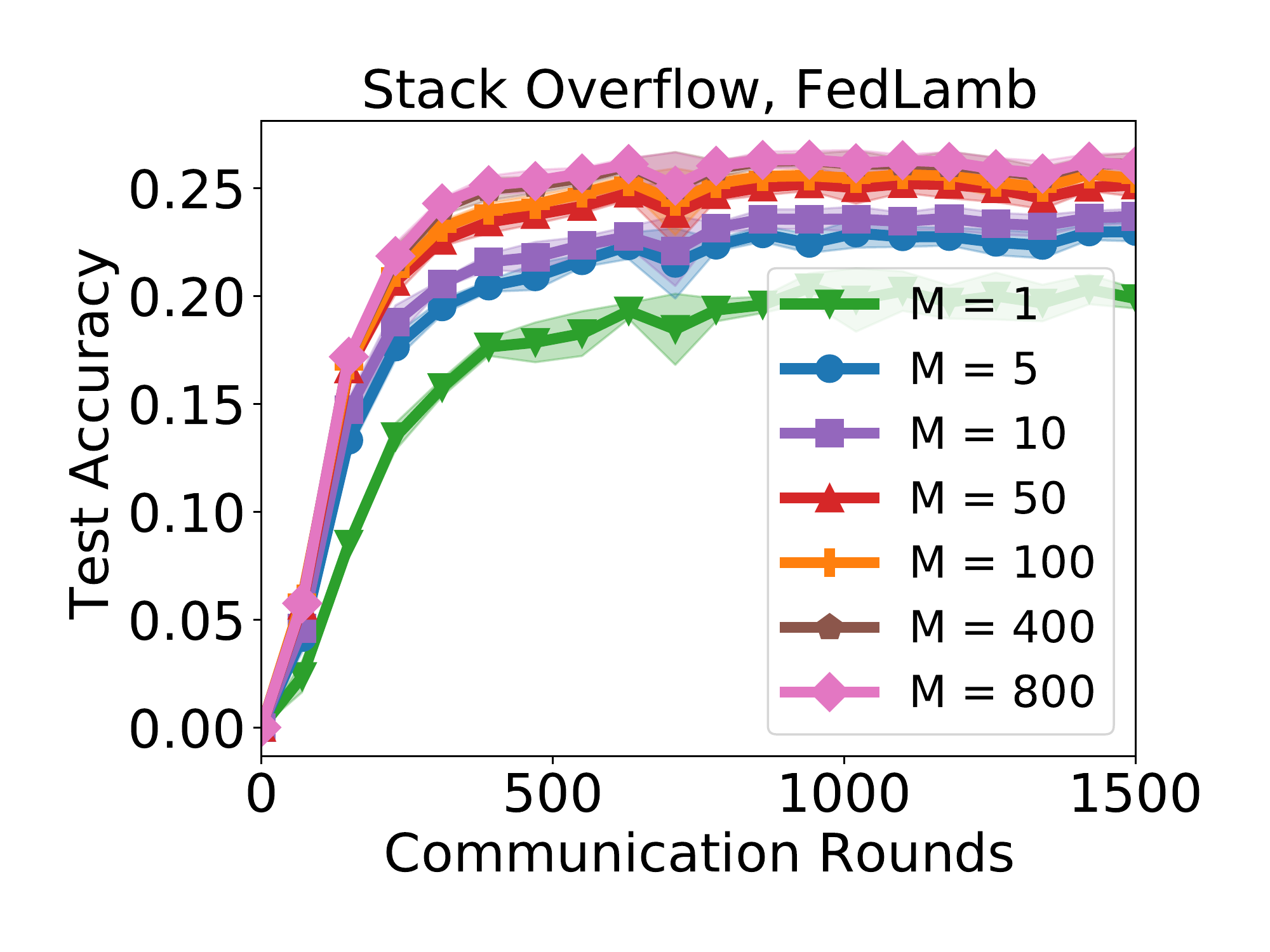}
\end{subfigure}%
\caption{Average test accuracy of \fedlamb versus the number of communication rounds, for various tasks and cohort sizes $M$.}
\label{fig:fedlamb_test_accuracy_versus_rounds}
\end{figure}

In this section, we present the test accuracy of various federated learning methods on various tasks, for various cohort sizes. The results are plotted in Figures \ref{fig:fedsgd_test_accuracy_versus_rounds}, \ref{fig:fedavg_test_accuracy_versus_rounds}, \ref{fig:fedavgm_test_accuracy_versus_rounds}, \ref{fig:fedadagrad_test_accuracy_versus_rounds}, \ref{fig:fedadam_test_accuracy_versus_rounds}, \ref{fig:fedlars_test_accuracy_versus_rounds}, and \ref{fig:fedlamb_test_accuracy_versus_rounds}, which give the results for \fedsgd, \fedavg, \fedavgm, \fedadagrad, \fedadam, \fedlars, and \fedlamb (respectively).

\FloatBarrier

\subsection{Accuracy Versus Cohort Size}\label{appendix:accuracy_versus_cohort}

\begin{figure}[ht!]
\centering
\begin{subfigure}{0.24\textwidth}
    \centering
    \includegraphics[width=1\linewidth]{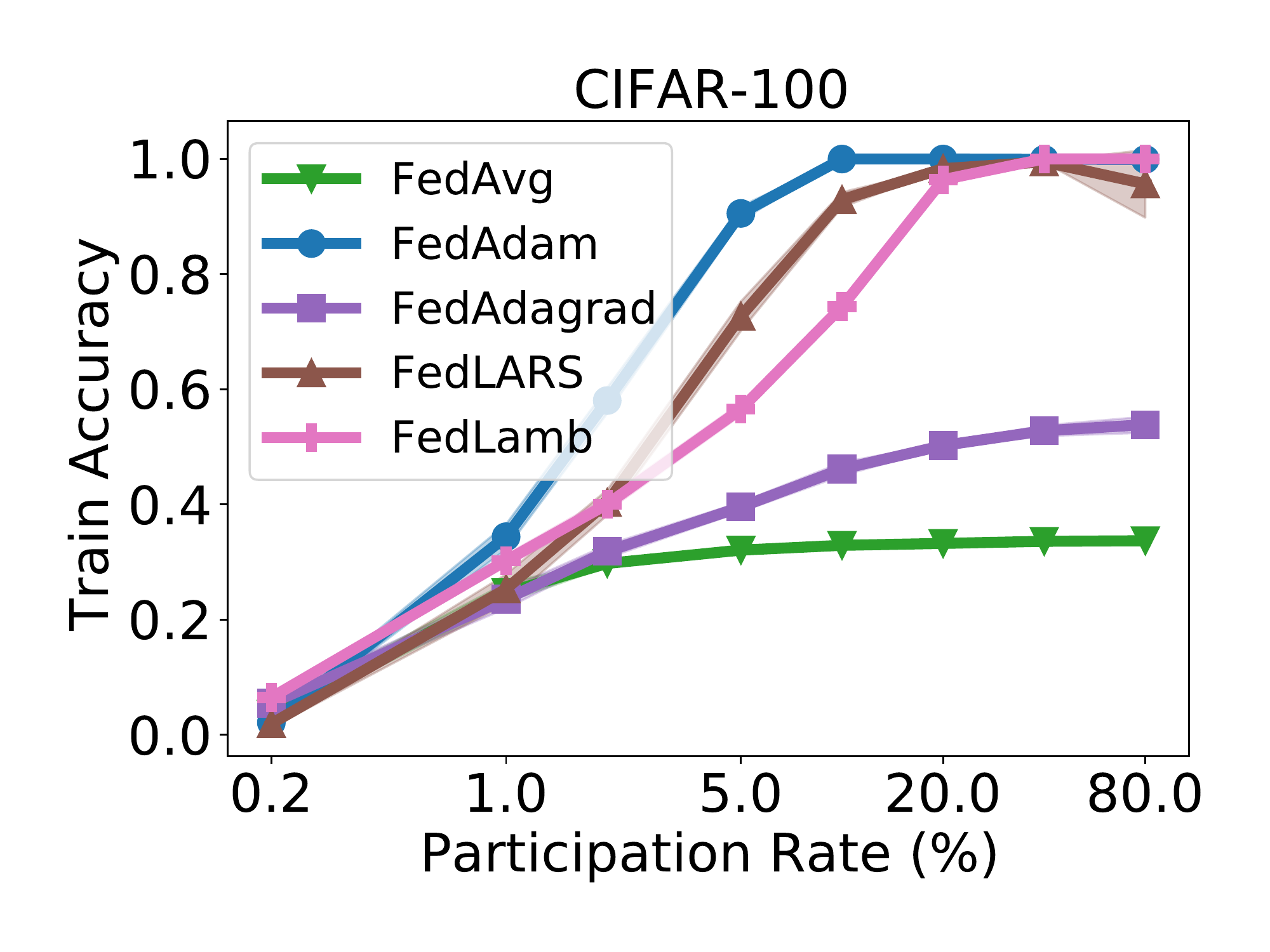}
\end{subfigure}%
\begin{subfigure}{0.24\textwidth}
    \centering
    \includegraphics[width=1\linewidth]{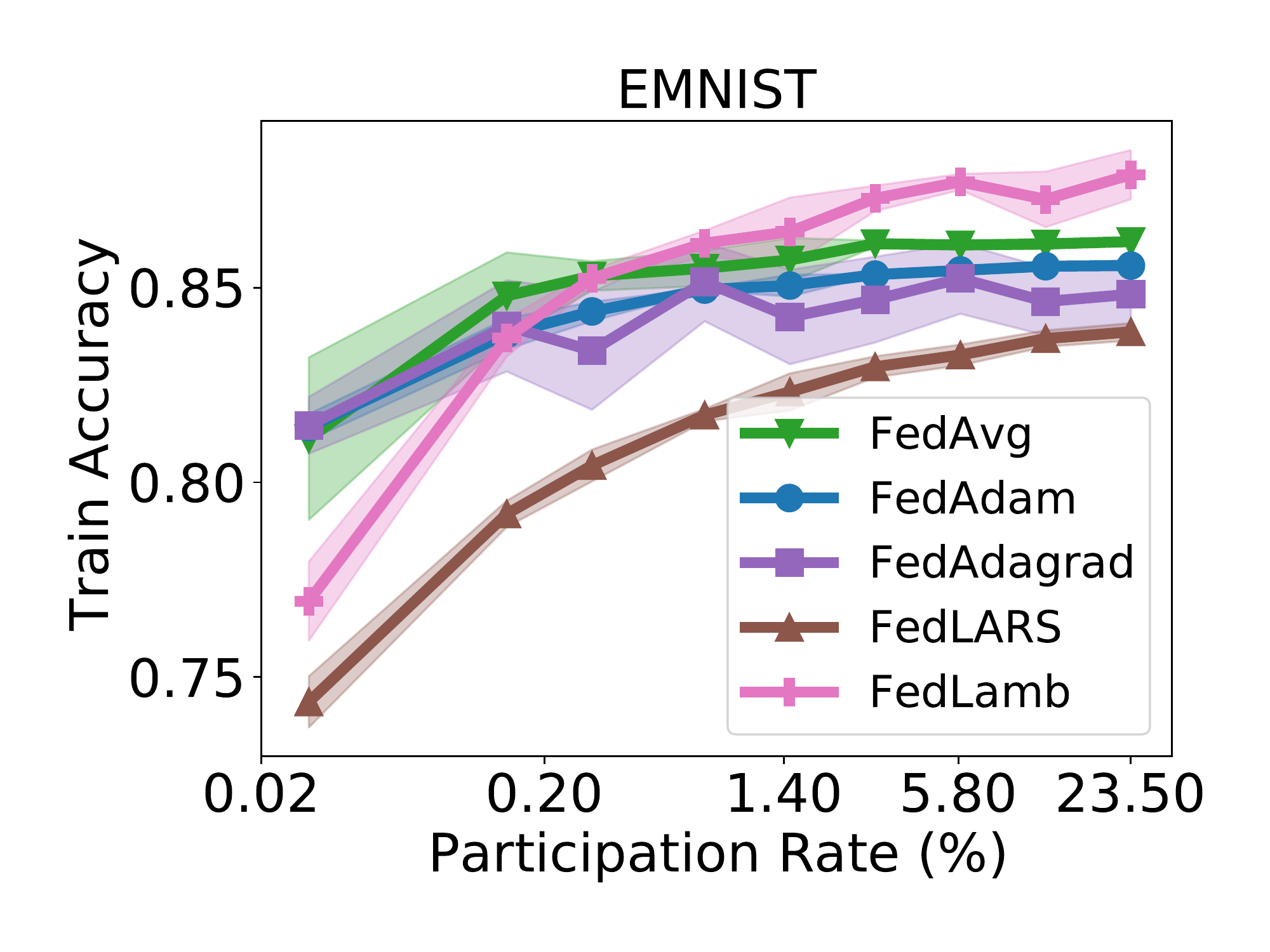}
\end{subfigure}%
\begin{subfigure}{0.24\textwidth}
    \centering
    \includegraphics[width=1\linewidth]{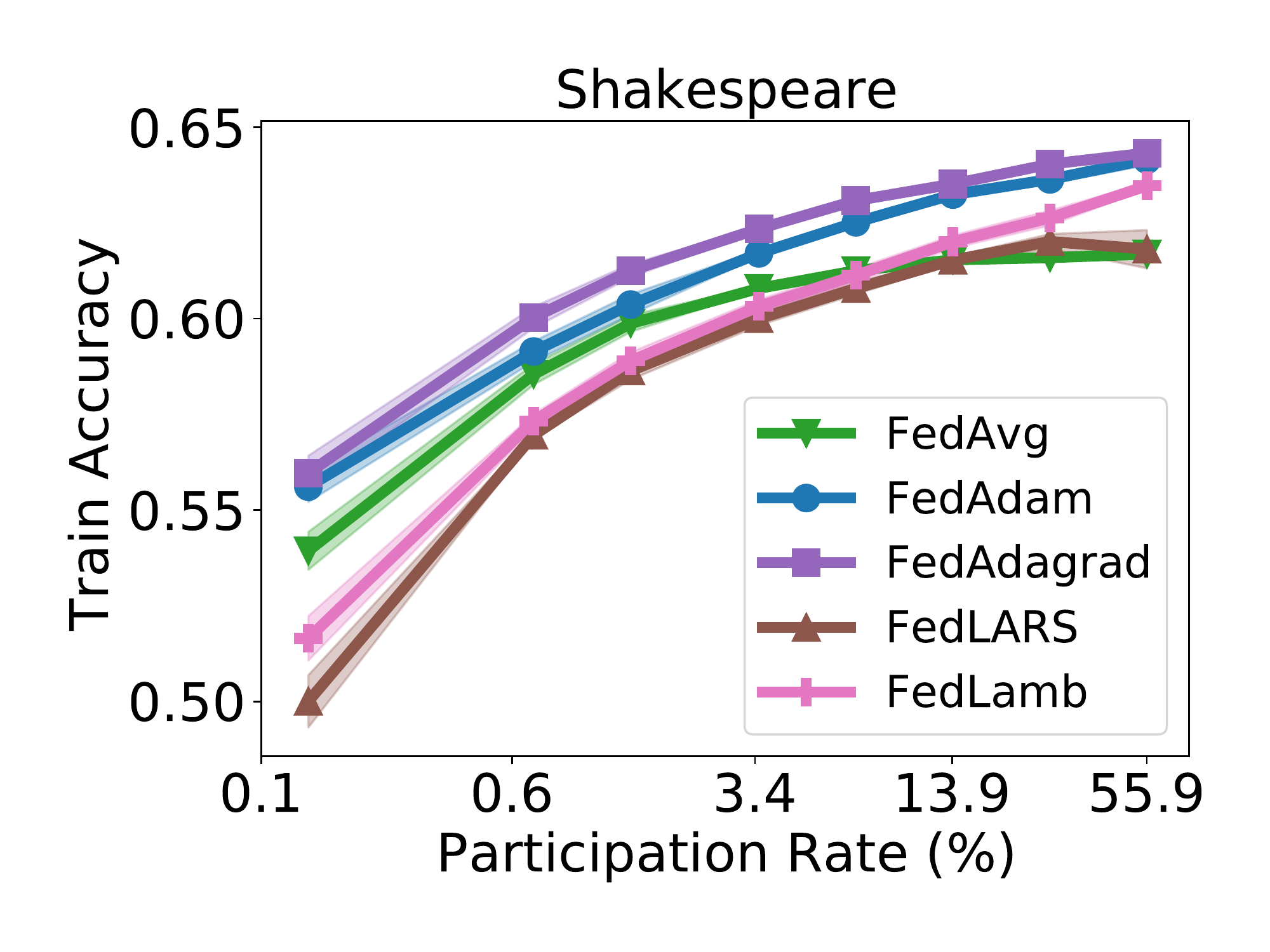}
\end{subfigure}%
\begin{subfigure}{0.24\textwidth}
    \centering
    \includegraphics[width=1\linewidth]{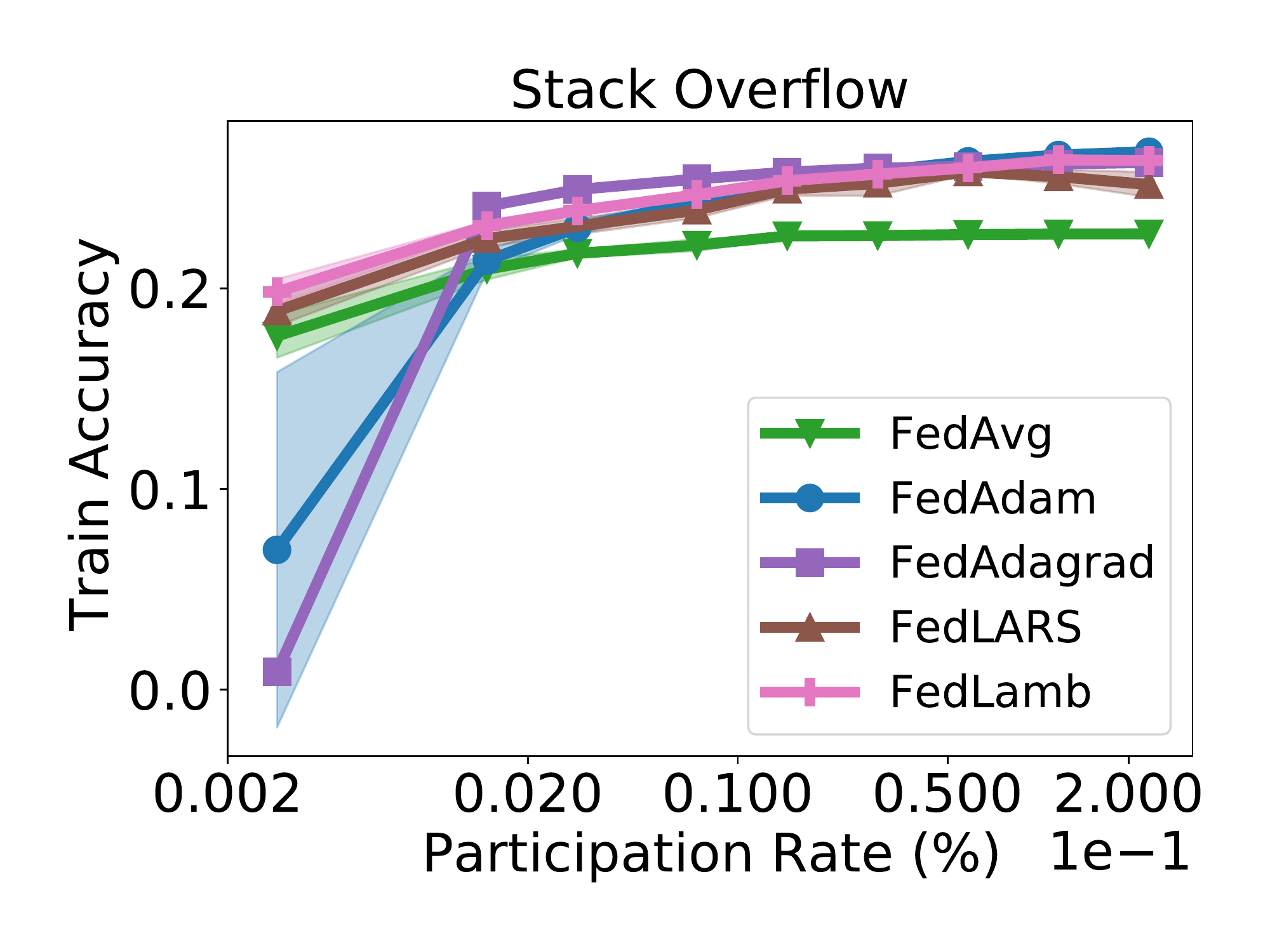}
\end{subfigure}
\caption{Train accuracy of \fedavg, \fedadam, \fedadagrad, \fedlars, and \fedlamb after 1500 rounds, using varying cohort sizes and tasks. The $x$-axis denotes the percentage of training clients in each cohort.}
\label{fig:train_accuracy_versus_cohort_size}
\end{figure}

\begin{figure}[ht!]
\centering
\begin{subfigure}{0.24\textwidth}
    \centering
    \includegraphics[width=1\linewidth]{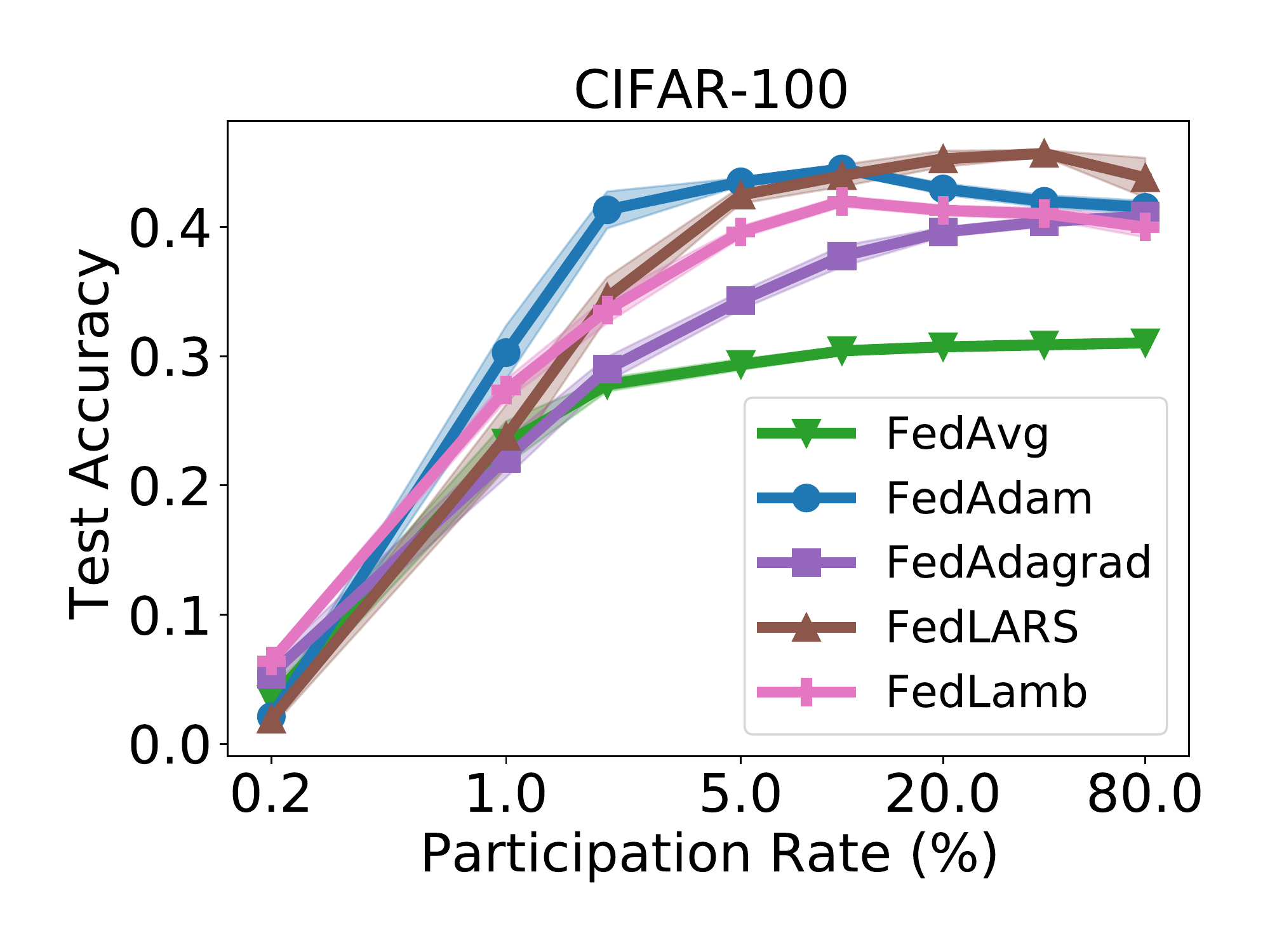}
\end{subfigure}%
\begin{subfigure}{0.24\textwidth}
    \centering
    \includegraphics[width=1\linewidth]{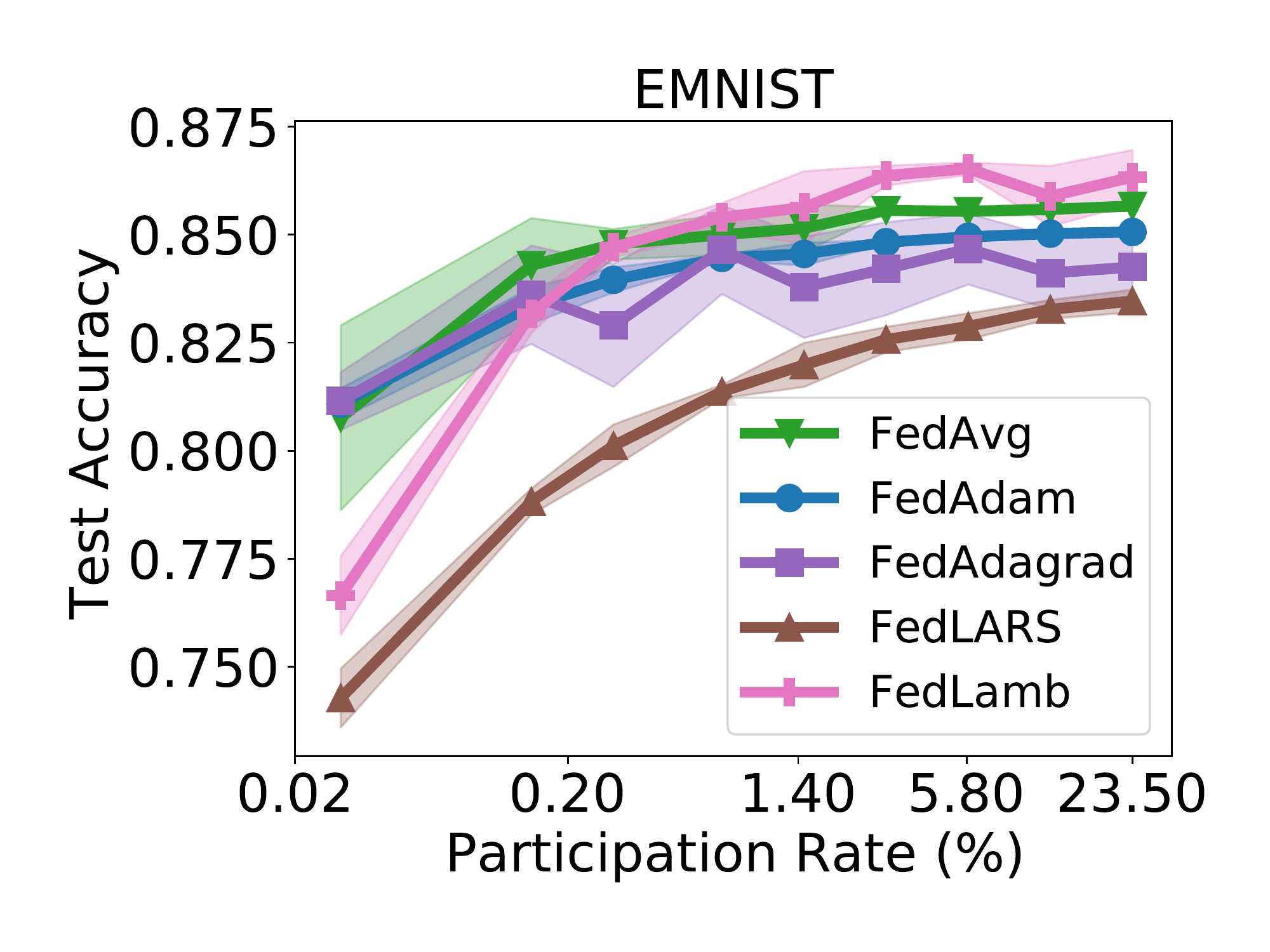}
\end{subfigure}%
\begin{subfigure}{0.24\textwidth}
    \centering
    \includegraphics[width=1\linewidth]{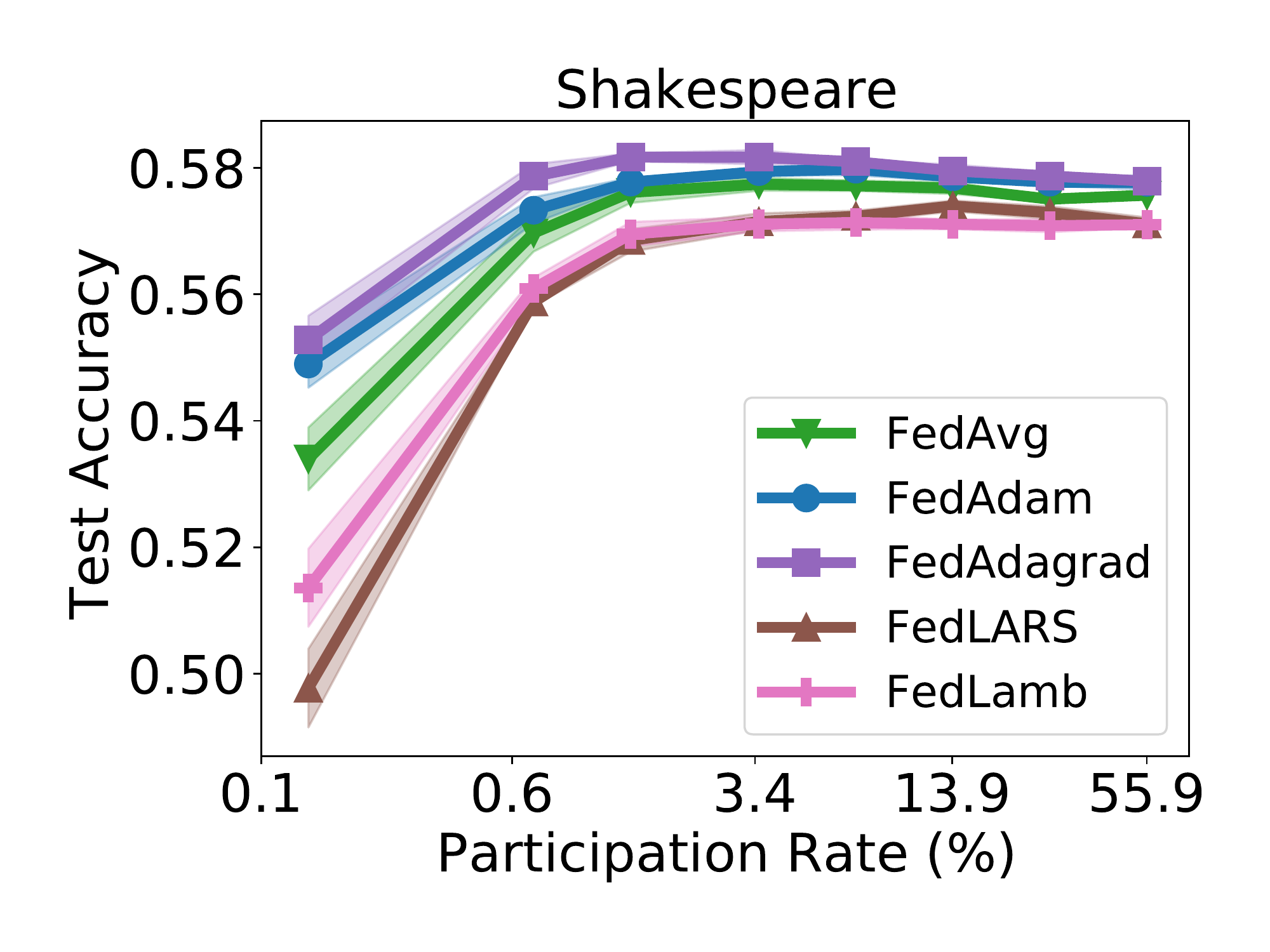}
\end{subfigure}%
\begin{subfigure}{0.24\textwidth}
    \centering
    \includegraphics[width=1\linewidth]{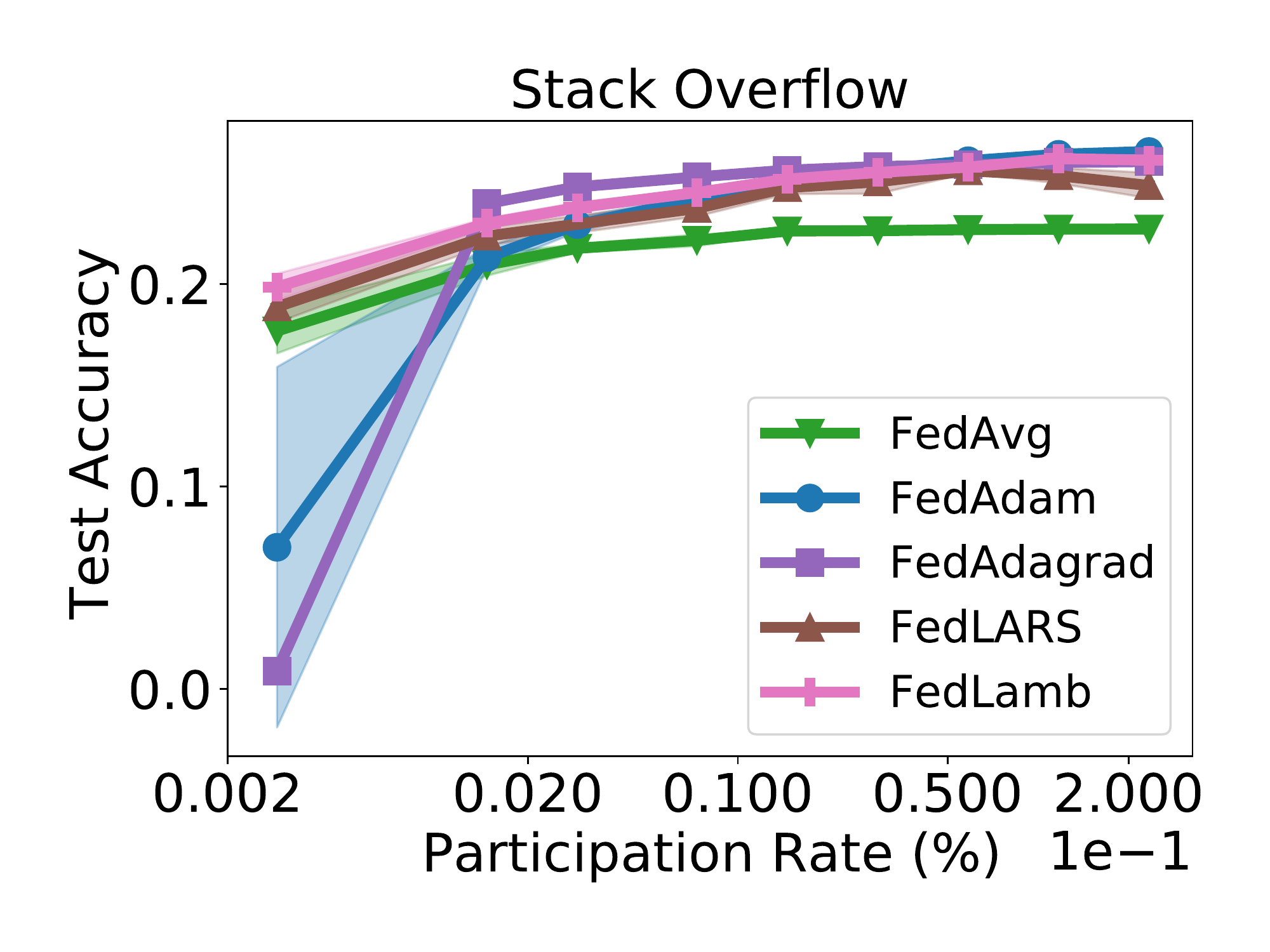}
\end{subfigure}
\caption{Test accuracy of \fedavg, \fedadam, \fedadagrad, \fedlars, and \fedlamb after 1500 rounds, using varying cohort sizes and tasks. The $x$-axis denotes the percentage of training clients in each cohort.}
\label{fig:test_accuracy_versus_cohort_size}
\end{figure}

In this section, we showcase the train and test accuracy of various methods, as a function of the cohort size. The results are given in Figures \ref{fig:train_accuracy_versus_cohort_size} and \ref{fig:test_accuracy_versus_cohort_size}, which correspond to the train and test accuracy, respectively. Both plots give the accuracy of \fedavg, \fedadam, \fedadagrad, \fedlars, and \fedlamb as a function of the \emph{participation rate}. That is, the percentage of training clients used in each cohort.

\FloatBarrier

\subsection{Cohort Size Speedups}\label{appendix:cohort_size_speedups}

\begin{figure}[ht!]
\centering
\begin{subfigure}{0.24\textwidth}
    \centering
    \includegraphics[width=1\linewidth]{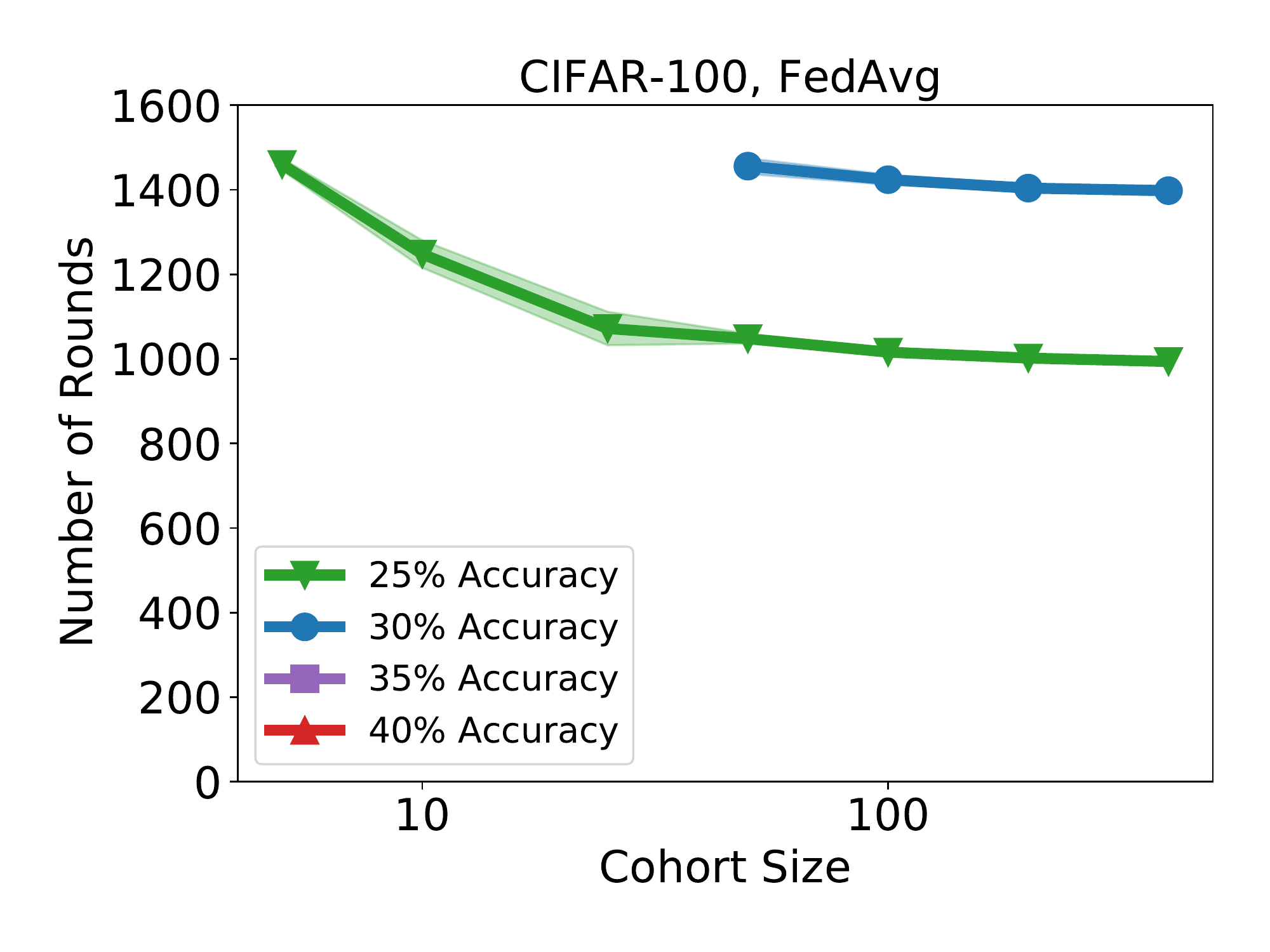}
\end{subfigure}%
\begin{subfigure}{0.24\textwidth}
    \centering
    \includegraphics[width=1\linewidth]{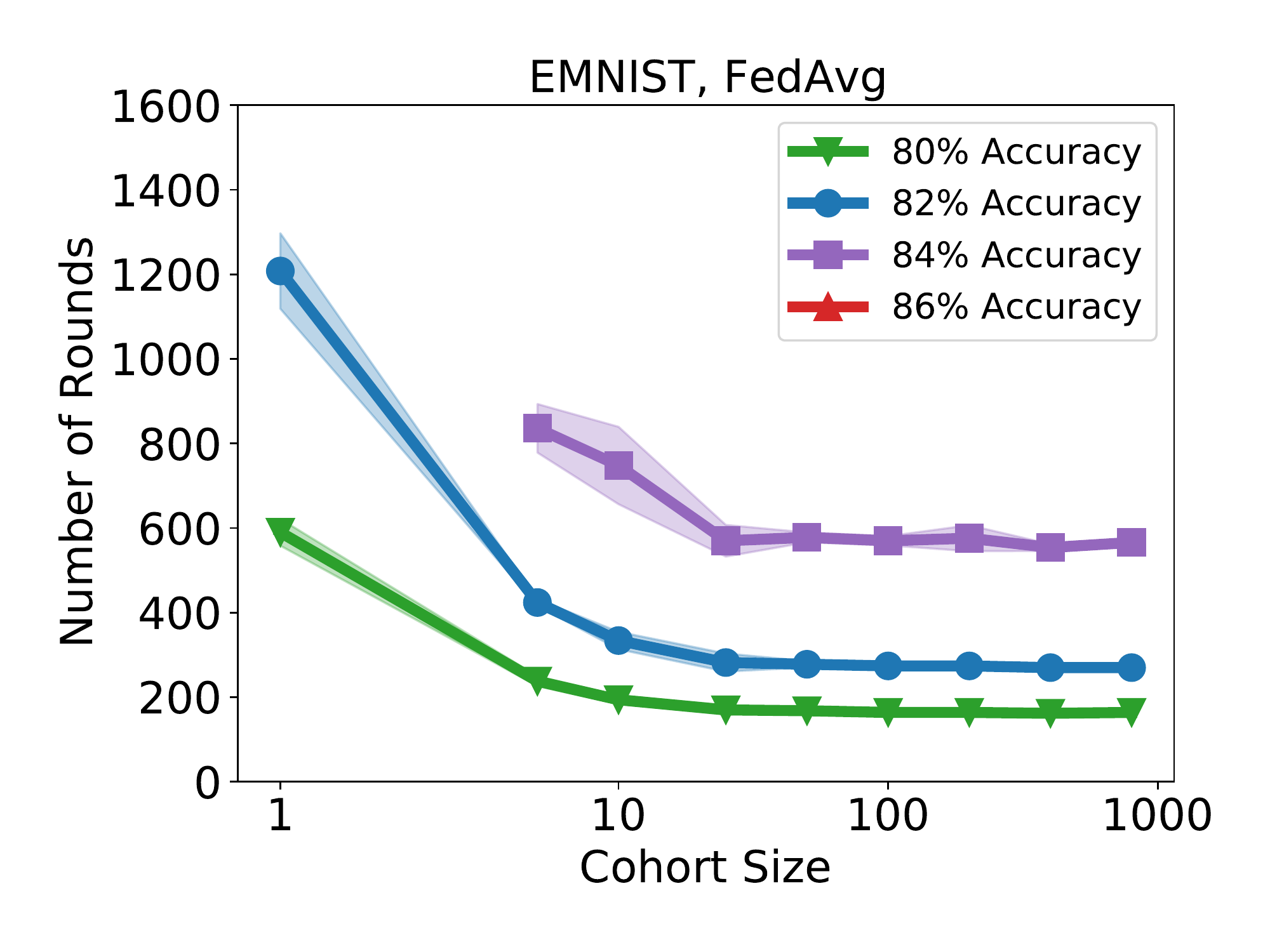}
\end{subfigure}%
\begin{subfigure}{0.24\textwidth}
    \centering
    \includegraphics[width=1\linewidth]{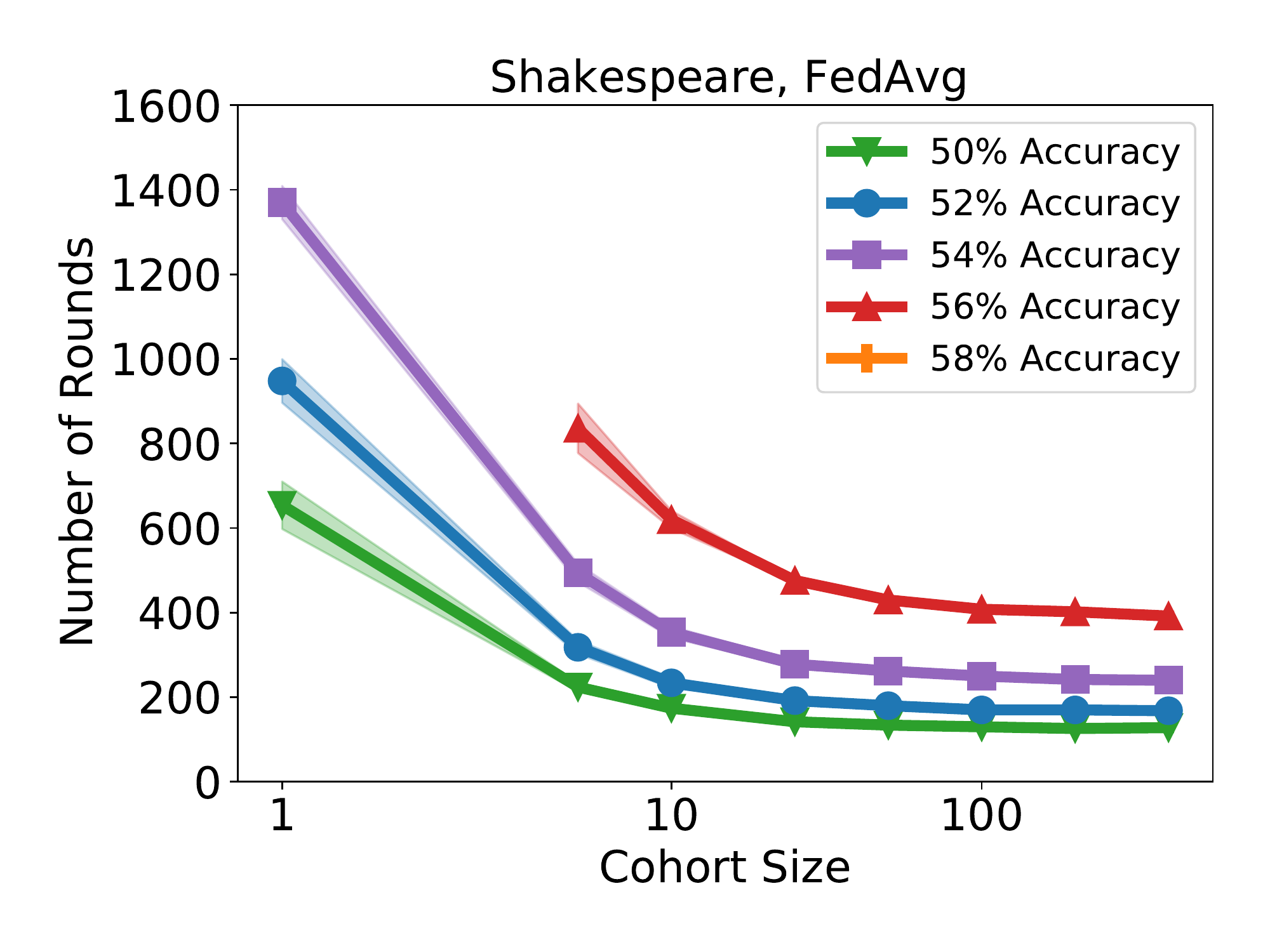}
\end{subfigure}%
\begin{subfigure}{0.24\textwidth}
    \centering
    \includegraphics[width=1\linewidth]{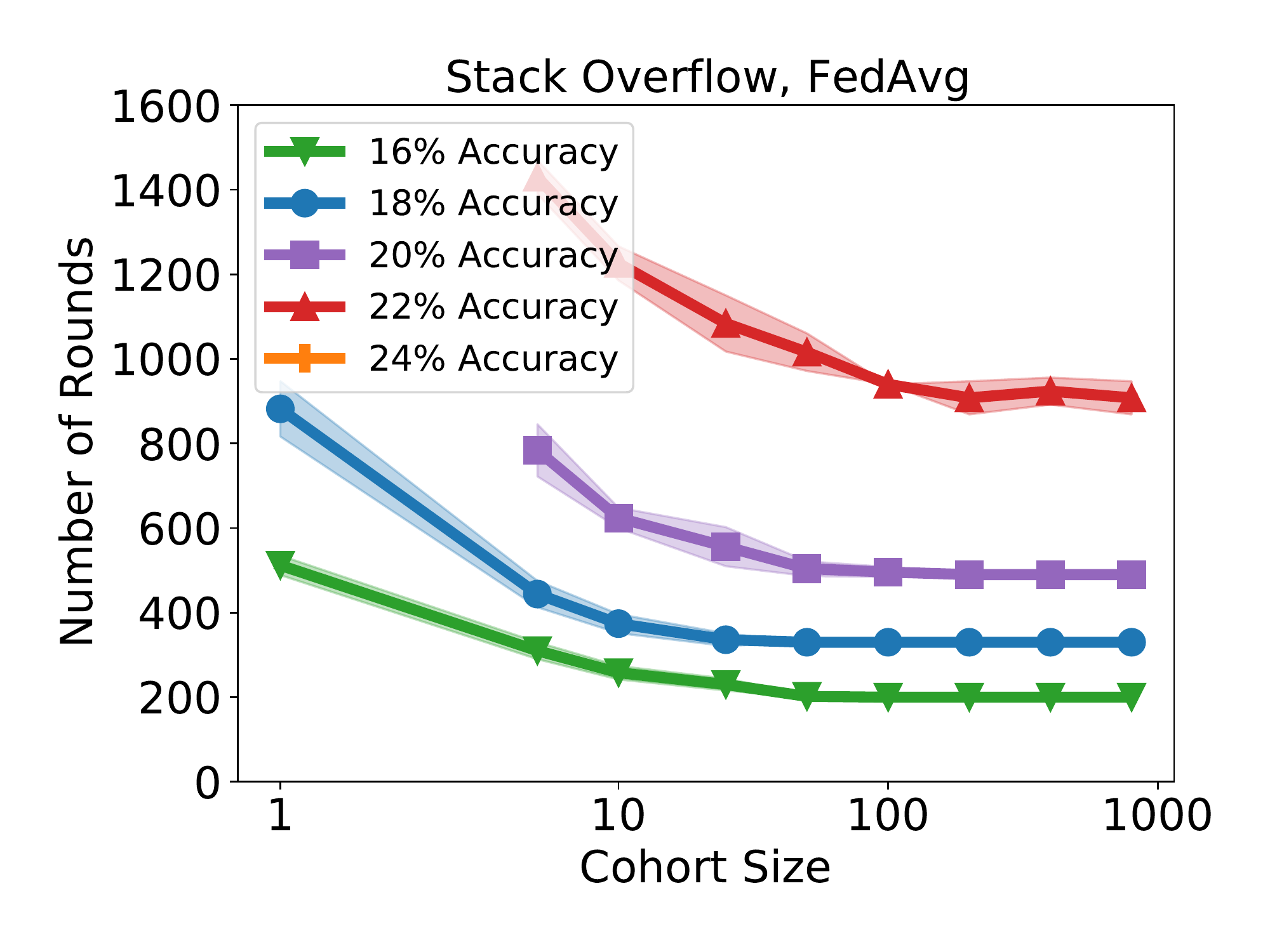}
\end{subfigure}
\caption{Number of communication rounds for \fedavg to obtain certain test accuracy thresholds. The $x$-axis denotes the cohort size.}
\label{fig:fedavg_rounds_to_test_accuracy}
\end{figure}

\begin{figure}[ht!]
\centering
\begin{subfigure}{0.24\textwidth}
    \centering
    \includegraphics[width=1\linewidth]{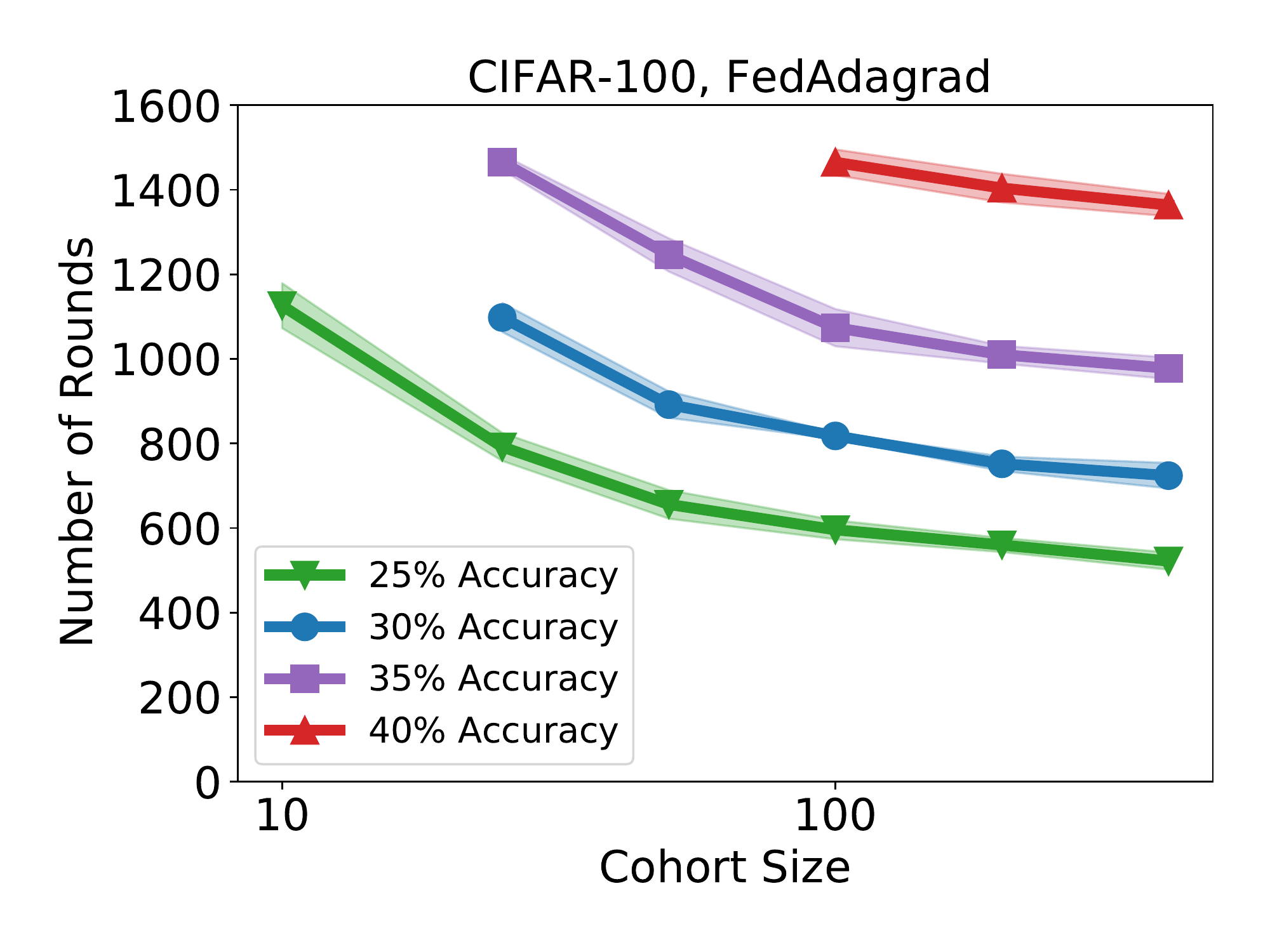}
\end{subfigure}%
\begin{subfigure}{0.24\textwidth}
    \centering
    \includegraphics[width=1\linewidth]{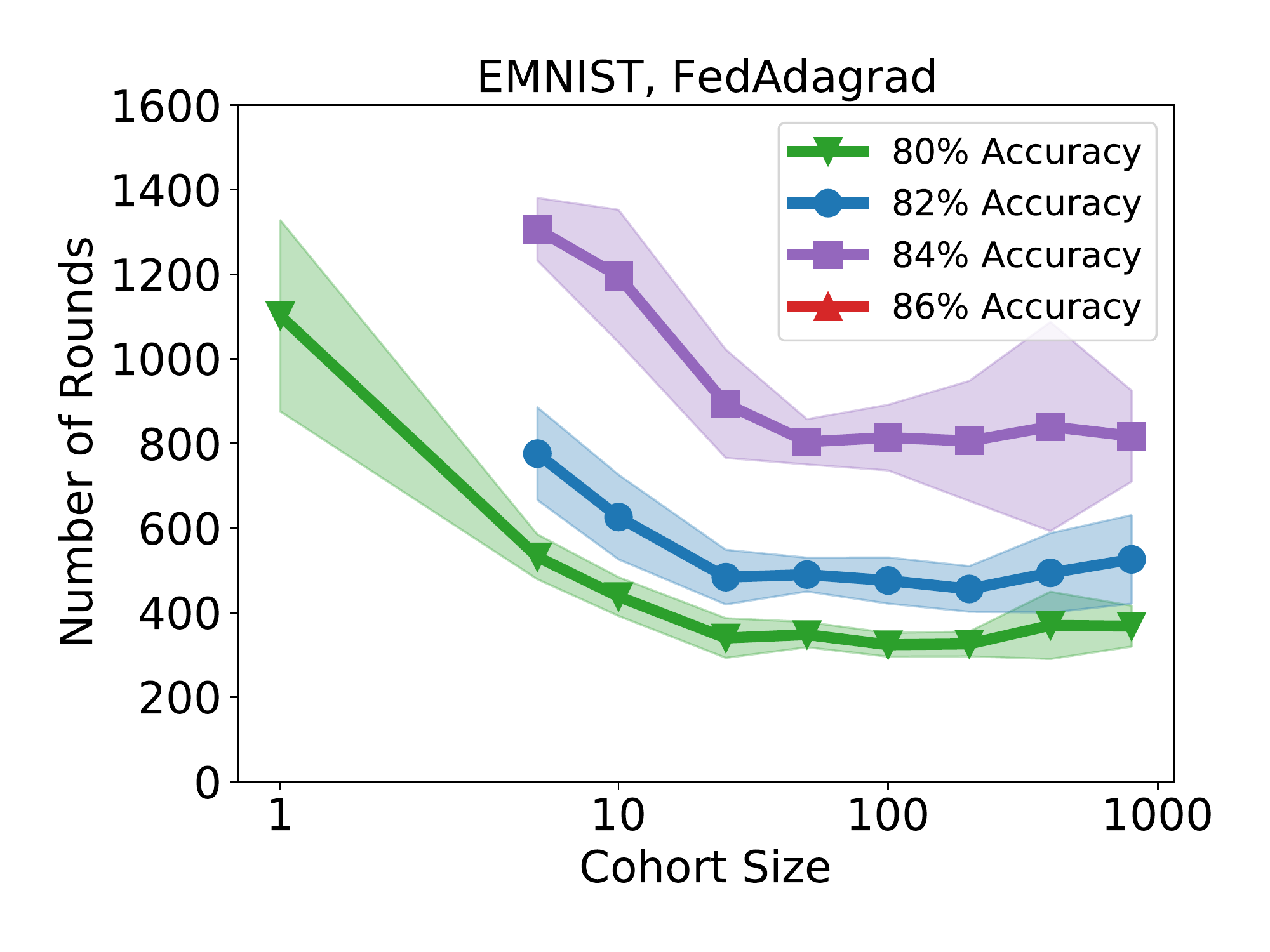}
\end{subfigure}%
\begin{subfigure}{0.24\textwidth}
    \centering
    \includegraphics[width=1\linewidth]{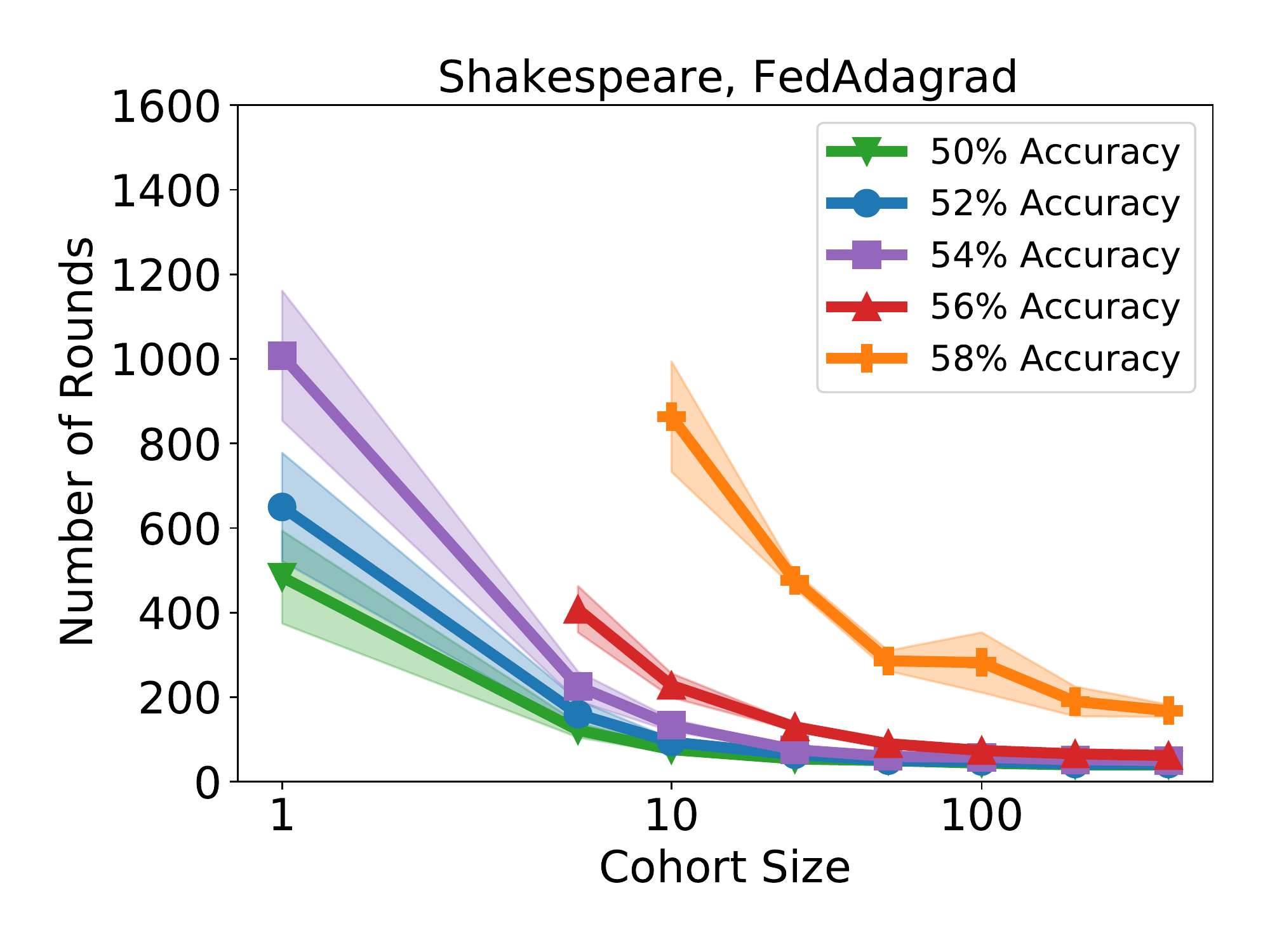}
\end{subfigure}%
\begin{subfigure}{0.24\textwidth}
    \centering
    \includegraphics[width=1\linewidth]{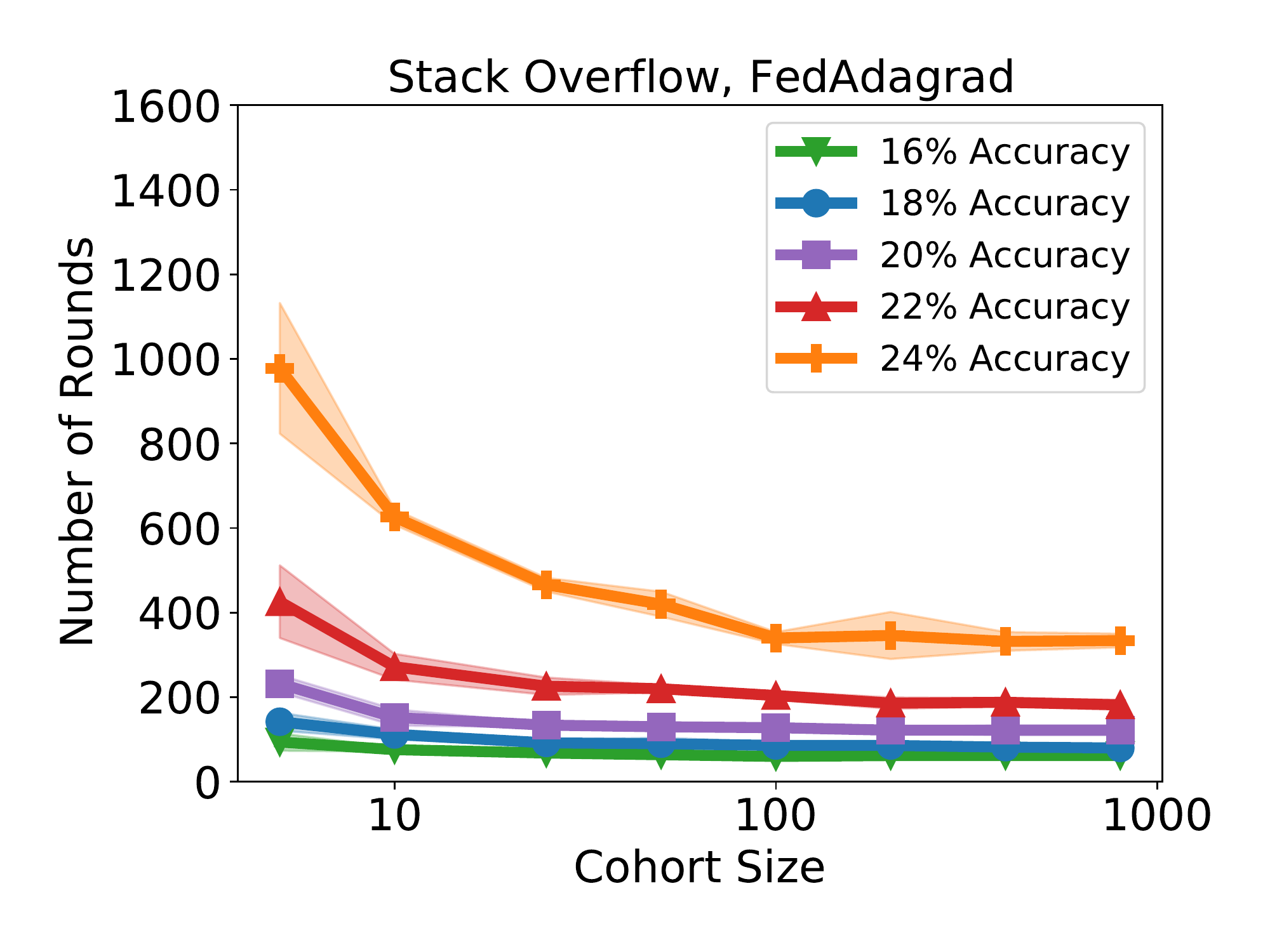}
\end{subfigure}
\caption{Number of communication rounds for \fedadagrad to obtain certain test accuracy thresholds. The $x$-axis denotes the cohort size.}
\label{fig:fedadagrad_rounds_to_test_accuracy}
\end{figure}

\begin{figure}[ht!]
\centering
\begin{subfigure}{0.24\textwidth}
    \centering
    \includegraphics[width=1\linewidth]{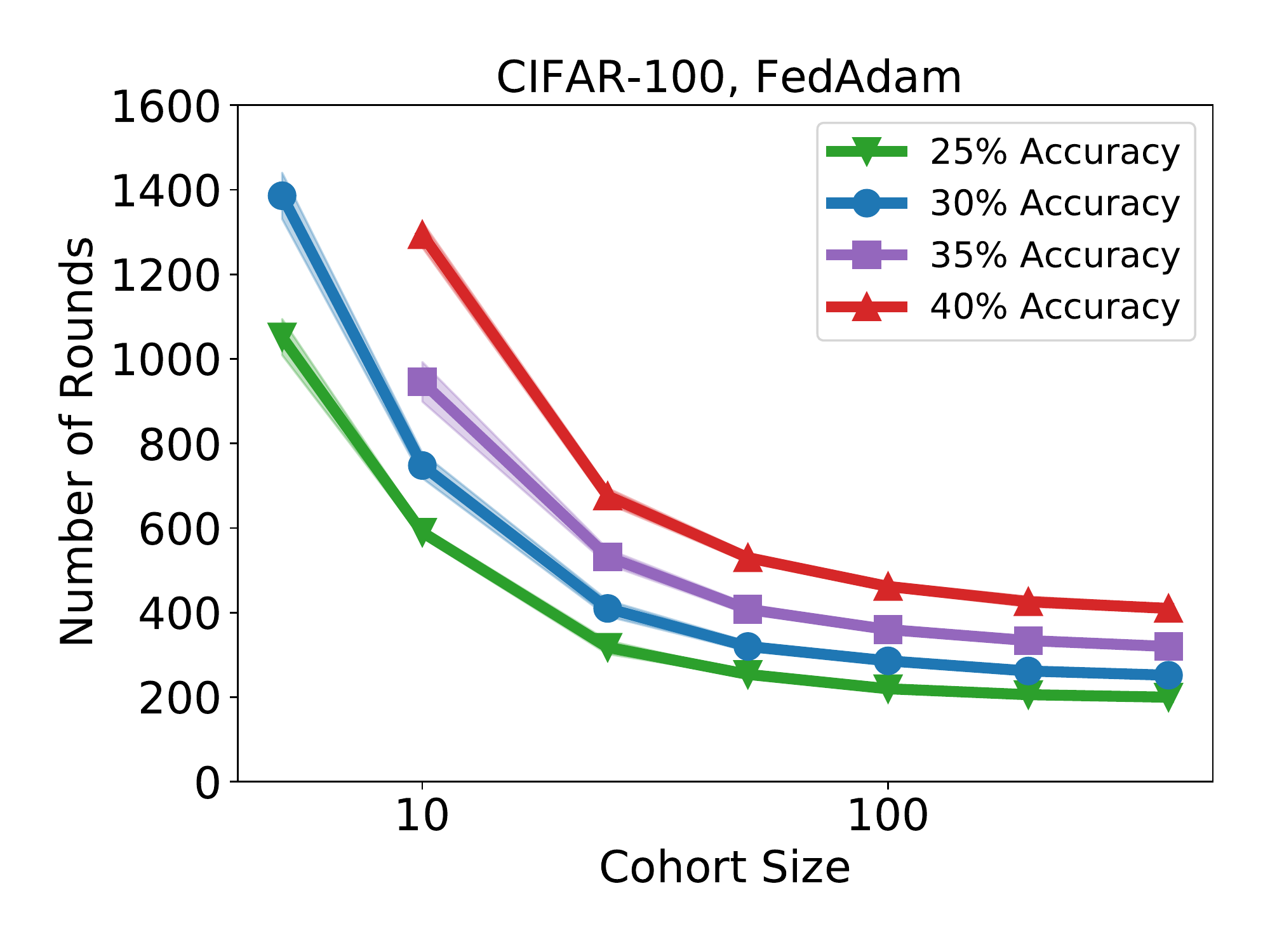}
\end{subfigure}%
\begin{subfigure}{0.24\textwidth}
    \centering
    \includegraphics[width=1\linewidth]{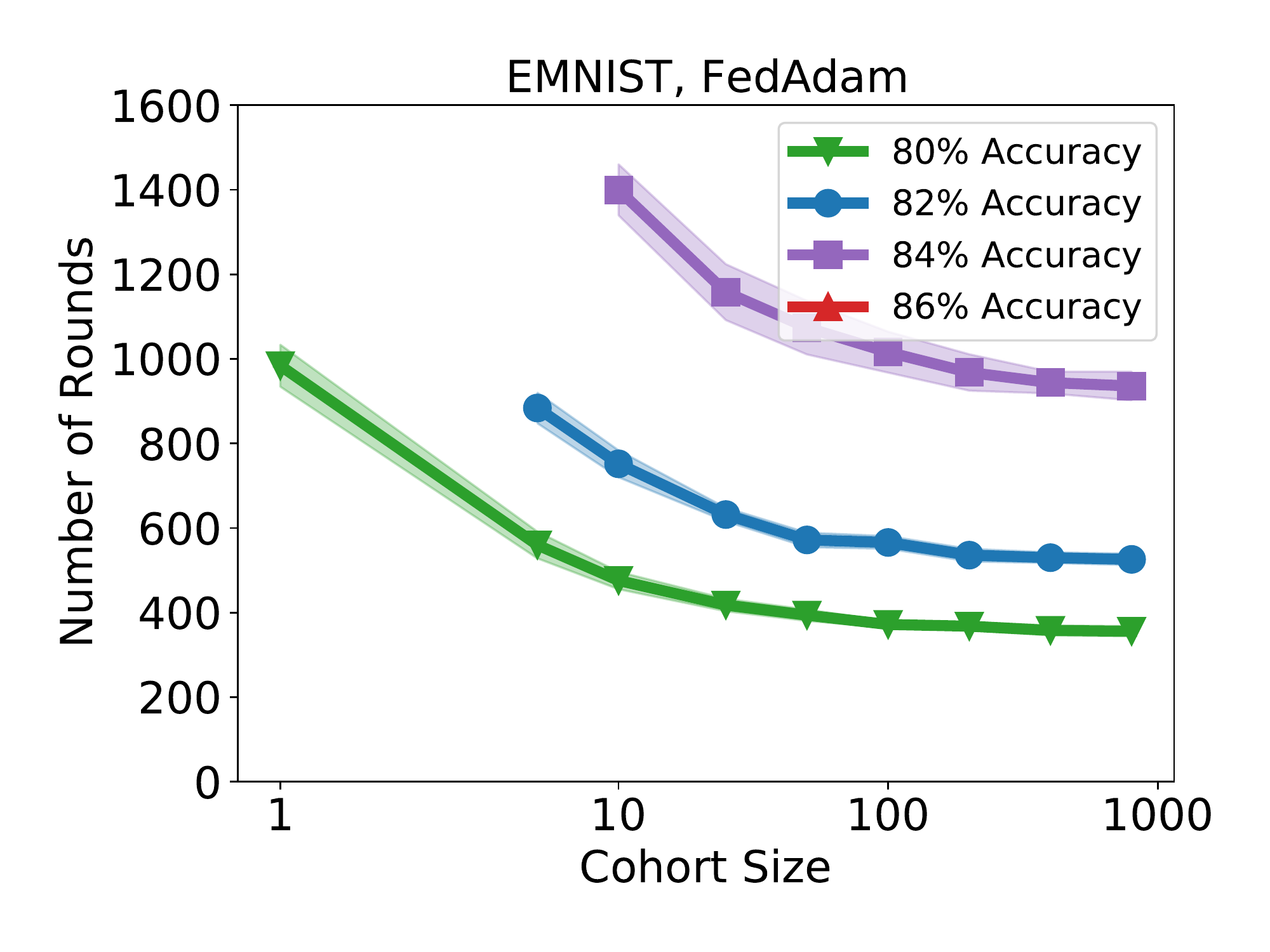}
\end{subfigure}%
\begin{subfigure}{0.24\textwidth}
    \centering
    \includegraphics[width=1\linewidth]{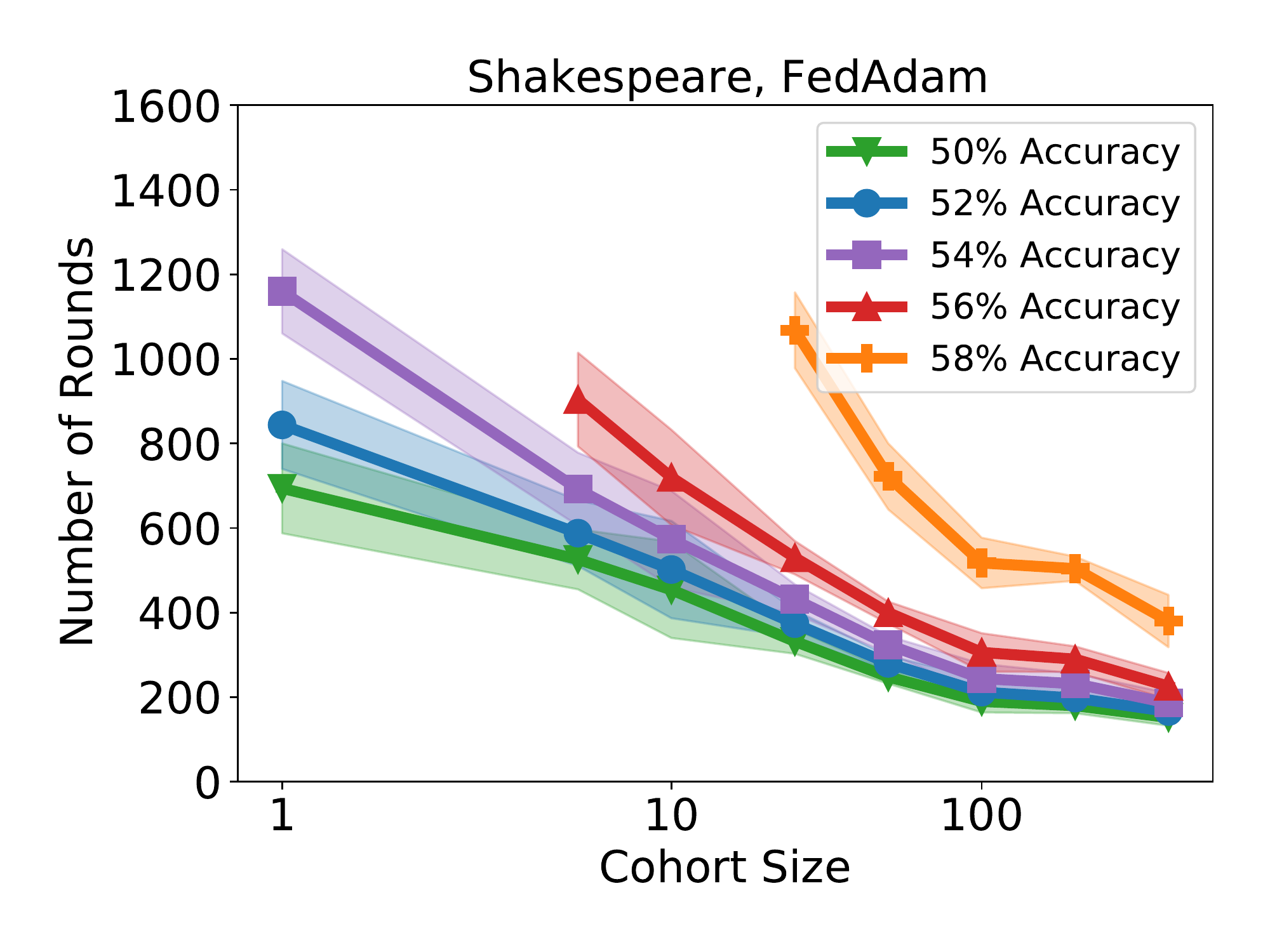}
\end{subfigure}%
\begin{subfigure}{0.24\textwidth}
    \centering
    \includegraphics[width=1\linewidth]{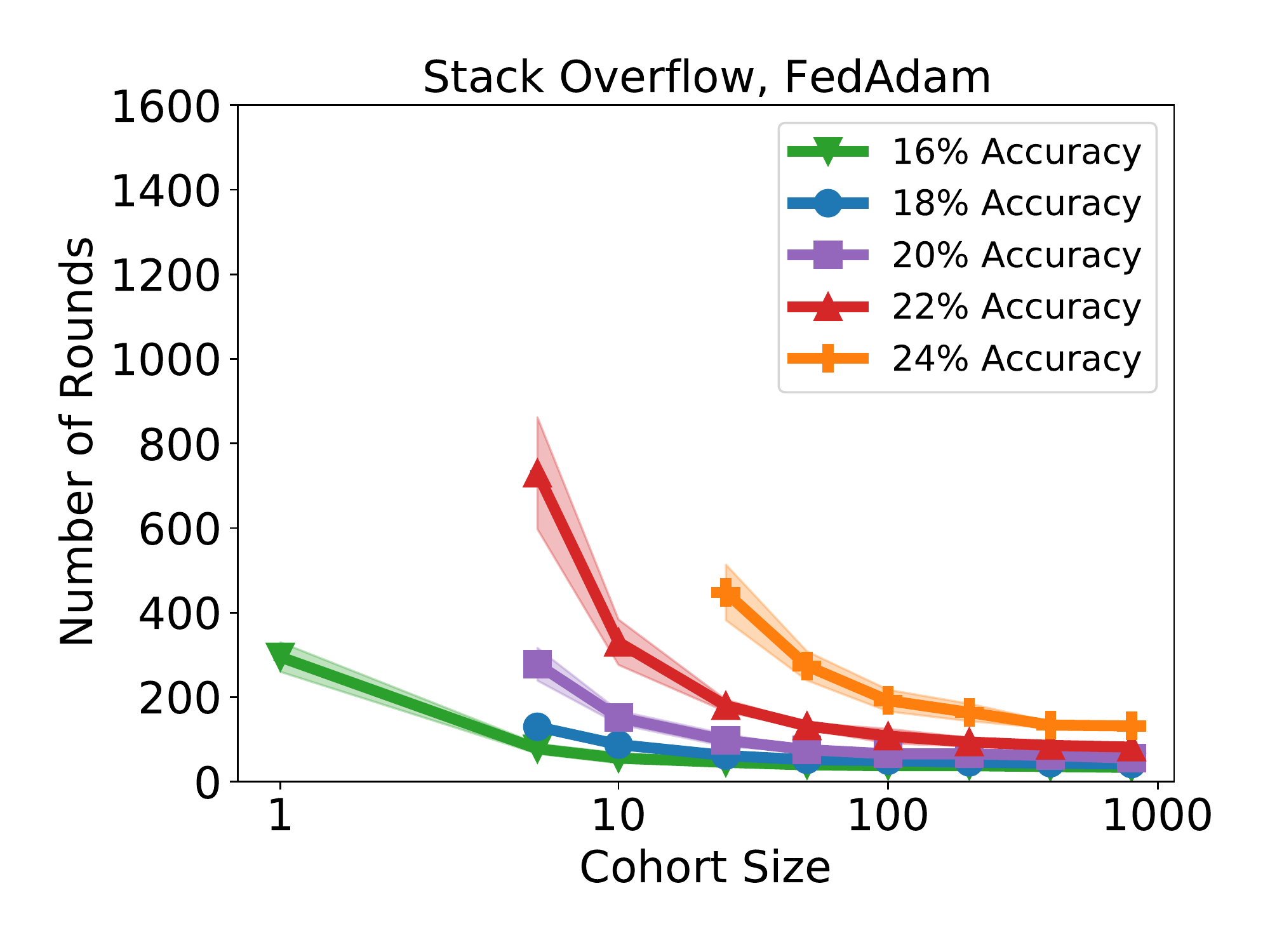}
\end{subfigure}
\caption{Number of communication rounds for \fedadam to obtain certain test accuracy thresholds. The $x$-axis denotes the cohort size.}
\label{fig:fedadam_rounds_to_test_accuracy}
\end{figure}

\begin{figure}[ht!]
\centering
\begin{subfigure}{0.24\textwidth}
    \centering
    \includegraphics[width=1\linewidth]{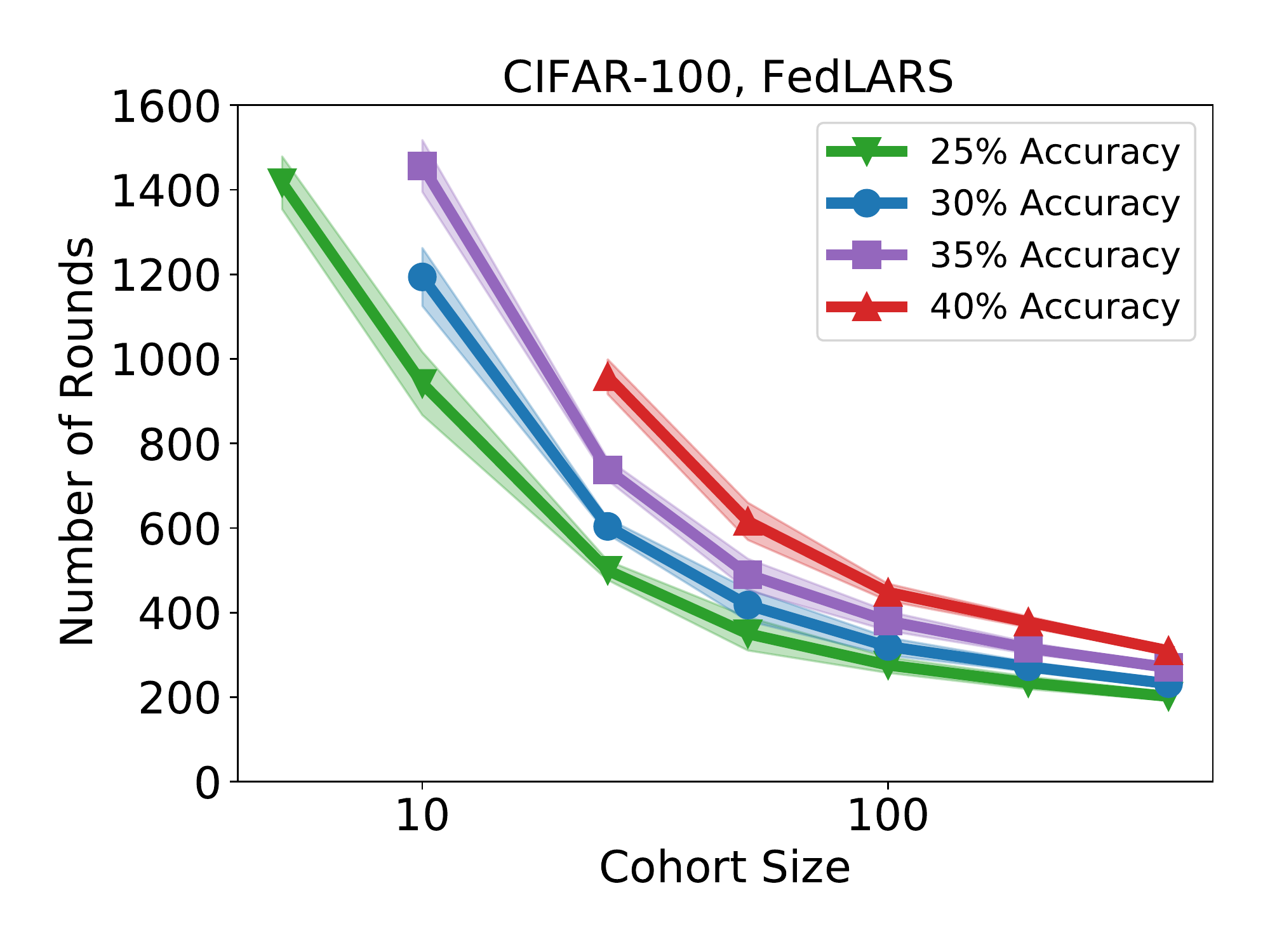}
\end{subfigure}%
\begin{subfigure}{0.24\textwidth}
    \centering
    \includegraphics[width=1\linewidth]{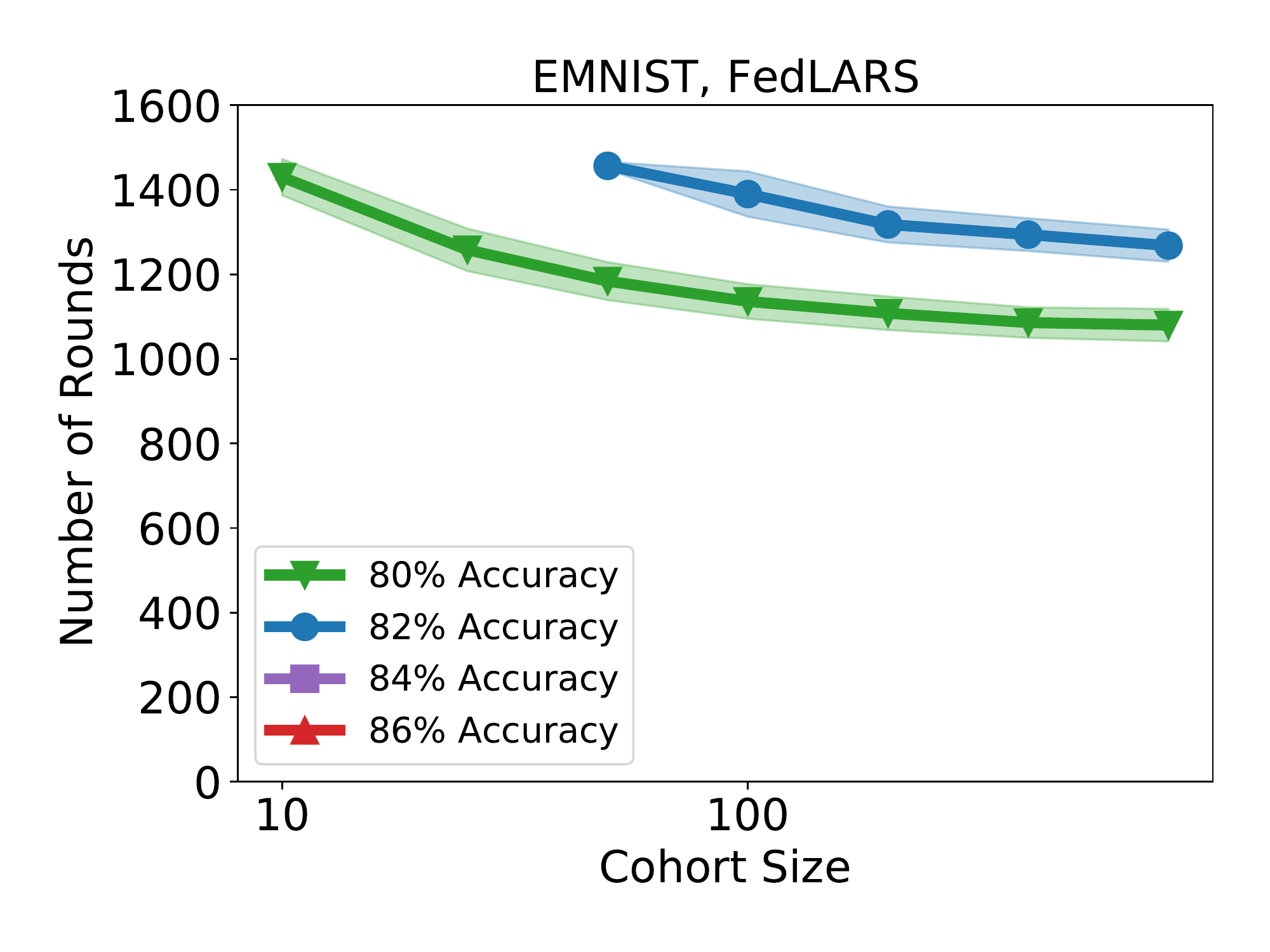}
\end{subfigure}%
\begin{subfigure}{0.24\textwidth}
    \centering
    \includegraphics[width=1\linewidth]{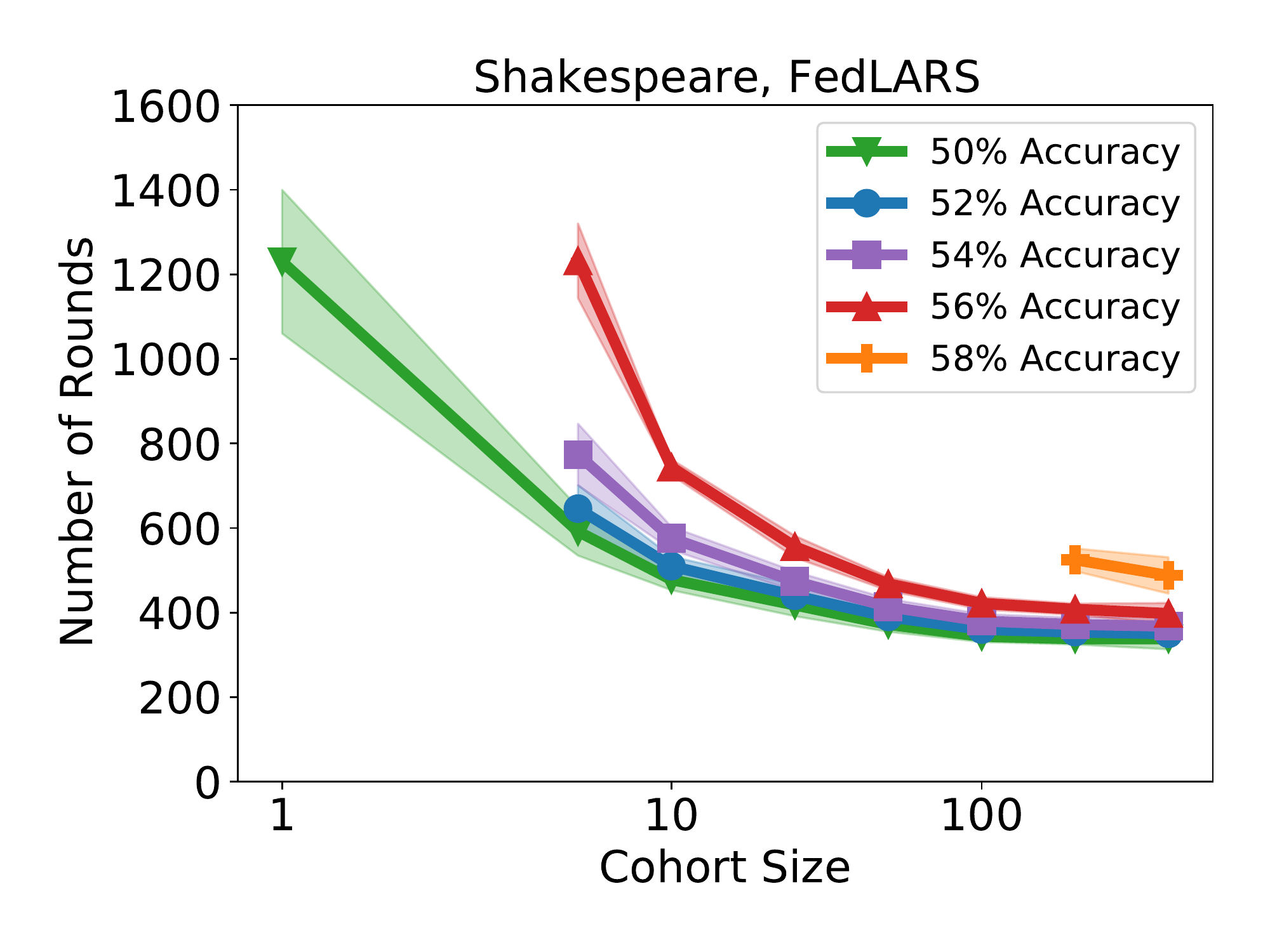}
\end{subfigure}%
\begin{subfigure}{0.24\textwidth}
    \centering
    \includegraphics[width=1\linewidth]{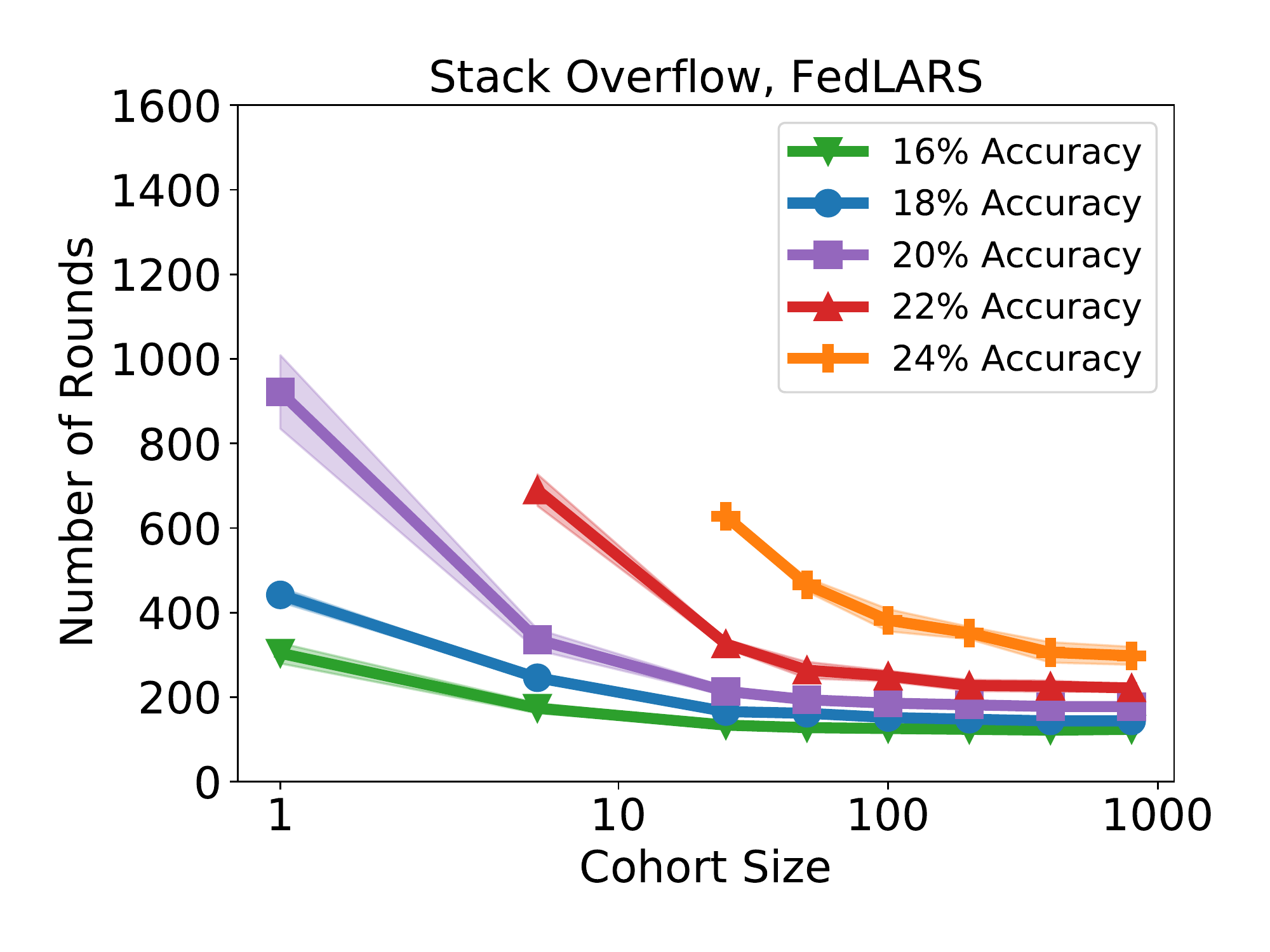}
\end{subfigure}
\caption{Number of communication rounds for \fedlars to obtain certain test accuracy thresholds. The $x$-axis denotes the cohort size.}
\label{fig:fedlars_rounds_to_test_accuracy}
\end{figure}

\begin{figure}[ht!]
\centering
\begin{subfigure}{0.24\textwidth}
    \centering
    \includegraphics[width=1\linewidth]{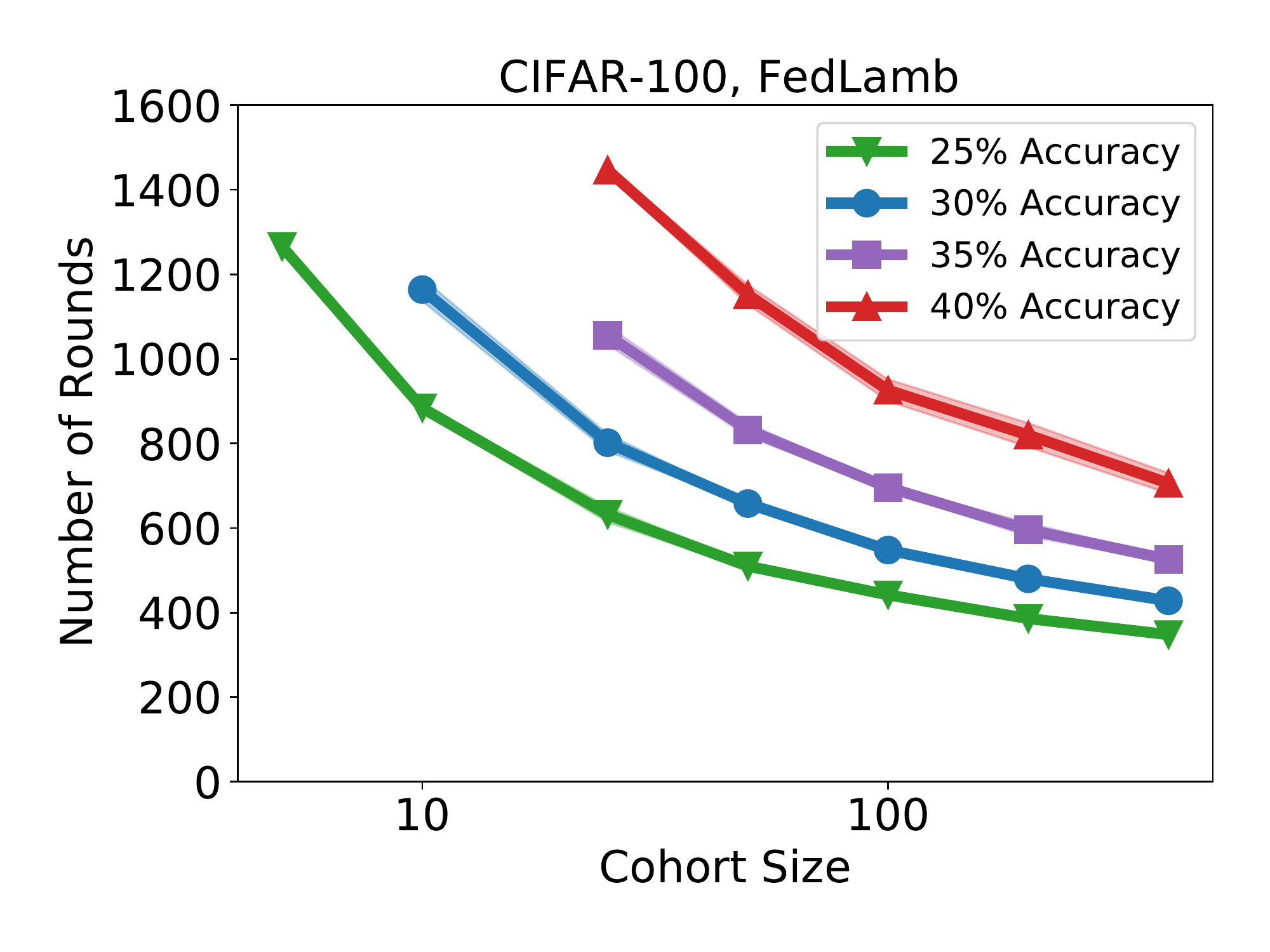}
\end{subfigure}%
\begin{subfigure}{0.24\textwidth}
    \centering
    \includegraphics[width=1\linewidth]{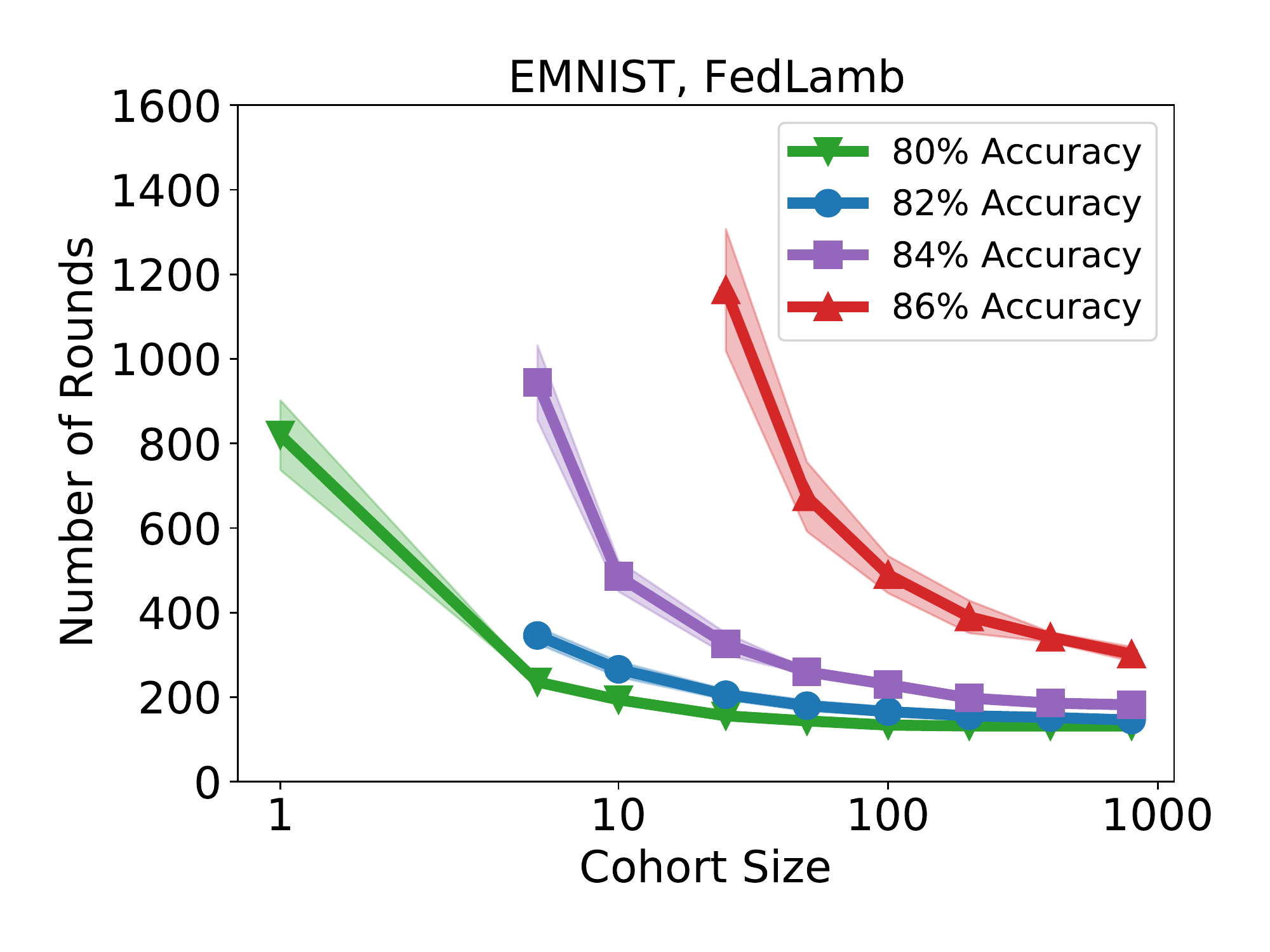}
\end{subfigure}%
\begin{subfigure}{0.24\textwidth}
    \centering
    \includegraphics[width=1\linewidth]{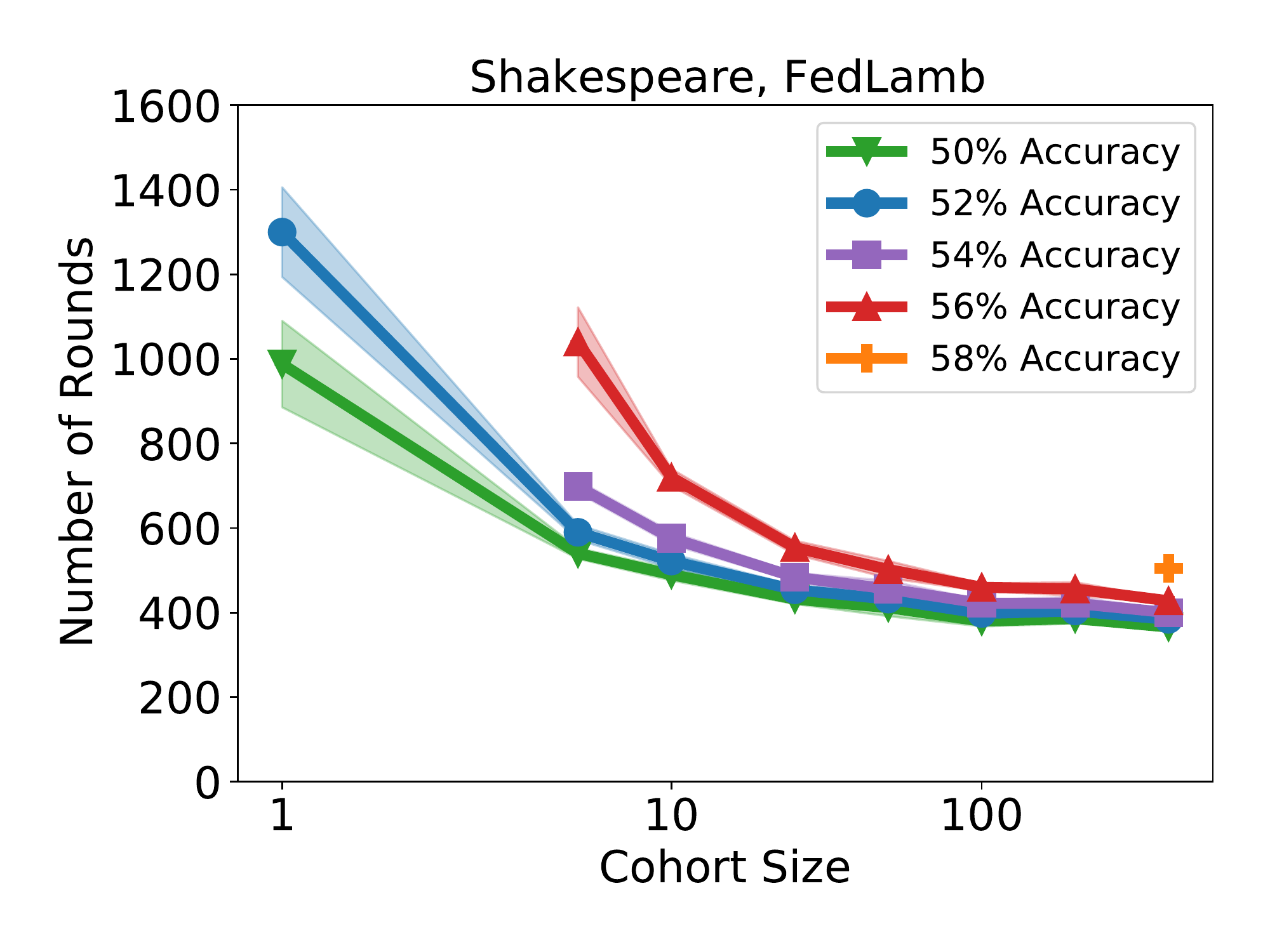}
\end{subfigure}%
\begin{subfigure}{0.24\textwidth}
    \centering
    \includegraphics[width=1\linewidth]{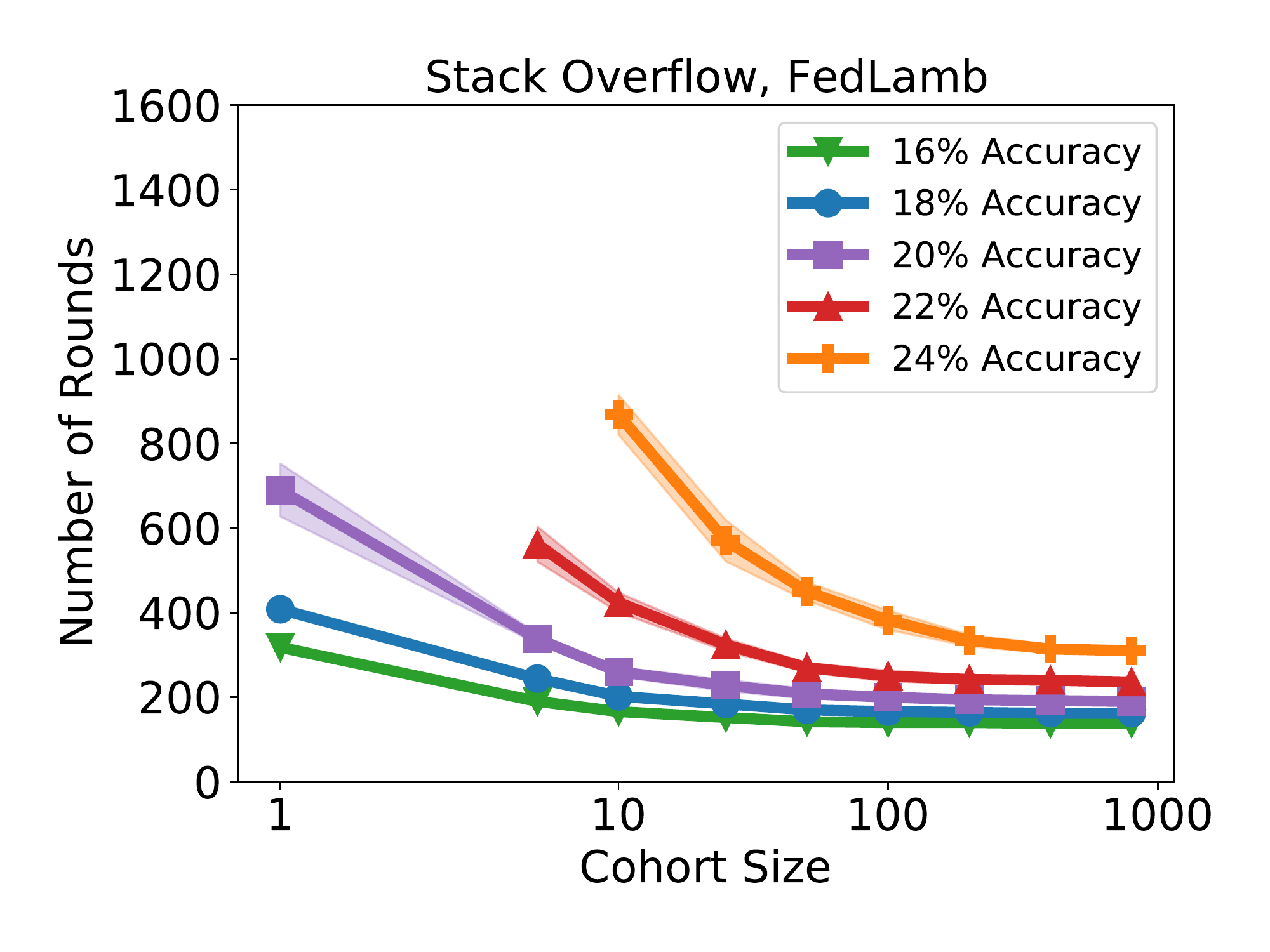}
\end{subfigure}
\caption{Number of communication rounds for \fedlamb to obtain certain test accuracy thresholds. The $x$-axis denotes the cohort size.}
\label{fig:fedlamb_rounds_to_test_accuracy}
\end{figure}

In this section, we attempt to see how much increasing the cohort size can speed up a federated algorithm. In particular, we plot the number of rounds needed to obtain a given accuracy threshold versus the cohort size. The results are given in Figures \ref{fig:fedavg_rounds_to_test_accuracy}, \ref{fig:fedadagrad_rounds_to_test_accuracy}, \ref{fig:fedadam_rounds_to_test_accuracy}, \ref{fig:fedlars_rounds_to_test_accuracy}, and \ref{fig:fedlamb_rounds_to_test_accuracy}. We see that in just about all cases, the speedups incurred by increasing the cohort size do not scale linearly. That being said, we still see that increasing the cohort size generally always leads to a reduction in the number of rounds needed to obtain a given test accuracy, and can lead to accuracy thresholds unobtainable by small-cohort training in communication-limited settings.

While theory shows that in the worst-case, the cohort size leads to linear speedups, we find that this is generally not the case in practice.

\FloatBarrier

\subsection{Measures of Accuracy Across Clients}\label{appendix:fairness}

In this section, we expand on the fairness results in \cref{sec:challenges}. In Tables \ref{table:fedadam_cifar100_client_accuracy_appendix}, \ref{table:fedadam_emnist_client_accuracy_appendix}, \ref{table:fedadam_shakespeare_client_accuracy_appendix}, and \ref{table:fedadam_stackoverflow_word_client_accuracy_appendix}, we present percentiles of accuracy of \fedadam across all test clients (after training for 1500 rounds, with varying cohort sizes and on varying tasks). For example, the 50th percentile of accuracy is the median accuracy of the learned model across all test clients.

The results show that in nearly all cases, the cohort size impacts all percentiles of accuracy in the same manner. For example, in \cref{table:fedadam_cifar100_client_accuracy_appendix}, we see that a cohort size of $M = 50$ is better than other cohort sizes, for all percentiles of test accuracy. Notably, this does not support the notion that larger cohorts learn more fair models. Instead, it seems that large cohorts can lead to generalization failures across all percentiles, as it does on CIFAR-100 and Shakespeare (Tables \ref{table:fedadam_cifar100_client_accuracy_appendix} and \ref{table:fedadam_shakespeare_client_accuracy_appendix}). However, this does not occur on EMNIST and Stack Overflow (Tables \ref{table:fedadam_emnist_client_accuracy_appendix} and \ref{table:fedadam_stackoverflow_word_client_accuracy_appendix}), which have many more train and test clients.

\begin{table}[ht!]
  \caption{Percentiles of accuracy across test clients for \fedadam on CIFAR-100 after 1500 communication rounds.  We present the mean and standard deviation across 5 random trials, with the largest accuracy values for each percentile in bold.}
  \label{table:fedadam_cifar100_client_accuracy_appendix}
  \centering
  \begin{tabular}{lccccc}
    \toprule
    \text{Percentile} & \multicolumn{5}{c}{Cohort Size} \\
    \cmidrule(r){2-6}
     & 10 & 50 & 100 & 200 & 400 \\
    5 & $27.0 \pm 3.3$ & \bm{$32.2 \pm 1.1$} & $30.6 \pm 0.9$ & $29.6 \pm 1.1$ & $29.0 \pm 1.2$ \\
    25 & $35.3 \pm 1.5$ & \bm{$39.0 \pm 0.7$} & $37.5 \pm 0.9$ & $37.2 \pm 1.1$ & $36.4 \pm 1.1$ \\
    50 & $41.1 \pm 1.3$ & \bm{$44.5 \pm 0.5$} & $43.1 \pm 0.7$ & $42.4 \pm 0.5$ & $41.6 \pm 0.7$ \\
    75 & $47.5 \pm 1.1$ & \bm{$50.1 \pm 0.7$} & $48.4 \pm 0.5$ & $47.2 \pm 0.4$ & $47.0 \pm 1.0$ \\
    95 & $54.2 \pm 1.5$ & \bm{$55.6 \pm 1.5$} & $54.6 \pm 1.5$ & $53.8 \pm 1.3$ & $53.6 \pm 1.5$ \\
    \bottomrule
  \end{tabular}
\end{table}

\begin{table}[ht!]
  \caption{Percentiles of accuracy across test clients for \fedadam on EMNIST after 1500 communication rounds.  We present the mean and standard deviation across 5 random trials, with the largest accuracy values for each percentile in bold.}
  \label{table:fedadam_emnist_client_accuracy_appendix}
  \centering
  \small
  \begin{tabular}{lcccccc}
    \toprule
    \text{Percentile} & \multicolumn{6}{c}{Cohort Size} \\
    \cmidrule(r){2-7}
     & 10 & 50 & 100 & 200 & 400 & 800 \\
    5 & $61.9 \pm 2.1$ & $62.5 \pm 2.4$ & $64.3 \pm 1.2$ & $63.8 \pm 1.4$ & $64.3 \pm 1.0$ & \bm{$65.0 \pm 0.9$} \\
    25 & $77.3 \pm 0.7$ & $77.4 \pm 1.4$ & $77.9 \pm 0.4$ & $78.2 \pm 0.4$ & $78.6 \pm 0.5$ & \bm{$78.7 \pm 0.3$} \\
    50 & $84.5 \pm 0.5$ & $85.4 \pm 0.5$ & $85.9 \pm 0.2$ & $85.9 \pm 0.2$ & $86.1 \pm 0.2$ & \bm{$86.2 \pm 0.1$} \\
    75 & $91.2 \pm 0.2$ & $92.0 \pm 0.1$ & $92.0 \pm 0.2$ & $92.3 \pm 0.1$ & $92.3 \pm 0.1$ & \bm{$92.3 \pm 0.0$} \\
    95 & $100.0 \pm 0.0$ & $100.0 \pm 0.0$ & $100.0 \pm 0.0$ & $100.0 \pm 0.0$ & $100.0 \pm 0.0$ & \bm{$100.0 \pm 0.0$} \\
    \bottomrule
  \end{tabular}
\end{table}

\begin{table}[ht!]
  \caption{Percentiles of accuracy across test clients for \fedadam on Shakespeare after 1500 communication rounds.  We present the mean and standard deviation across 5 random trials, with the largest accuracy values for each percentile in bold.}
  \label{table:fedadam_shakespeare_client_accuracy_appendix}
  \centering
  \begin{tabular}{lccccc}
    \toprule
    \text{Percentile} & \multicolumn{5}{c}{Cohort Size} \\
    \cmidrule(r){2-6}
     & 10 & 50 & 100 & 200 & 400 \\
    5 & \bm{$39.9 \pm 0.8$} & $39.2 \pm 1.8$ & $39.4 \pm 1.6$ & $37.8 \pm 0.7$ & $38.6 \pm 1.2$ \\
    25 & $54.9 \pm 0.1$ & \bm{$55.0 \pm 0.2$} & $54.9 \pm 0.2$ & $54.9 \pm 0.1$ & $54.9 \pm 0.2$ \\
    50 & $58.3 \pm 0.2$ & $58.5 \pm 0.2$ & \bm{$58.5 \pm 0.2$} & $58.3 \pm 0.1$ & $58.4 \pm 0.1$ \\
    75 & $61.8 \pm 0.2$ & \bm{$62.4 \pm 0.2$} & $62.2 \pm 0.2$ & $62.1 \pm 0.2$ & $62.1 \pm 0.3$ \\
    95 & $70.9 \pm 0.6$ & \bm{$71.2 \pm 0.6$} & $71.2 \pm 0.4$ & $71.1 \pm 0.2$ & $71.1 \pm 0.2$ \\
    \bottomrule
  \end{tabular}
\end{table}

\begin{table}[ht!]
  \caption{Percentiles of accuracy across test clients for \fedadam on Stack Overflow after 1500 communication rounds.  We present the mean and standard deviation across 5 random trials, with the largest accuracy values for each percentile in bold.}
  \label{table:fedadam_stackoverflow_word_client_accuracy_appendix}
  \centering
  \begin{tabular}{lcccccc}
    \toprule
    \text{Percentile} & \multicolumn{6}{c}{Cohort Size} \\
    \cmidrule(r){2-7}
     & 10 & 50 & 100 & 200 & 400 & 800 \\
    5 & $16.7 \pm 0.2$ & $18.7 \pm 0.2$ & $19.1 \pm 0.2$ & $19.5 \pm 0.1$ & $19.8 \pm 0.1$ & \bm{$19.9 \pm 0.1$} \\
    25 & $21.0 \pm 0.3$ & $23.2 \pm 0.2$ & $23.6 \pm 0.2$ & $24.1 \pm 0.1$ & $24.4 \pm 0.1$ & \bm{$24.5 \pm 0.1$} \\
    50 & $23.5 \pm 0.3$ & $25.8 \pm 0.2$ & $26.3 \pm 0.3$ & $26.7 \pm 0.1$ & $27.0 \pm 0.1$ & \bm{$27.2 \pm 0.1$} \\
    75 & $26.1 \pm 0.3$ & $28.4 \pm 0.2$ & $29.0 \pm 0.3$ & $29.4 \pm 0.1$ & $29.7 \pm 0.1$ & \bm{$29.9 \pm 0.1$} \\
    95 & $30.8 \pm 0.3$ & $33.2 \pm 0.2$ & $33.7 \pm 0.4$ & $34.2 \pm 0.1$ & $34.6 \pm 0.1$ & \bm{$34.8 \pm 0.1$} \\
    \bottomrule
  \end{tabular}
\end{table}

\FloatBarrier

\subsection{Simulating Straggler Effects}\label{appendix:stragglers}

As shown in \cref{sec:data_efficiency}, large-cohort training methods seem to face data-efficiency issues, where training with large cohorts requires processing many more examples to reach accuracy thresholds than small-cohort training. While this is related to diminishing returns (\cref{sec:diminishing_returns}) and occurs in large-batch training as well~\citep{golmant2018computational}, we highlight this issue due to its consequences in federated learning.

Unlike centralized learning, federated learning faces fundamental limits on parallelization. Since data cannot be shared, we typically cannot scale up to arbitrarily large cohort sizes. Instead, the parallelization is limited by the available training clients. In order to learn on a client's local dataset, that client must actually perform the training on its examples. Unfortunately, since clients are often lightweight in cross-device settings~\citep{kairouz2019advances}, clients with many examples may require longer compute times, becoming stragglers in a given communication round. If an algorithm is data-inefficient, these straggler clients may have to participate many times throughout training, causing the overall runtime to be greater. In short, data inefficiency can dramatically slow down large-cohort training algorithms.

To exemplify this, we compute simulated runtimes of federated algorithms under a version of the probabilistic straggler model from \citep{lee2017speeding}. We model each client's runtime as a random variable drawn from a shifted exponential distribution. Such models were found to be good models of runtimes for file queries in cloud storage systems~\citep{liang2014tofec} and mini-batch \sgd on distributed compute systems~\citep{lee2017speeding}.

In our model, we assume that the time a client requires to perform local training is some constant proportional to the number of examples the client has, plus an exponential random variable. More formally, let $N_k$ denote the number of examples held by some client $k$, and let $X_k$ denote the amount of time required by client $k$ to perform their local training in \cref{alg:fedopt}. Then we assume that there are constants $\alpha, \lambda > 0$ such that
\[
X_k - \alpha N_k \sim \text{Exp}\left(\frac{1}{\lambda N_k}\right).
\]
Here $\lambda$ is the \emph{straggler parameter}. Recall that if $X \sim \text{Exp}(1/\lambda)$, then $\E[X] = \lambda$. Therefore, we assume that the expected runtime of client $k$ equal $\alpha N$ plus some random variable whose expected value is $\lambda N$. Thus, larger $\lambda$ means larger expected client runtimes. By convention, we can also use $\lambda = 0$ in which case $X_k = \alpha N_k$. For a given round $t$ of \cref{alg:fedopt}, let $C_t$ denote the cohort sampled. Since \cref{alg:fedopt} requires all clients to finish before updating its global model, we model the runtime $Y_t$ of round $t$ as
\[
Y_t = \max_{k \in C_t}\left\{X_k \right\}.
\]

\begin{figure}[!ht]
\centering
\begin{subfigure}{0.24\textwidth}
    \centering
    \includegraphics[width=1\linewidth]{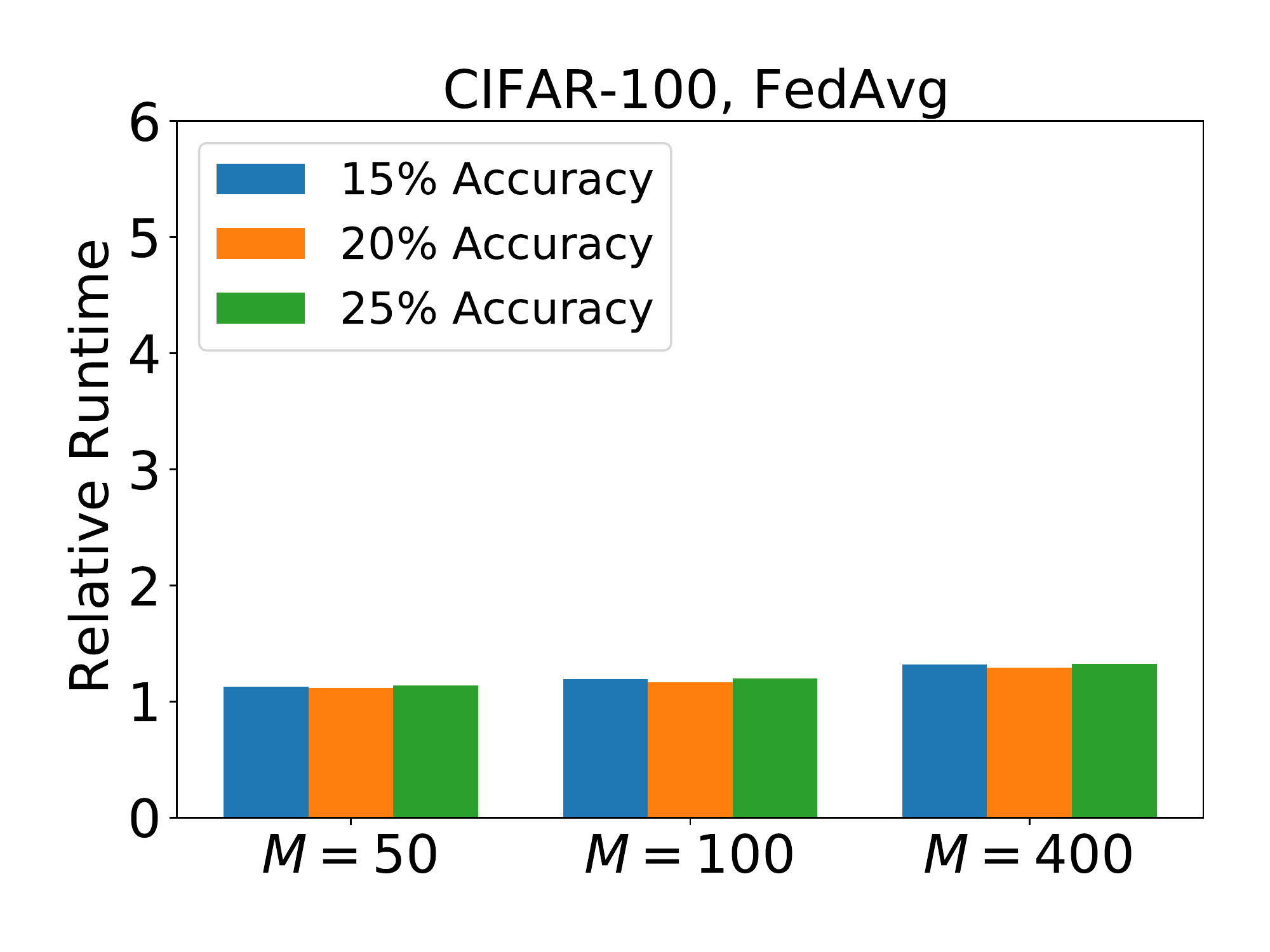}
    \caption{$\lambda = 0.1$}
\end{subfigure}%
\begin{subfigure}{0.24\textwidth}
    \centering
    \includegraphics[width=1\linewidth]{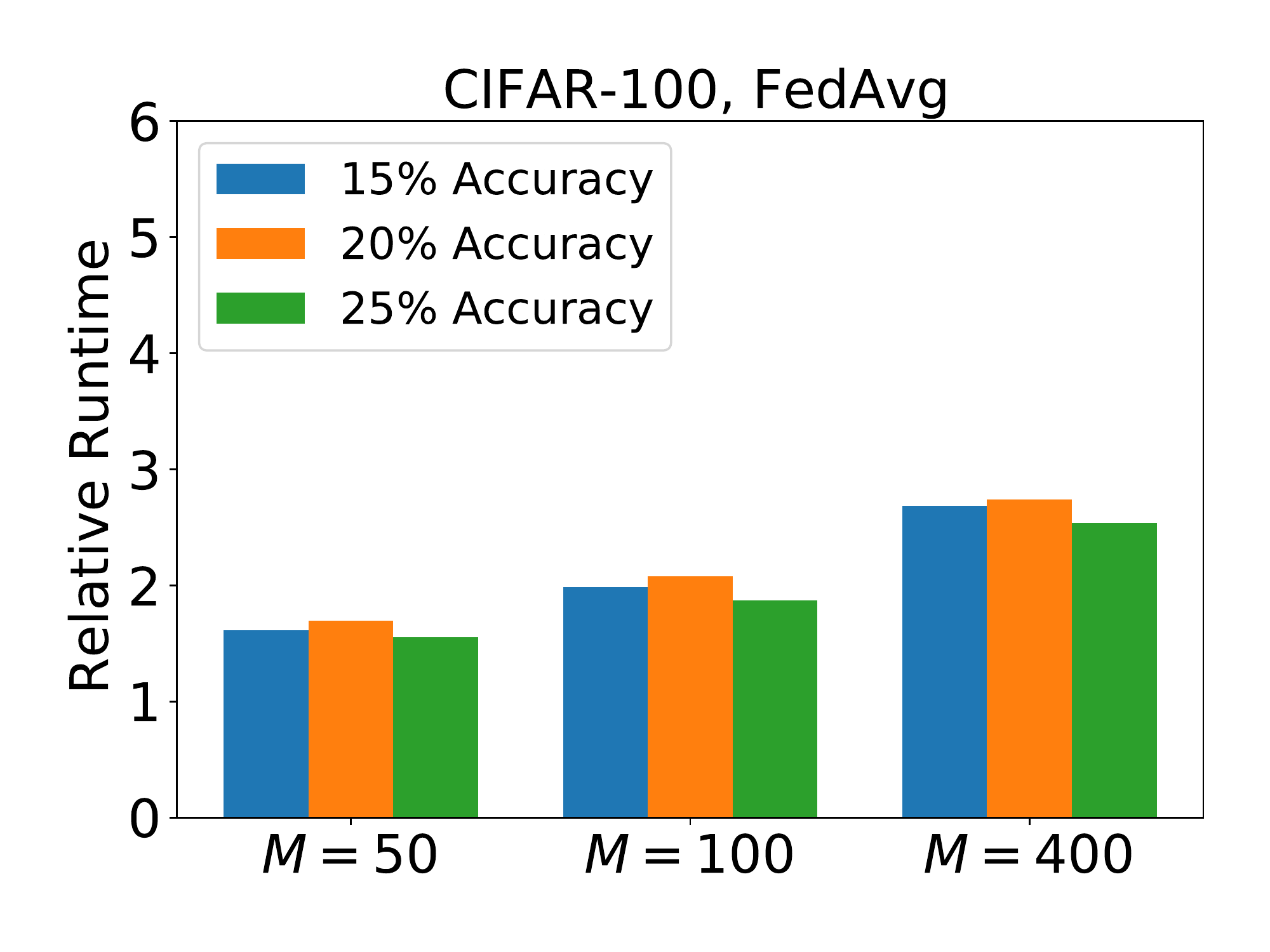}
    \caption{$\lambda = 1$}
\end{subfigure}%
\begin{subfigure}{0.24\textwidth}
    \centering
    \includegraphics[width=1\linewidth]{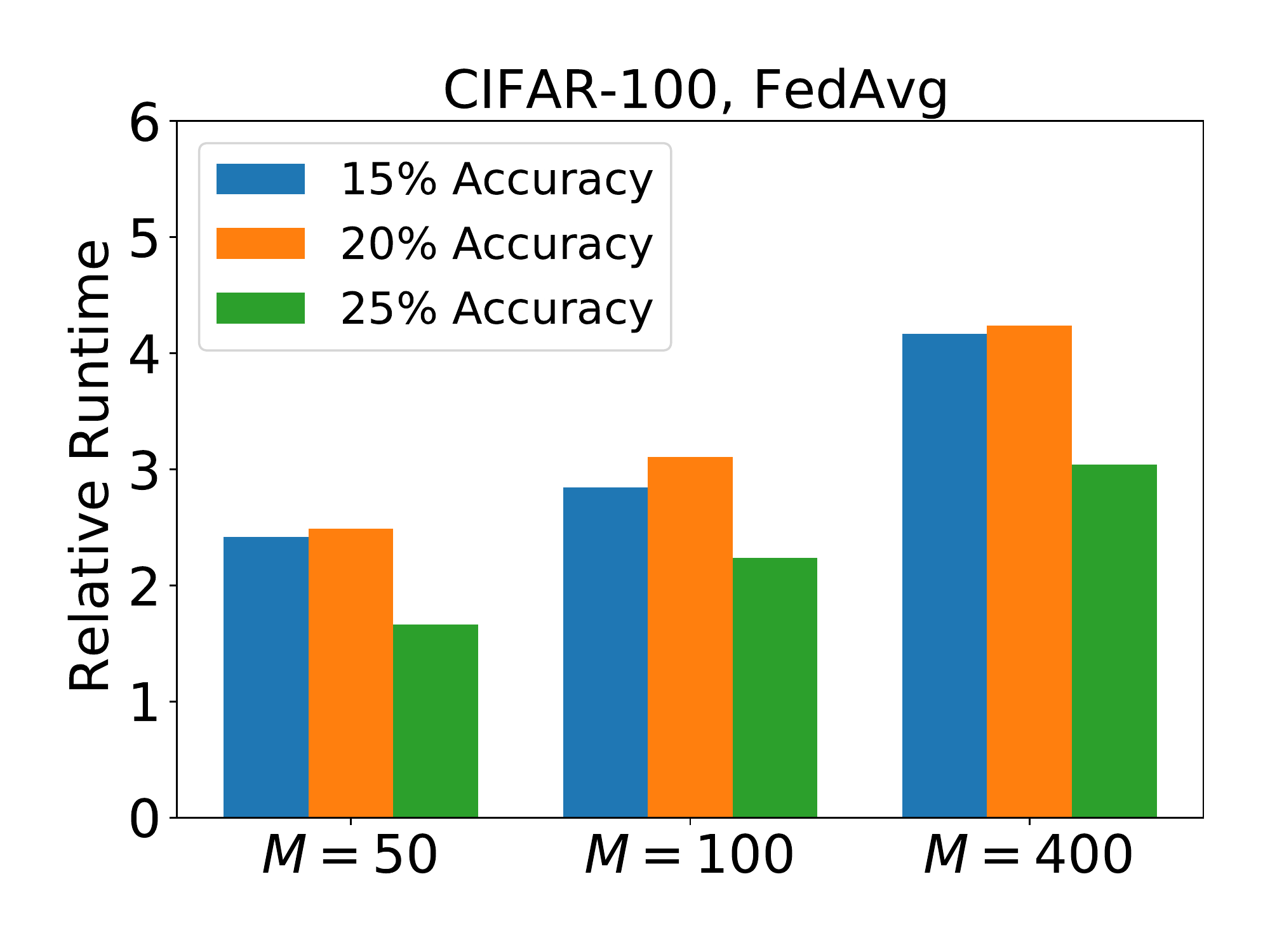}
    \caption{$\lambda = 10$}
\end{subfigure}%
\begin{subfigure}{0.24\textwidth}
    \centering
    \includegraphics[width=1\linewidth]{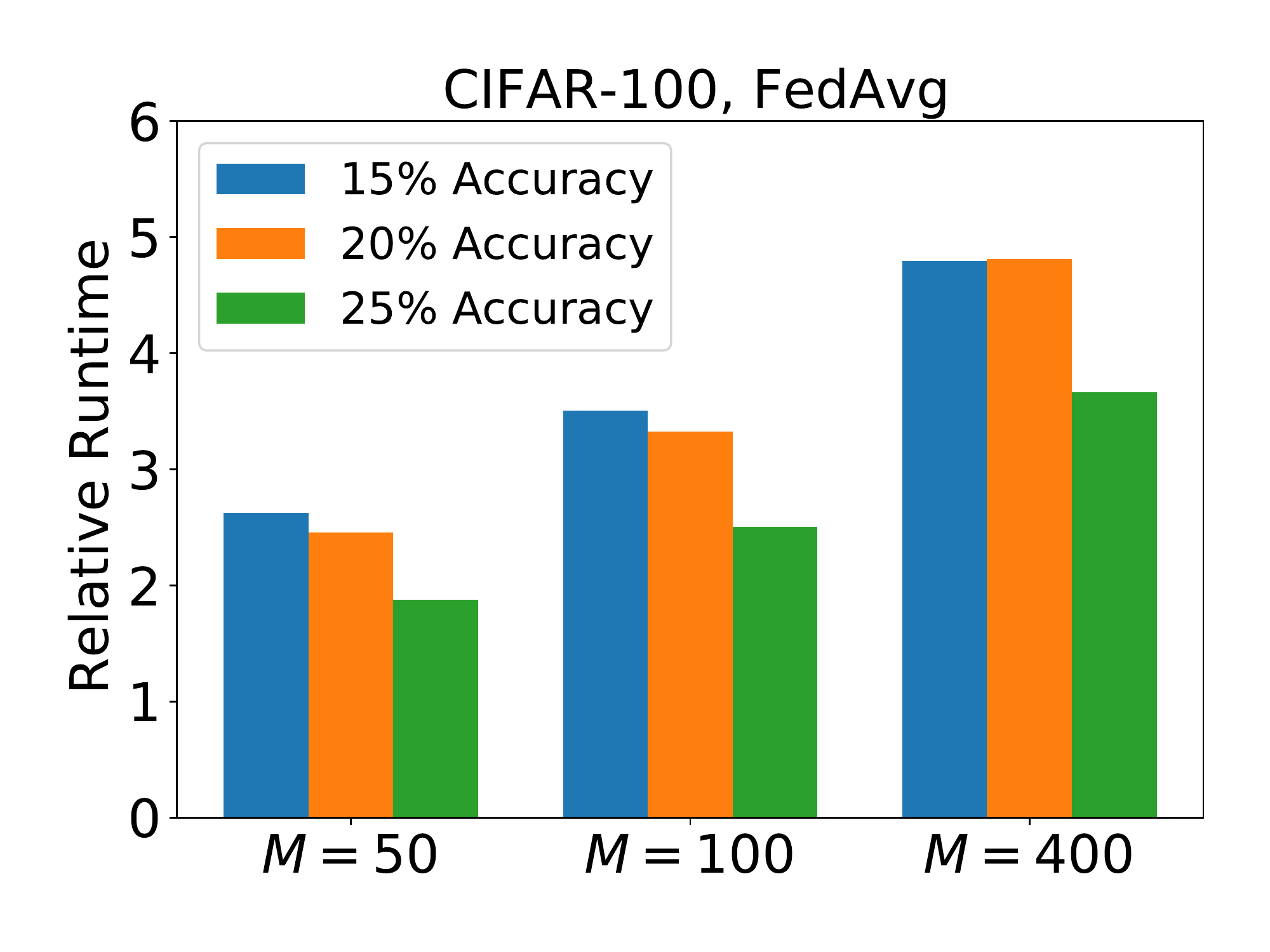}
    \caption{$\lambda = 100$}
\end{subfigure}%
\caption{The relative amount of time required to reach given test accuracies on CIFAR-100 with varying cohort sizes. We present the ratio of the runtime needed for $M > 10$ with respect to the time needed for $M = 10$. Runtimes are simulated under a shifted exponential model with $\alpha = 1$ and varying $\lambda$.}
\label{fig:cifar100_runtime}
\end{figure}

\begin{figure}[!ht]
\centering
\begin{subfigure}{0.24\textwidth}
    \centering
    \includegraphics[width=1\linewidth]{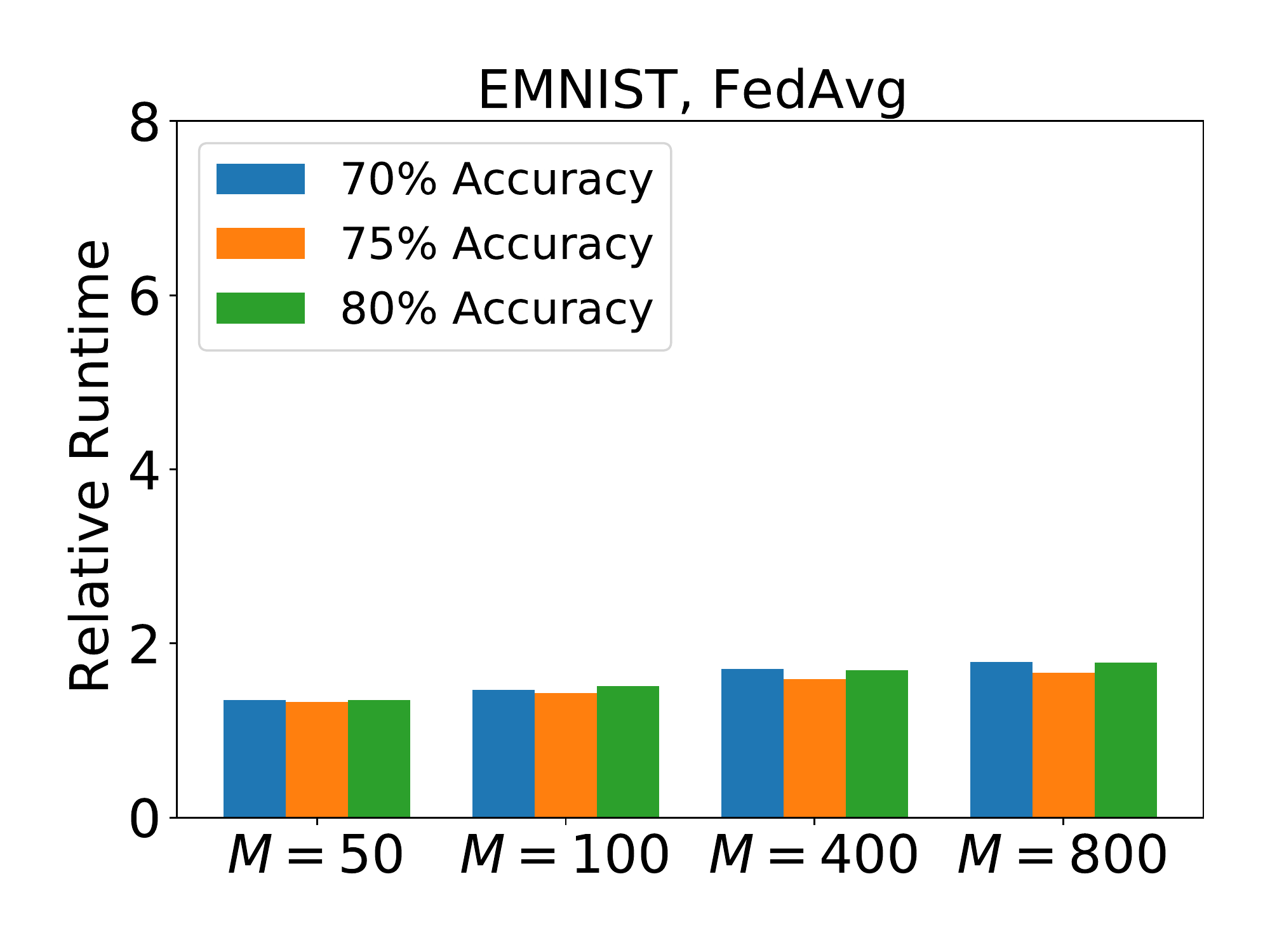}
    \caption{$\lambda = 0.1$}
\end{subfigure}%
\begin{subfigure}{0.24\textwidth}
    \centering
    \includegraphics[width=1\linewidth]{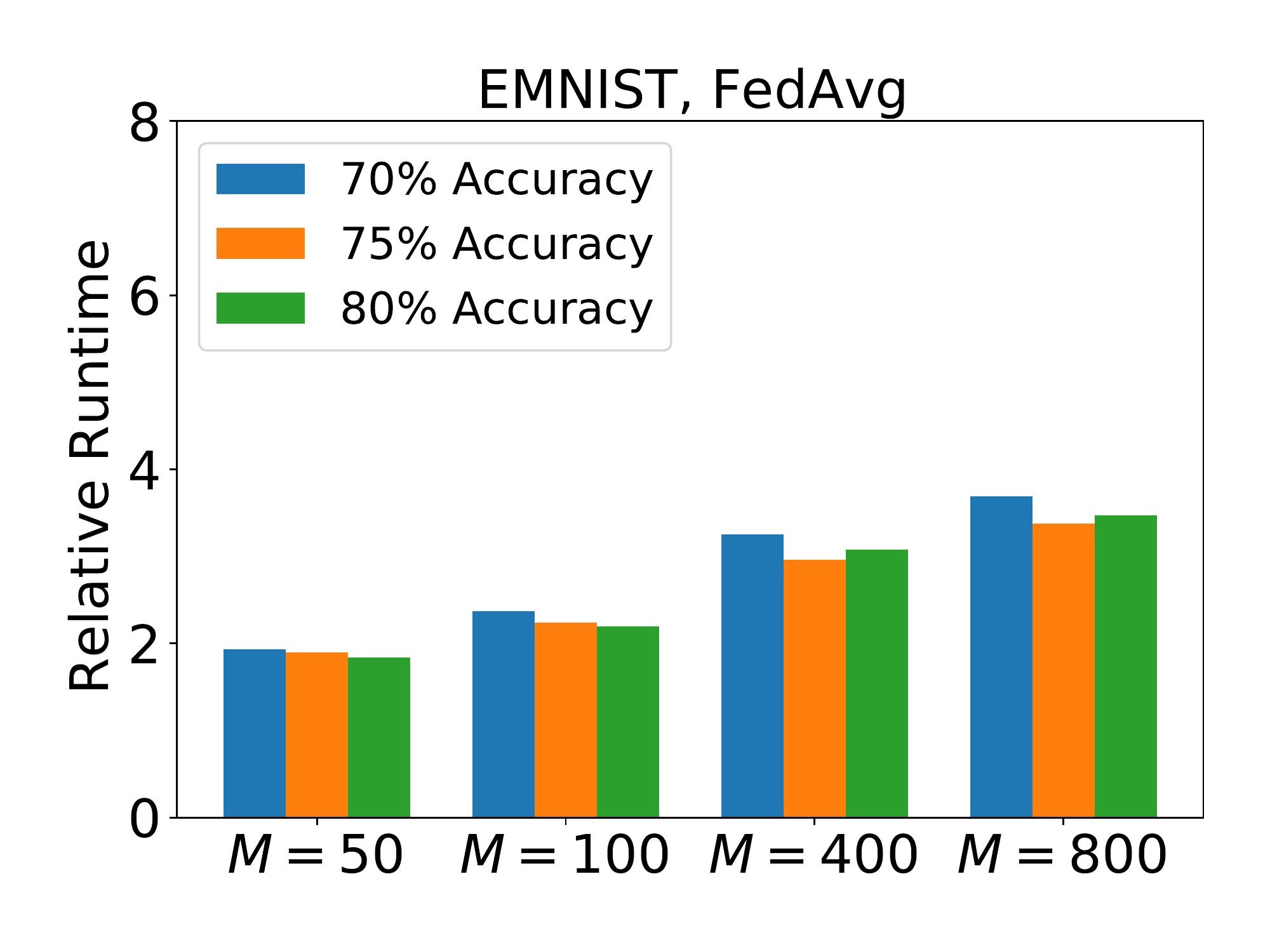}
    \caption{$\lambda = 1$}
\end{subfigure}%
\begin{subfigure}{0.24\textwidth}
    \centering
    \includegraphics[width=1\linewidth]{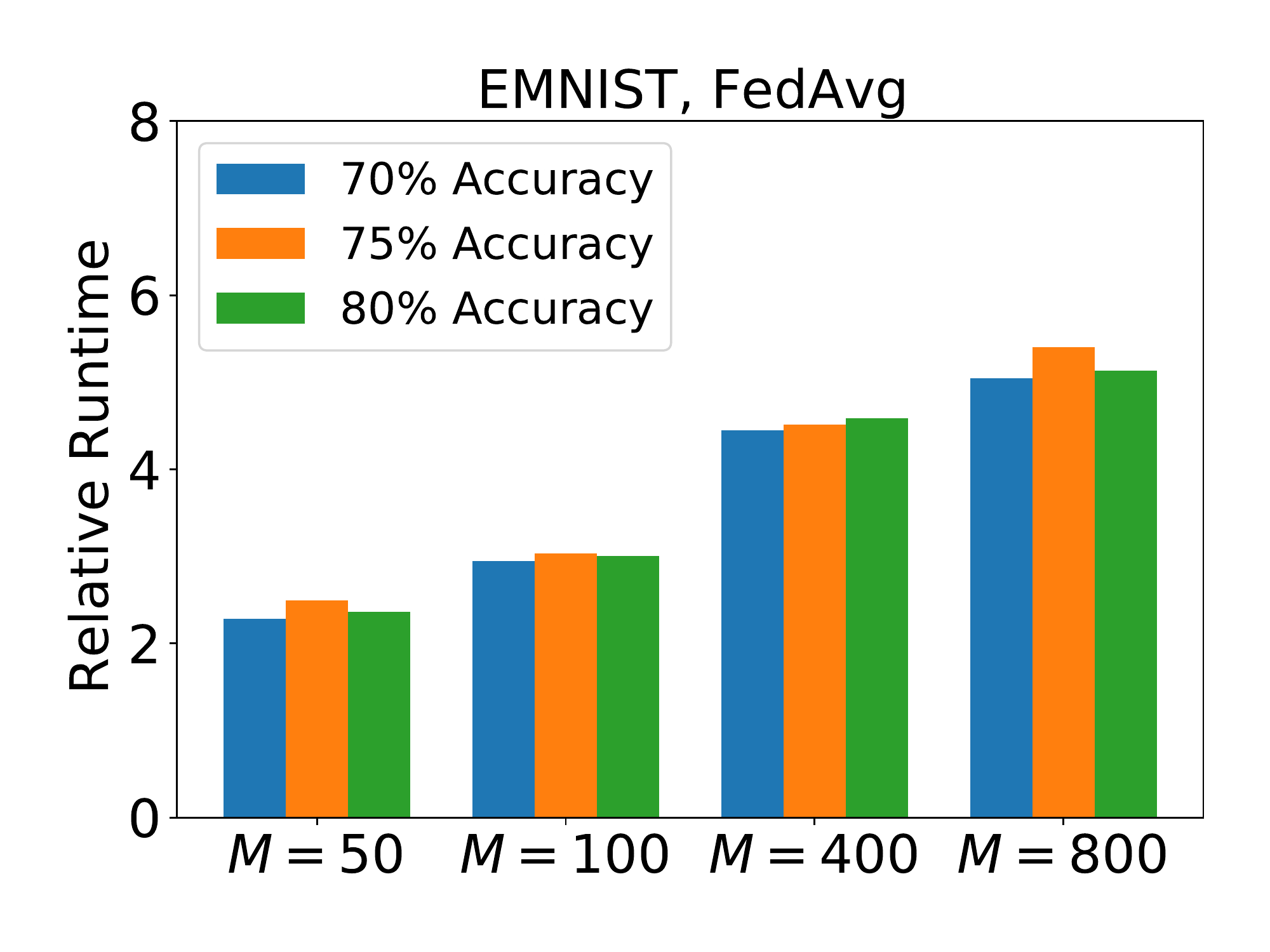}
    \caption{$\lambda = 10$}
\end{subfigure}%
\begin{subfigure}{0.24\textwidth}
    \centering
    \includegraphics[width=1\linewidth]{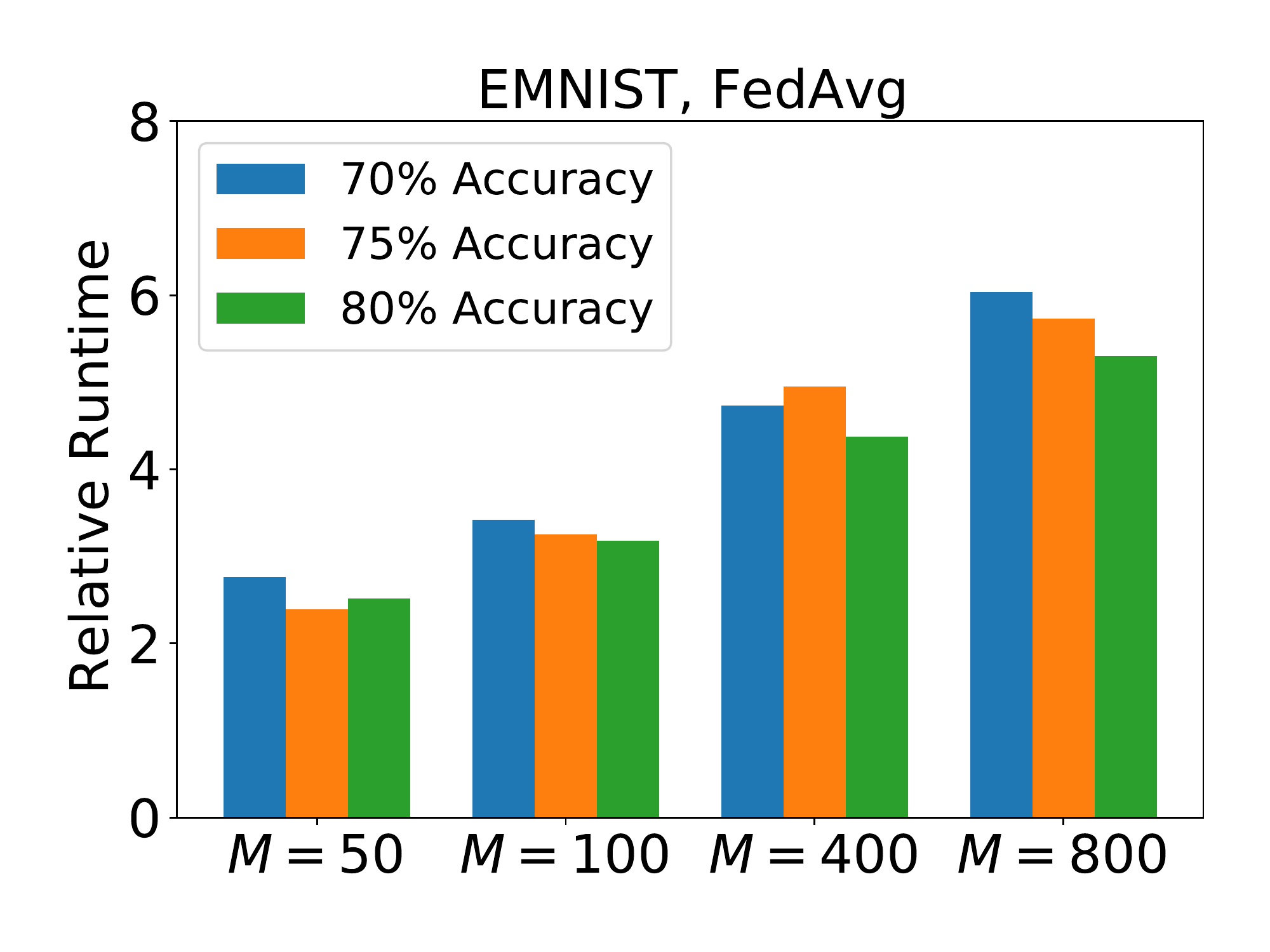}
    \caption{$\lambda = 100$}
\end{subfigure}%
\caption{The relative amount of time required to reach given test accuracies on EMNIST with varying cohort sizes. We present the ratio of the runtime needed for $M > 10$ with respect to the time needed for $M = 10$. Runtimes are simulated under a shifted exponential model with $\alpha = 1$ and varying $\lambda$.}
\label{fig:emnist_runtime}
\end{figure}

\begin{figure}[!ht]
\centering
\begin{subfigure}{0.24\textwidth}
    \centering
    \includegraphics[width=1\linewidth]{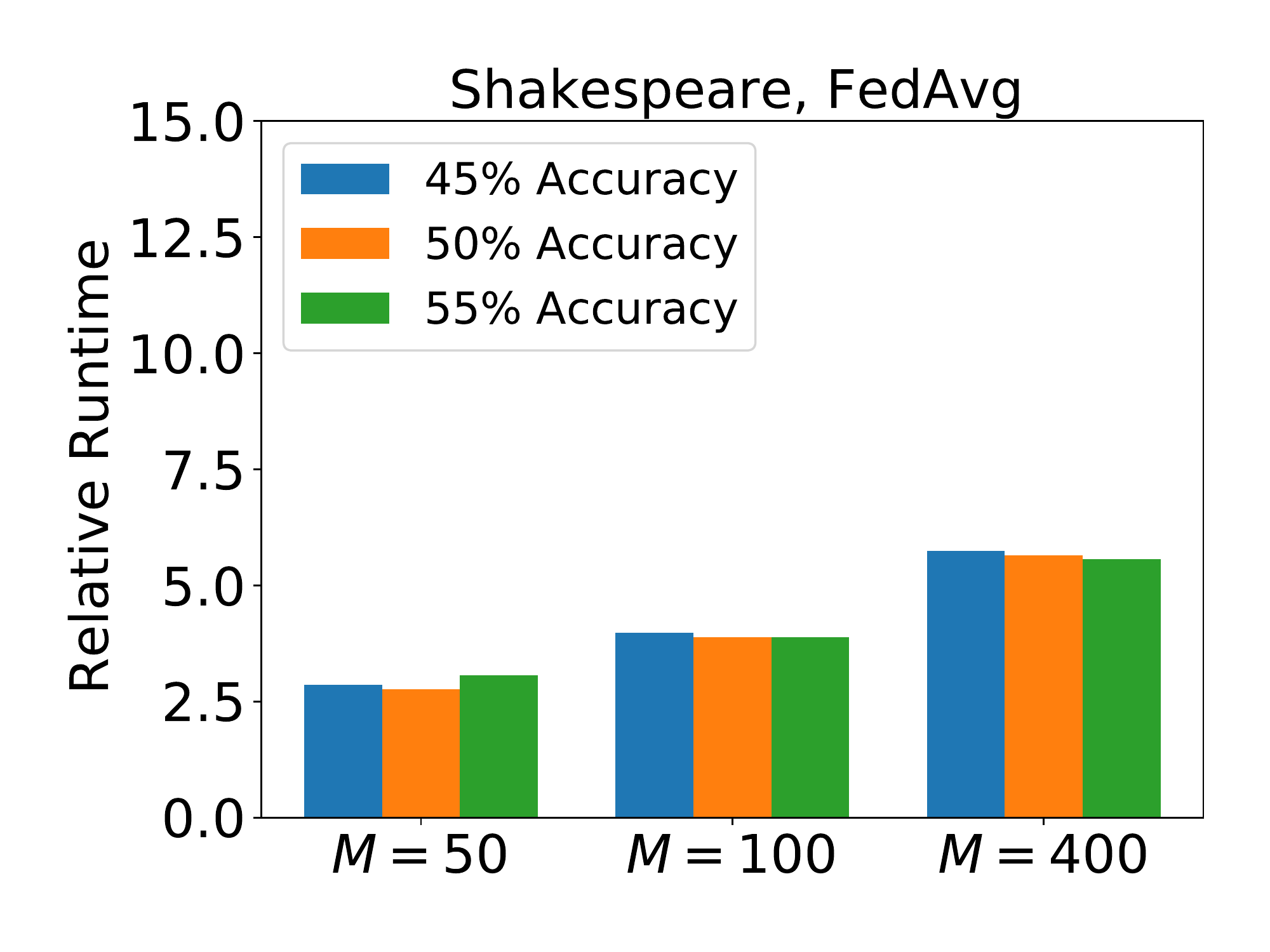}
    \caption{$\lambda = 0.1$}
\end{subfigure}%
\begin{subfigure}{0.24\textwidth}
    \centering
    \includegraphics[width=1\linewidth]{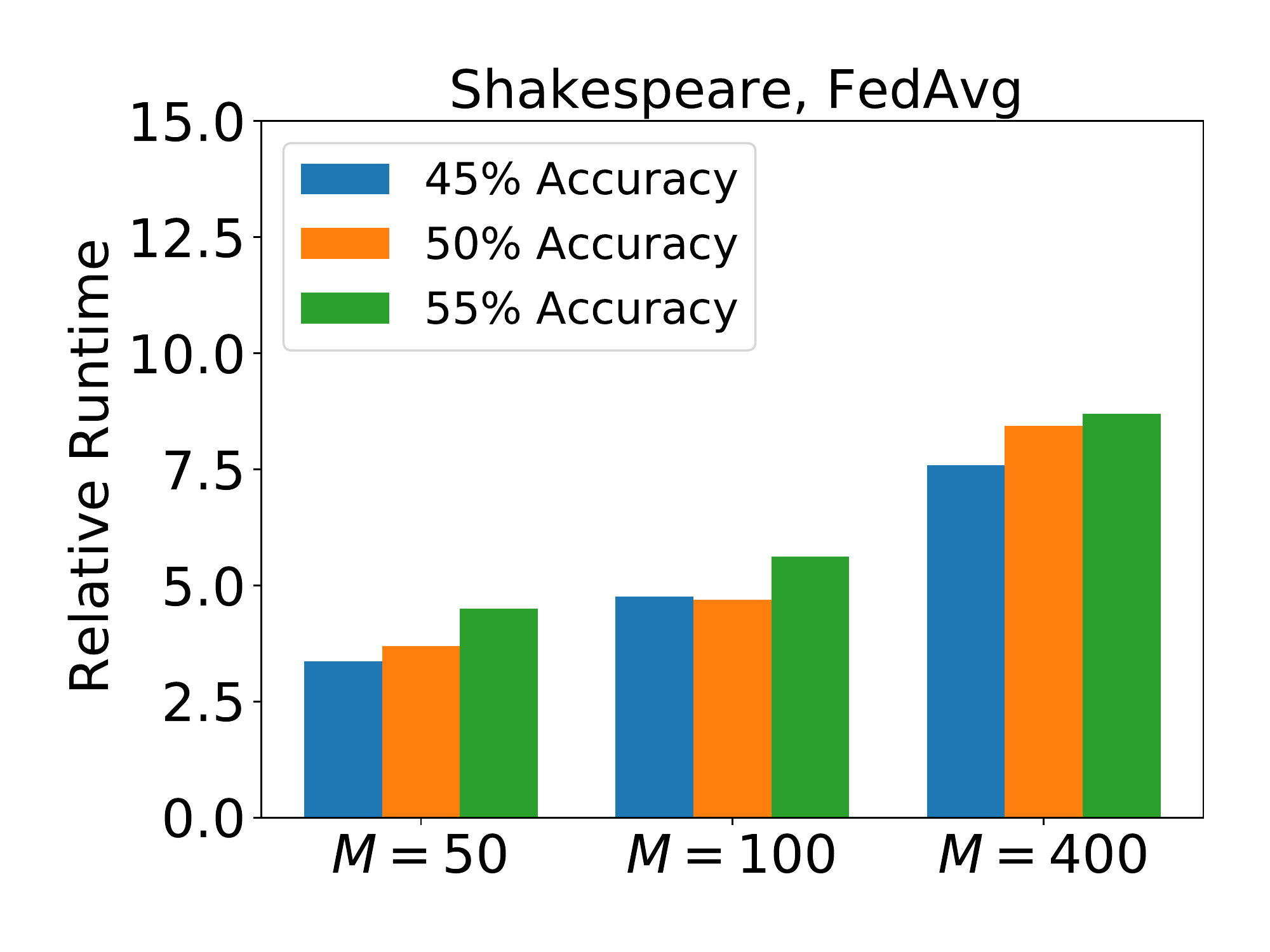}
    \caption{$\lambda = 1$}
\end{subfigure}%
\begin{subfigure}{0.24\textwidth}
    \centering
    \includegraphics[width=1\linewidth]{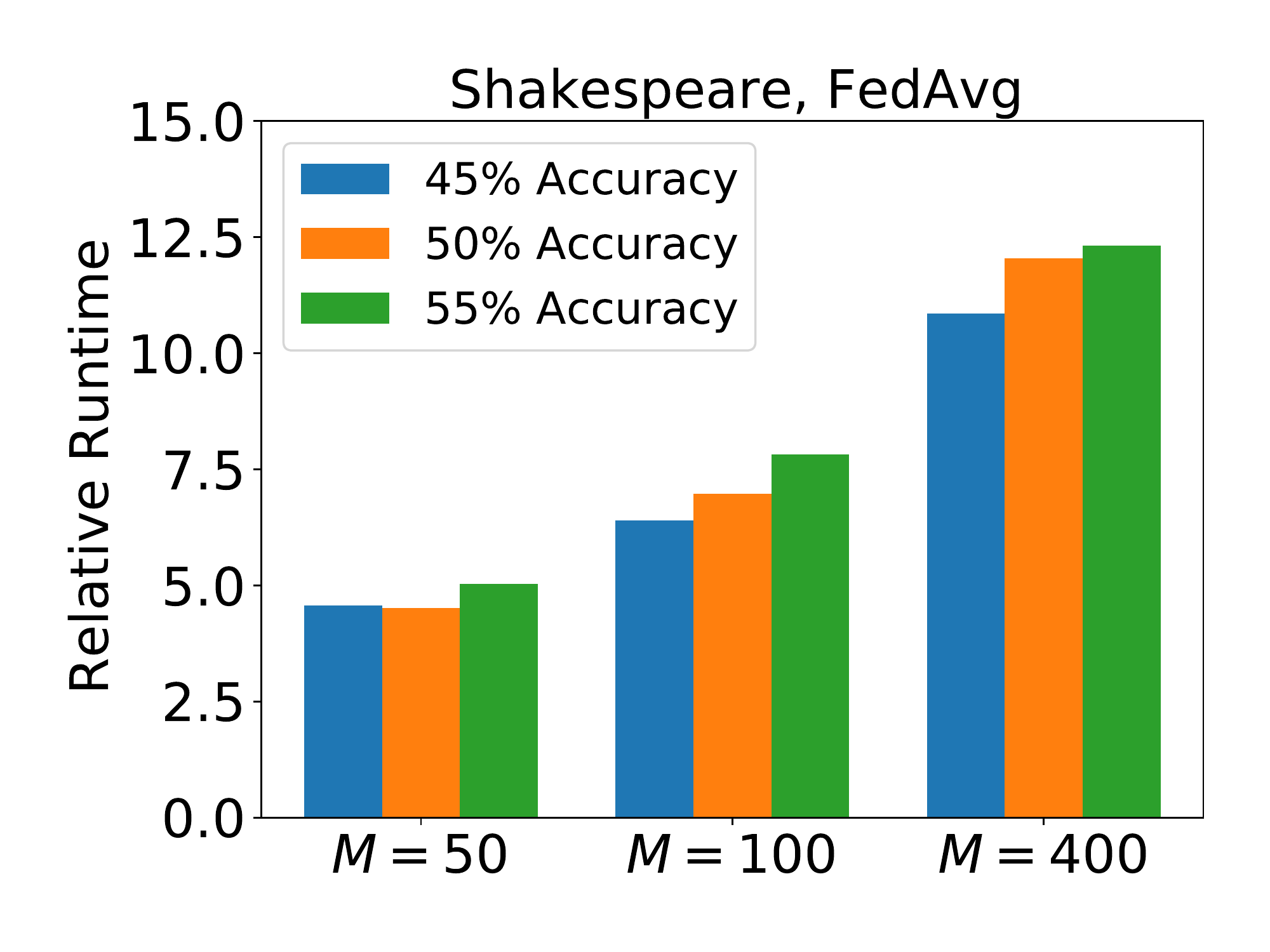}
    \caption{$\lambda = 10$}
\end{subfigure}%
\begin{subfigure}{0.24\textwidth}
    \centering
    \includegraphics[width=1\linewidth]{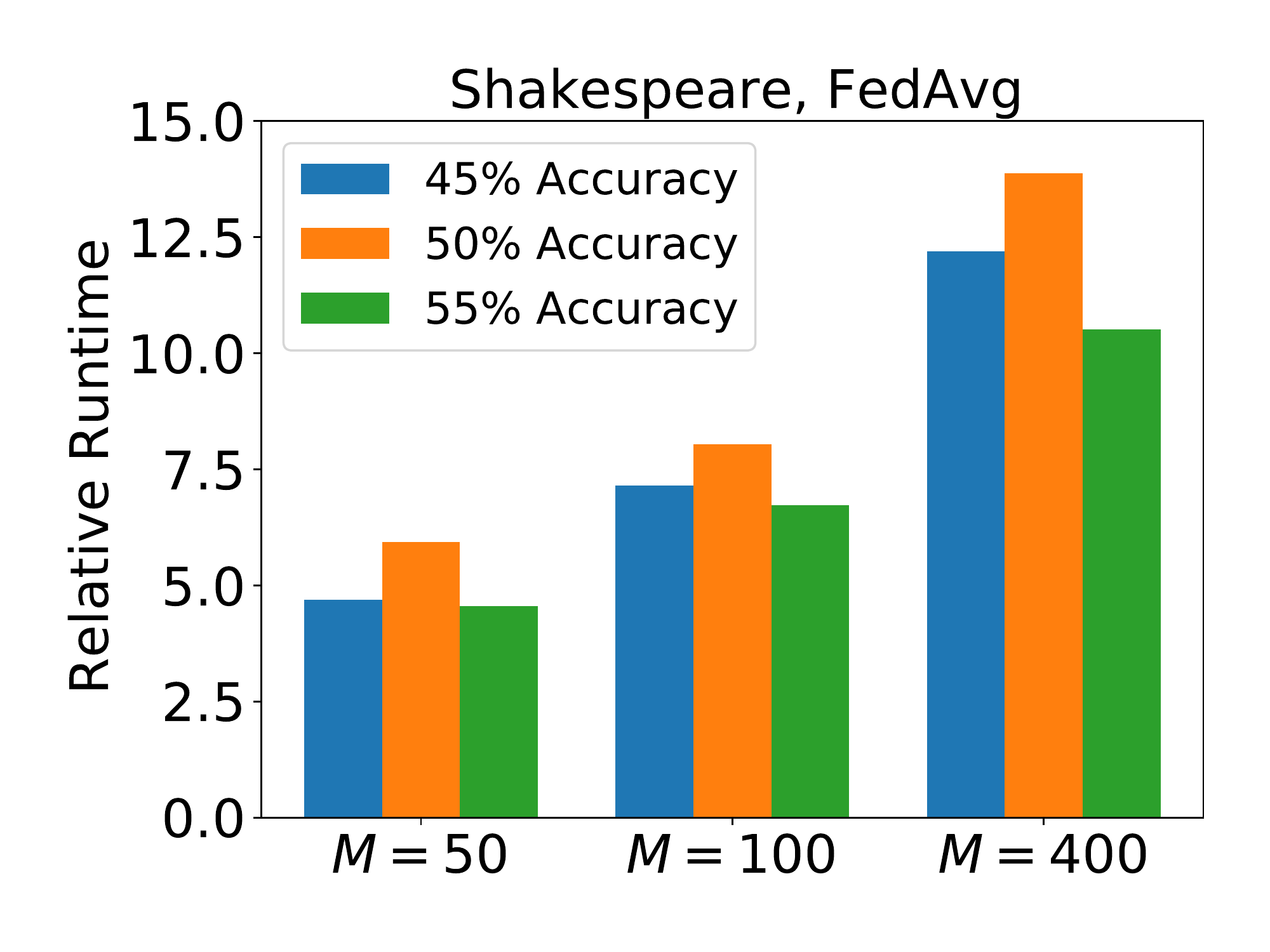}
    \caption{$\lambda = 100$}
\end{subfigure}%
\caption{The relative amount of time required to reach given test accuracies on Shakespeare with varying cohort sizes. We present the ratio of the runtime needed for $M > 10$ with respect to the time needed for $M = 10$. Runtimes are simulated under a shifted exponential model with $\alpha = 1$ and varying $\lambda$.}
\label{fig:shakespeare_runtime}
\end{figure}

\begin{figure}[!ht]
\centering
\begin{subfigure}{0.24\textwidth}
    \centering
    \includegraphics[width=1\linewidth]{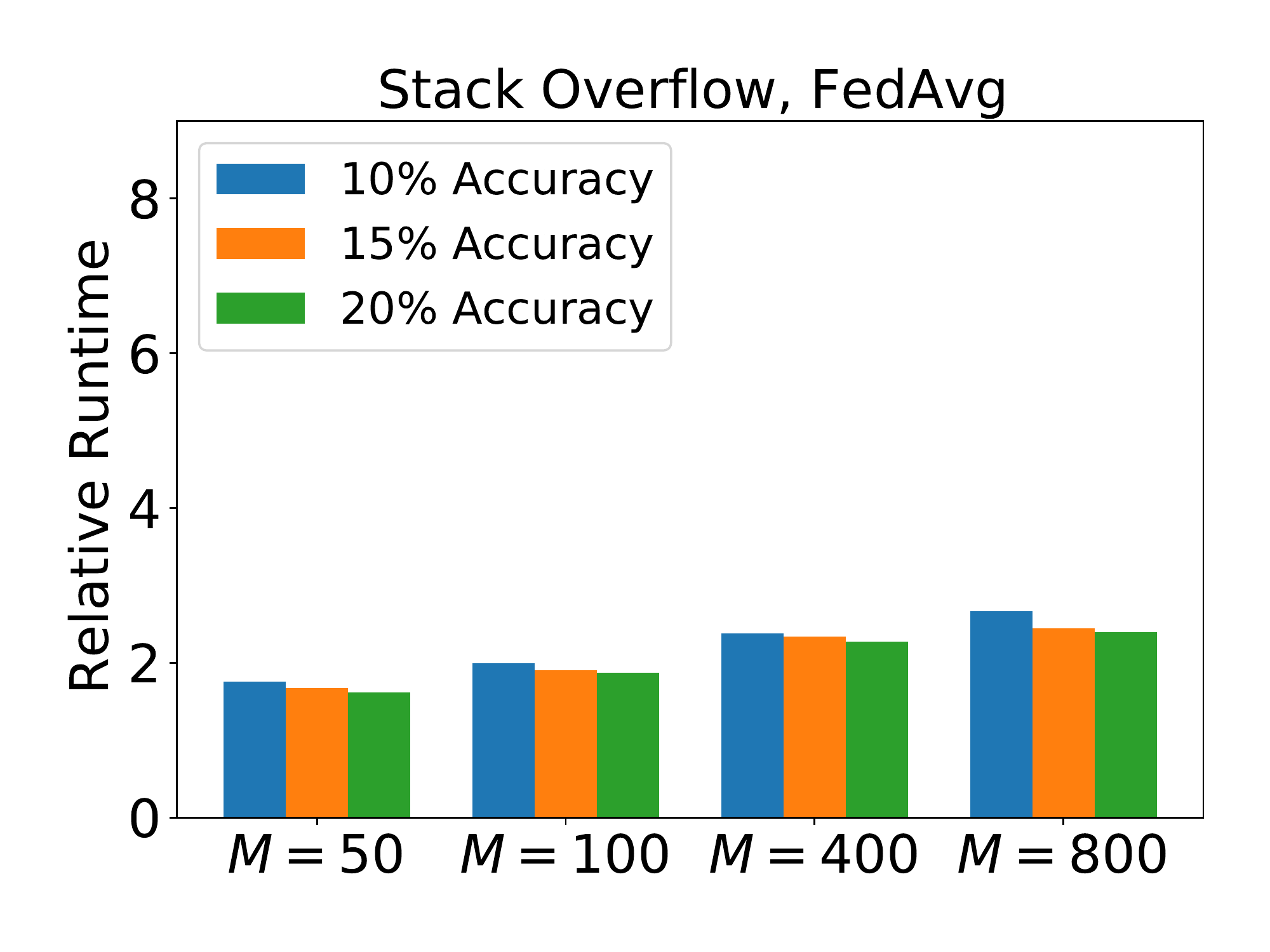}
    \caption{$\lambda = 0.1$}
\end{subfigure}%
\begin{subfigure}{0.24\textwidth}
    \centering
    \includegraphics[width=1\linewidth]{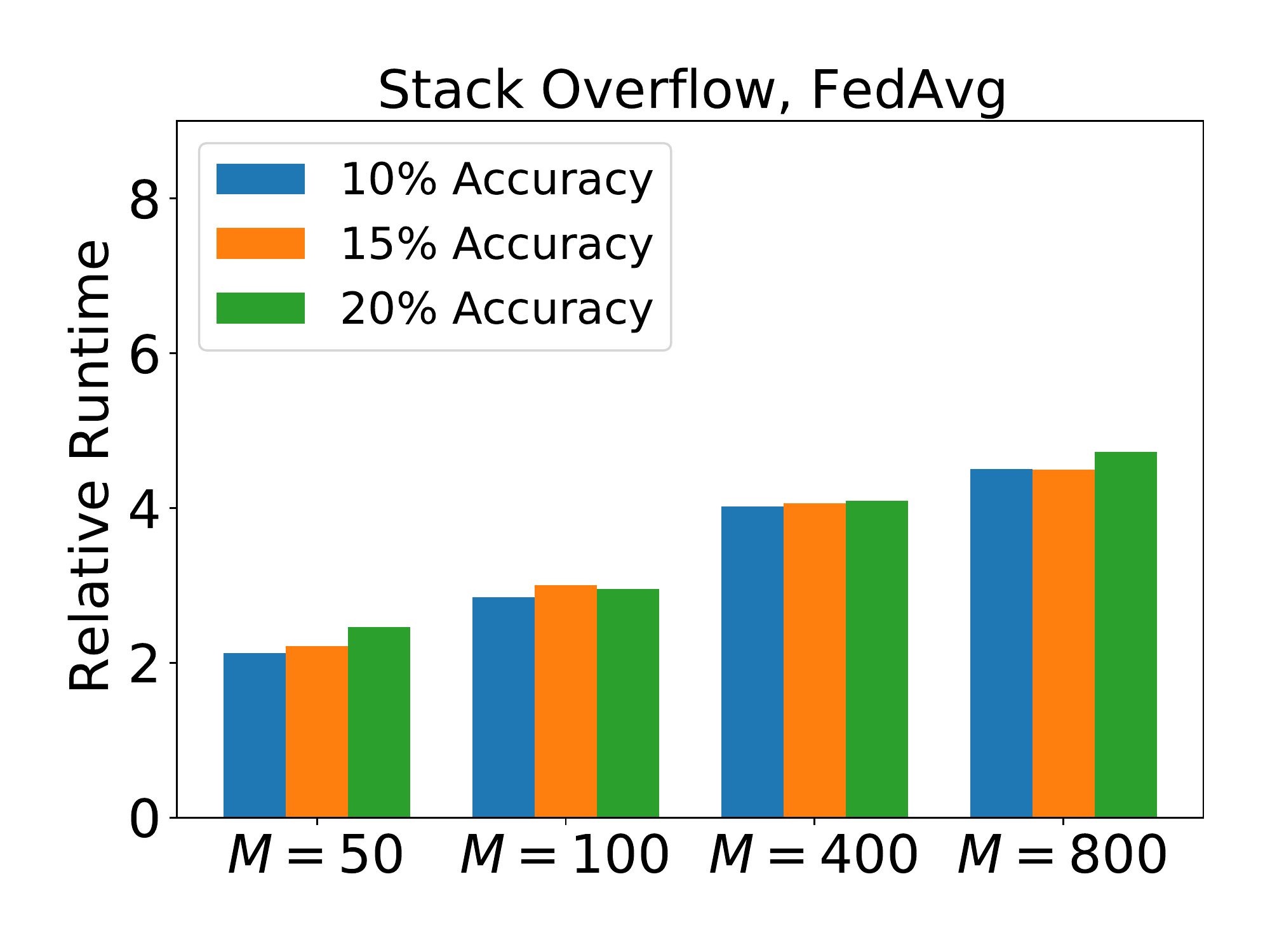}
    \caption{$\lambda = 1$}
\end{subfigure}%
\begin{subfigure}{0.24\textwidth}
    \centering
    \includegraphics[width=1\linewidth]{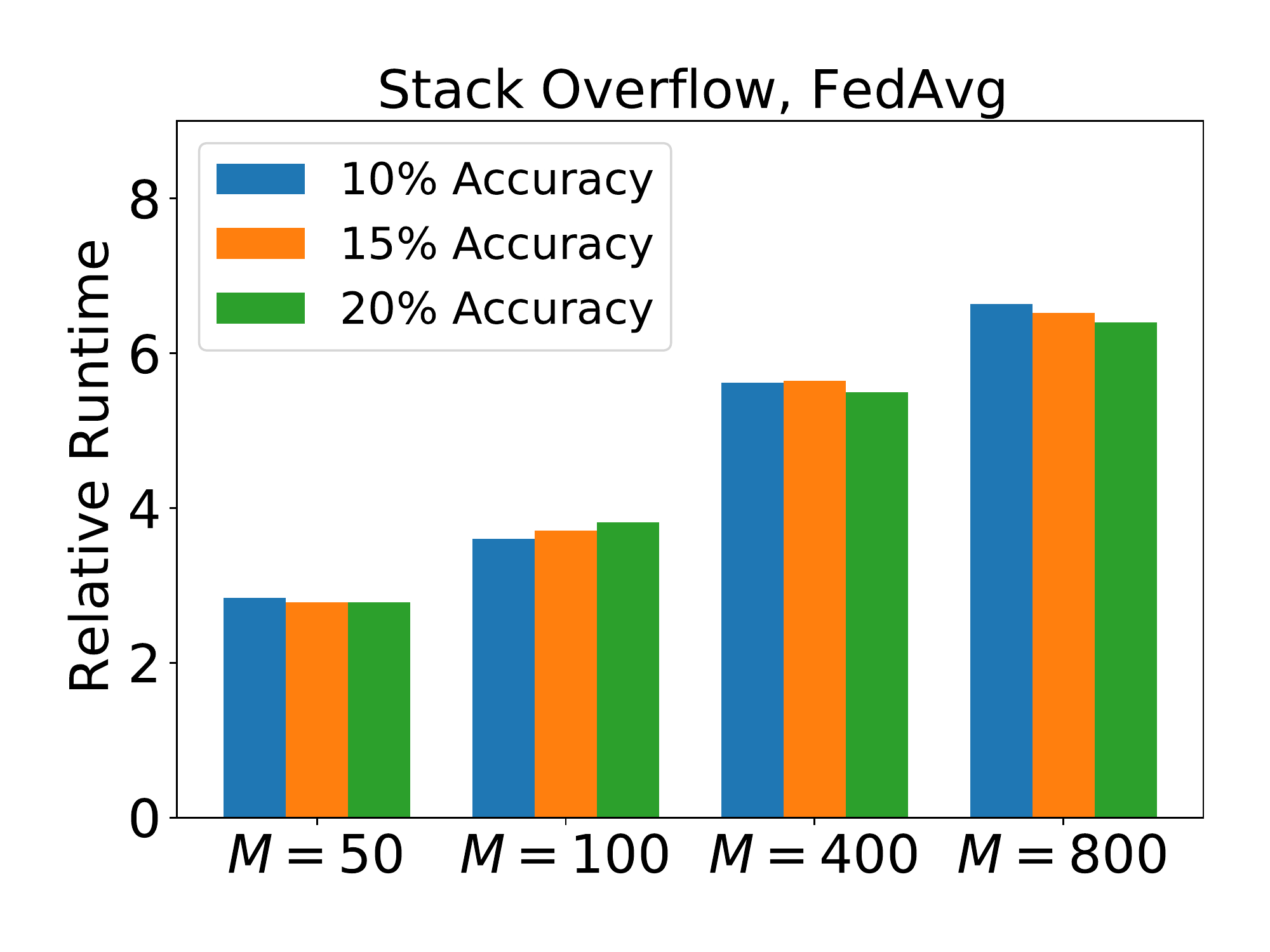}
    \caption{$\lambda = 10$}
\end{subfigure}%
\begin{subfigure}{0.24\textwidth}
    \centering
    \includegraphics[width=1\linewidth]{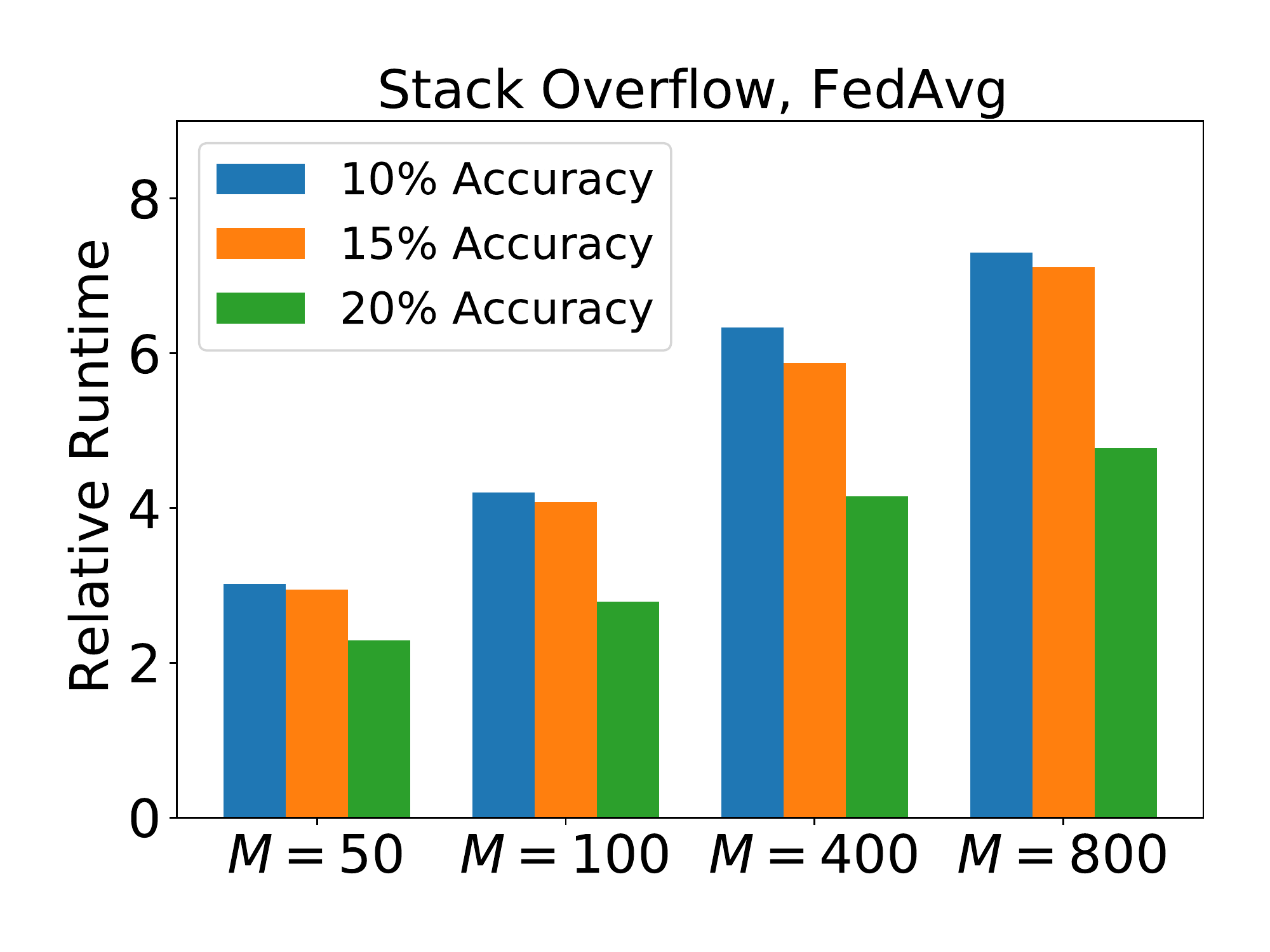}
    \caption{$\lambda = 100$}
\end{subfigure}%
\caption{The relative amount of time required to reach given test accuracies on Stack Overflow with varying cohort sizes. We present the ratio of the runtime needed for $M > 10$ with respect to the time needed for $M = 10$. Runtimes are simulated under a shifted exponential model with $\alpha = 1$ and varying $\lambda$.}
\label{fig:stackoverflow_runtime}
\end{figure}

Thus, the round runtime is the maximum of $M$ shifted exponential random variables, where $M$ is the cohort size. Note that this only models the client computation time, not the server computation time or communication time. Using this model, we plot the simulated runtime of \fedavg on various tasks, for varying cohort sizes. For simplicity, we assume $\alpha = 1$ in all experiments, and vary $\lambda$ over $\{0.1, 1, 10, 100\}$. To showcase how much longer the runtime of large-cohort training may be, we present the simulated runtime, \emph{relative} to $M = 10$. For $a \in [0, 1]$, we plot the ratio of how long it takes to reach a test accuracy of $a$ with a cohort size of $M$, versus how long it takes to reach $a$ with $M = 10$.
We give the results for CIFAR-100, EMNIST, Shakespeare, and Stack Overflow in Figures \ref{fig:cifar100_runtime}, \ref{fig:emnist_runtime}, \ref{fig:shakespeare_runtime}, \ref{fig:stackoverflow_runtime}, respectively.

When $\lambda$ is small, we see that larger cohorts can obtain higher test accuracy in a comparable amount of time to $M = 10$. However, when $\lambda$ is large, large-cohort training may require anywhere from 5-10 times more client compute time. This is particularly important in cross-device settings with lightweight edge devices, as the straggler effect (which essentially increases with $\lambda$) may be larger. Note that we see particularly large increases in relative runtimes for smaller accuracy thresholds, which suggests that the dynamic cohort strategy from \cref{sec:better_methods} may be useful in helping mitigate such issues.

\FloatBarrier

\subsection{Pseudo-Gradient Norms}\label{appendix:pseudo_gradient_norms}

\begin{figure}[ht!]
\centering
\begin{subfigure}{0.24\textwidth}
     \centering
     \includegraphics[width=1\linewidth]{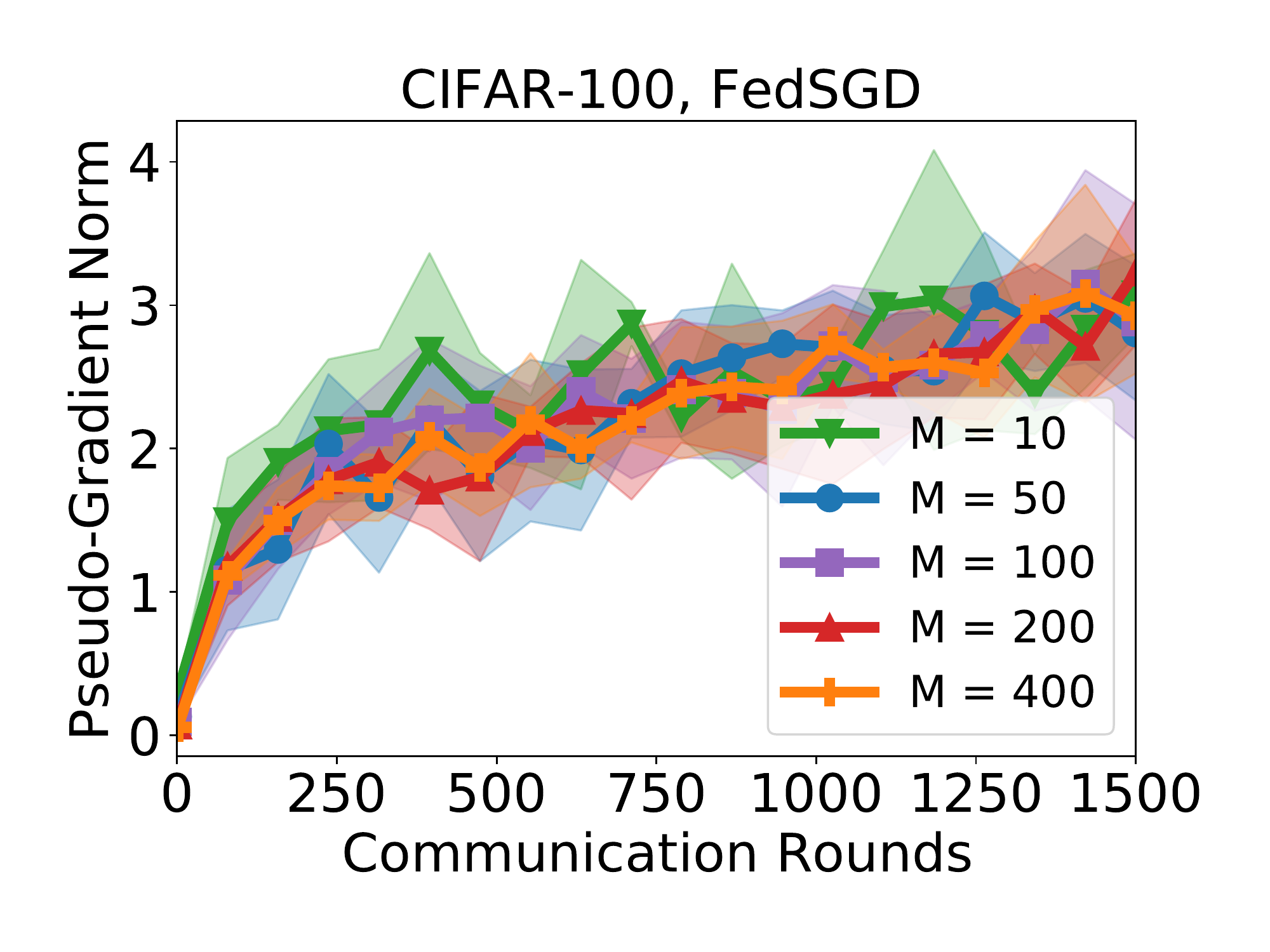}
\end{subfigure}%
\begin{subfigure}{0.24\textwidth}
     \centering
     \includegraphics[width=1\linewidth]{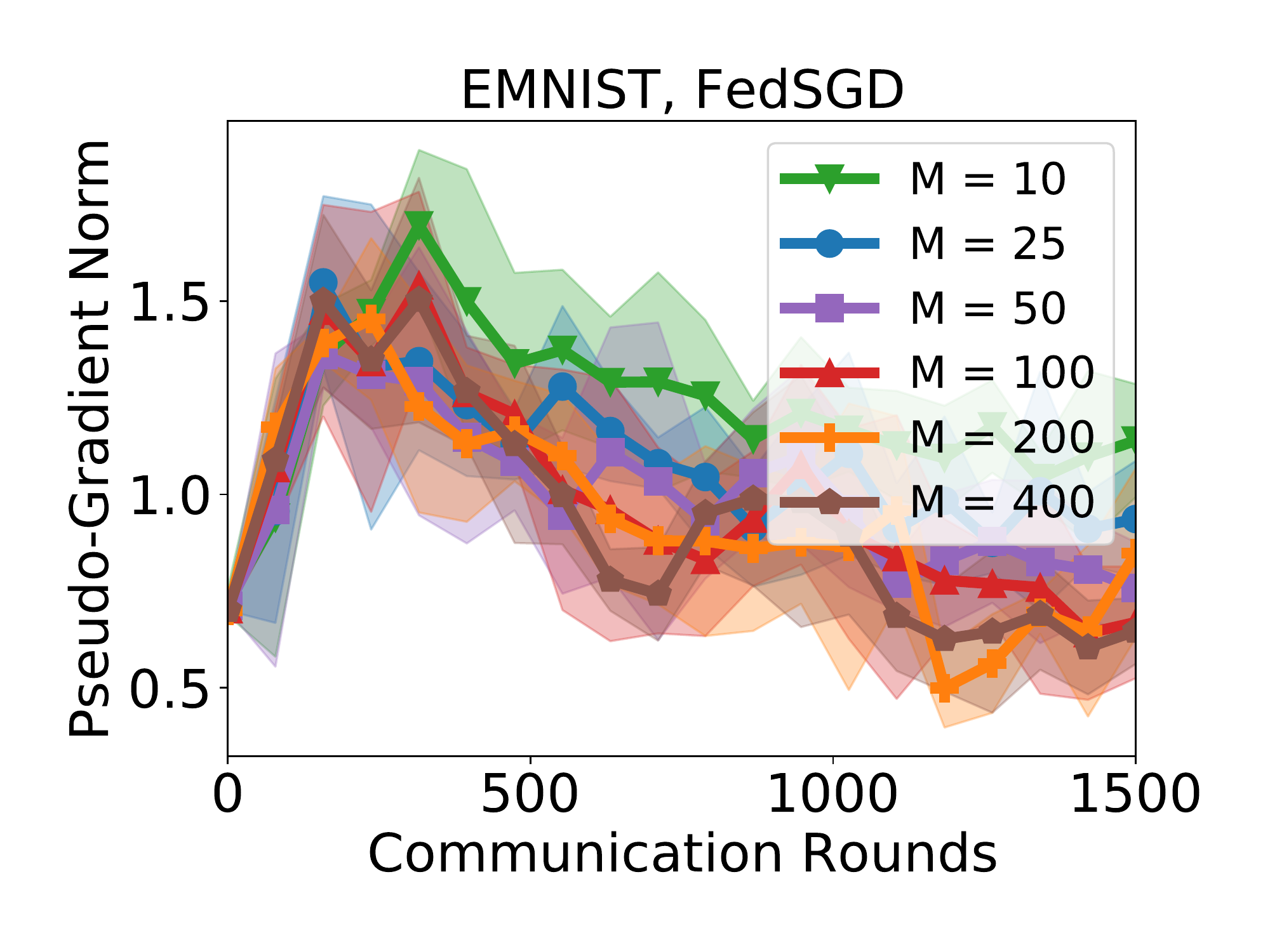}
\end{subfigure}%
\begin{subfigure}{0.24\textwidth}
     \centering
     \includegraphics[width=1\linewidth]{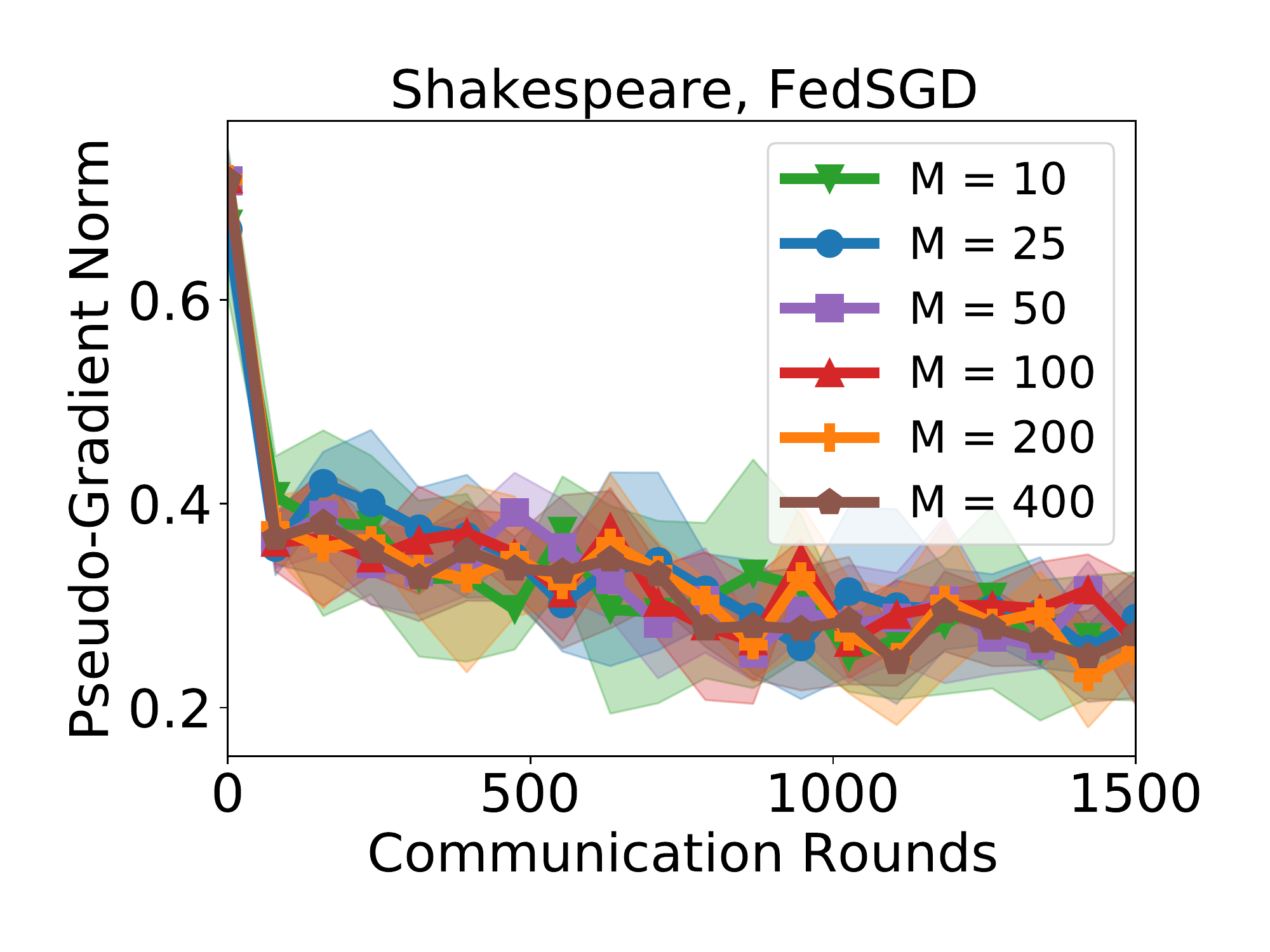}
\end{subfigure}%
\begin{subfigure}{0.24\textwidth}
     \centering
     \includegraphics[width=1\linewidth]{figures/pseudo_gradients/stackoverflow_word_fedsgd_actual_pseudogradient_norm.pdf}
\end{subfigure}%
\caption{Average pseudo-gradient norm of \fedsgd versus the number of communication rounds, for various tasks and cohort sizes $M$.}
\label{fig:fedsgd_pseudogradient_norm}
\end{figure}

\begin{figure}[ht!]
\centering
\begin{subfigure}{0.24\textwidth}
     \centering
     \includegraphics[width=1\linewidth]{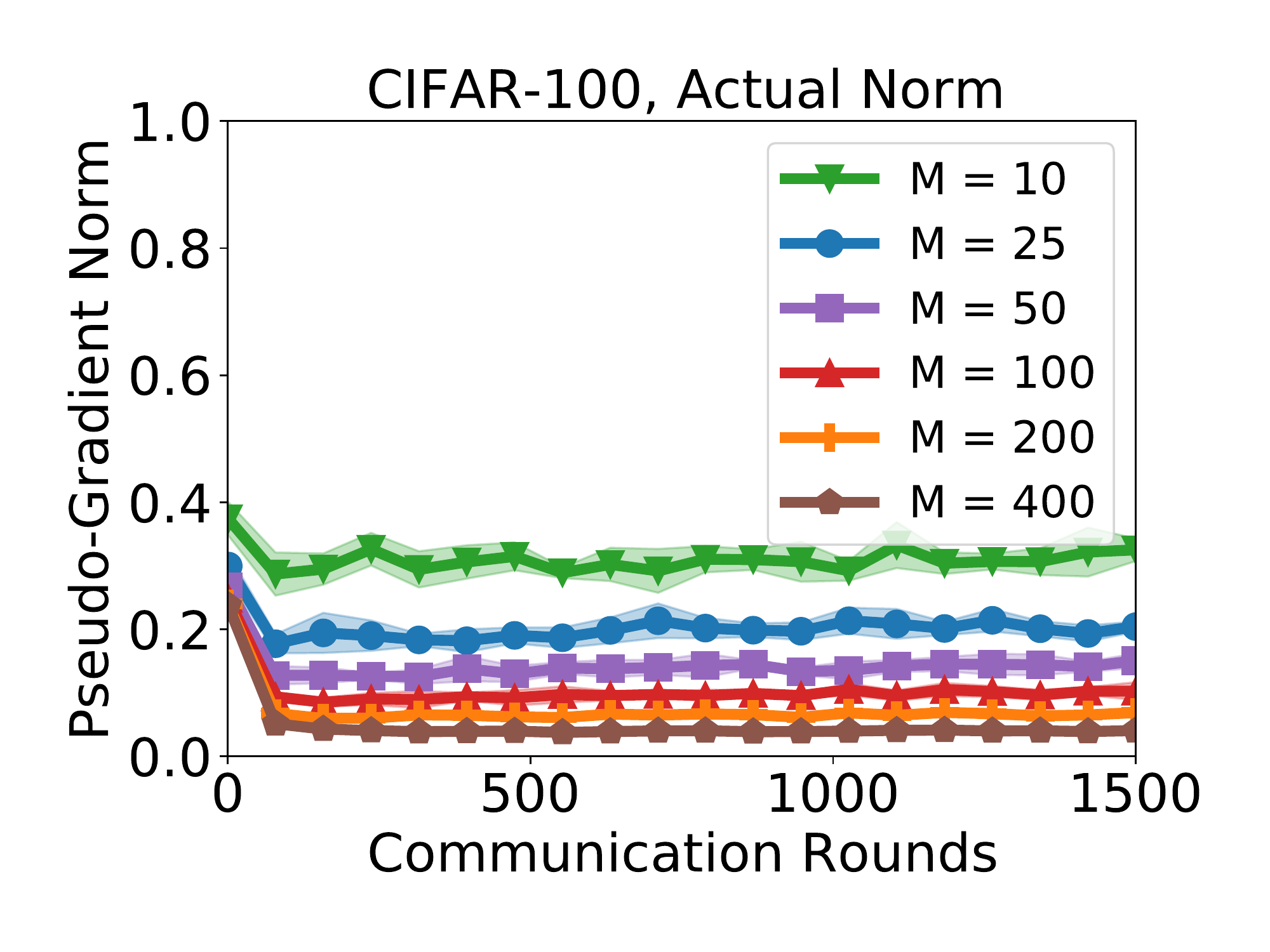}
\end{subfigure}%
\begin{subfigure}{0.24\textwidth}
     \centering
     \includegraphics[width=1\linewidth]{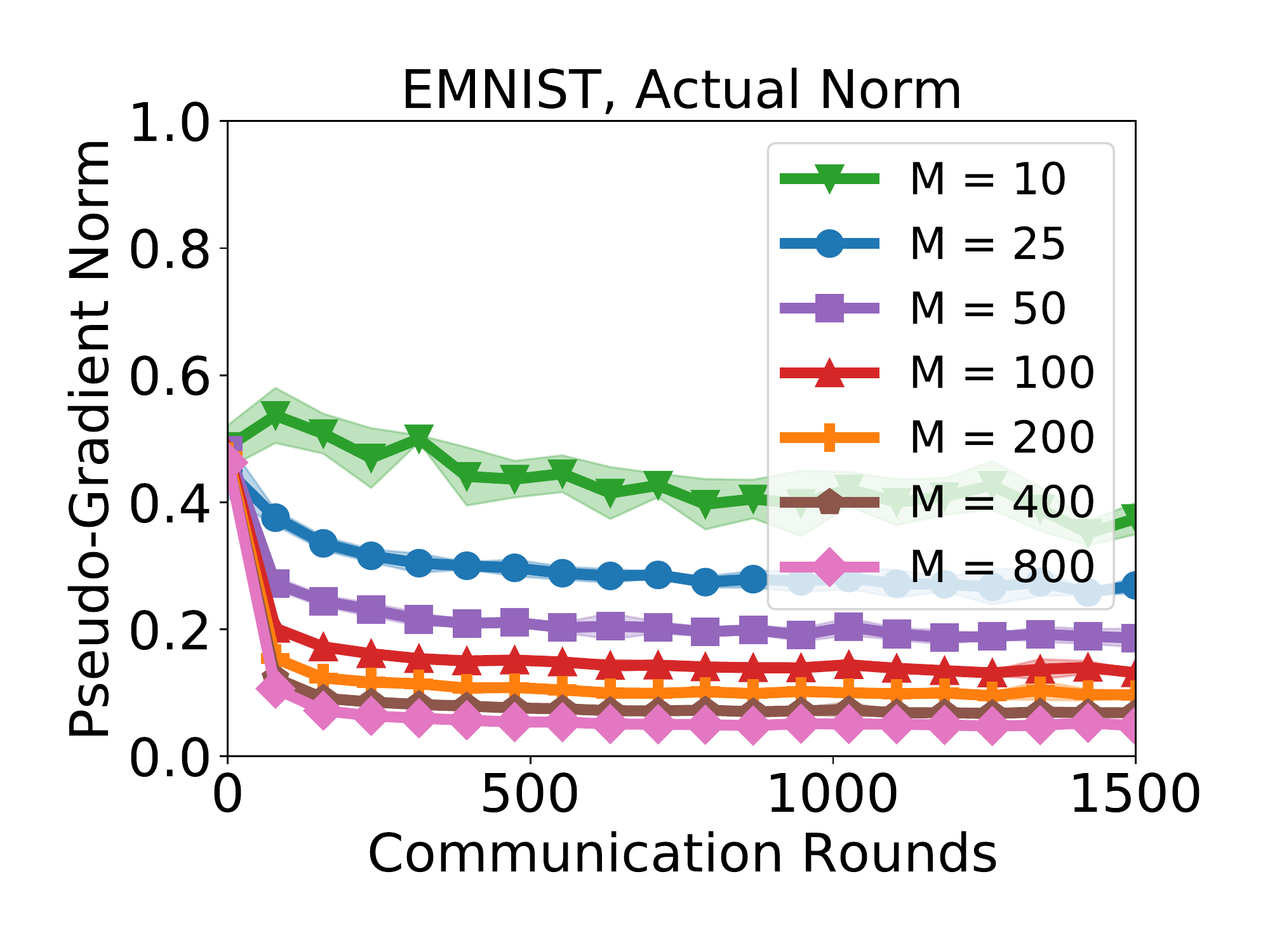}
\end{subfigure}%
\begin{subfigure}{0.24\textwidth}
     \centering
     \includegraphics[width=1\linewidth]{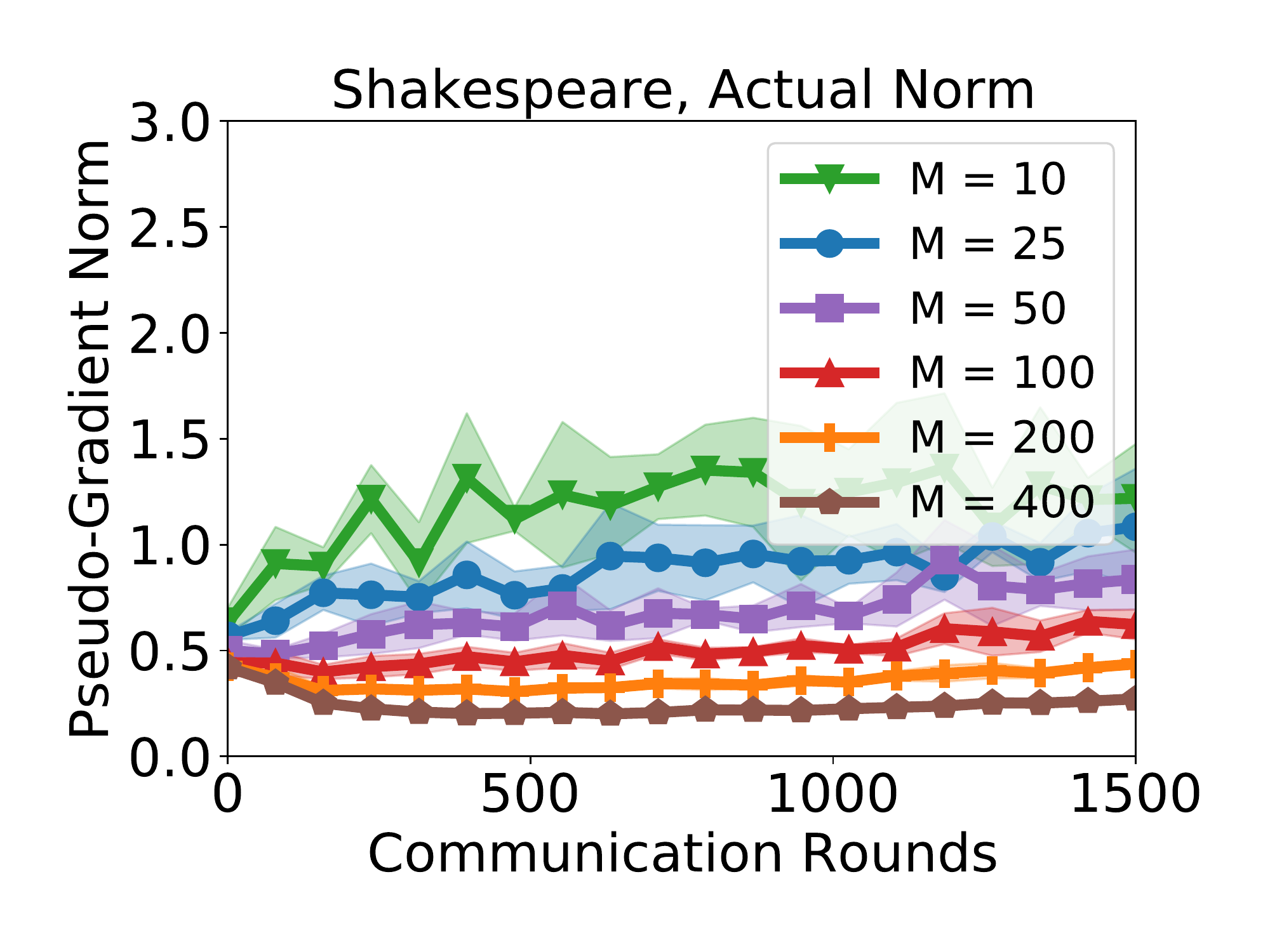}
\end{subfigure}%
\begin{subfigure}{0.24\textwidth}
     \centering
     \includegraphics[width=1\linewidth]{figures/pseudo_gradients/stackoverflow_word_fedavg_actual_pseudogradient_norm.pdf}
\end{subfigure}%
\caption{Average pseudo-gradient norm of \fedavg versus the number of communication rounds, for various tasks and cohort sizes $M$.}
\label{fig:fedavg_pseudogradient_norm}
\end{figure}

\begin{figure}[ht!]
\centering
\begin{subfigure}{0.24\textwidth}
     \centering
     \includegraphics[width=1\linewidth]{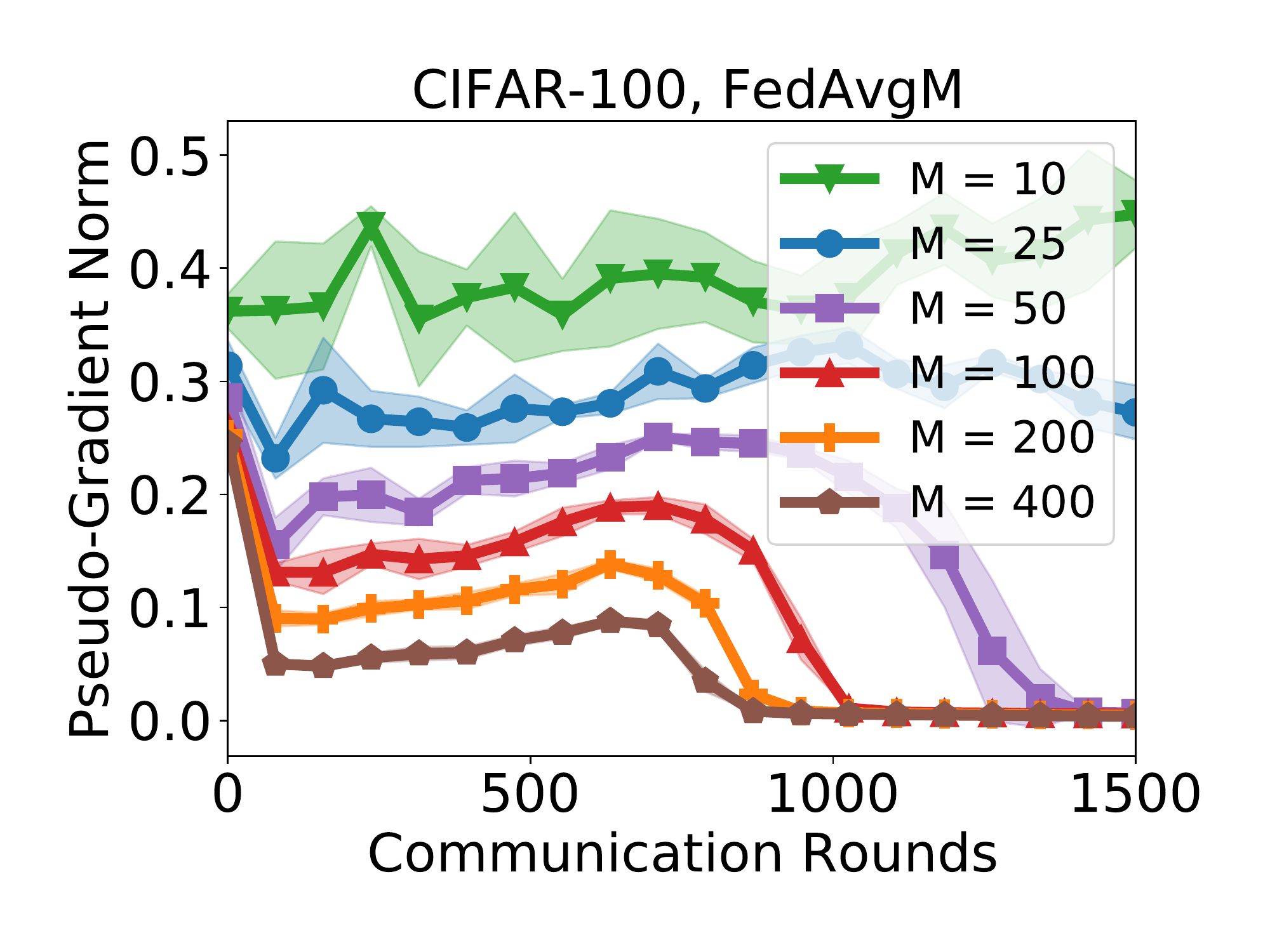}
\end{subfigure}%
\begin{subfigure}{0.24\textwidth}
     \centering
     \includegraphics[width=1\linewidth]{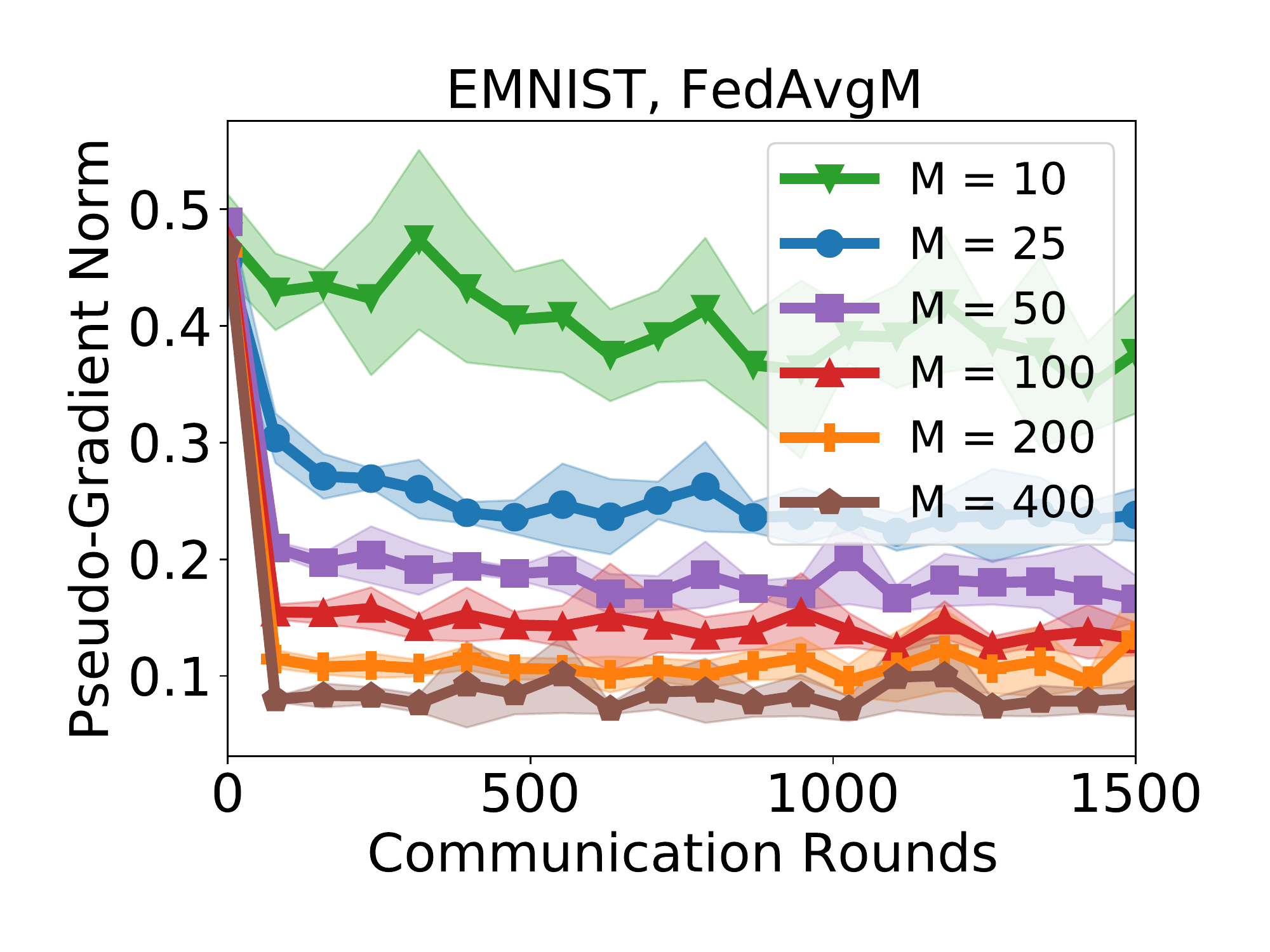}
\end{subfigure}%
\begin{subfigure}{0.24\textwidth}
     \centering
     \includegraphics[width=1\linewidth]{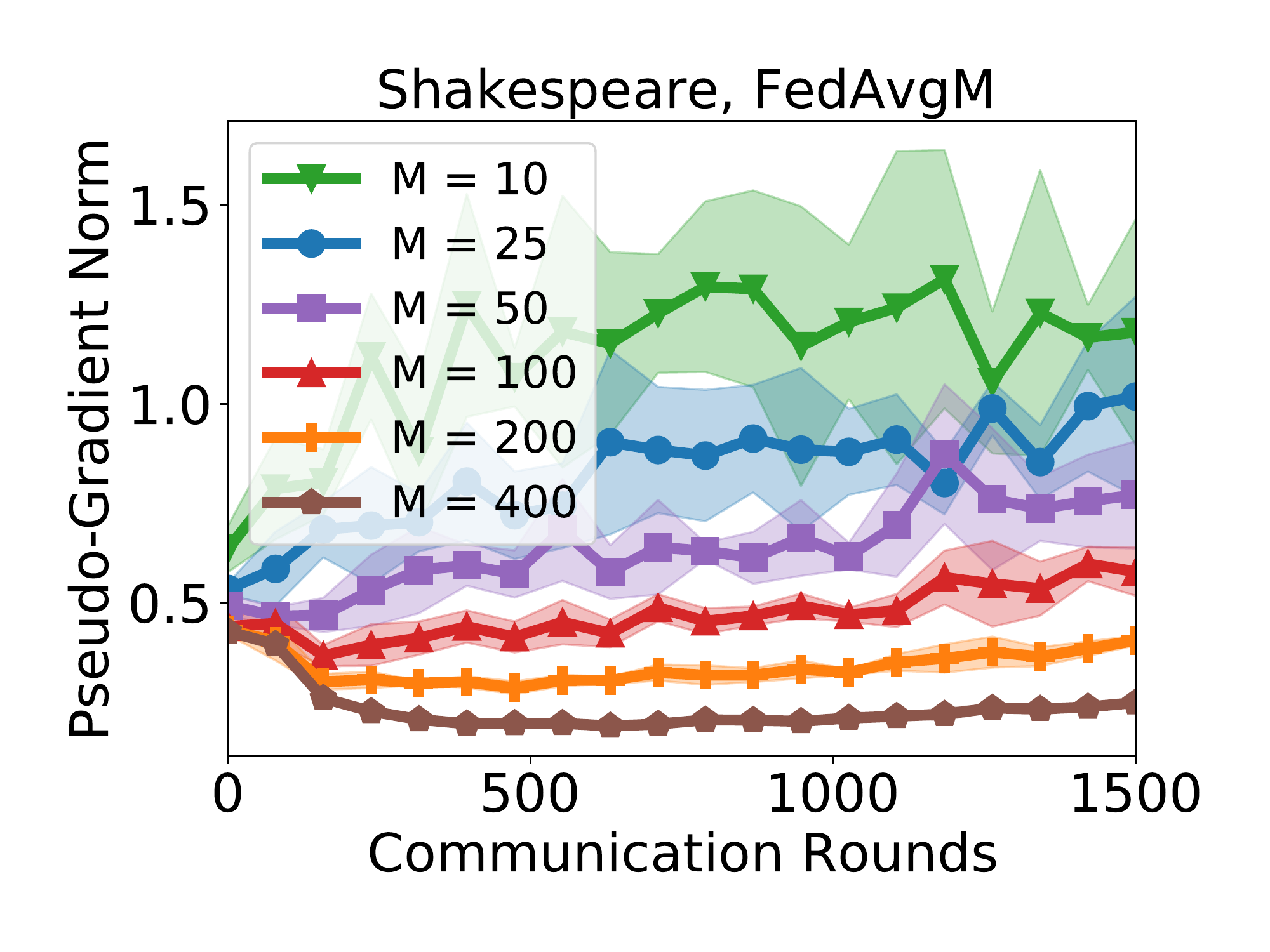}
\end{subfigure}%
\begin{subfigure}{0.24\textwidth}
     \centering
     \includegraphics[width=1\linewidth]{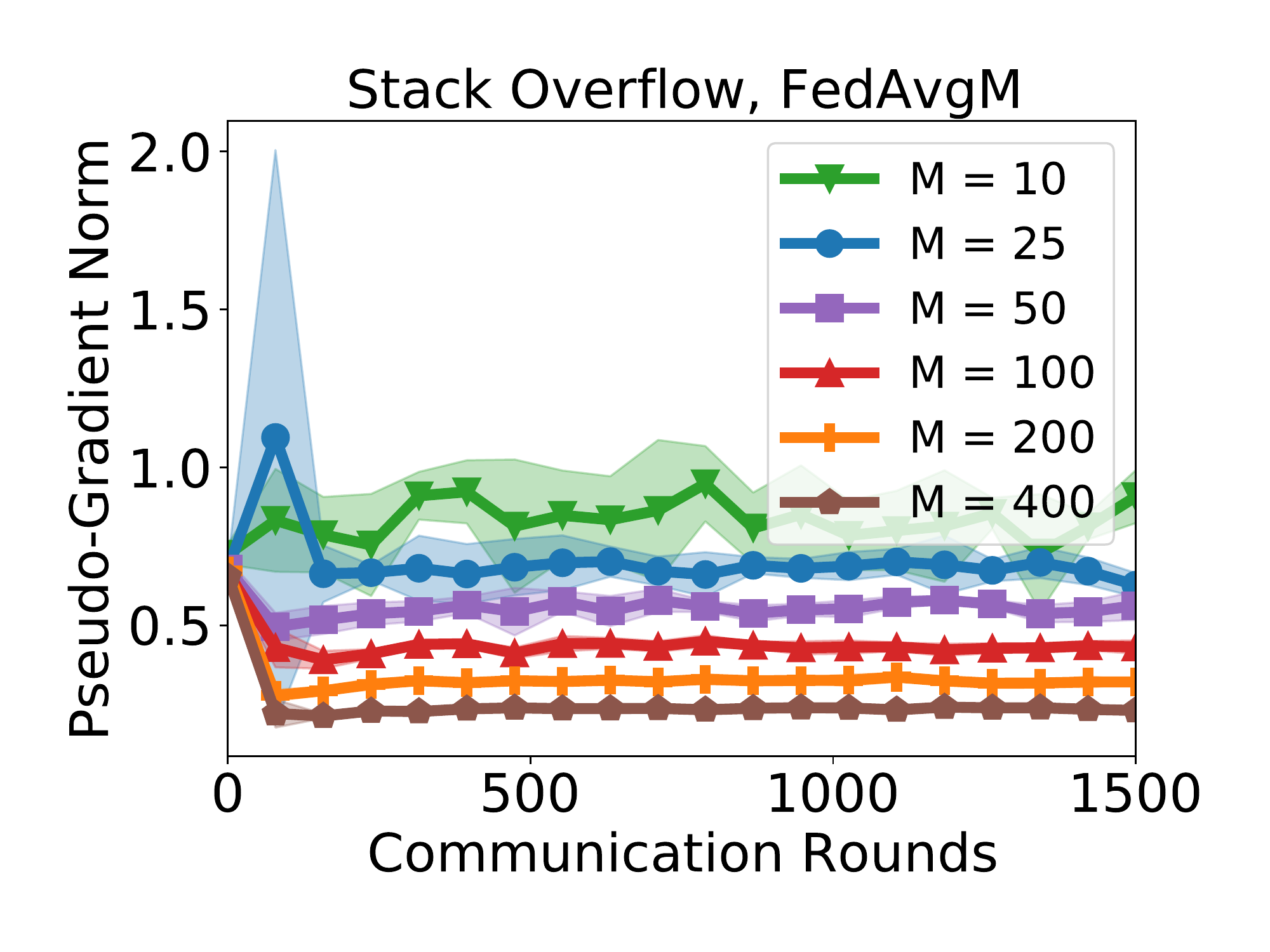}
\end{subfigure}%
\caption{Average pseudo-gradient norm of \fedavgm versus the number of communication rounds, for various tasks and cohort sizes $M$.}
\label{fig:fedavgm_pseudogradient_norm}
\end{figure}

\begin{figure}[ht!]
\centering
\begin{subfigure}{0.24\textwidth}
     \centering
     \includegraphics[width=1\linewidth]{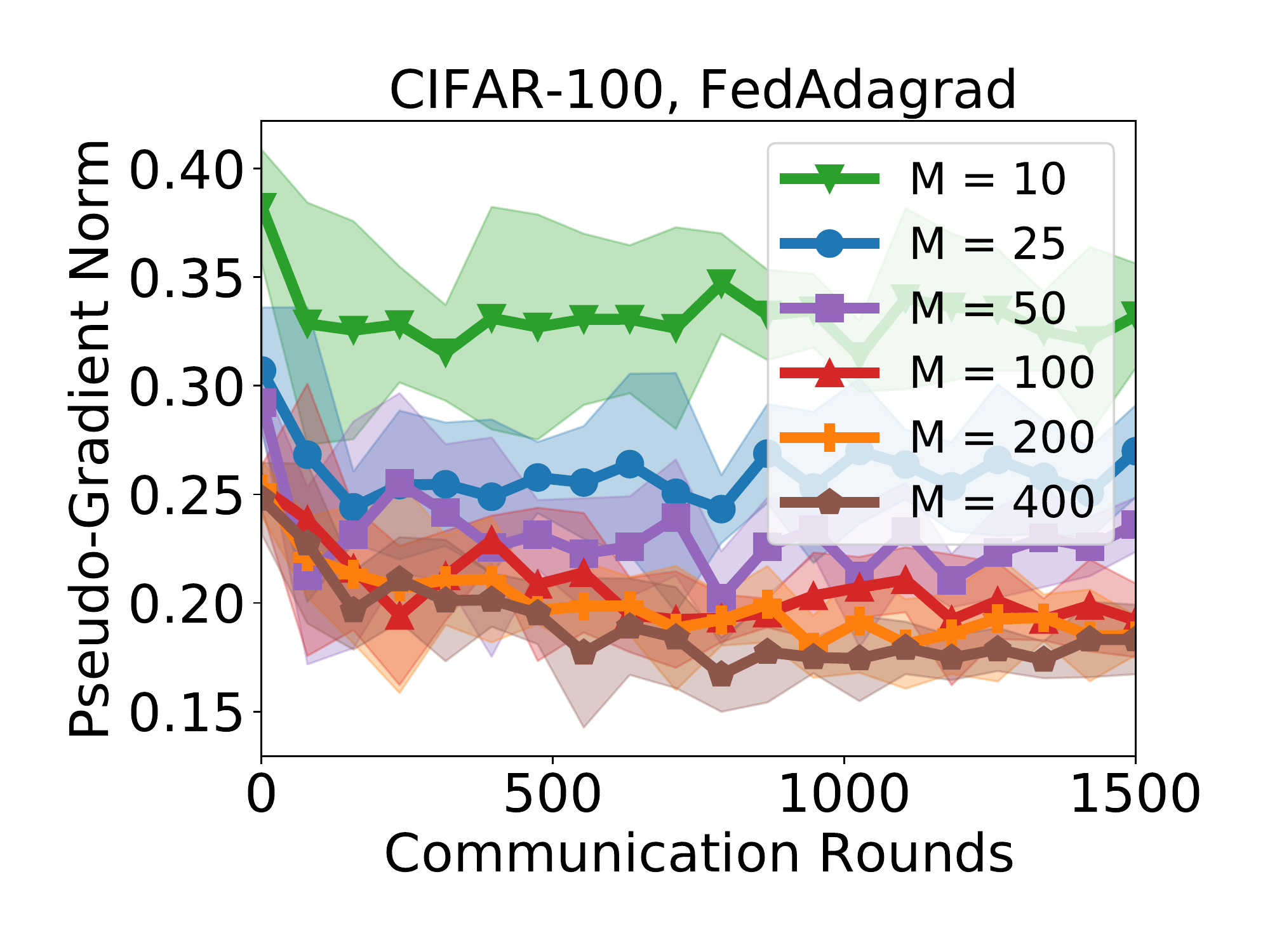}
\end{subfigure}%
\begin{subfigure}{0.24\textwidth}
     \centering
     \includegraphics[width=1\linewidth]{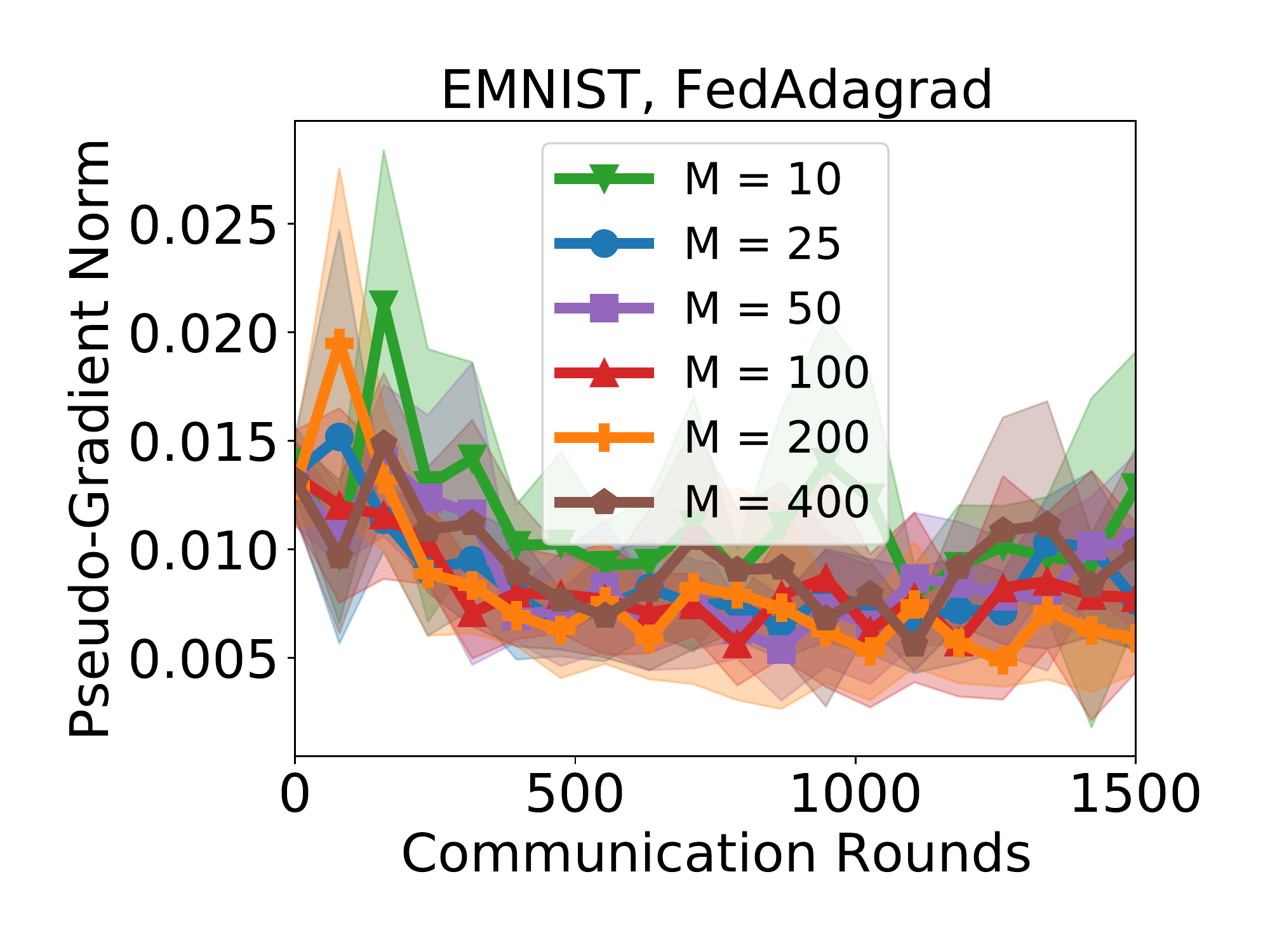}
\end{subfigure}%
\begin{subfigure}{0.24\textwidth}
     \centering
     \includegraphics[width=1\linewidth]{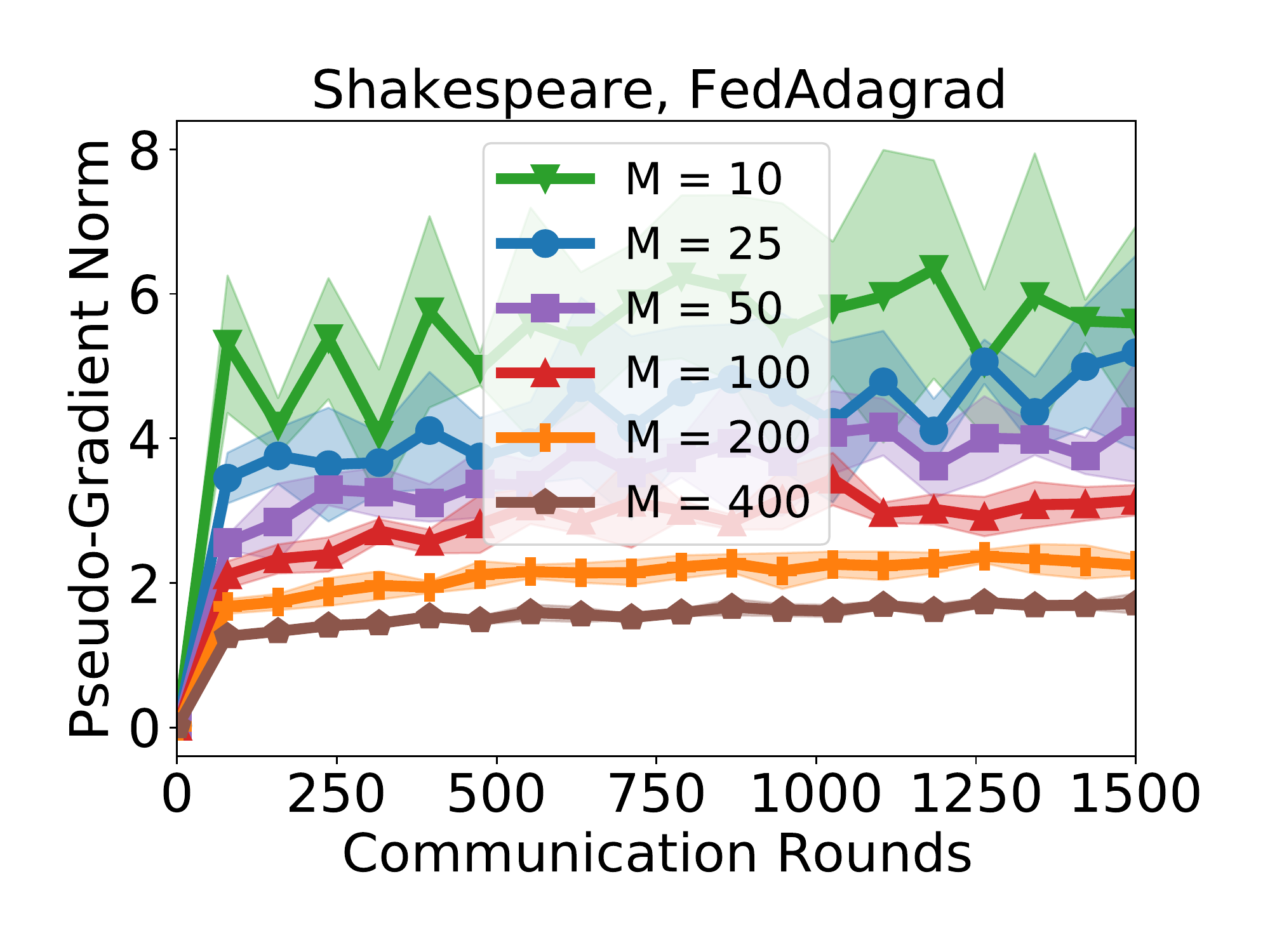}
\end{subfigure}%
\begin{subfigure}{0.24\textwidth}
     \centering
     \includegraphics[width=1\linewidth]{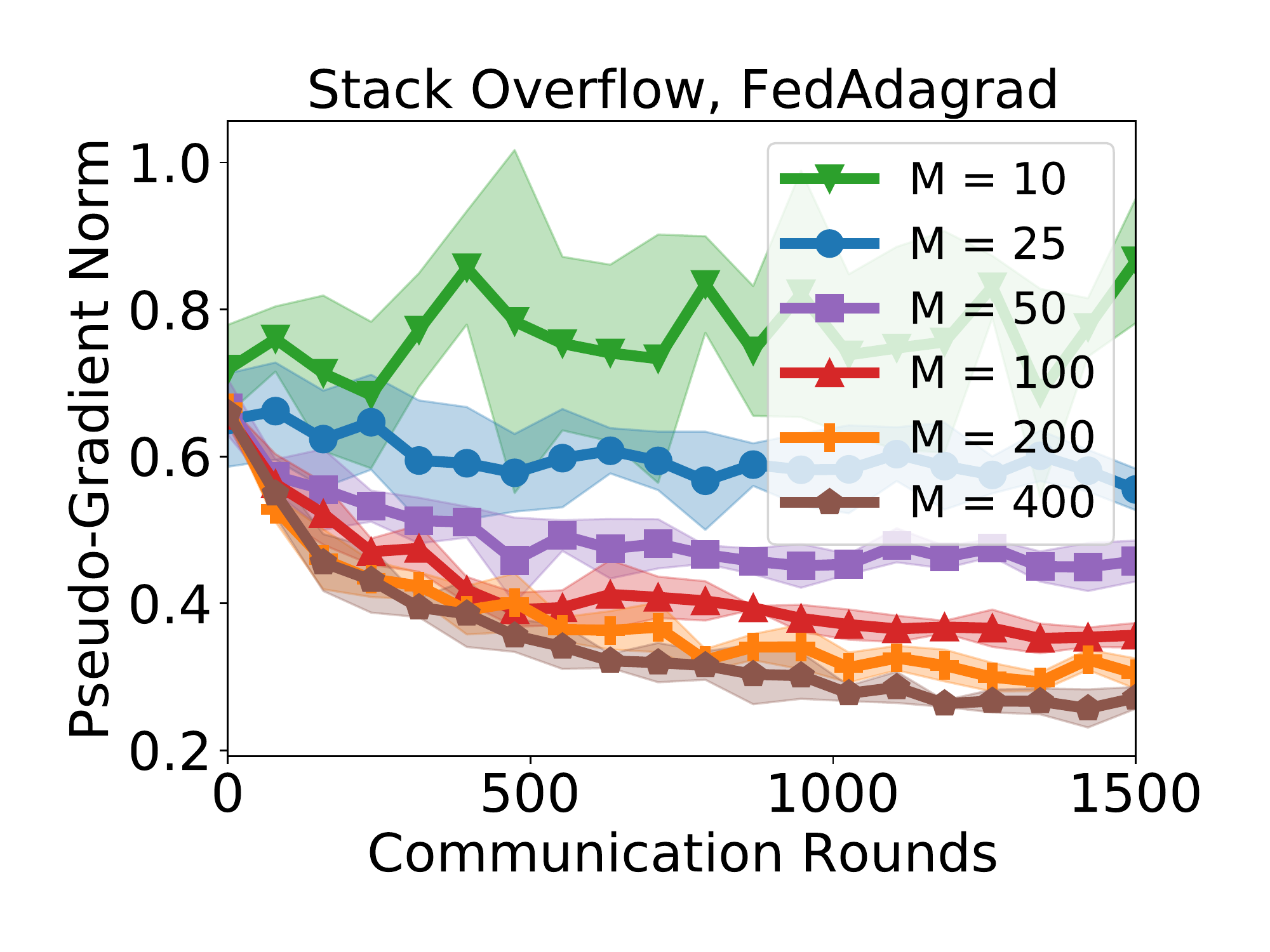}
\end{subfigure}%
\caption{Average pseudo-gradient norm of \fedadagrad versus the number of communication rounds, for various tasks and cohort sizes $M$.}
\label{fig:fedadagrad_pseudogradient_norm}
\end{figure}

\begin{figure}[ht!]
\centering
\begin{subfigure}{0.24\textwidth}
     \centering
     \includegraphics[width=1\linewidth]{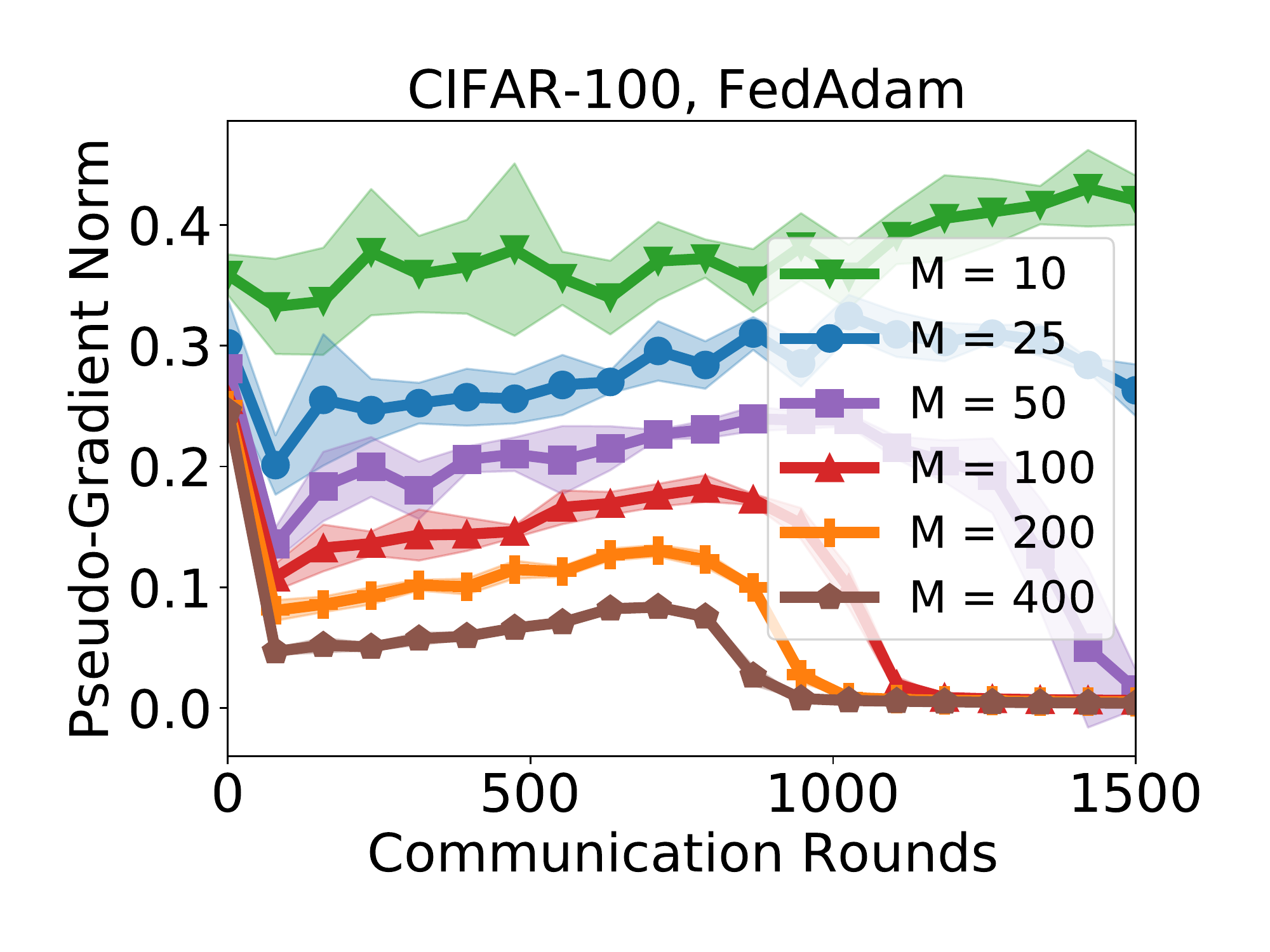}
\end{subfigure}%
\begin{subfigure}{0.24\textwidth}
     \centering
     \includegraphics[width=1\linewidth]{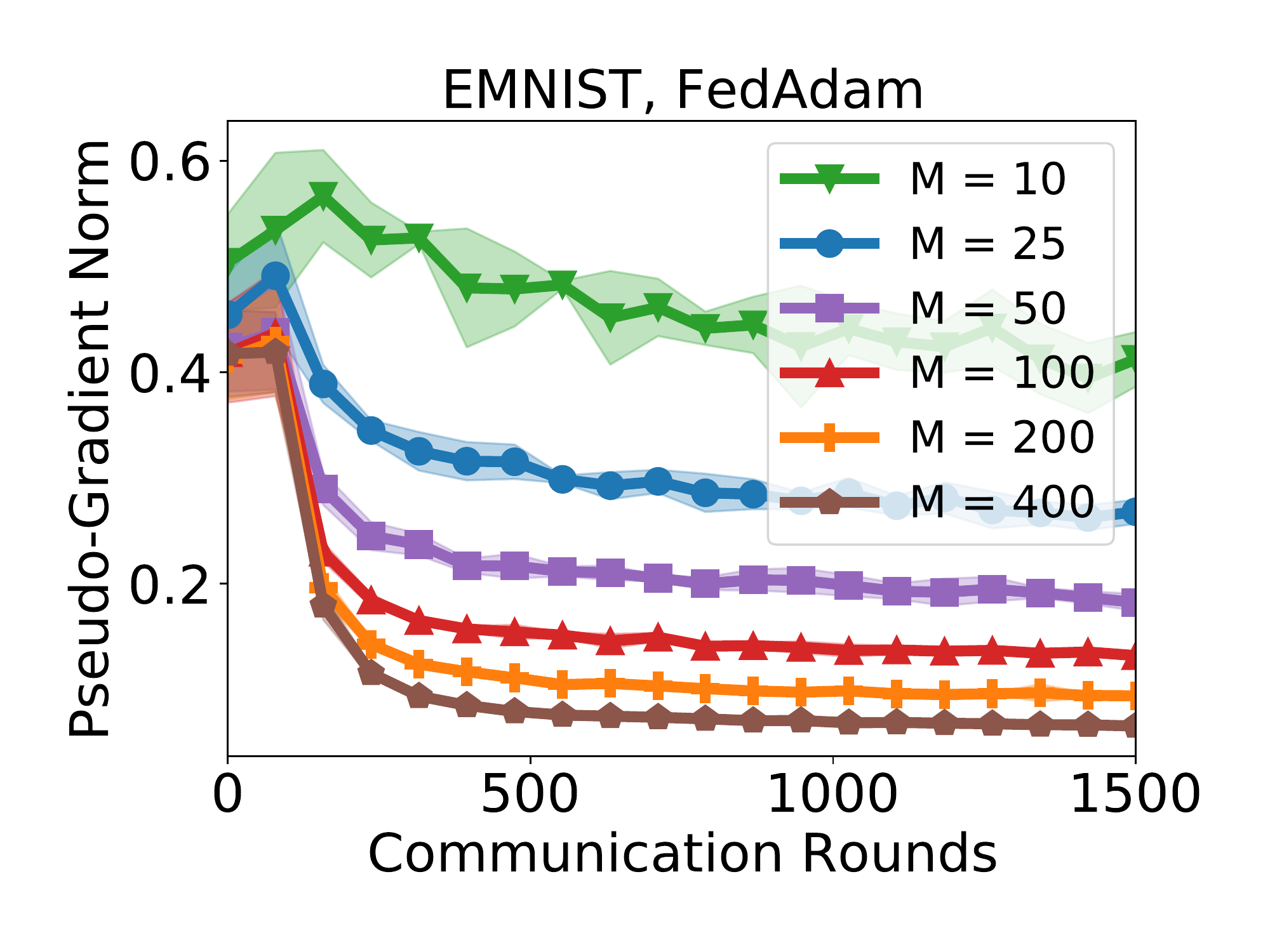}
\end{subfigure}%
\begin{subfigure}{0.24\textwidth}
     \centering
     \includegraphics[width=1\linewidth]{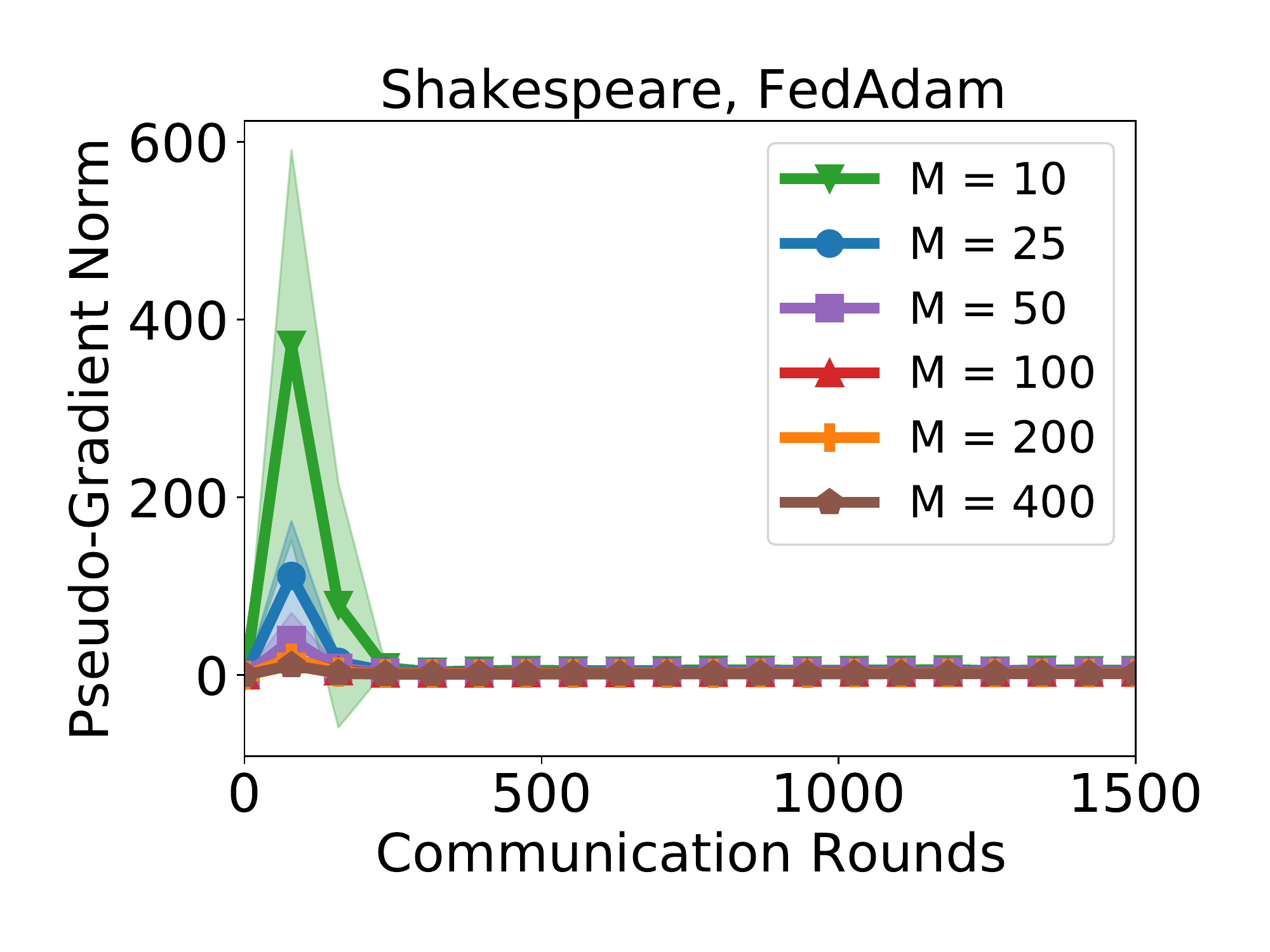}
\end{subfigure}%
\begin{subfigure}{0.24\textwidth}
     \centering
     \includegraphics[width=1\linewidth]{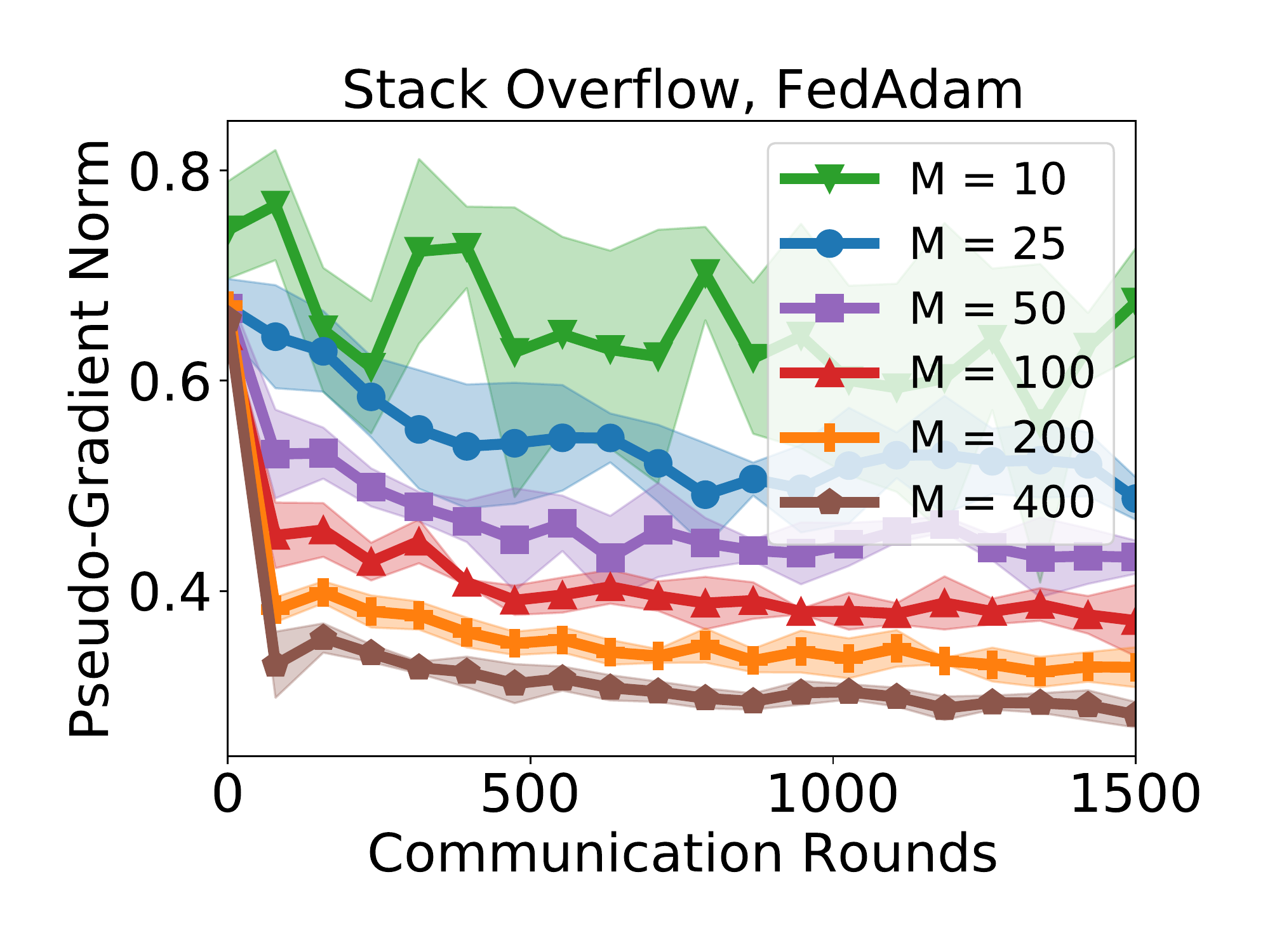}
\end{subfigure}%
\caption{Average pseudo-gradient norm of \fedadam versus the number of communication rounds, for various tasks and cohort sizes $M$.}
\label{fig:fedadam_pseudogradient_norm}
\end{figure}

\begin{figure}[ht!]
\centering
\begin{subfigure}{0.24\textwidth}
     \centering
     \includegraphics[width=1\linewidth]{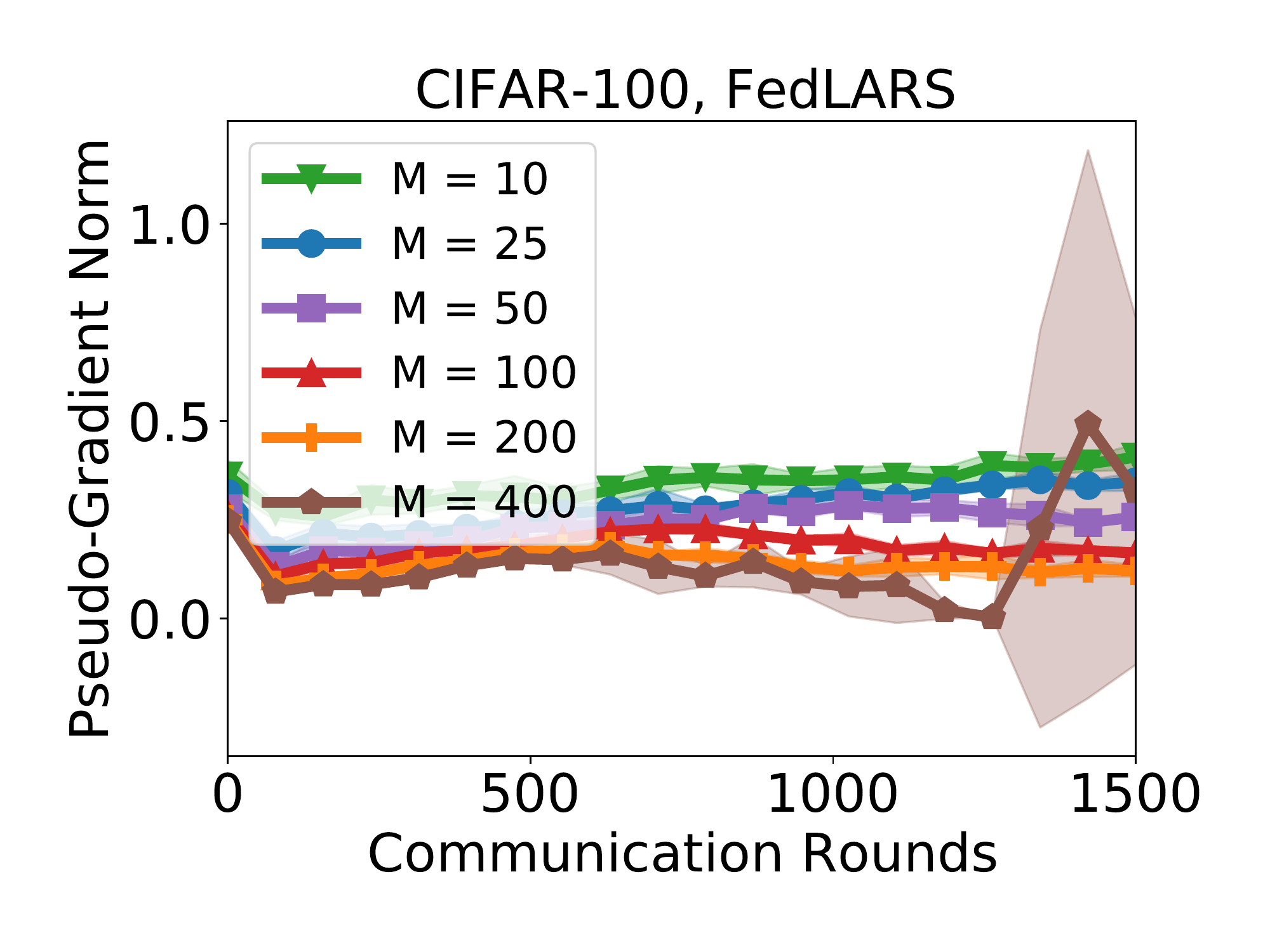}
\end{subfigure}%
\begin{subfigure}{0.24\textwidth}
     \centering
     \includegraphics[width=1\linewidth]{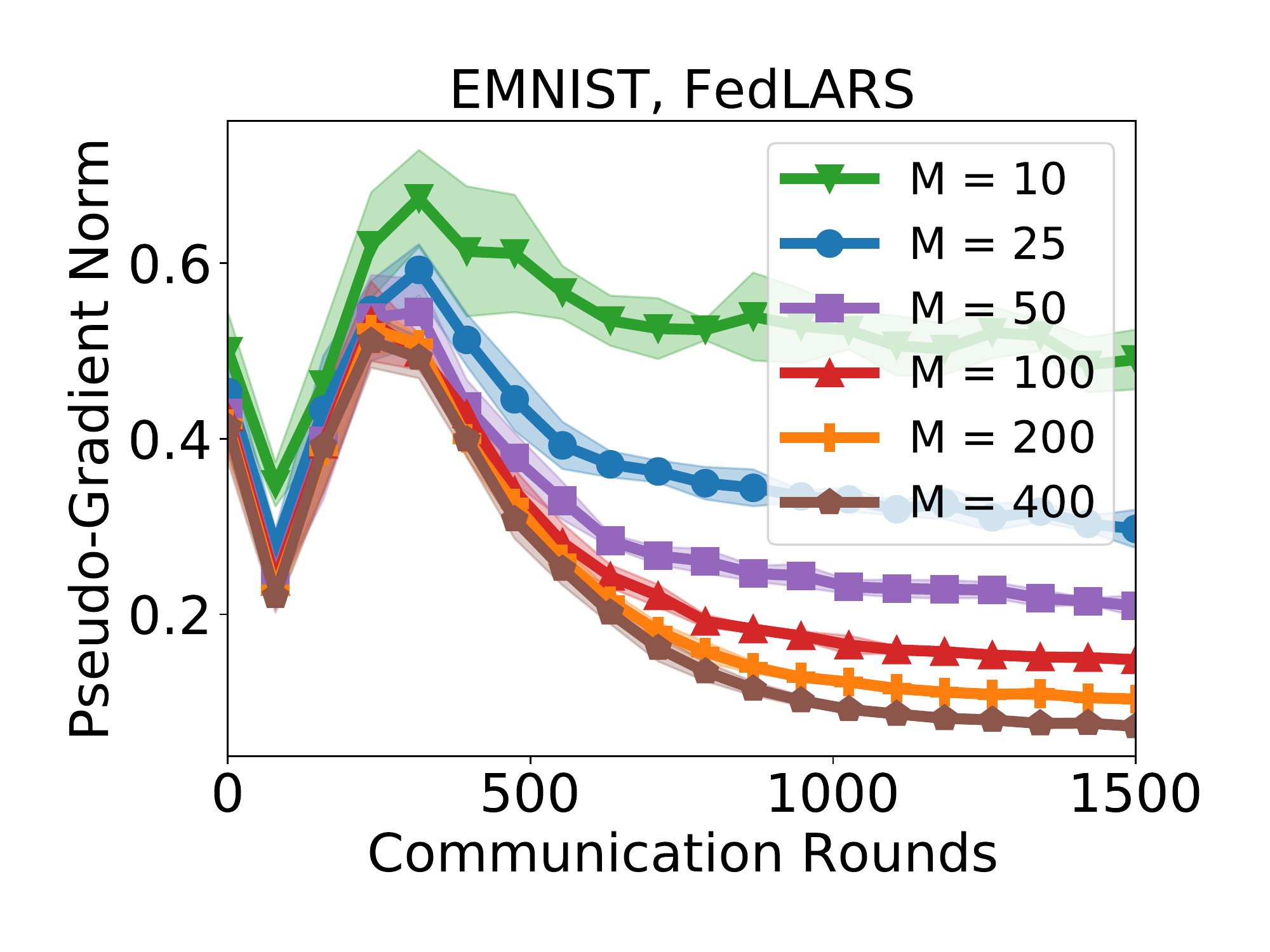}
\end{subfigure}%
\begin{subfigure}{0.24\textwidth}
     \centering
     \includegraphics[width=1\linewidth]{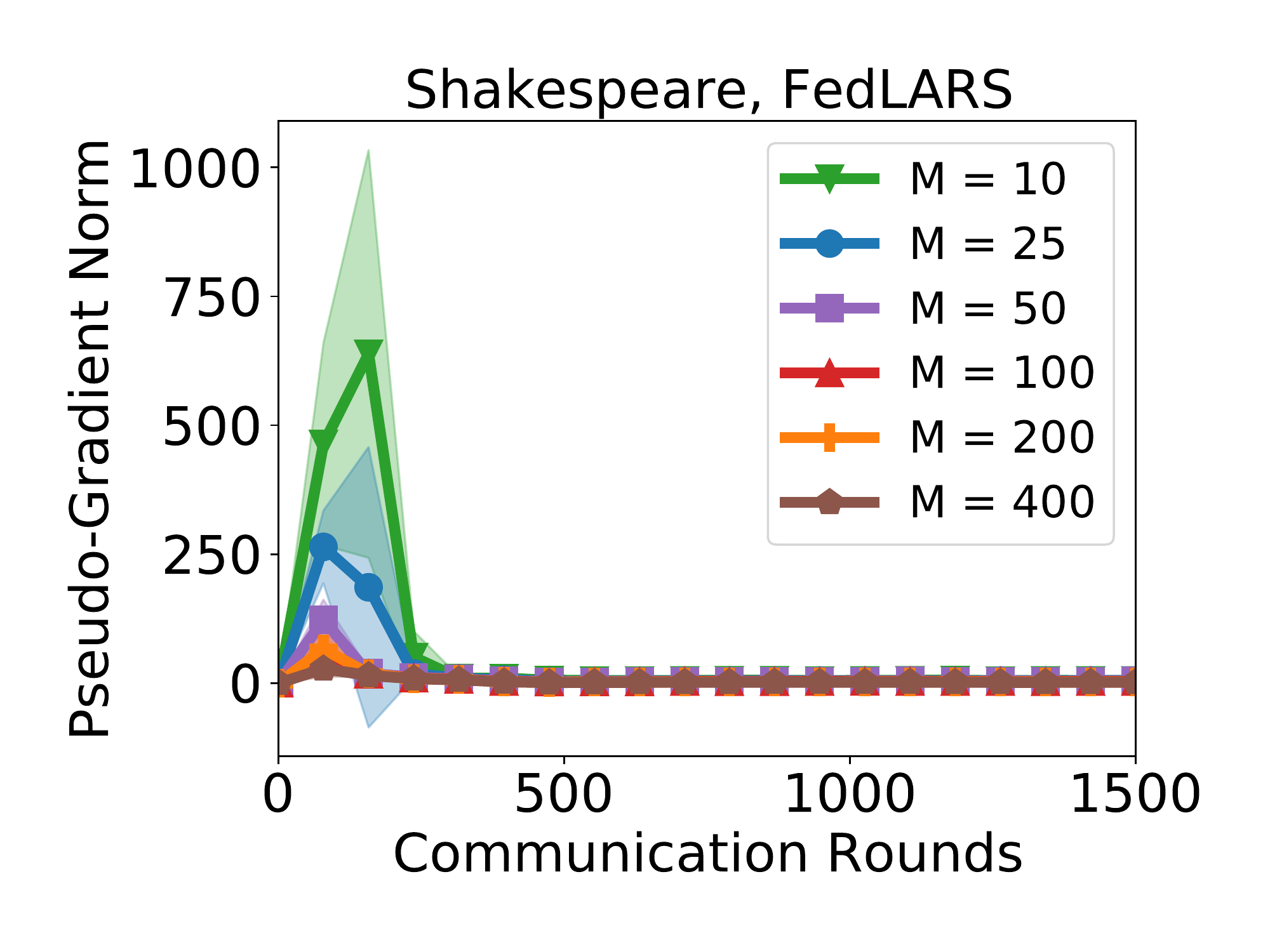}
\end{subfigure}%
\begin{subfigure}{0.24\textwidth}
     \centering
     \includegraphics[width=1\linewidth]{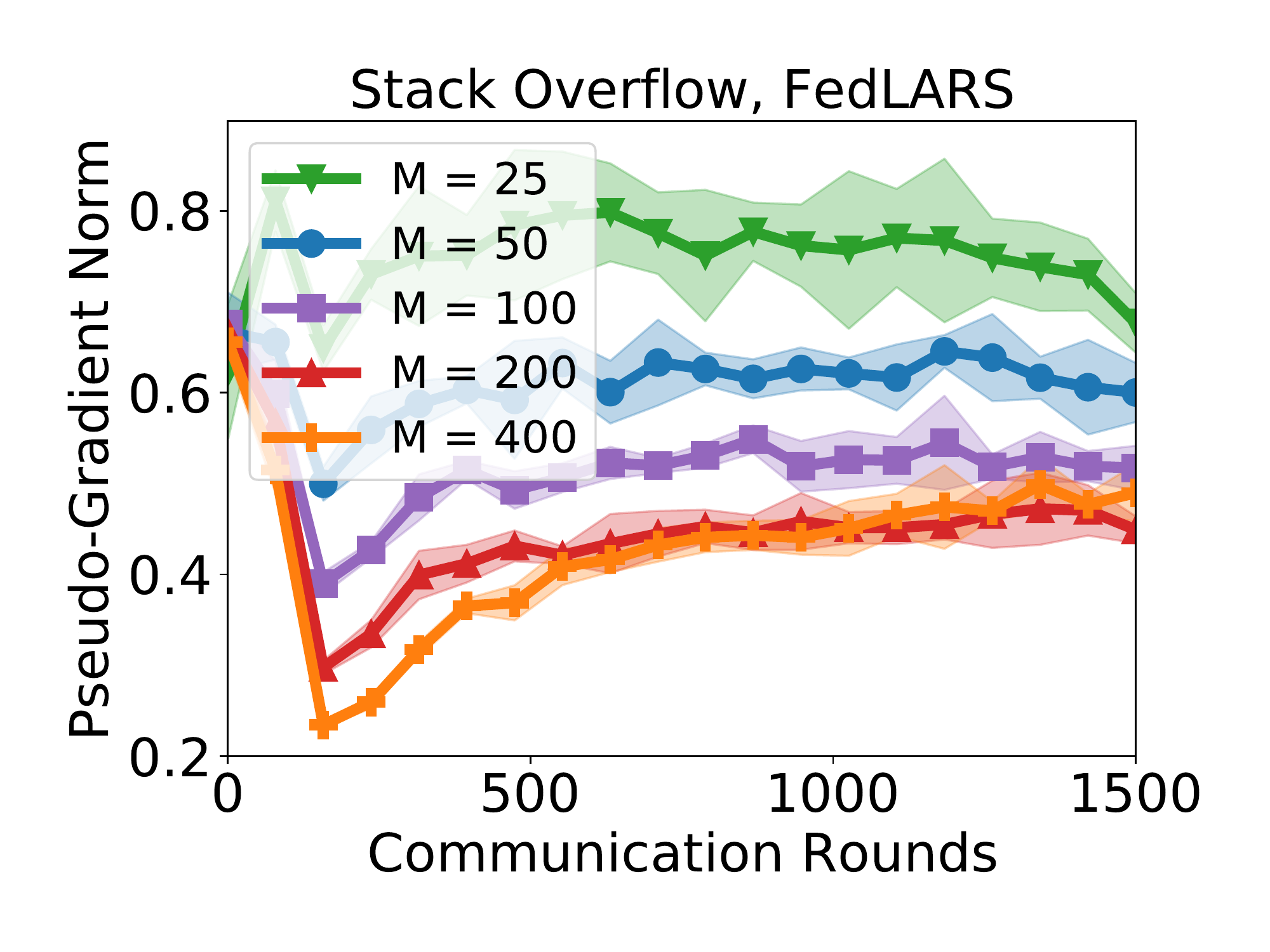}
\end{subfigure}%
\caption{Average pseudo-gradient norm of \fedlars versus the number of communication rounds, for various tasks and cohort sizes $M$.}
\label{fig:fedlars_pseudogradient_norm}
\end{figure}

\begin{figure}[ht!]
\centering
\begin{subfigure}{0.24\textwidth}
     \centering
     \includegraphics[width=1\linewidth]{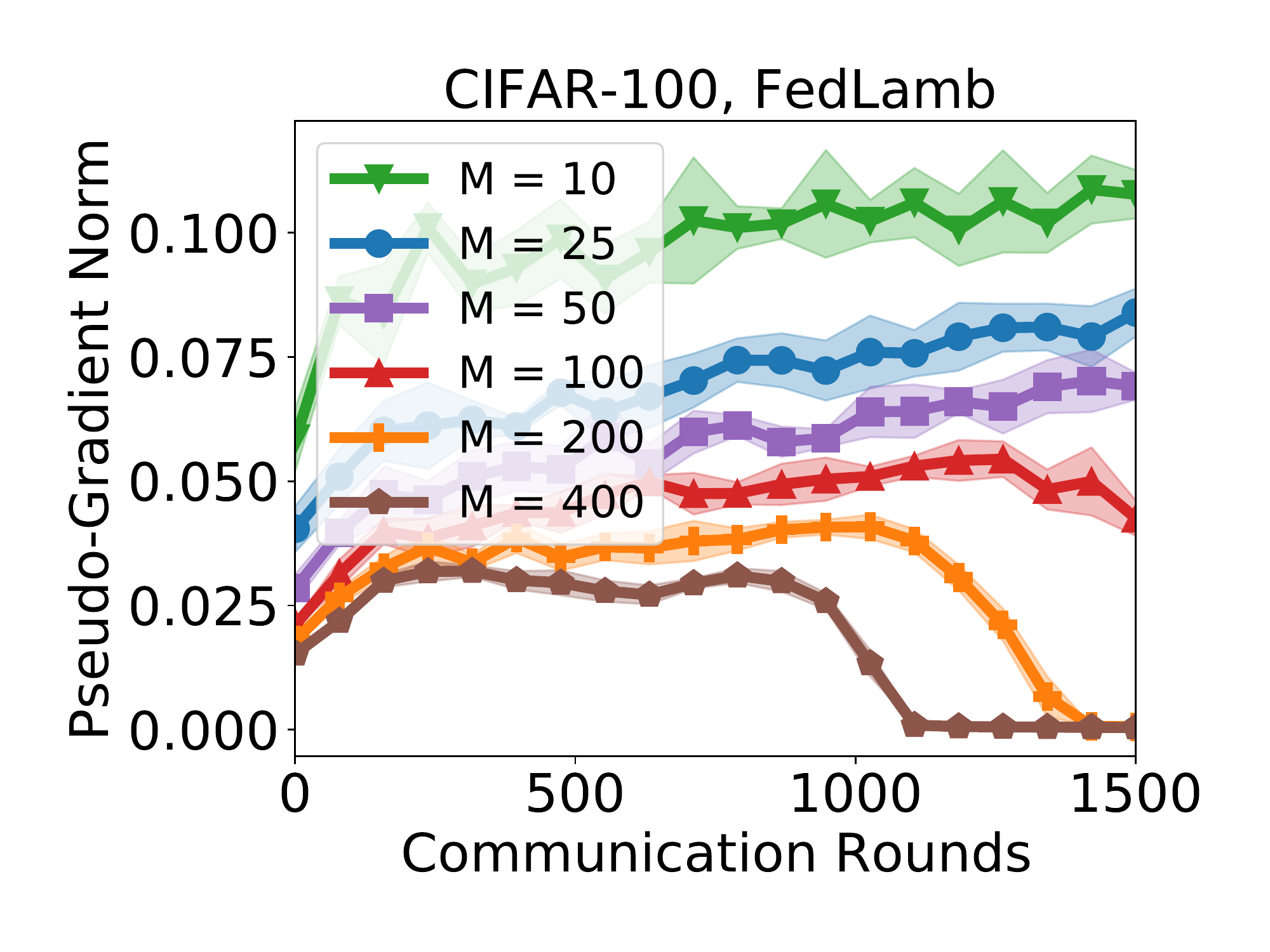}
\end{subfigure}%
\begin{subfigure}{0.24\textwidth}
     \centering
     \includegraphics[width=1\linewidth]{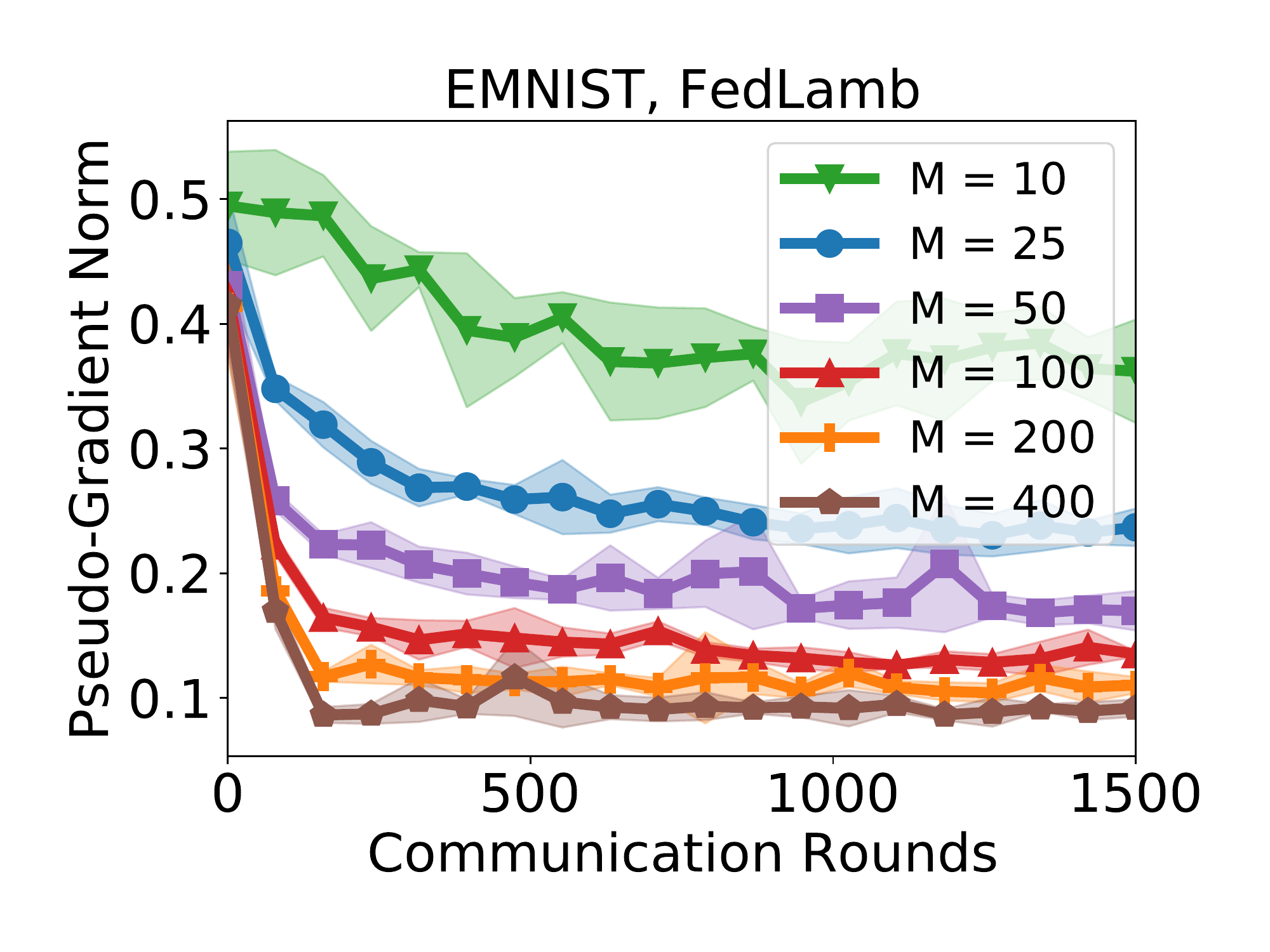}
\end{subfigure}%
\begin{subfigure}{0.24\textwidth}
     \centering
     \includegraphics[width=1\linewidth]{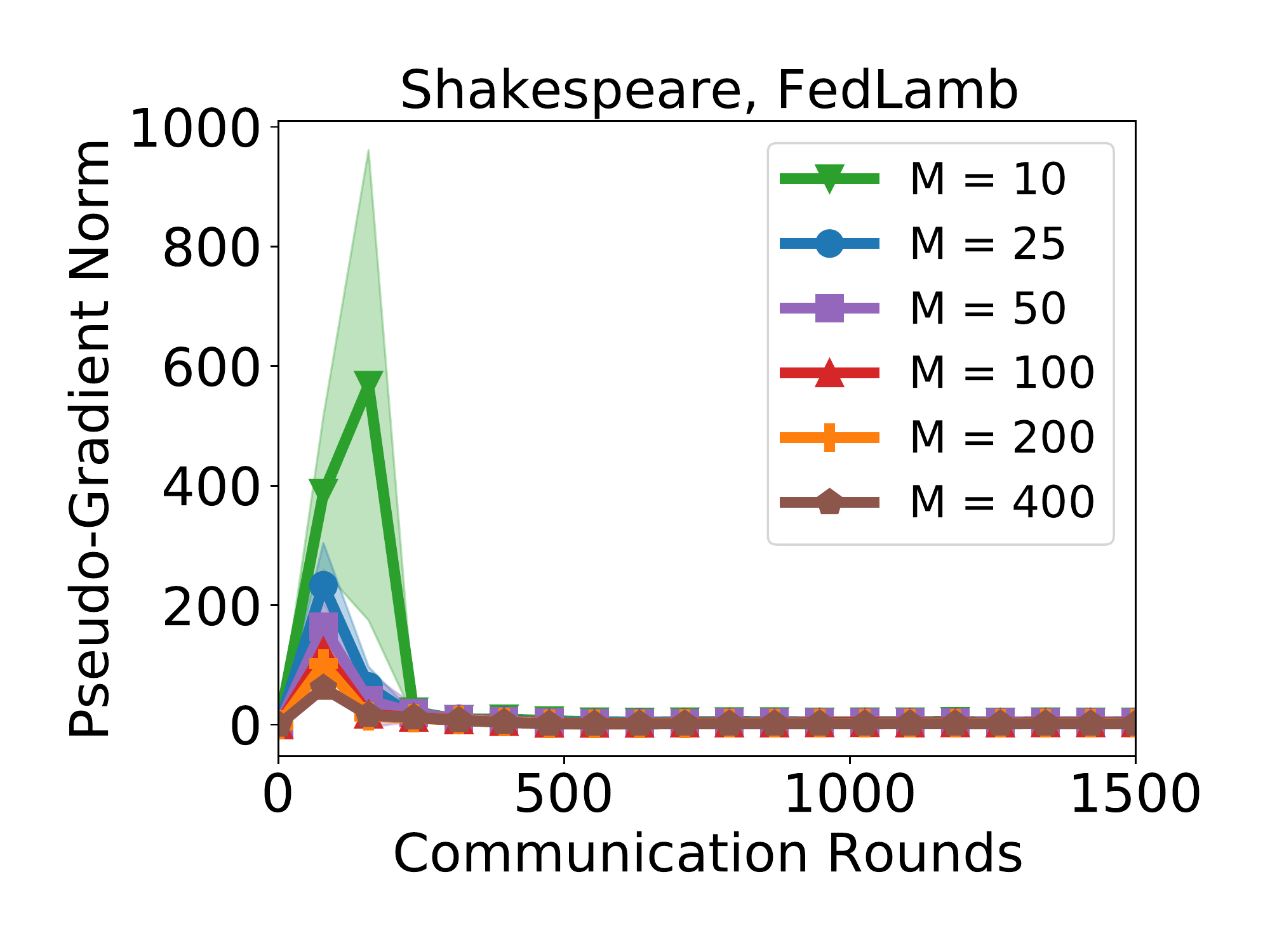}
\end{subfigure}%
\begin{subfigure}{0.24\textwidth}
     \centering
     \includegraphics[width=1\linewidth]{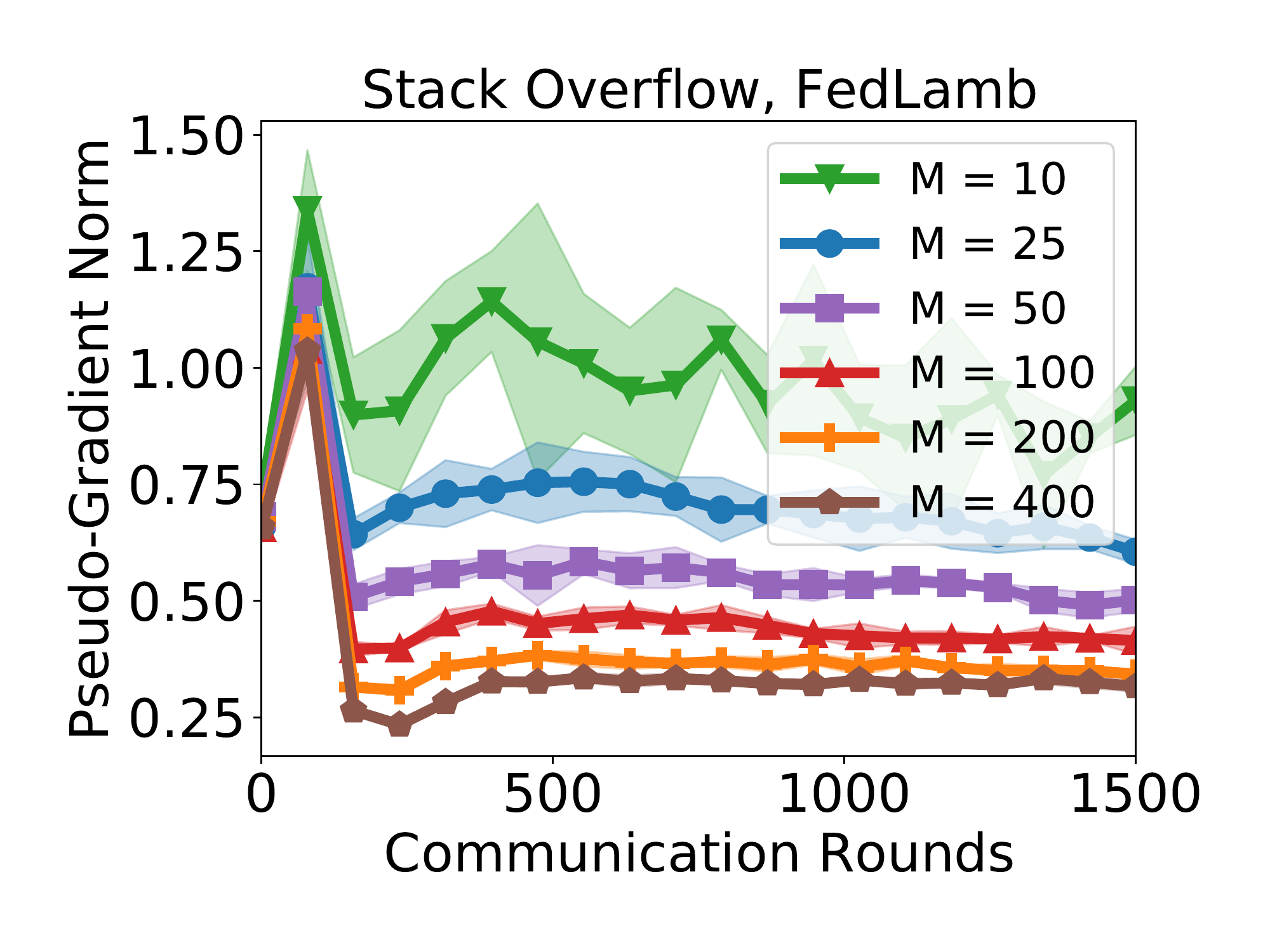}
\end{subfigure}%
\caption{Average pseudo-gradient norm of \fedlamb versus the number of communication rounds, for various tasks and cohort sizes $M$.}
\label{fig:fedlamb_pseudogradient_norm}
\end{figure}

In this section, we present the norm of the server pseudo-gradient $\Delta$ in \cref{alg:fedopt} with respect to the number of communication rounds. We do this for varying cohort sizes and tasks across 1500 communication rounds. All plots give the $\ell_2$ norm of $\Delta$. The results are given in Figures \ref{fig:fedsgd_pseudogradient_norm}, \ref{fig:fedavg_pseudogradient_norm}, \ref{fig:fedavgm_pseudogradient_norm}, \ref{fig:fedadagrad_pseudogradient_norm}, \ref{fig:fedadam_pseudogradient_norm}, \ref{fig:fedlars_pseudogradient_norm}, and \ref{fig:fedlamb_pseudogradient_norm}. These gives the results for \fedsgd, \fedavg, \fedavgm, \fedadagrad, \fedadam, \fedlars, and \fedlamb (respectively).

We find that in nearly all cases, the results for \fedsgd differ from all other algorithms. While there is significant overlap in the pseudo-gradient norm for \fedsgd across all cohort sizes (Figure \ref{fig:fedsgd_pseudogradient_norm}), any method that uses multiple local training steps generally does not see such behavior. The only notable counter-example is \fedadagrad on EMNIST (\cref{fig:fedadagrad_pseudogradient_norm}). Otherwise, both non-adaptive and adaptive federated methods that use local training (such as \fedavg, \fedadam, and \fedlamb) see similar behavior: The pseudo-gradient norm is effectively stratified by the cohort size. Larger cohort sizes lead to smaller pseudo-gradient norms, with little overlap. Moreover, as discussed in \cref{sec:diagnosis}, we see that after enough communication rounds occur, the pseudo-gradient norm obeys an inverse square root scaling rule.

\FloatBarrier

\subsubsection{Cosine Similarity of Client Updates}\label{appendix:cosine_similarity}

\begin{figure}[ht!]
\centering
\begin{subfigure}{0.24\textwidth}
     \centering
     \includegraphics[width=1\linewidth]{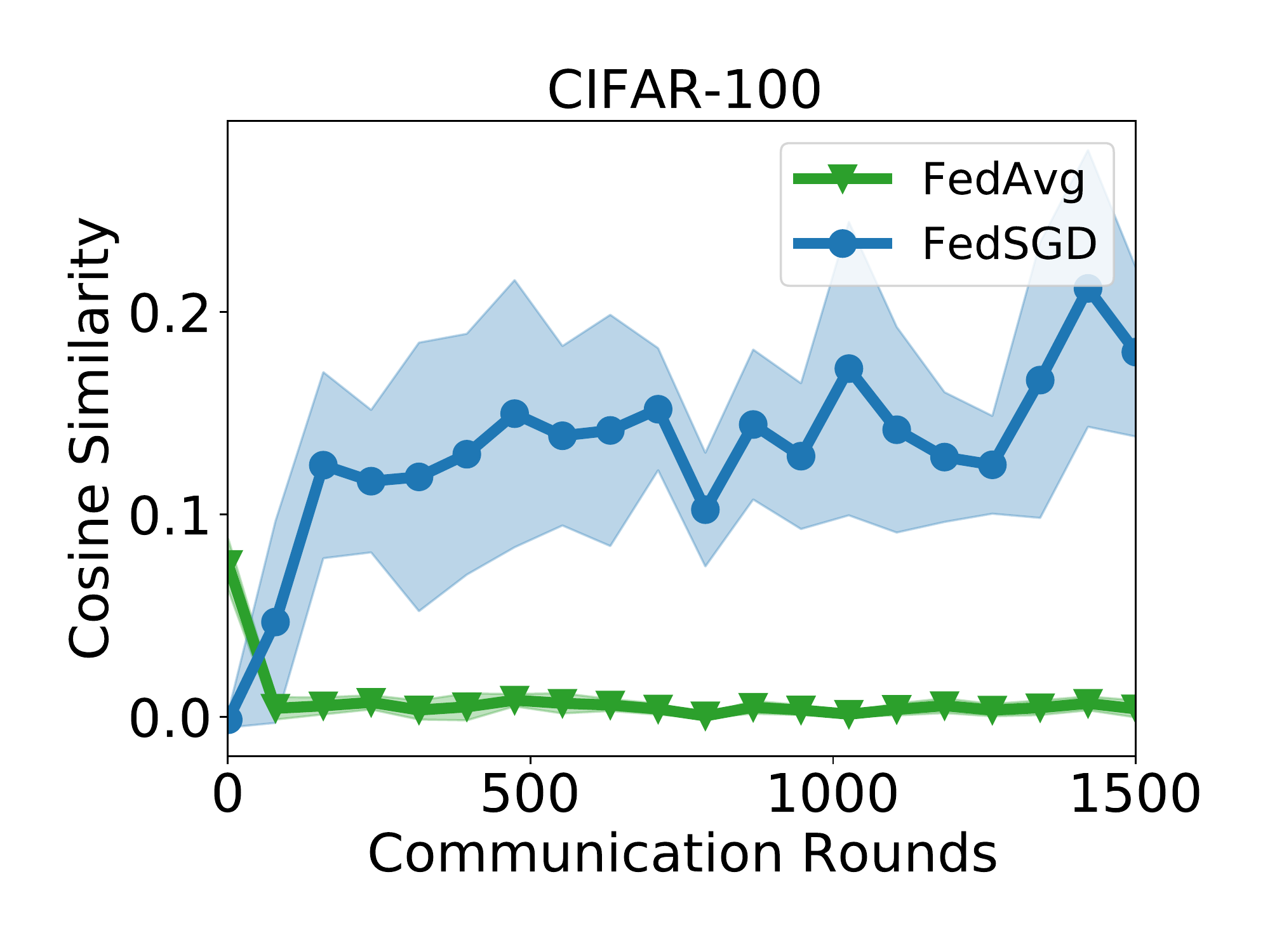}
\end{subfigure}%
\begin{subfigure}{0.24\textwidth}
     \centering
     \includegraphics[width=1\linewidth]{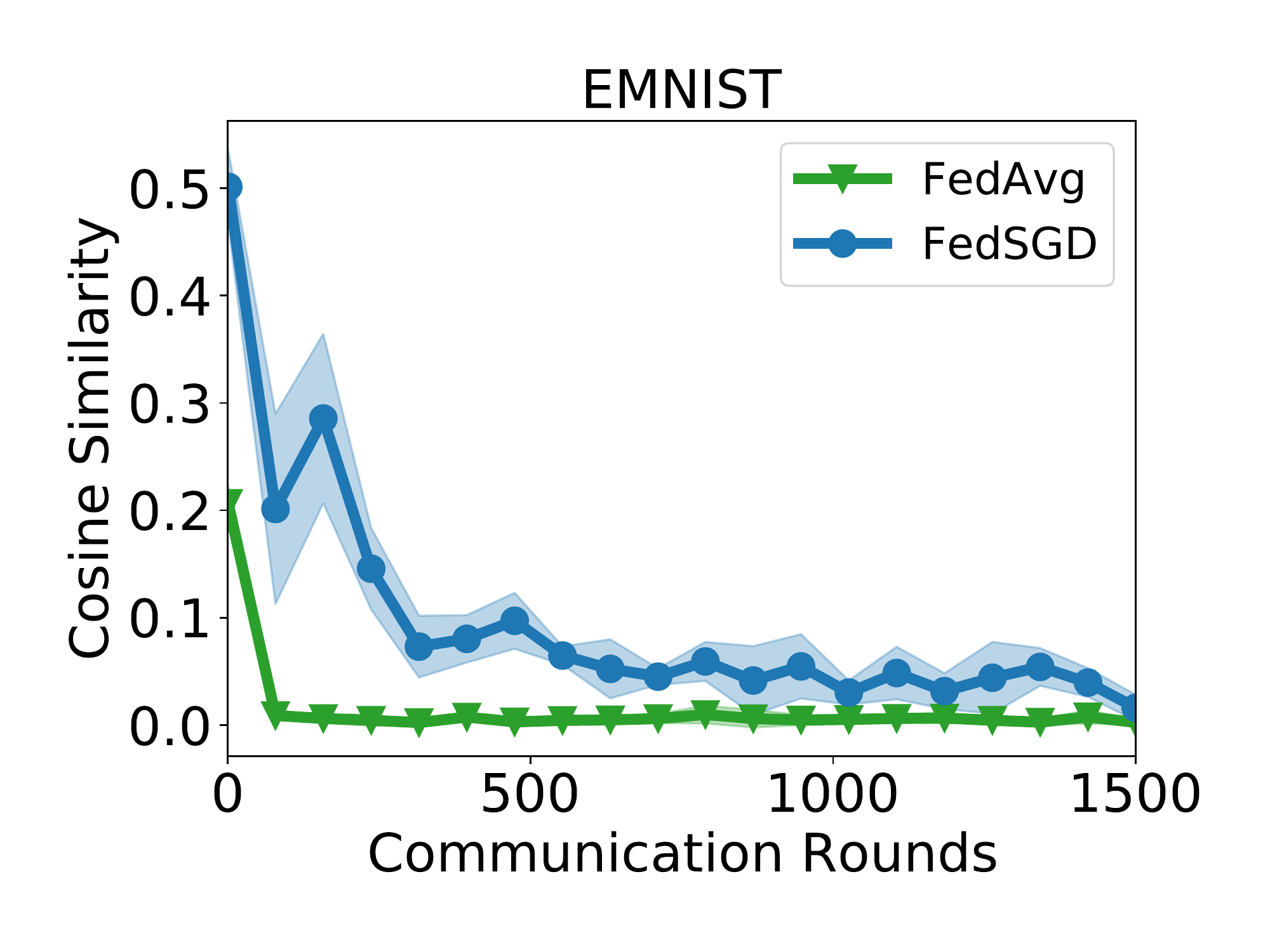}
\end{subfigure}%
\begin{subfigure}{0.24\textwidth}
     \centering
     \includegraphics[width=1\linewidth]{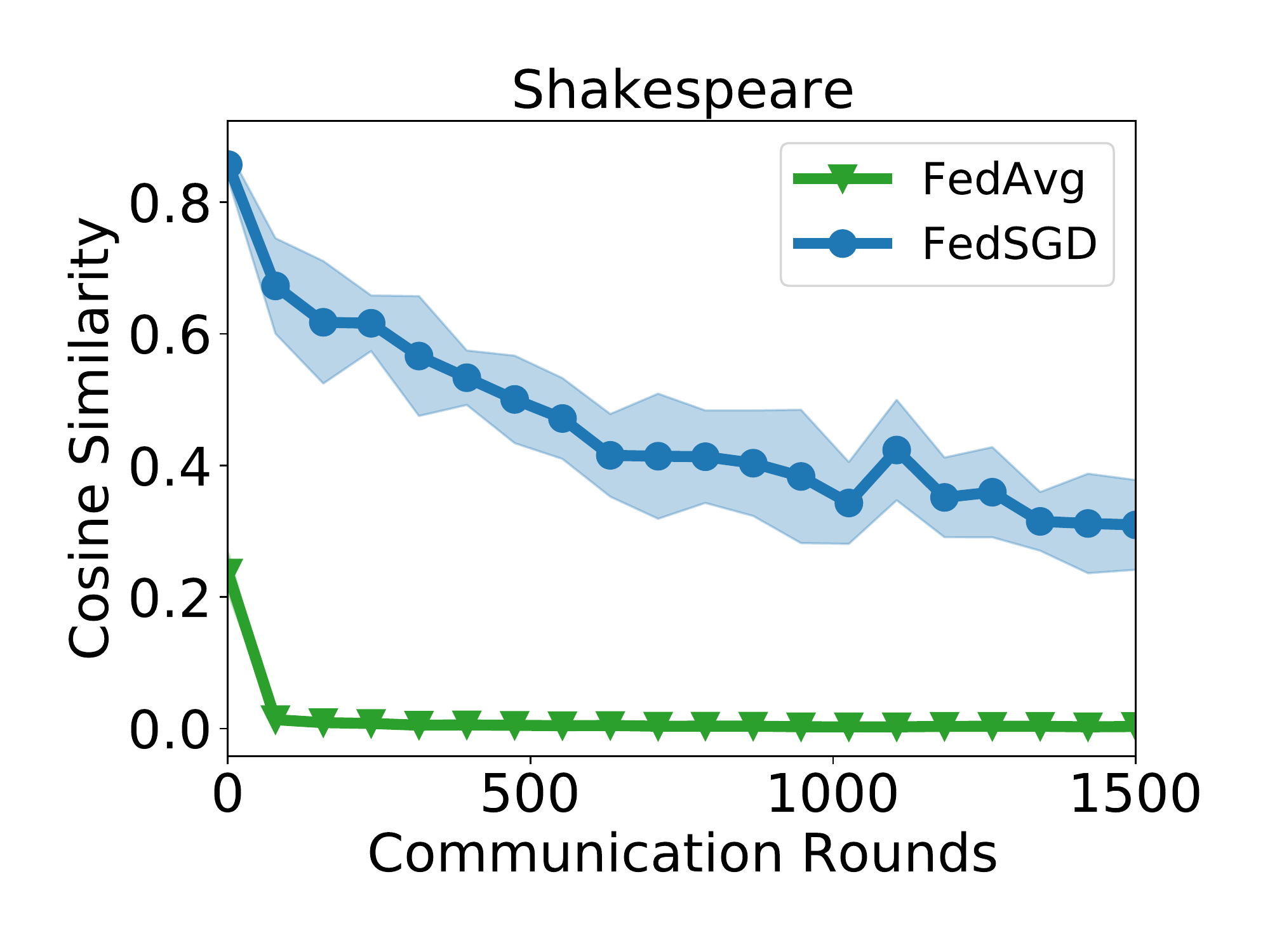}
\end{subfigure}%
\begin{subfigure}{0.24\textwidth}
     \centering
     \includegraphics[width=1\linewidth]{figures/cosine_similarity/stackoverflow_word_cosine_similarity.pdf}
\end{subfigure}%
\caption{Average cosine similarity $\theta_t$ (as in \eqref{eq:cosine_similarity}) between client updates $\Delta_k^t$ with respect to the number of communication rounds, for \fedavg on EMNIST with a cohort size of $M = 50$.}
\label{fig:cosine_similarities}
\end{figure}

Recall that in \cref{sec:diagnosis}, we showed that for \fedavg, client updates are nearly orthogonal on the Stack Overflow task. In this section, we show that this holds across tasks. In \cref{fig:cosine_similarities}, we present the average cosine similarity between distinct clients in each training round, for \fedavg and \fedsgd. Thus, given a cohort size $M$, at each round $t$ we compute $\binom{|M|}{2}$ cosine similarities between client updates, and take the average over all pairs. Formally, we compute, for each round $t$,
\begin{equation}\label{eq:cosine_similarity}
\theta_t := \binom{|C_t|}{2}^{-1} \sum_{\substack{i, j \in C_t \\ i \neq j}} \dfrac{\left\langle \Delta_i^t, \Delta_j^t \right \rangle}{\|\Delta_i^t\|_2\|\Delta_j^t\|_2}
\end{equation}
where $C_t$ is the cohort of sampled clients in round $t$, and $\Delta_k^t$ denotes the client update of client $k \in C_t$ (see \cref{alg:fedopt}). Note that because we normalize, it does not matter whether we use clipping or not (\cref{alg:fedopt_with_clipping}). The results for $\theta_t$ with cohort size $M = 50$ are given in \cref{fig:cosine_similarities}.

We see that in all cases, after a small number of communication rounds, $\theta_t$ becomes close to zero for \fedavg. By contrast, $\theta_t$ is not nearly as small for \fedsgd, especially in intermediate rounds. We note that for EMNIST, the cosine similarity for \fedsgd approaches that of \fedavg as $T \to 1500$.

\FloatBarrier

\subsection{Server Learning Rate Scaling}\label{appendix:server_lr_scaling}

In this section, we present our full results using the learning rate scaling methods proposed in \cref{sec:better_methods}. Recall that our methods increase the server learning rate $\eta_s$ in accordance with the cohort size. To do so, we fix a learning rate $\eta_s$ for some cohort size $M$. As in \eqref{eq:server_lr_scaling}, for $M ' \geq M$, we use a server learning rate $\eta_s'$
\[
\eta_s' = r\left(\frac{M'}{M}\right)\eta_s
\]
where $r: \mathbb{R}_{\geq 0} \to \mathbb{R}_{\geq 0}$ determines the scaling rate. In particular, we focus on $r(a) = \sqrt{a}$ (square root scaling) and $r(a) = a$ (linear scaling). These rules both can be viewed as federated analogs of learning rate scaling techniques used for large-batch training~\citep{krizhevsky2014one, goyal2017accurate}. We use them with a federated version of the warmup technique proposed by \citet{goyal2017accurate}, where we linearly increase the server learning rate from $0$ to $\eta_s'$ over the first $W = 100$ communication rounds.

\begin{figure}[ht!]
\centering
\includegraphics[width=0.5\linewidth]{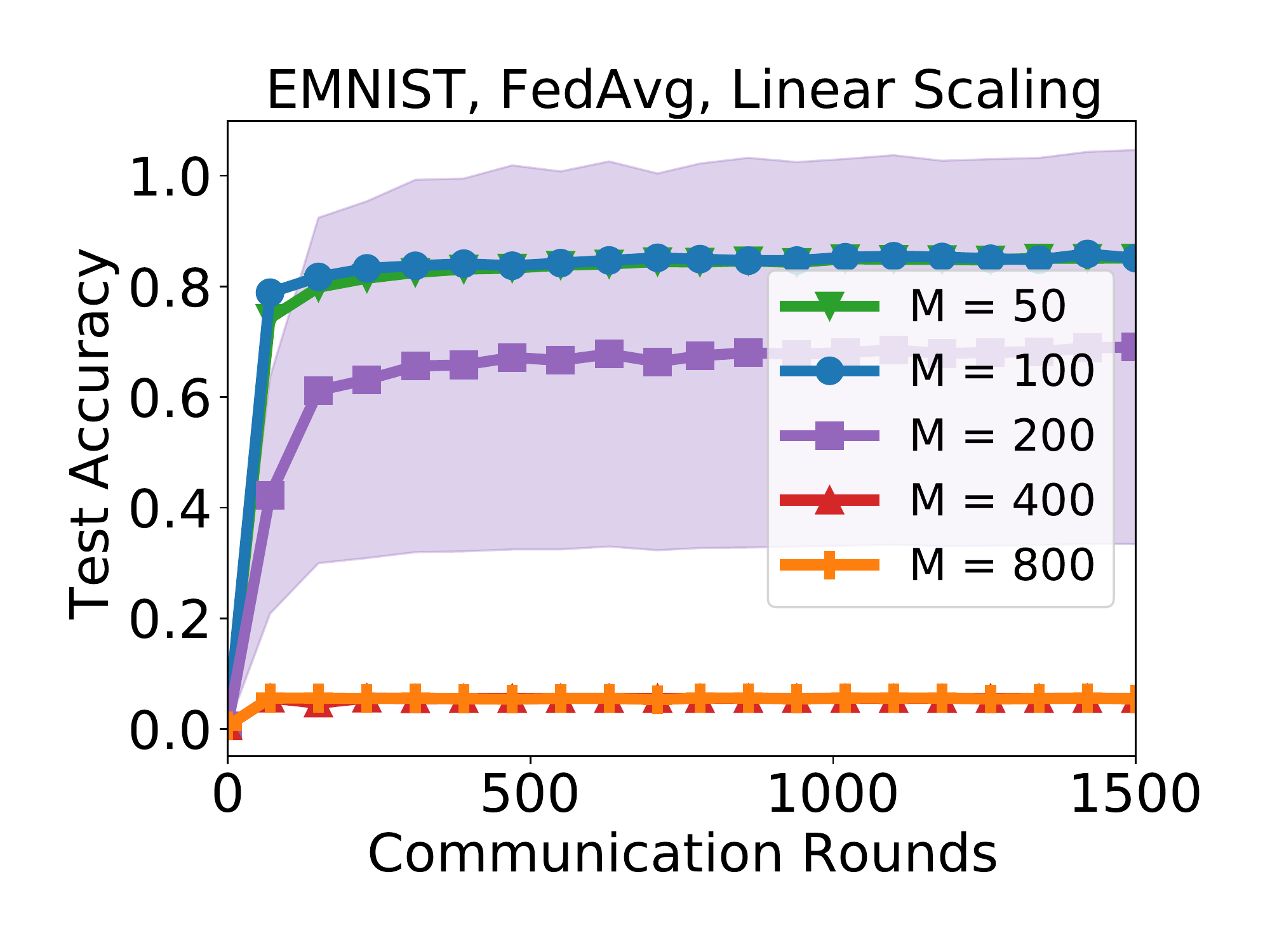}
\caption{The test accuracy of \fedavg on EMNIST with the linear scaling rule, across 5 random trials. The mean accuracy is given in bold, with the standard deviation indicated by the pale region. We see that for $M = 200$, there are a number of catastrophic training failures, while for $M \geq 400$, all trials experienced catastrophic training failures.}
\label{fig:linear_scaling_failure}
\end{figure}

Despite the historical precedent for the linear scaling rule~\citep{goyal2017accurate}, we find that it leads to catastrophic training failures in the federated regime, even with adaptive clipping. To showcase this, we plot the accuracy of \fedavg on EMNIST with the linear scaling rule in \cref{fig:linear_scaling_failure}. We plot the test accuracy over time, averaged across 5 random trials, for various cohort sizes $M$. While $M = 50, 100$ see similar convergence as in \cref{fig:fedavg_test_accuracy_versus_rounds}, for $M = 200$, we saw one catastrophic training failure across all 5 trials. Using $M \geq 400$, we found that all trials resulted in catastrophic training failures. In short, linear scaling can be too aggressive in federated settings, potentially due to heterogeneity among clients (which intuitively requires some amount of conservatism in server model updates).

\begin{figure}[ht!]
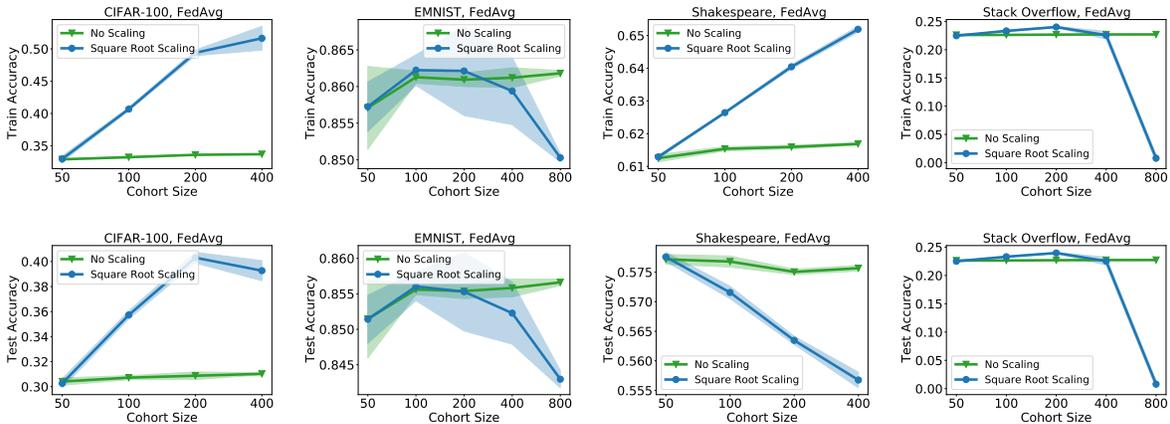

\centering
\begin{subfigure}{0.24\textwidth}
    \centering
    \includegraphics[width=1\linewidth]{figures/scaling/cifar100_fedavg_scaling_train_accuracy_versus_cohort_size.pdf}
\end{subfigure}%
\begin{subfigure}{0.24\textwidth}
    \centering
    \includegraphics[width=1\linewidth]{figures/scaling/emnist_fedavg_scaling_train_accuracy_versus_cohort_size.pdf}
\end{subfigure}%
\begin{subfigure}{0.24\textwidth}
    \centering
    \includegraphics[width=1\linewidth]{figures/scaling/shakespeare_fedavg_scaling_train_accuracy_versus_cohort_size.pdf}
\end{subfigure}%
\begin{subfigure}{0.24\textwidth}
    \centering
    \includegraphics[width=1\linewidth]{figures/scaling/stackoverflow_word_fedavg_scaling_train_accuracy_versus_cohort_size.pdf}
\end{subfigure}
\begin{subfigure}{0.24\textwidth}
    \centering
    \includegraphics[width=1\linewidth]{figures/scaling/cifar100_fedavg_scaling_test_accuracy_versus_cohort_size.pdf}
\end{subfigure}%
\begin{subfigure}{0.24\textwidth}
    \centering
    \includegraphics[width=1\linewidth]{figures/scaling/emnist_fedavg_scaling_test_accuracy_versus_cohort_size.pdf}
\end{subfigure}%
\begin{subfigure}{0.24\textwidth}
    \centering
    \includegraphics[width=1\linewidth]{figures/scaling/shakespeare_fedavg_scaling_test_accuracy_versus_cohort_size.pdf}
\end{subfigure}%
\begin{subfigure}{0.24\textwidth}
    \centering
    \includegraphics[width=1\linewidth]{figures/scaling/stackoverflow_word_fedavg_scaling_test_accuracy_versus_cohort_size.pdf}
\end{subfigure}
\caption{The train accuracy (top) and test accuracy (bottom) of \fedavg using square root scaling with warmup, versus no scaling, after training for 1500 communication rounds. Results are given for various cohort sizes and tasks}
\label{fig:accuracy_sqrt_scaling}
\end{figure}

By contrast, the square root scaling rule did not lead to such training failures. We plot the training accuracy and test accuracy of \fedavg using the square root scaling rule in \cref{fig:accuracy_sqrt_scaling}. We plot this with respect to the cohort size, with and without the scaling rule. We see that the performance of the scaling rule is decidedly mixed. While it leads to significant improvements in training accuracy for CIFAR-100 and Shakespeare, it leads to only minor improvements (or a degradation in training accuracy) for EMNIST and Stack Overflow. Notably, while the training accuracy improvement also led to a test accuracy improvement for CIFAR-100, the same is not true for Shakespeare. In fact, the training benefits of the square root scaling there led to worse generalization across the board.

\FloatBarrier

\subsection{Dynamic Cohort Sizes}\label{appendix:dyanmic_cohort_sizes}

In this section, we plot the full results of using the dynamic cohort size strategy from \cref{sec:better_methods}. Recall that there, we use an analog of dynamic batch size methods for centralized learning, where the cohort size is increased over time. We specifically start with a cohort size of $M = 50$, and double every 300 communication rounds. If doubling would ever make the cohort size larger than the number of training clients, we simply use the full set of training clients in a cohort.

\begin{figure}[ht!]
\centering

\begin{subfigure}{0.24\textwidth}
    \centering
    \includegraphics[width=1\linewidth]{figures/variable_cohorts/cifar100_fedavg_test_accuracy_versus_examples_variable.pdf}
\end{subfigure}%
\begin{subfigure}{0.24\textwidth}
    \centering
    \includegraphics[width=1\linewidth]{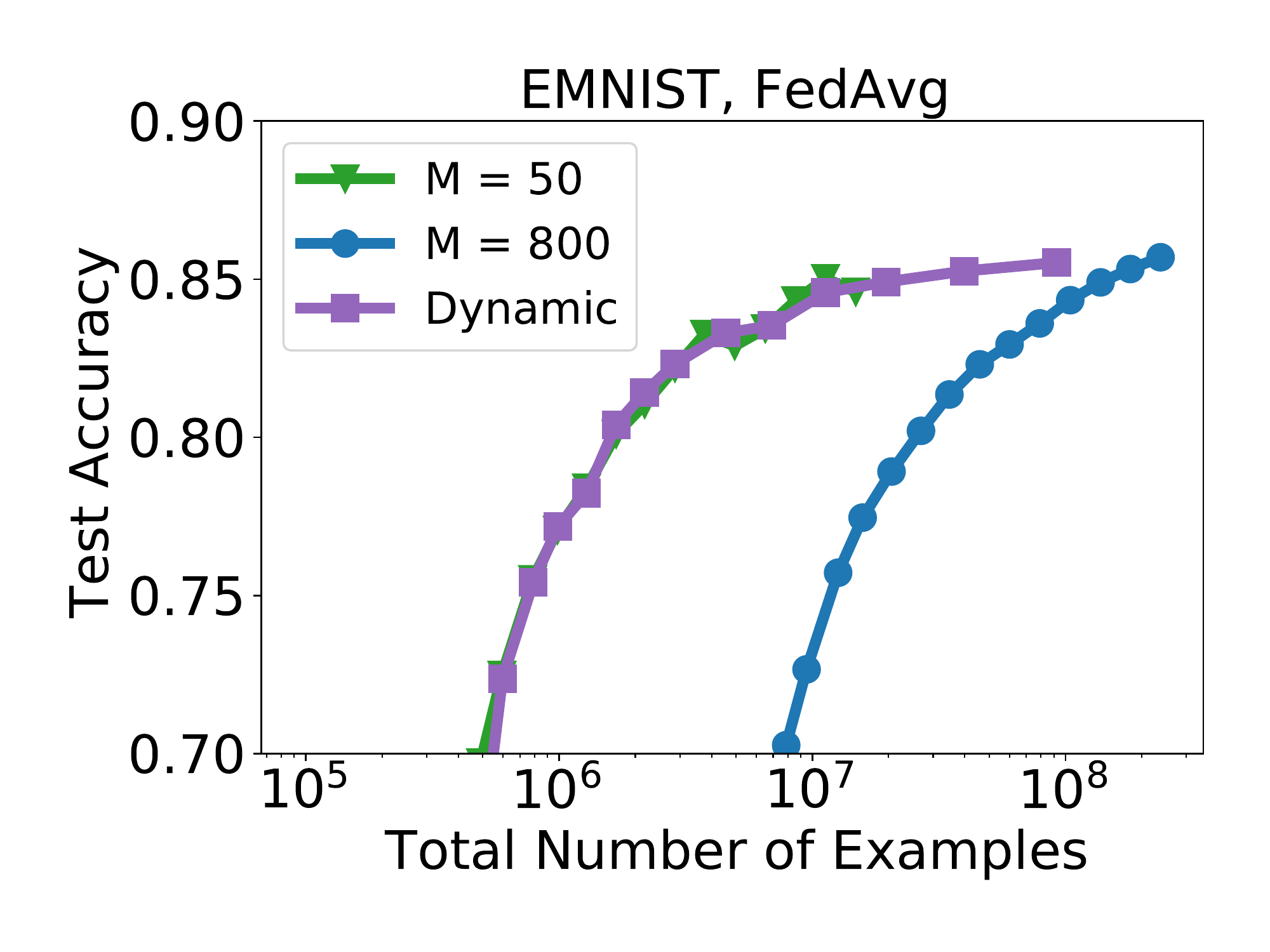}
\end{subfigure}%
\begin{subfigure}{0.24\textwidth}
    \centering
    \includegraphics[width=1\linewidth]{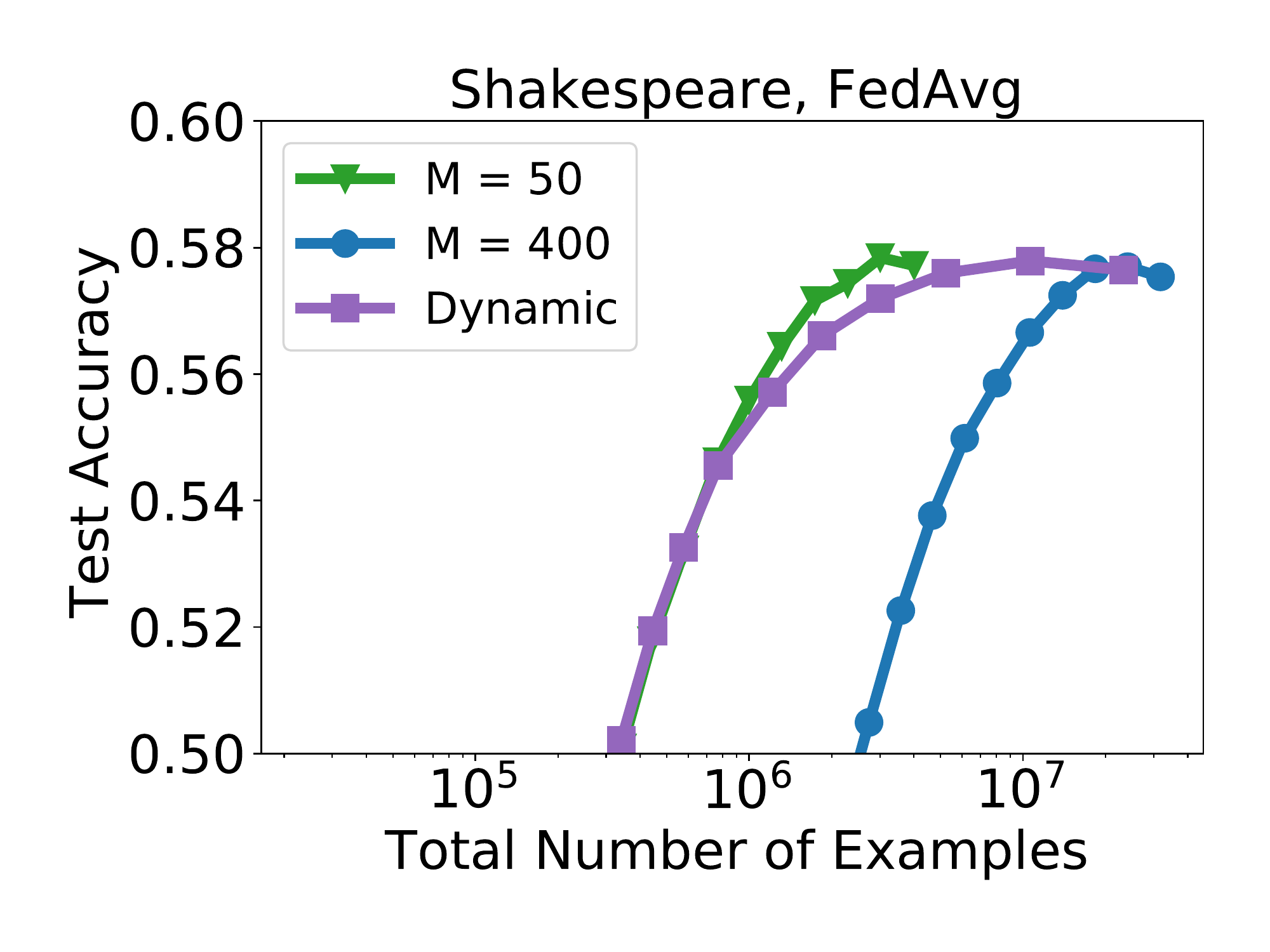}
\end{subfigure}%
\begin{subfigure}{0.24\textwidth}
    \centering
    \includegraphics[width=1\linewidth]{figures/variable_cohorts/stackoverflow_word_fedavg_test_accuracy_versus_examples_variable.pdf}
\end{subfigure}
\begin{subfigure}{0.24\textwidth}
    \centering
    \includegraphics[width=1\linewidth]{figures/variable_cohorts/cifar100_fedadam_test_accuracy_versus_examples_variable.pdf}
\end{subfigure}%
\begin{subfigure}{0.24\textwidth}
    \centering
    \includegraphics[width=1\linewidth]{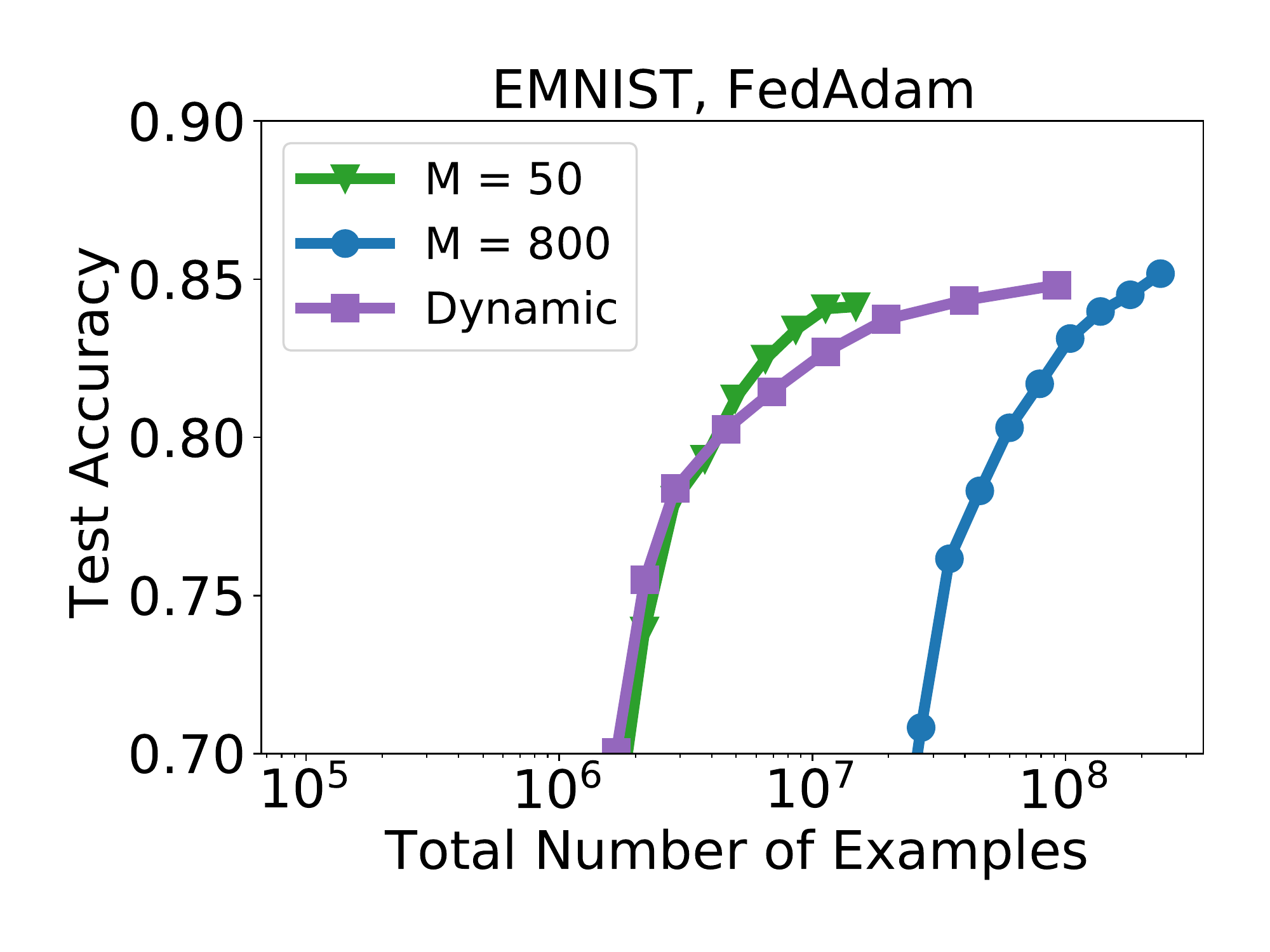}
\end{subfigure}%
\begin{subfigure}{0.24\textwidth}
    \centering
    \includegraphics[width=1\linewidth]{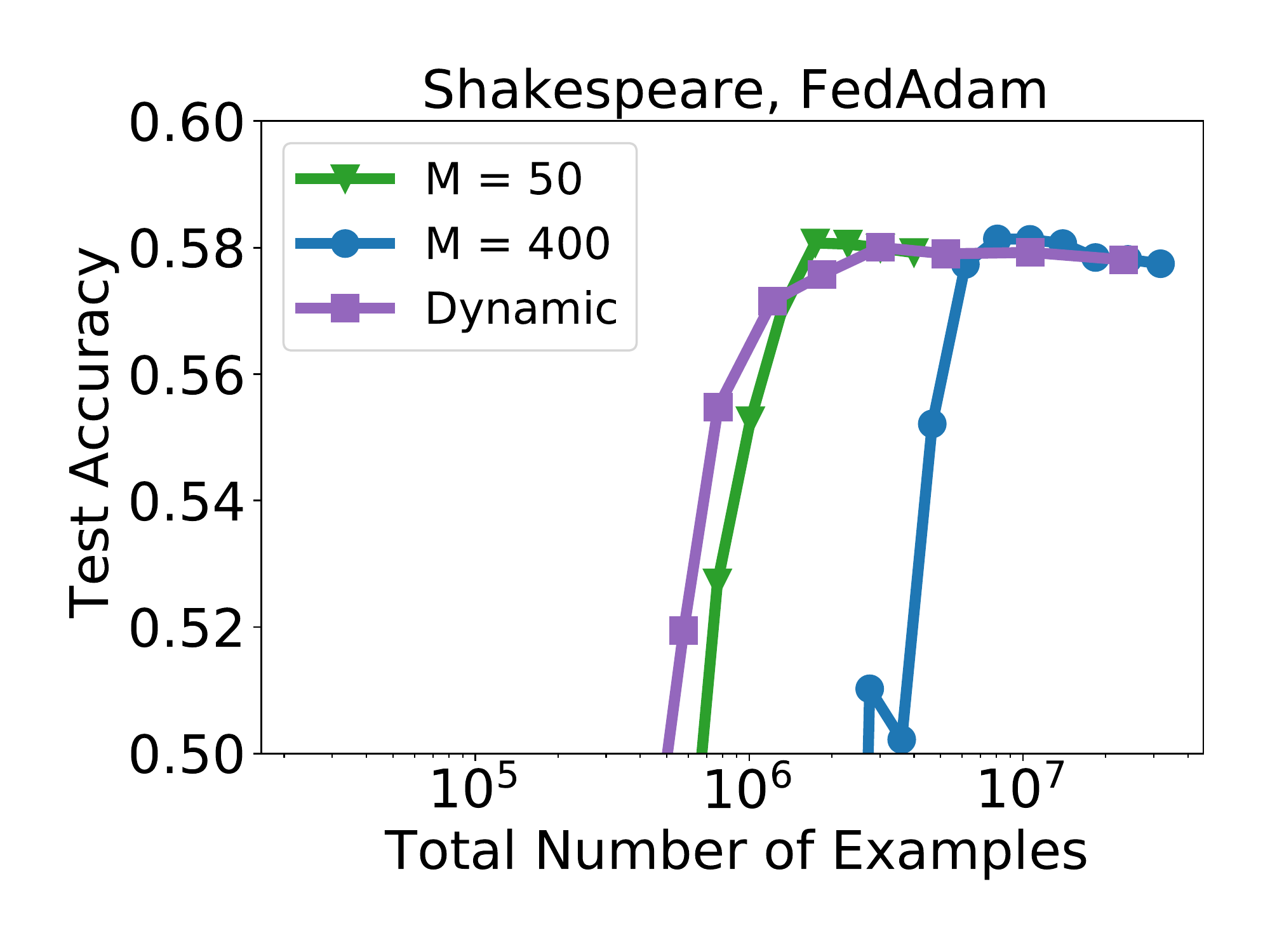}
\end{subfigure}%
\begin{subfigure}{0.24\textwidth}
    \centering
    \includegraphics[width=1\linewidth]{figures/variable_cohorts/stackoverflow_word_fedadam_test_accuracy_versus_examples_variable.pdf}
\end{subfigure}
\caption{Test accuracy of \fedavg (top) and \fedadam (bottom) with respect to the total number of examples. Both algorithms are applied to various tasks, with various fixed cohort sizes, and the dynamically increasing cohort strategy.}
\label{fig:dynamic_accuracy_versus_examples}
\end{figure}

We plot the test accuracy of \fedadam and \fedavg using the dynamic cohort size, as well as fixed cohort sizes of $M = 50$ and $M = 400$ (for CIFAR-100 and Shakespeare) or $M = 800$. In \cref{fig:dynamic_accuracy_versus_examples}, the test accuracy is plotted with respect to the number of examples processed by the clients, in order to measure the data-efficiency of the various methods. We find that while the dynamic cohort strategy can help interpolate the data efficiency between small and large cohort sizes, obtaining the same data efficiency as $M = 50$ for most accuracy thresholds, then transitioning to the data efficiency of larger $M$.

\begin{figure}[ht!]
\centering

\begin{subfigure}{0.24\textwidth}
    \centering
    \includegraphics[width=1\linewidth]{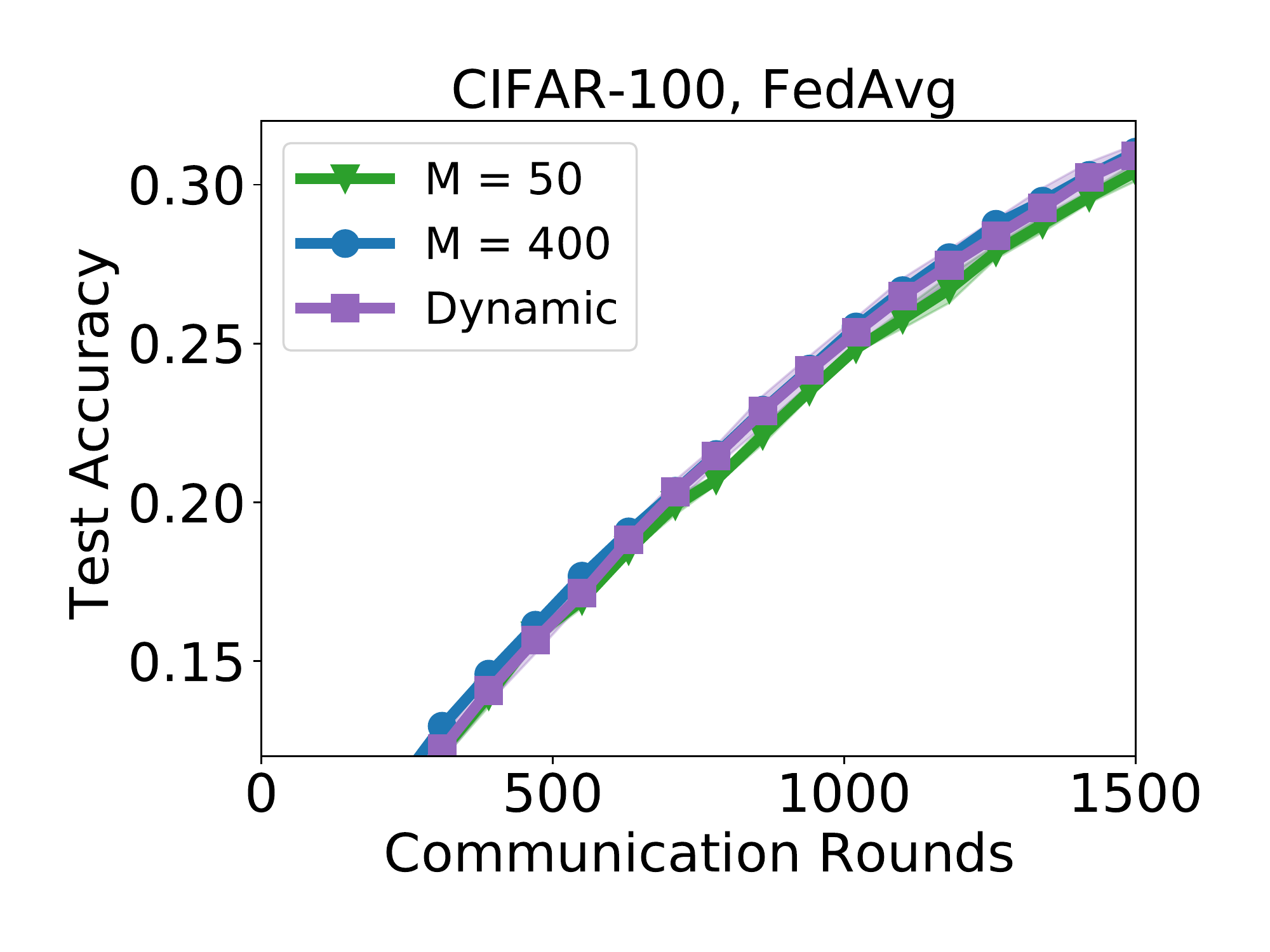}
\end{subfigure}%
\begin{subfigure}{0.24\textwidth}
    \centering
    \includegraphics[width=1\linewidth]{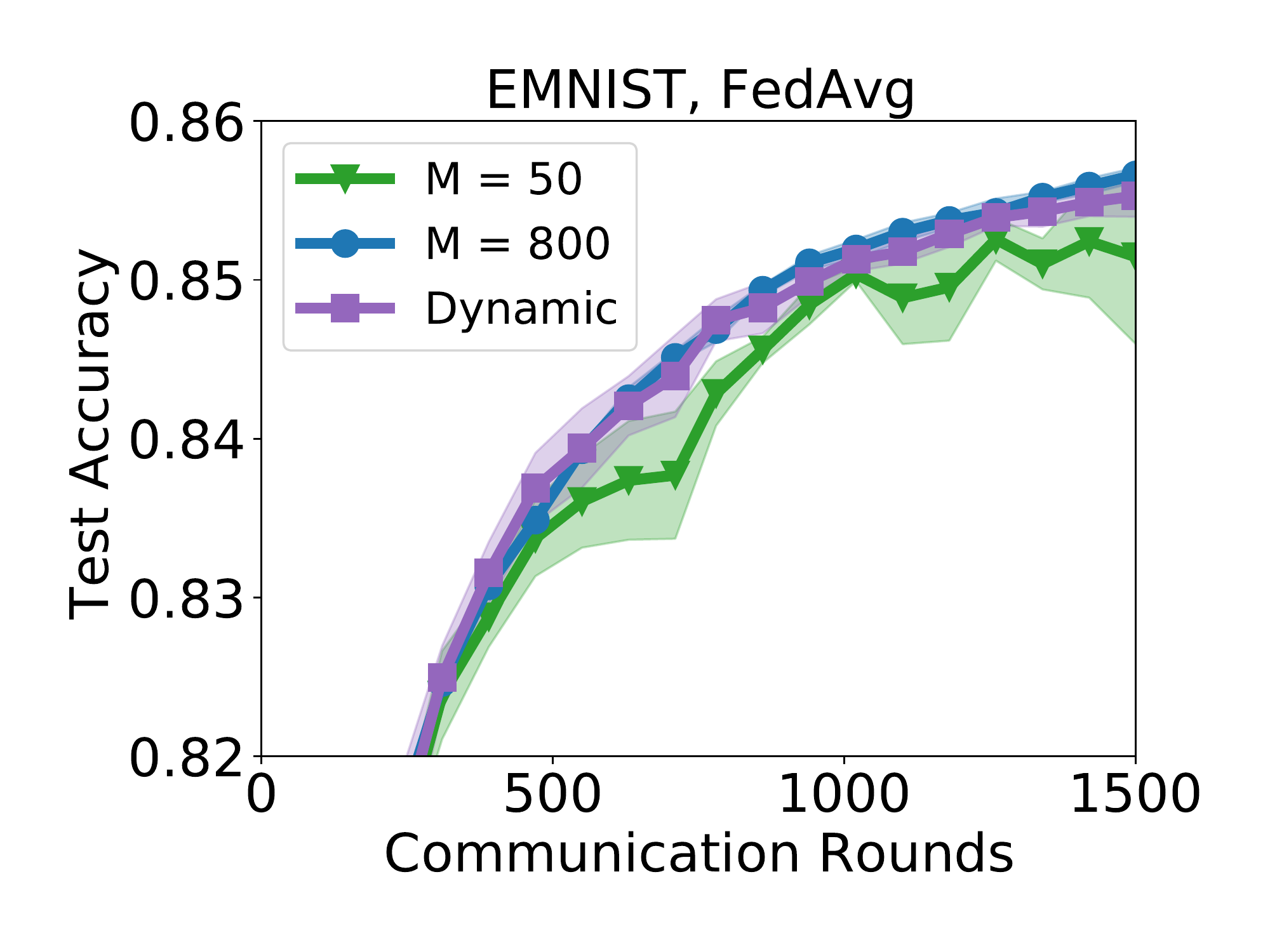}
\end{subfigure}%
\begin{subfigure}{0.24\textwidth}
    \centering
    \includegraphics[width=1\linewidth]{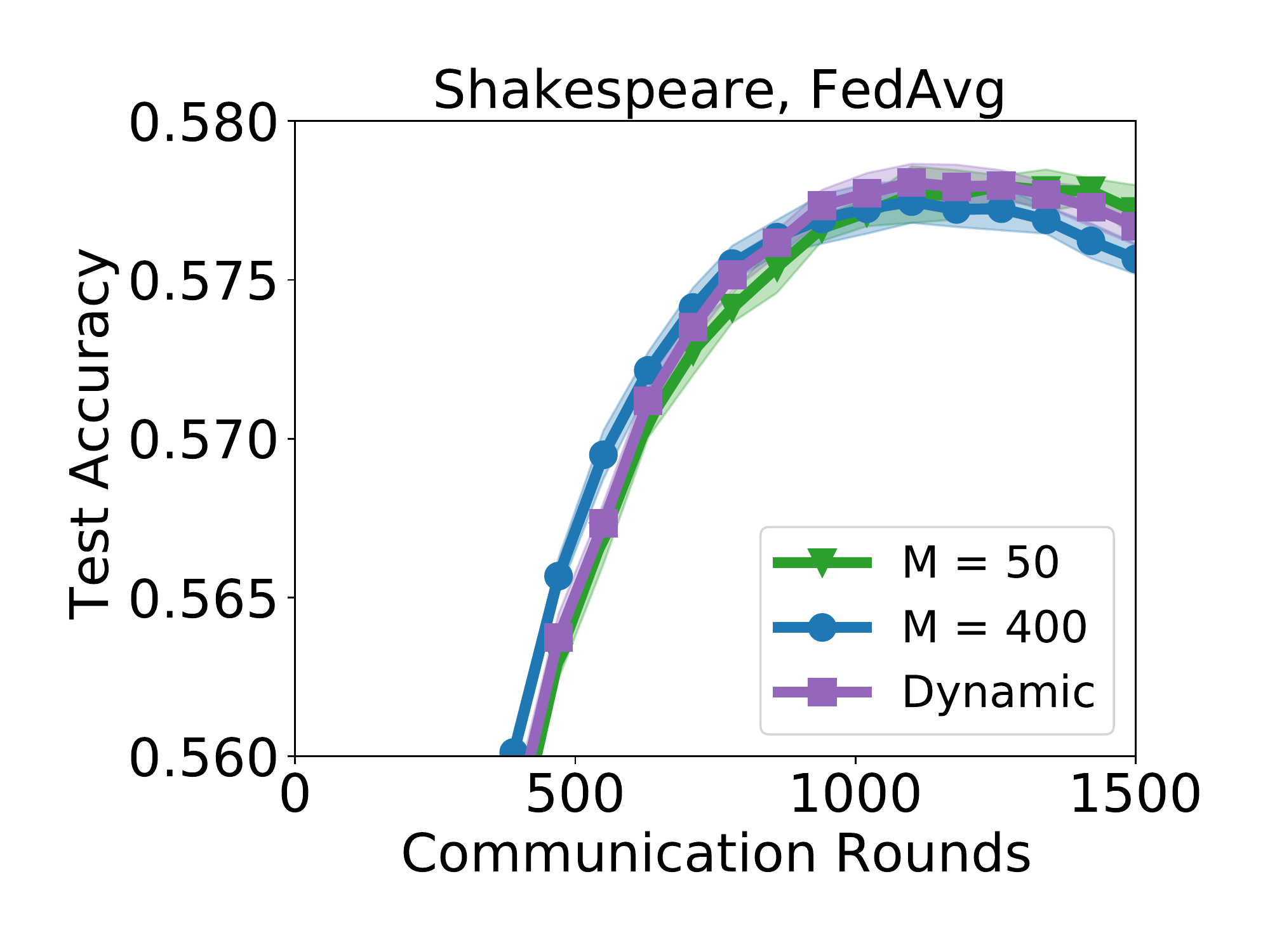}
\end{subfigure}%
\begin{subfigure}{0.24\textwidth}
    \centering
    \includegraphics[width=1\linewidth]{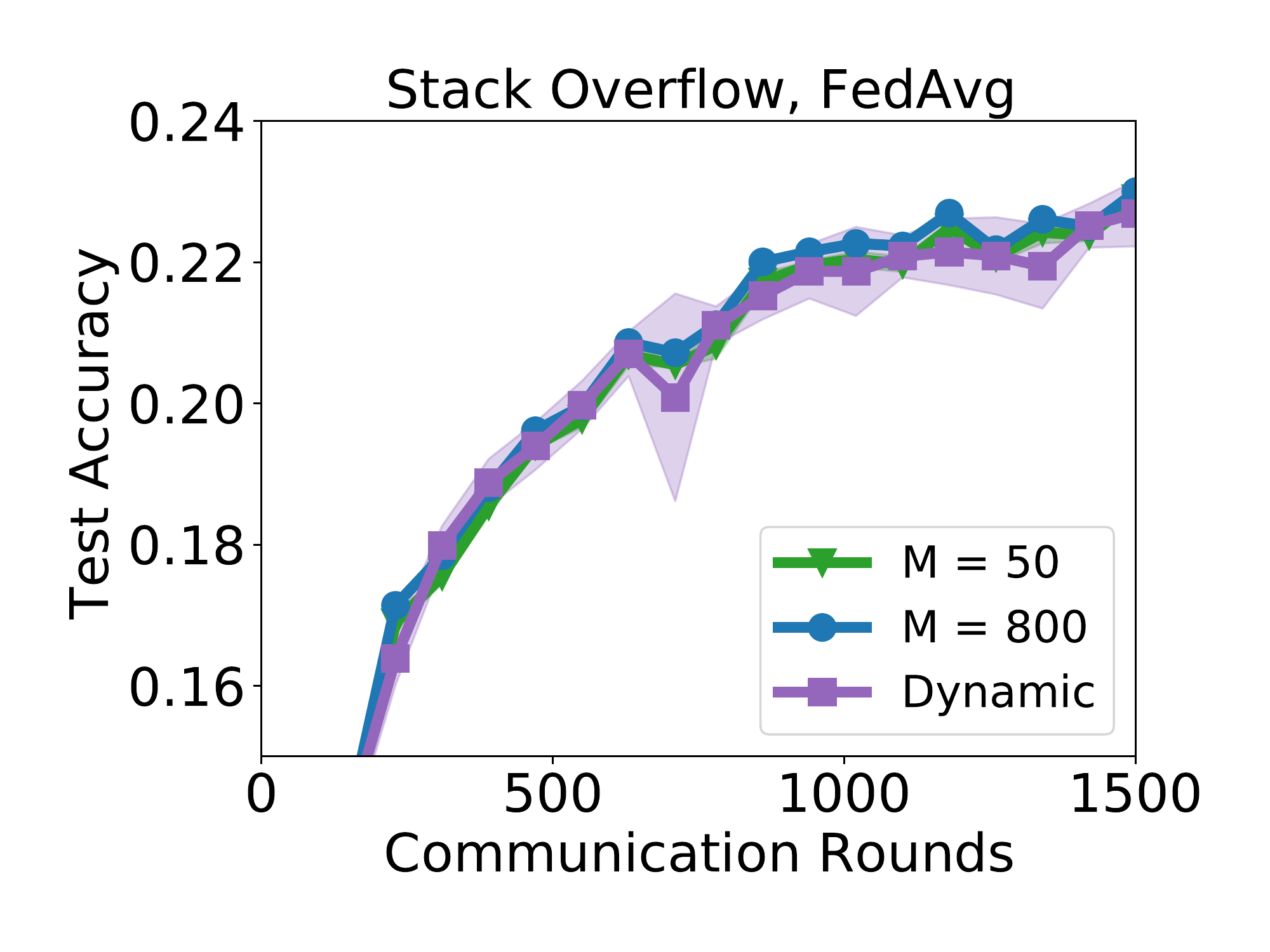}
\end{subfigure}
\begin{subfigure}{0.24\textwidth}
    \centering
    \includegraphics[width=1\linewidth]{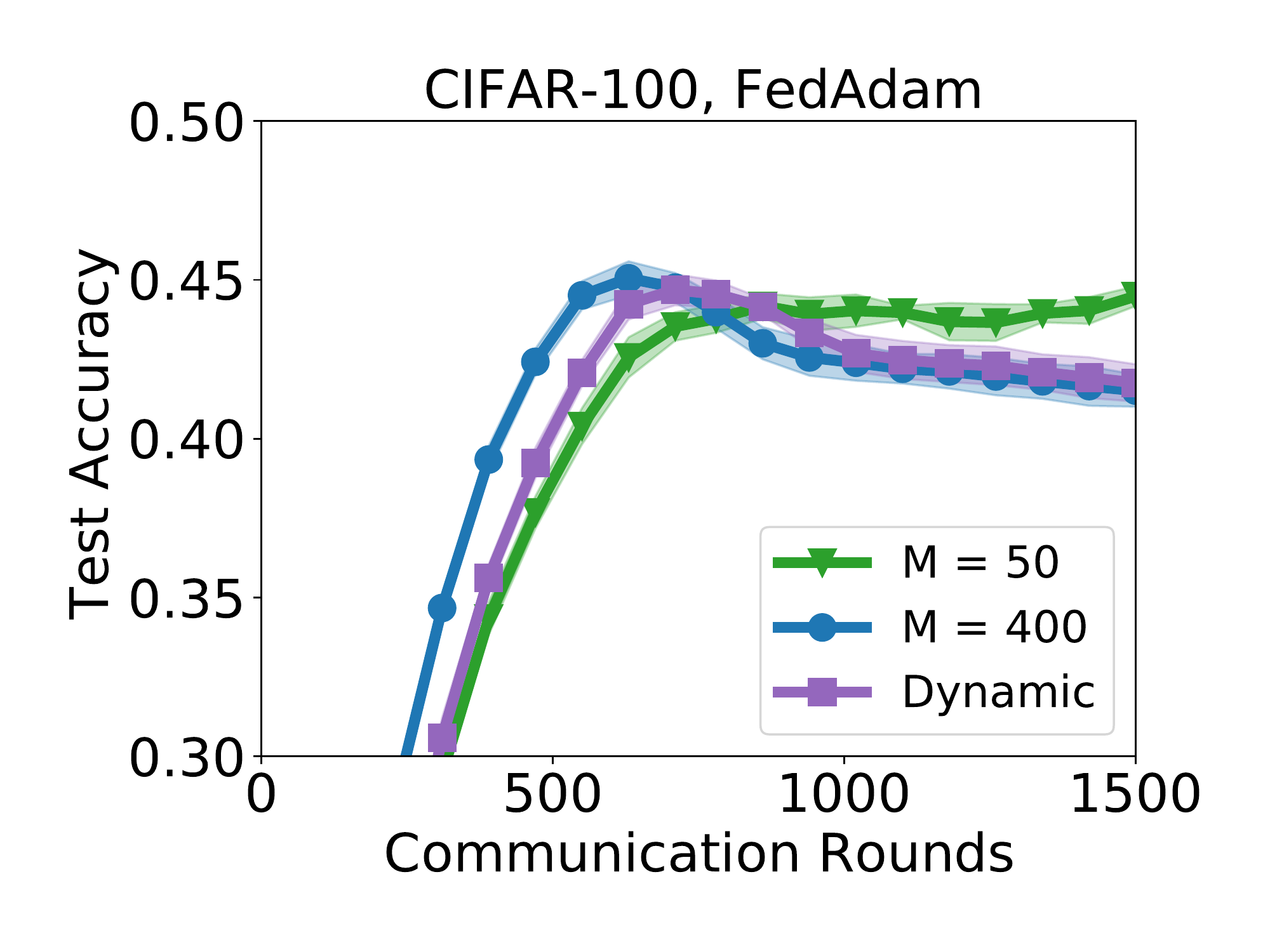}
\end{subfigure}%
\begin{subfigure}{0.24\textwidth}
    \centering
    \includegraphics[width=1\linewidth]{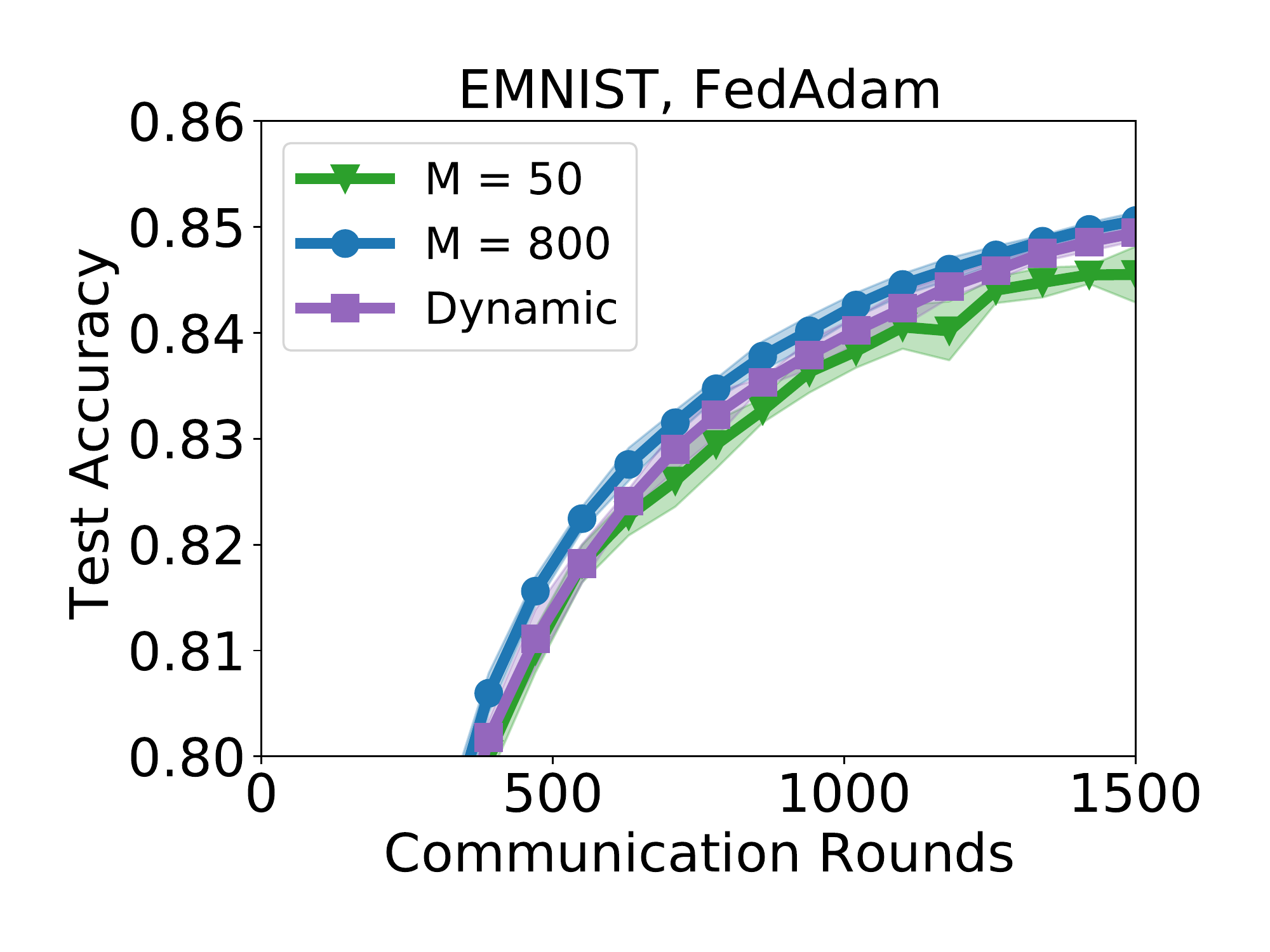}
\end{subfigure}%
\begin{subfigure}{0.24\textwidth}
    \centering
    \includegraphics[width=1\linewidth]{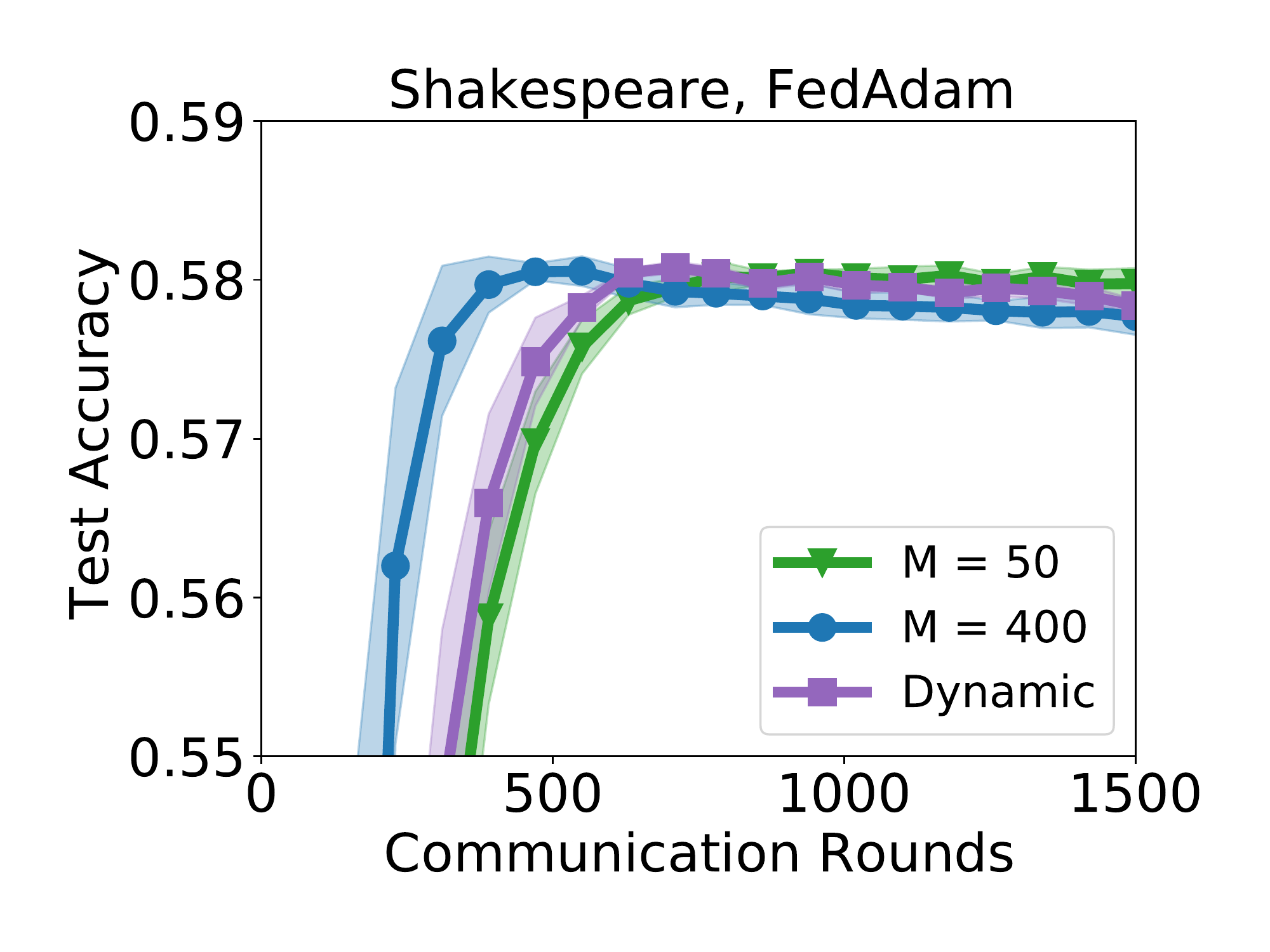}
\end{subfigure}%
\begin{subfigure}{0.24\textwidth}
    \centering
    \includegraphics[width=1\linewidth]{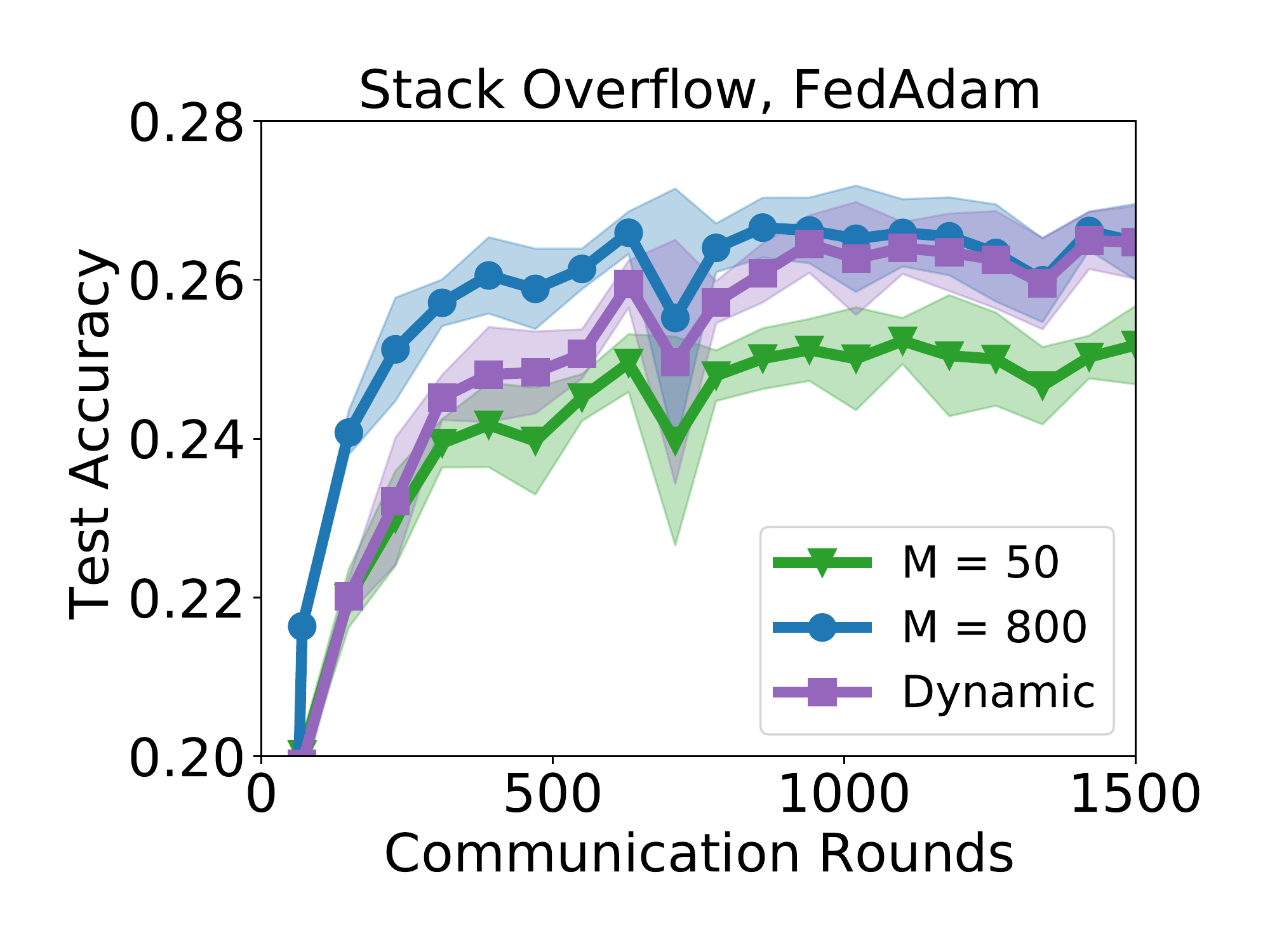}
\end{subfigure}
\caption{Test accuracy of \fedavg (top) and \fedadam (bottom) with respect to the total number of communication rounds. Both algorithms are applied to various tasks, with various fixed cohort sizes, and the dynamically increasing cohort strategy.}
\label{fig:dynamic_accuracy_versus_rounds}
\end{figure}

In \cref{fig:dynamic_accuracy_versus_rounds}, we plot the test accuracy of the methods discussed above with respect to the number of communication rounds, in order to better visualize the generalization behavior of the dynamic cohort strategy. We see that for \fedavg, there is little to no difference between the test accuracy for $M = 50$ and $M = 400$ or $M = 800$, and that the dynamic cohort strategy generally lays in-between these two. This is partially a consequence of the diminishing returns discussed in \cref{sec:diminishing_returns}. For \fedadam, we see that there are more returns to be had for increasing the cohort size. Moreover, we see that the dynamic cohort strategy typically begins at the accuracy level of $M = 50$, and later matches that of the larger cohort. This can be beneficial such as in the case of Stack Overflow, or it can be detrimental as in the case of CIFAR-100, where we see that the dynamic cohort strategy faces the generalization issues in \cref{subsec:generalization_failures}. Thus, we see that the dynamic cohort strategy can help improve the data efficiency of large cohort training, but cannot remedy issues of diminishing returns or generalization failures.

\end{document}